\pdfoutput=1
\documentclass{article}

\PassOptionsToPackage{numbers, compress}{natbib}

\usepackage[final]{neurips_data_2021}

\usepackage[utf8]{inputenc} 
\usepackage[T1]{fontenc}    
\usepackage[colorlinks=true,citecolor=blue]{hyperref}       
\usepackage{url}            
\usepackage{booktabs}       
\usepackage{amsfonts}       
\usepackage{nicefrac}       
\usepackage{microtype}      
\usepackage{comment}        
\usepackage{amsmath}
\usepackage{multirow}
\usepackage{bm}
\usepackage{amsthm}
\usepackage{hhline}
\usepackage{amssymb}
\usepackage{makecell}
\usepackage{mathtools}
\usepackage{color}
\usepackage{xcolor}
\usepackage{MnSymbol}
\usepackage{makecell}
\usepackage{graphicx}
\usepackage{arydshln}
\usepackage{booktabs}
\usepackage{framed}
\usepackage[font=small]{caption}
\usepackage[scientific-notation=true]{siunitx}
\usepackage{wrapfig}
\usepackage{algorithm}
\usepackage{algorithmic}
\usepackage{lipsum}
\usepackage{sidecap}
\usepackage{pifont}
\usepackage{fixltx2e}
\usepackage[flushleft]{threeparttable}
\usepackage{diagbox}
\usepackage{listings}

\newcounter{oldtocdepth}

\newcommand{\hidefromtoc}{%
  \setcounter{oldtocdepth}{\value{tocdepth}}%
  \addtocontents{toc}{\protect\setcounter{tocdepth}{-10}}%
}

\newcommand{\unhidefromtoc}{%
  \addtocontents{toc}{\protect\setcounter{tocdepth}{\value{oldtocdepth}}}%
}

\newcommand{\cmark}{\ding{51}}%
\newcommand{\xmark}{\ding{55}}%

\newcommand{\namel}{\textsc{Multiscale Multimodal Benchmark}}
\newcommand{\names}{\textsc{MultiBench}}

\newcommand{\codes}{\textsc{MultiZoo}}

\newcommand{\dataurl}{\url{https://github.com/pliang279/MultiBench}}
\newcommand{\weburl}{\url{https://cmu-multicomp-lab.github.io/multibench/}}

\definecolor{gg}{RGB}{15,150,15}
\definecolor{rr}{RGB}{230,45,45}

\makeatletter
\def\maketag@@@#1{\hbox{\m@th\normalfont\normalsize#1}}
\makeatother

\usepackage{amsmath,amsfonts,bm}

\def\eqref#1{eq~(\ref{#1})}

\def\1{\bm{1}}

\DeclareMathAlphabet{\mathsfit}{\encodingdefault}{\sfdefault}{m}{sl}
\SetMathAlphabet{\mathsfit}{bold}{\encodingdefault}{\sfdefault}{bx}{n}

\usepackage{hyperref}

\usepackage{booktabs}       %
\usepackage{amsfonts}       %
\usepackage{nicefrac}       %
\usepackage{microtype}      %
\usepackage{subcaption}

\usepackage{enumitem}
\setlist{nolistsep}
\setlist[itemize]{noitemsep, topsep=0pt}

\usepackage{times}

\usepackage{graphicx} %
\usepackage{color}

\usepackage{array}
\usepackage{amssymb}
\usepackage{amsmath}
\usepackage{xspace}
\usepackage{fancyhdr}
\usepackage{comment}

\newcolumntype{H}{>{\setbox0=\hbox\bgroup}c<{\egroup}@{}}

\newcommand{\noaistats}[1]{}  %

\definecolor{darkgreen}{rgb}{0,0.4,0.0}
\definecolor{darkblue}{rgb}{0,0.1,0.3}
\definecolor{darkred}{rgb}{0.7,0.0,0.0}

\newcommand\norm[1]{\left\lVert#1\right\rVert}

\AtBeginDocument{%
  \addtolength\abovedisplayskip{-0.3\baselineskip}%
  \addtolength\belowdisplayskip{-0.3\baselineskip}%
}

\title{\names: Multiscale Benchmarks for\\Multimodal Representation Learning}

\author{%
    Paul Pu Liang$^1$, Yiwei Lyu$^1$, Xiang Fan$^1$, Zetian Wu$^2$, Yun Cheng$^1$,\\
    \textbf{Jason Wu$^1$, Leslie Chen$^3$, Peter Wu$^1$, Michelle A. Lee$^4$, Yuke Zhu$^5$,}\\
    \textbf{Ruslan Salakhutdinov$^1$, Louis-Philippe Morency$^1$}\\
    $^1$CMU, $^2$Johns Hopkins, $^3$Northeastern, $^4$Stanford, $^5$UT Austin\\
    \weburl
}

\begin{document}

\maketitle

\vspace{-6mm}
\begin{abstract}
\vspace{-2mm}
Learning multimodal representations involves integrating information from multiple heterogeneous sources of data. It is a challenging yet crucial area with numerous real-world applications in multimedia, affective computing, robotics, finance, human-computer interaction, and healthcare. Unfortunately, multimodal research has seen limited resources to study (1) generalization across domains and modalities, (2) complexity during training and inference, and (3) robustness to noisy and missing modalities.
In order to accelerate progress towards understudied modalities and tasks while ensuring real-world robustness, we release \names, a systematic and unified large-scale benchmark for multimodal learning spanning $15$ datasets, $10$ modalities, $20$ prediction tasks, and $6$ research areas. \names\ provides an automated end-to-end machine learning pipeline that simplifies and standardizes data loading, experimental setup, and model evaluation. To enable holistic evaluation, \names\ offers a comprehensive methodology to assess (1) generalization, (2) time and space complexity, and (3) modality robustness.
\names\ introduces impactful challenges for future research, including scalability to large-scale multimodal datasets and robustness to realistic imperfections.
To accompany this benchmark, we also provide a standardized implementation of $20$ core approaches in multimodal learning spanning innovations in fusion paradigms, optimization objectives, and training approaches. Simply applying methods proposed in different research areas can improve the state-of-the-art performance on $9/15$ datasets.
Therefore, \names\ presents a milestone in unifying disjoint efforts in multimodal machine learning research and paves the way towards a better understanding of the capabilities and limitations of multimodal models, all the while ensuring ease of use, accessibility, and reproducibility. \names, our standardized implementations, and leaderboards are publicly available, will be regularly updated, and welcomes inputs from the community.
\end{abstract}

\hidefromtoc

\vspace{-6mm}
\section{Introduction}
\vspace{-3mm}

Our perception of the natural world surrounding us involves multiple sensory modalities: we see objects, hear audio signals, feel textures, smell fragrances, and taste flavors. A \textit{modality} refers to a way in which a signal exists or is experienced. Multiple modalities then refer to a combination of multiple signals each expressed in heterogeneous manners~\citep{baltruvsaitis2018multimodal}. Many real-world research problems are inherently multimodal: from the early research on audio-visual speech recognition~\citep{dupont2000audio} to the recent explosion of interest in language, vision, and video understanding~\citep{dupont2000audio} for applications such as multimedia~\citep{liang2018multimodal,1667983}, affective computing~\citep{liang2019tensor,PORIA201798}, robotics~\citep{kirchner2019embedded,lee2019making}, finance~\citep{doi:10.1177/0170840618765019}, dialogue~\citep{Pittermann2010}, human-computer interaction~\citep{dumas2009multimodal,obrenovic2004modeling}, and healthcare~\citep{medical,xu2019multimodal}. The research field of multimodal machine learning (ML) brings unique challenges for both computational and theoretical research given the heterogeneity of various data sources~\citep{baltruvsaitis2018multimodal}. At its core lies the learning of \textit{multimodal representations} that capture correspondences between modalities for prediction, and has emerged as a vibrant interdisciplinary field of immense importance and with extraordinary potential.

\textbf{Limitations of current multimodal datasets:} Current multimodal research has led to impressive advances in benchmarking and modeling for specific domains such as language and vision~\cite{agrawal2017vqa,liang2018computational,lin2014microsoft,ramesh2021zero}. However, other domains, modalities, and tasks are relatively understudied. Many of these tasks are crucial for real-world intelligence such as improving accessibility to technology for diverse populations~\cite{hamisu2011accessible}, accelerating healthcare diagnosis to aid doctors~\citep{MIMIC}, and building reliable robots that can engage in human-AI interactions~\cite{belpaeme2018social,kim2013social,scassellati2012robots}.
Furthermore, current benchmarks typically focus on performance without quantifying the potential drawbacks involved with increased time and space complexity~\citep{tan2016multimodal}, and the risk of decreased robustness from imperfect modalities~\cite{liang2019tensor,pham2019found}. In real-world deployment, a balance between performance, robustness, and complexity is often required.

\begin{figure*}[tbp]
\centering
\vspace{-0mm}
\includegraphics[width=0.9\linewidth]{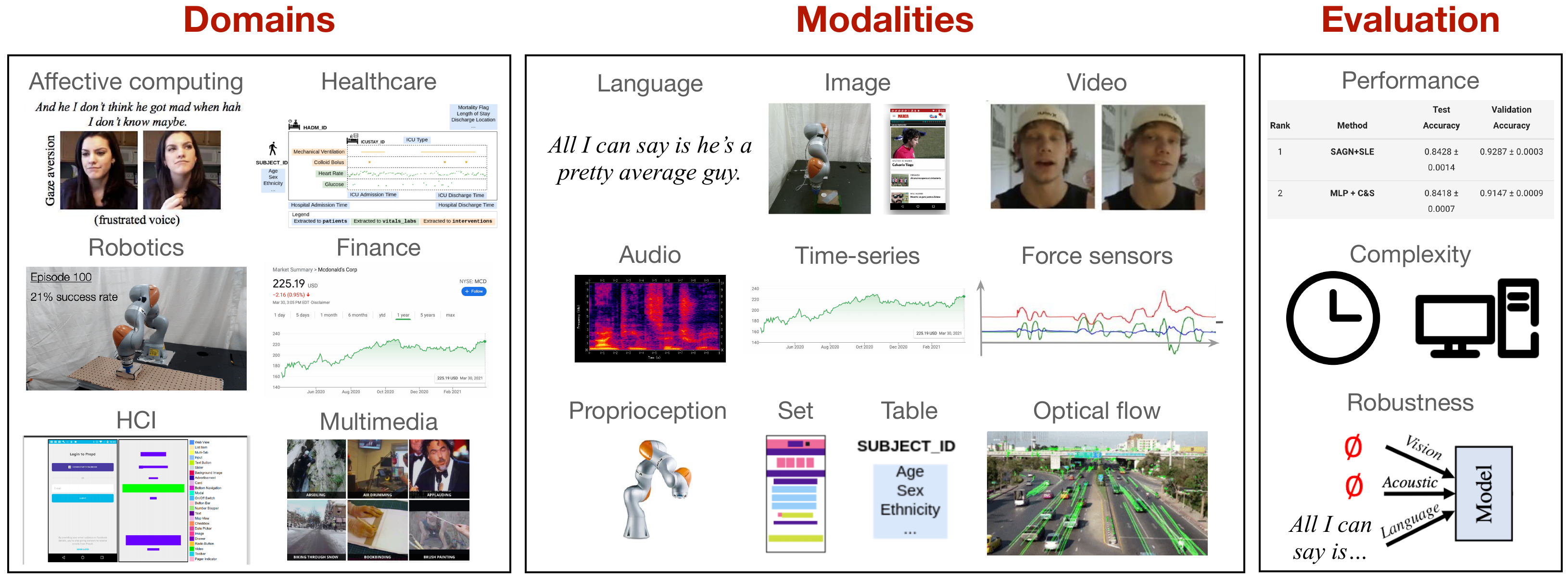}
\vspace{-0mm}
\caption{\names\ contains a diverse set of $15$ datasets spanning $10$ modalities and testing for more than $20$ prediction tasks across $6$ distinct research areas, thereby enabling standardized, reliable, and reproducible large-scale benchmarking of multimodal models. To reflect real-world requirements, \names\ is designed to holistically evaluate (1) performance across domains and modalities, (2) complexity during training and inference, and (3) robustness to noisy and missing modalities.\vspace{-2mm}}
\label{figs:overview}
\end{figure*}

\textbf{\names:} In order to accelerate research in building general-purpose multimodal models, our main contribution is \names\ (Figure~\ref{figs:overview}), a systematic and unified large-scale benchmark that brings us closer to the requirements of real-world multimodal applications. \names\ is designed to comprehensively evaluate $3$ main components: generalization across domains and modalities, complexity during training and inference, and robustness to noisy and missing modalities:

\vspace{-1mm}
1. \textit{Generalization across domains and modalities:} \names\ contains a diverse set of $15$ datasets spanning $10$ modalities and testing for $20$ prediction tasks across $6$ distinct research areas. These research areas include important tasks understudied from a multimodal learning perspective, such as healthcare, finance, and HCI. Building upon extensive data-collection efforts by domain experts, we worked with them to adapt datasets that reflect real-world relevance, present unique challenges to multimodal learning, and enable opportunities in algorithm design and evaluation.

\vspace{-1mm}
2. \textit{Complexity during training and inference:} \names\ also quantifies potential drawbacks involving increased time and space complexity of multimodal learning. Together, these metrics summarize the tradeoffs of current models as a step towards efficiency in real-world settings~\citep{strubell2019energy}.

\vspace{-1mm}
3. \textit{Robustness to noisy and missing modalities:} Different modalities often display different noise topologies, and real-world multimodal signals possibly suffer from missing or noisy data in at least one of the modalities~\cite{baltruvsaitis2018multimodal}. \names\ provides a standardized way to assess the risk of decreased robustness from imperfect modalities through a set of modality-specific and multimodal imperfections that reflect real-world noise, thereby providing a benchmark towards safe and robust deployment.

\vspace{-1mm}
Together, \names\ unifies efforts across separate research areas in multimodal learning to enable quick and accurate benchmarking across a wide range of datasets and metrics. 

To help the community accurately compare performance and ensure reproducibility, \names\ includes an end-to-end pipeline including data preprocessing, dataset splits, multimodal algorithms, evaluation metrics, and cross-validation protocols. This includes an implementation of $20$ core multimodal approaches spanning innovations in fusion paradigms, optimization objectives, and training approaches in a standard public toolkit called \codes.
We perform a systematic evaluation and show that directly applying these methods can improve the state-of-the-art performance on $9$ out of the $15$ datasets. Therefore, \names\ presents a step towards unifying disjoint efforts in multimodal research and paves a way towards a deeper understanding of multimodal models. Most importantly, our public zoo of multimodal benchmarks and models will ensure ease of use, accessibility, and reproducibility. Finally, we outline our plans to ensure the continual availability, maintenance, and expansion of \names, including using it as a theme for future workshops and competitions and to support the multimodal learning courses taught around the world.

\begin{table*}[]
\fontsize{9}{11}\selectfont
\setlength\tabcolsep{3.0pt}
\vspace{-0mm}
\caption{\names\ provides a comprehensive suite of $15$ multimodal datasets to benchmark current and proposed approaches in multimodal representation learning. It covers a diverse range of research areas, dataset sizes, input modalities (in the form of $\ell$: language, $i$: image, $v$: video, $a$: audio, $t$: time-series, $ta$: tabular, $f$: force sensor, $p$: proprioception sensor, $s$: set, $o$: optical flow), and prediction tasks. We provide a standardized data loader for datasets in \names, along with a set of state-of-the-art multimodal models.}
\centering
\footnotesize
\vspace{-0mm}
\begin{tabular}{l|lccccccc}
\Xhline{3\arrayrulewidth}
\multicolumn{1}{l|}{Research Area} & Size & Dataset & Modalities & \# Samples & Prediction task &   \\
\Xhline{0.5\arrayrulewidth}
\multirow{4}{*}{Affective Computing} & \multirow{1}{*}{S} & \textsc{MUStARD}~\cite{castro2019towards} & $\{\ell,v,a\}$ & $690$ & sarcasm \\
& \multirow{1}{*}{M} & \textsc{CMU-MOSI}~\cite{zadeh2016mosi} & $\{\ell,v,a\}$ & $2,199$ & sentiment \\
& \multirow{1}{*}{L} & \textsc{UR-FUNNY}~\cite{hasan2019ur} & $\{\ell,v,a\}$ & $16,514$ & humor \\
& \multirow{1}{*}{L} & \textsc{CMU-MOSEI}~\cite{zadeh2018multimodal} & $\{\ell,v,a\}$ & $22,777$ & sentiment, emotions \\
\Xhline{0.5\arrayrulewidth}\\
\Xhline{0.5\arrayrulewidth}
\multirow{1}{*}{Healthcare} & L & \textsc{MIMIC}~\cite{MIMIC} & $\{t,ta\}$ & $36,212$ & mortality, ICD-$9$ codes\\
\Xhline{0.5\arrayrulewidth}\\
\Xhline{0.5\arrayrulewidth}
\multirow{2}{*}{Robotics} & M & \textsc{MuJoCo Push}~\citep{lee2020multimodal} & $\{i,f,p\}$ & $37,990$ & object pose \\ 
& L & \textsc{Vision\&Touch}~\citep{lee2019making_tro} & $\{i,f,p\}$ & $147,000$ & contact, robot pose \\
\Xhline{0.5\arrayrulewidth}\\
\Xhline{0.5\arrayrulewidth}
\multirow{3}{*}{Finance} & M & \textsc{Stocks-F\&B} & $\{t \times 18\}$ & $5,218$ & stock price, volatility \\
& M & \textsc{Stocks-Health} & $\{t \times 63\}$ & $5,218$ & stock price, volatility \\
& M & \textsc{Stocks-Tech} & $\{t \times 100\}$ & $5,218$ & stock price, volatility \\
\Xhline{0.5\arrayrulewidth}\\
\Xhline{0.5\arrayrulewidth}
\multirow{1}{*}{HCI} & S & \textsc{ENRICO}~\citep{leiva2020enrico} & $\{i,s\}$ & $1,460$ & design interface \\
\Xhline{0.5\arrayrulewidth}\\
\Xhline{0.5\arrayrulewidth}
\multirow{4}{*}{Multimedia} & \multirow{1}{*}{S} & \textsc{Kinetics400-S}~\citep{kay2017kinetics} & $\{v,a,o\}$ & $2,624$ & human action \\
& \multirow{1}{*}{M} & \textsc{MM-IMDb}~\citep{arevalo2017gated} & $\{\ell,i\}$ & $25,959$ & movie genre \\
& \multirow{1}{*}{M} & \textsc{AV-MNIST}~\citep{vielzeuf2018centralnet} & $\{i,a\}$ & $70,000$ & digit \\

& \multirow{1}{*}{L} & \textsc{Kinetics400-L}~\citep{kay2017kinetics} & $\{v,a,o\}$ & $306,245$ & human action \\
\Xhline{3\arrayrulewidth}
\end{tabular}
\vspace{-2mm}
\label{data:overview}
\end{table*}



\vspace{-3mm}
\section{\names: The \namel}
\label{dataset}
\vspace{-3mm}

\textbf{Background:} We define a modality as a single particular mode in which a signal is expressed or experienced. Multiple modalities then refer to a combination of multiple heterogeneous signals~\citep{baltruvsaitis2018multimodal}.
The first version of \names\ focuses on benchmarking algorithms for \textit{multimodal fusion}, where the main challenge is to join information from two or more modalities to perform a prediction (e.g., classification, regression). Classic examples for multimodal fusion include audio-visual speech recognition where visual lip motion is fused with speech signals to predict spoken words~\citep{dupont2000audio}. Multimodal fusion can be contrasted with multimodal translation where the goal is to generate a new and different modality~\citep{vinyals2016show}, grounding and question answering where one modality is used to query information in another (e.g., visual question answering~\citep{agrawal2017vqa}), and unsupervised or self-supervised multimodal representation learning~\citep{lu2019vilbert,Su2020VLBERT}. We plan future versions of \names\ to study these important topics in multimodal research in Appendix~\ref{appendix:future}.

\vspace{-1mm}
Each of the following $15$ datasets in \names\ contributes a unique perspective to the various technical challenges in multimodal learning involving learning and aligning complementary information, scalability to a large number of modalities, and robustness to realistic real-world imperfections.

\vspace{-2mm}
\subsection{Datasets}
\label{dataset_details}
\vspace{-2mm}

Table~\ref{data:overview} shows an overview of the datasets provided in \names. We provide a brief overview of the modalities and tasks for each of these datasets and refer the reader to Appendix~\ref{appendix:data} for details.

\vspace{-1mm}
\textbf{Affective computing} studies the perception of human affective states (emotions, sentiment, and personalities) from our natural display of multimodal signals spanning language (spoken words), visual (facial expressions, gestures), and acoustic (prosody, speech tone)~\citep{picard2000affective}. It has broad impacts towards building emotionally intelligent computers, human behavior analysis, and AI-assisted education. \names\ contains $4$ datasets involving fusing \textit{language}, \textit{video}, and \textit{audio} time-series data to predict sentiment (\textsc{CMU-MOSI}~\citep{zadeh2016mosi}), emotions (\textsc{CMU-MOSEI}~\citep{zadeh2018multimodal}), humor (\textsc{UR-FUNNY}~\cite{hasan2019ur}), and sarcasm (\textsc{MUStARD}~\citep{castro2019towards}). Complementary information may occurs at different moments, requiring models to address the multimodal challenges of grounding and alignment.

\vspace{-1mm}
\textbf{Healthcare:} Modern medical decision-making often involves integrating complementary information and signals from several sources such as lab tests, imaging reports, and patient-doctor conversations. Multimodal models can help doctors make sense of high-dimensional data and assist them in the diagnosis process~\cite{amisha2019overview}. \names\ includes the large-scale \textsc{MIMIC} dataset~\citep{MIMIC} which records ICU patient data including \textit{time-series} data measured every hour and other demographic variables (e.g., age, gender, ethnicity in the form of \textit{tabular numerical} data). These are used to predict the disease ICD-$9$ code and mortality rate. \textsc{MIMIC} poses unique challenges in integrating time-varying and static modalities, reinforcing the need of aligning multimodal information at correct granularities.

\vspace{-1mm}
\textbf{Robotics:} Modern robot systems are equipped with multiple sensors to aid in their decision-making. We include the large-scale \textsc{MuJoCo Push}~\citep{lee2020multimodal} and \textsc{Vision\&Touch}~\citep{lee2019making_tro} datasets which record the manipulation of simulated and real robotic arms equipped with \textit{visual} (RGB and depth), \textit{force}, and \textit{proprioception} sensors. In \textsc{MuJoCo Push}, the goal is to predict the pose of the object being pushed by the robot end-effector. In \textsc{Vision\&Touch}, the goal is to predict action-conditional learning objectives that capture forward dynamics of the different modalities (contact prediction and robot end-effector pose). Robustness is important due to the risk of real-world sensor failures~\citep{lee2020detect}.

\vspace{-1mm}
\textbf{Finance:} We gathered historical stock data from the internet to create our own dataset for financial time-series prediction across $3$ groups of correlated stocks: \textsc{Stocks-F\&B}, \textsc{Stocks-Health}, and \textsc{Stocks-Tech}. Within each group, the previous stock prices of a set of stocks are used as multimodal \textit{time-series} inputs to predict the price and volatility of a related stock (e.g., using Apple, Google, and Microsoft data to predict future Microsoft prices). Multimodal stock prediction~\citep{sardelich2018multimodal} presents scalability issues due to a large number of modalities ($18/63/100$ vs $2/3$ in most datasets), as well as robustness challenges arising from real-world data with an inherently low signal-to-noise ratio.

\vspace{-1mm}
\textbf{Human Computer Interaction (HCI)} studies the design of computer technology and interactive interfaces between humans and computers~\citep{dix2000human}. Many real-world problems involve multimodal inputs such as language, visual, and audio interfaces. We use the \textsc{Enrico} (Enhanced Rico) dataset~\cite{deka2017rico,leiva2020enrico} of Android app screens (consisting of an \textit{image} as well as a \textit{set} of apps and their locations) categorized by their design motifs and collected for data-driven design applications such as design search, user interface (UI) layout generation, UI code generation, and user interaction modeling.

\vspace{-1mm}
\textbf{Multimedia:} A significant body of research in multimodal learning has been fueled by the large availability of multimedia data (language, image, video, and audio) on the internet. \names\ includes $3$ popular large-scale multimedia datasets with varying sizes and levels of difficulty: (1) \textsc{AV-MNIST}~\citep{vielzeuf2018centralnet} is assembled from \textit{images} of handwritten digits~\cite{mnist} and \textit{audio} samples of spoken digits~\cite{tidigits}, (2) \textsc{MM-IMDb}~\citep{arevalo2017gated} uses movie \textit{titles}, \textit{metadata}, and movie \textit{posters} to perform multi-label classification of movie genres, and (3) \textsc{Kinetics}~\citep{kay2017kinetics} contains \textit{video}, \textit{audio}, and \textit{optical flow} of $306,245$ video clips annotated for $400$ human actions. To ease experimentation, we split \textsc{Kinetics} into small and large partitions (see Appendix~\ref{appendix:data}).

\vspace{-2mm}
\subsection{Evaluation Protocol}
\label{eval}
\vspace{-2mm}

\names\ contains evaluation scripts for the following holistic desiderata in multimodal learning:

\vspace{-1mm}
\textbf{Performance:} We standardize evaluation using metrics designed for each dataset, including MSE and MAE for regression to accuracy, micro \& macro F1-score, and AUPRC for classification.

\vspace{-1mm}
\textbf{Complexity:} Modern ML research unfortunately causes significant impacts to energy consumption~\citep{strubell2019energy}, a phenomenon often exacerbated in processing high-dimensional multimodal data. As a step towards quantifying energy complexity and recommending lightweight multimodal models, \names\ records the amount of information taken in bits (i.e., data size), number of model parameters, as well as time and memory resources required during the entire training process. Real-world models may also need to be small and compact to run on mobile devices~\cite{radu2016towards} so we also report inference time and memory on CPU and GPU (see Appendix~\ref{appendix:complexity}).

\vspace{-1mm}
\textbf{Robustness:} Real-world multimodal data is often imperfect as a result of missing entries, noise corruption, or missing modalities entirely, which calls for robust models that can still make accurate predictions despite only having access to noisy and missing signals~\citep{liang2019tensor,pham2019found}. To standardize efforts in evaluating robustness, \names\ includes the following tests: (1) \textit{Modality-specific imperfections} are independently applied to each modality taking into account its unique noise topologies (i.e., flips and crops of images, natural misspellings in text, abbreviations in spoken audio). (2) \textit{Multimodal imperfections} capture correlations in imperfections across modalities (e.g., missing modalities, or a chunk of time missing in multimodal time-series data). We use both qualitative measures (performance-imperfection curve) and quantitative metrics~\citep{taori2020measuring} that summarize (1) \textit{relative robustness} measuring accuracy under imperfections and (2) \textit{effective robustness} measuring the \textit{rate} of accuracy drops after equalizing for initial accuracy on clean test data (see Appendix~\ref{appendix:robustness} for details).

\vspace{-3mm}
\section{\codes: A Zoo of Multimodal Algorithms}
\label{algorithms}
\vspace{-3mm}

\begin{figure*}[tbp]
\centering
\vspace{-0mm}
\includegraphics[width=\linewidth]{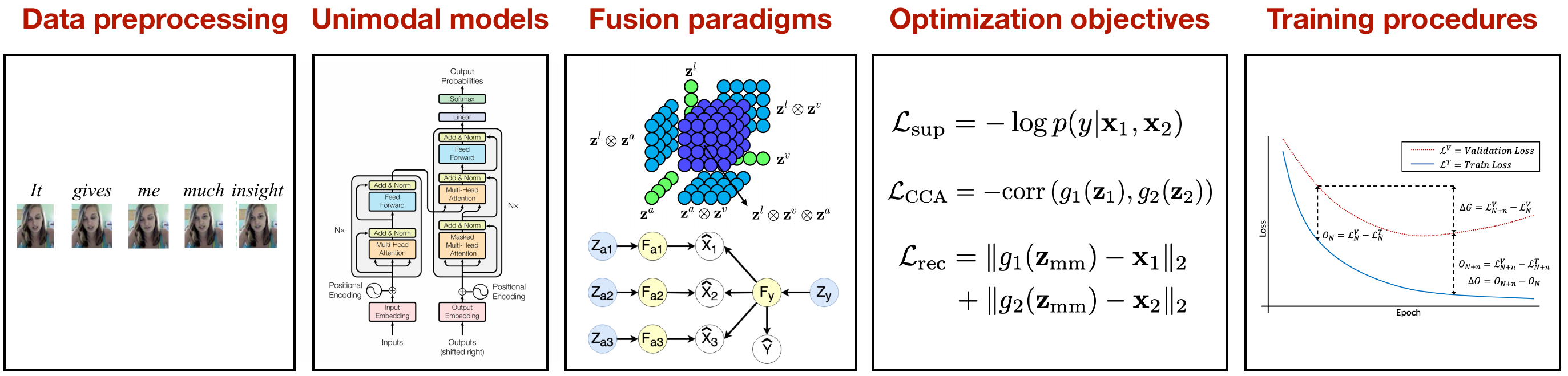}
\vspace{-2mm}
\caption{\codes\ provides a standardized implementation of a suite of multimodal methods in a modular fashion to enable accessibility for new researchers, compositionality of approaches, and reproducibility of results.\vspace{-2mm}}
\label{figs:multizoo}
\end{figure*}

To complement \names, we release a comprehensive toolkit, \codes, as starter code for multimodal algorithms which implements $20$ methods spanning different methodological innovations in (1) data preprocessing, (2) fusion paradigms, (3) optimization objectives, and (4) training procedures (see Figure~\ref{figs:multizoo}). To introduce these algorithms, we use the simple setting with $2$ modalities for notational convenience but refer the reader to Appendix~\ref{appendix:algos} for detailed descriptions and implementations. We use $\mathbf{x}_1, \mathbf{x}_2$ for input modalities, $\mathbf{z}_1, \mathbf{z}_2$ for unimodal representations, $\mathbf{z}_\textrm{mm}$ for the multimodal representation, and $\hat{y}$ for the predicted label.

\vspace{-2mm}
\subsection{Data Preprocessing}
\vspace{-2mm}

\textbf{Temporal alignment}~\cite{chen2017multimodal} has been shown to help tackle the multimodal alignment problem for time-series data. This approach assumes a temporal granularity of the modalities (e.g., at the level of words for text) and aligns information from the remaining modalities to the same granularity. We call this approach \textsc{WordAlign}~\cite{chen2017multimodal} for temporal data where text is one of the modalities.

\vspace{-2mm}
\subsection{Fusion Paradigms}
\label{model_design}
\vspace{-2mm}

\textbf{Early and late fusion:} Early fusion performs concatenation of input data before using a model (i.e., $\mathbf{z}_\textrm{mm} = \left[ \mathbf{x}_1, \mathbf{x}_2\right]$) while late fusion applies suitable unimodal models to each modality to obtain their feature representations, concatenates these features, and defines a classifier to the label (i.e., $\mathbf{z}_\textrm{mm} = \left[ \mathbf{z}_1, \mathbf{z}_2\right]$)~\citep{baltruvsaitis2018multimodal}. \codes\ includes their implementations denoted as \textsc{EF} and \textsc{LF} respectively.

\vspace{-1mm}
\textbf{Tensors} are specifically designed to tackle the multimodal complementarity challenge by explicitly capturing higher-order interactions across modalities~\cite{zadeh2017tensor}. Given unimodal representations $\mathbf{z}_1, \mathbf{z}_2$, tensors are defined as $\mathbf{z}_\textrm{mm} = \begin{bmatrix} \mathbf{z}_{1} \\ 1 \end{bmatrix} \otimes \begin{bmatrix} \mathbf{z}_{2} \\ 1 \end{bmatrix}$ where $\otimes$ denotes an outer product. However, computing tensor products is expensive since their dimension scales exponentially with the number of modalities so several efficient approximations have been proposed~\cite{hou2019deep,liang2019tensor,liu2018efficient}. \codes\ includes Tensor Fusion (\textsc{TF})~\citep{zadeh2017tensor} as well as the approximate Low-rank Tensor Fusion (\textsc{LRTF})~\citep{liu2018efficient}.

\vspace{-1mm}
\textbf{Multiplicative Interactions (MI)} generalize tensor products to include learnable parameters that capture multimodal interactions~\citep{Jayakumar2020Multiplicative}. In its most general form, MI defines a bilinear product $\mathbf{z}_\textrm{mm} = \mathbf{z}_1 \mathbb{W} \mathbf{z}_2 + \mathbf{z}_1^\top \mathbf{U} + \mathbf{V} \mathbf{z}_2 + \mathbf{b}$ where $\mathbb{W}, \mathbf{U}, \mathbf{Z}$, and $\mathbf{b}$ are trainable parameters. By appropriately constraining the rank and structure of these parameters, MI recovers HyperNetworks~\citep{ha2016hypernetworks} (unconstrained parameters resulting in a matrix output), Feature-wise linear modulation (FiLM)~\cite{perez2018film,zhong2019rtfm} (diagonal parameters resulting in vector output), and 
Sigmoid units~\cite{dauphin2017language} (scalar parameters resulting in scalar output). \codes\ includes all $3$ as \textsc{MI-Matrix}, \textsc{MI-Vector}, and \textsc{MI-Scalar} respectively.

\vspace{-1mm}
\textbf{Multimodal gated units} learn representations that dynamically change for every input~\cite{chaplot2017gated,wang2020makes,xu2015show}. Its general form can be written as $\mathbf{z}_\textrm{mm} = \mathbf{z}_1 \odot h(\mathbf{z}_2)$, where $h$ represents a function with sigmoid activation and $\odot$ denotes element-wise product. $h(\mathbf{z}_2)$ is commonly referred to as ``attention weights'' learned from $\mathbf{z}_2$ to attend on $\mathbf{z}_1$. Attention is conceptually similar to \textsc{MI-Vector} but recent work has explored more expressive forms of $h$ such as using a Query-Key-Value mechanism~\cite{wang2020makes} or fully-connected layers~\citep{chaplot2017gated}. We implement the Query-Key-Value mechanism as \textsc{NL Gate}~\cite{wang2020makes}.

\vspace{-1mm}
\textbf{Temporal attention models} tackle the challenge of multimodal alignment and complementarity. Transformer models~\citep{vaswani2017attention} are useful for temporal data by automatically aligning and capturing complementary features at different time-steps~\cite{tsai2019multimodal,yao-wan-2020-multimodal}. We include the Multimodal Transformer (\textsc{MulT})~\citep{tsai2019multimodal} which applied a Crossmodal Transformer block using $\mathbf{z}_1$ to attend to $\mathbf{z}_2$ (and vice-versa) to obtain a multimodal representation $\mathbf{z}_\textrm{mm} = \left[ \mathbf{z}_{1 \rightarrow 2}, \mathbf{z}_{2 \rightarrow 1} \right] = \left[ \textsc{CM}(\mathbf{z}_1, \mathbf{z}_2), \textsc{CM}(\mathbf{z}_2, \mathbf{z}_1) \right]$.

\vspace{-1mm}
\textbf{Architecture search:} Instead of hand-designing architectures, several approaches define a set of atomic operations (e.g., linear transformation, activation, attention, etc.) and use architecture search to learn the best order of these operations for a given task~\cite{perez2019mfas,xu2021mufasa}, which we call \textsc{MFAS}.

\vspace{-2mm}
\subsection{Optimization Objectives}
\vspace{-2mm}

In addition to the standard supervised losses (e.g., cross entropy for classification, MSE/MAE for regression), several proposed methods have proposed new objective functions based on:

\vspace{-1mm}
\textbf{Prediction-level alignment} objectives tackle the challenge of alignment by capturing a representations where semantically similar concepts from different modalities are close together~\cite{bachman2019learning,cui2020unsupervised,lee2019making,tian2019contrastive}. Alignment objectives have been applied at both prediction and feature levels. In the former, we implement Canonical Correlation Analysis (\textsc{CCA})~\cite{andrew2013deep,sun2020learning,wang2015deep}, which maximizes correlation by adding a loss term $\mathcal{L}_\textrm{CCA} = - \textrm{corr} \left( g_1(\mathbf{z}_{1}), g_2(\mathbf{z}_{2}) \right)$ where $g_1,g_2$ are auxiliary classifiers mapping each unimodal representation to the label.

\vspace{-1mm}
\textbf{Feature-level alignment:} In the latter, contrastive learning has emerged as a popular approach to bring similar concepts close in feature space and different concepts far away~\cite{cui2020unsupervised,lee2019making,tian2019contrastive}. We include \textsc{ReFNet}~\cite{sankaran2021multimodal} which uses a self-supervised contrastive loss between unimodal representations $\mathbf{z}_{1}, \mathbf{z}_{2}$ and the multimodal representation $\mathbf{z}_\textrm{mm}$, i.e., $\mathcal{L}_\textrm{contrast} = 1 - \textrm{cos} (\mathbf{z}_\textrm{mm}, g_1(\mathbf{z}_{1})) + 1 - \textrm{cos} (\mathbf{z}_\textrm{mm}, g_2(\mathbf{z}_{2})) $ where $g_1,g_2$ is a layer mapping each modality's representation into the joint multimodal space.

\vspace{-1mm}
\textbf{Reconstruction objectives} based on generative-discriminative models (e.g., VAEs) aim to reconstruct the input (or some part of the input)~\cite{lee2019making,tsai2019learning}. These have been shown to better preserve task-relevant information learned in the representation, especially in settings with sparse supervised signals such as robotics~\cite{lee2019making} and long videos~\cite{tsai2019learning}. We include the Multimodal Factorized Model (\textsc{MFM})~\citep{tsai2019learning} that learns a representation $\mathbf{z}_\textrm{mm}$ that can reconstruct input data $\mathbf{x}_{1}, \mathbf{x}_{2}$ while also predicting the label, i.e., adding an objective $\mathcal{L}_\textrm{rec} = \norm{ g_1(\mathbf{z}_\textrm{mm}) - \mathbf{x}_{1}}_2 + \norm{ g_2(\mathbf{z}_\textrm{mm}) - \mathbf{x}_{2}}_2$ where $g_1,g_2$ are auxiliary decoders mapping $\mathbf{z}_\textrm{mm}$ to each raw input modality. \textsc{MFM} can be paired with any multimodal model from section~\ref{model_design} (e.g., learning $\mathbf{z}_\textrm{mm}$ via tensors and adding a term to reconstruct input data).

\newlength{\textfloatsepsave} \setlength{\textfloatsepsave}{\textfloatsep} \setlength{\textfloatsep}{0pt}

\begin{algorithm}[tb]
    \caption{PyTorch code integrating \names\ datasets and \codes\ models.}
    \label{alg:code}
   
    \definecolor{codeblue}{rgb}{0.25,0.5,0.5}
    \lstset{
      basicstyle=\fontsize{7.2pt}{7.2pt}\ttfamily\bfseries,
      commentstyle=\fontsize{7.2pt}{7.2pt}\color{codeblue},
      keywordstyle=\fontsize{7.2pt}{7.2pt},
    }
\begin{lstlisting}[language=python]
from datasets.get_data import get_dataloader
from unimodals.common_models import ResNet, Transformer
from fusions.common_fusions import MultInteractions
from training_structures.gradient_blend import train, test

# loading Multimodal IMDB dataset
traindata, validdata, testdata = get_dataloader('multimodal_imdb')
out_channels = 3
# defining ResNet and Transformer unimodal encoders
encoders = [ResNet(in_channels=1, out_channels, layers=5),
            Transformer(in_channels=1, out_channels, layers=3)]
# defining a Multiplicative Interactions fusion layer
fusion = MultInteractions([out_channels*8, out_channels*32], out_channels*32, 'matrix')
classifier = MLP(out_channels*32, 100, labels=23)
# training using Gradient Blend algorithm
model = train(encoders, fusion, classifier, traindata, validdata, 
        epochs=100, optimtype=torch.optim.SGD, lr=0.01, weight_decay=0.0001)
# testing
performance, complexity, robustness = test(model, testdata)
\end{lstlisting}
\vspace{-2mm}
\end{algorithm}

\setlength{\textfloatsep}{\textfloatsepsave}

\vspace{-1mm}
\textbf{Improving robustness:} These approaches modify the objective function to account for robustness to noisy~\cite{liang2019tensor} or missing~\cite{lee2020detect,ma2021smil,pham2019found} modalities. \codes\ includes \textsc{MCTN}~\cite{pham2019found} which uses cycle-consistent translation to predict the noisy/missing modality from present ones (i.e., a path $\mathbf{x}_{1} \rightarrow \mathbf{z}_\textrm{mm} \rightarrow \hat{\mathbf{x}}_{2} \rightarrow \mathbf{z}_\textrm{mm} \rightarrow \hat{\mathbf{x}}_{1}$, with additional reconstruction losses $\mathcal{L}_\textrm{rec} = \norm{ \mathbf{x}_{1} - \hat{\mathbf{x}}_{1} }_2 + \norm{ \mathbf{x}_{2} - \hat{\mathbf{x}}_{2} }_2$). While \textsc{MCTN} is trained with multimodal data, it only takes in one modality $\mathbf{x}_{1}$ at test-time which makes it robust to the remaining modalities.

\vspace{-2mm}
\subsection{Training Procedures}
\vspace{-2mm}

\textbf{Improving generalization:} Recent work has found that directly training a multimodal model is sub-optimal since different modalities overfit and generalize at different rates. \codes\ includes Gradient Blending (\textsc{GradBlend}), that computes generalization statistics for each modality to determine their weights during fusion~\citep{wang2020makes}, and Regularization by Maximizing Functional Entropies (\textsc{RMFE}), which uses functional entropy to balance the contribution of each modality to the result~\cite{gat2020removing}.

\vspace{-2mm}
\subsection{Putting Everything Together}
\vspace{-2mm}

In Algorithm~\ref{alg:code}, we show a sample code snippet in Python that loads a dataset from \names\ (section~\ref{appendix:dataset_details}), defines the unimodal and multimodal architectures, optimization objective, and training procedures (section~\ref{algorithms}), before running the evaluation protocol (section~\ref{eval}). Our \codes\ toolkit is easy to use and trains entire multimodal models in less than $10$ lines of code. By standardizing the implementation of each module and disentangling the individual effects of models, optimizations, and training, \codes\ ensures both accessibility and reproducibility of its algorithms.

\vspace{-3mm}
\section{Experiments and Discussion}
\vspace{-3mm}

\textbf{Setup:} Using \names, we load each of the datasets and test the multimodal approaches in \codes. We only vary the contributed method of interest and keep all other possibly confounding factors constant (i.e., using the exact same training loop when testing a new multimodal fusion paradigm), a practice unfortunately not consistent in previous work. Our code is available at \dataurl. Please refer to Appendix~\ref{appendix:setup} for experimental details. \names\ allows for careful analysis of multimodal models and we summarize the main take-away messages below (see Appendix~\ref{appendix:results} for full results and analysis).

\begin{table*}[]
\fontsize{9}{11}\selectfont
\setlength\tabcolsep{2.0pt}
\caption{\textbf{Standardizing methods and datasets} enables quick application of methods from different research areas which achieves stronger performance on $9/15$ datasets in \names, especially in healthcare, HCI, robotics, and finance. \textit{In-domain} refers to the best performance across methods previously proposed on that dataset and \textit{out-domain} shows best performance across remaining methods. $\uparrow$ indicates metrics where higher is better (Acc, AUPRC), $\downarrow$ indicates lower is better (MSE).}
\centering
\footnotesize
\vspace{-0mm}
\begin{tabular}{l|c|c|c|c|c}
\Xhline{3\arrayrulewidth}
Dataset & \textsc{MUStARD} $\uparrow$ & \textsc{CMU-MOSI} $\uparrow$ & \textsc{UR-FUNNY} $\uparrow$ & \textsc{CMU-MOSEI} $\uparrow$ & \textsc{MIMIC} $\uparrow$ \\
\Xhline{0.5\arrayrulewidth}
Unimodal & $68.6 \pm 0.4 $ & $74.2 \pm 0.5$ & $58.3 \pm 0.2$ & $78.8 \pm 1.5$ & $76.7 \pm 0.3$ \\
\Xhline{0.5\arrayrulewidth}
In-domain & $66.3 \pm 0.3$ & $\mathbf{83.0 \pm 0.1}$ & $62.9 \pm 0.2$ & $\mathbf{82.1 \pm 0.5}$ & $77.9 \pm 0.3$ \\
Out-domain & $\mathbf{71.8 \pm 0.3}$ & $75.5 \pm 0.5$ & $\mathbf{66.7 \pm 0.3}$ & $ 78.1 \pm 0.3$ & $\mathbf{78.2 \pm 0.2}$ \\
Improvement & \textcolor{gg}{$\mathbf{4.7\%}$} & - & \textcolor{gg}{$\mathbf{6.0\%}$} & - & \textcolor{gg}{$\mathbf{0.4\%}$} \\
\Xhline{3\arrayrulewidth}
\end{tabular}

\vspace{4mm}

\begin{tabular}{l|c|c|c|c|c}
\Xhline{3\arrayrulewidth}
Dataset & \textsc{MuJoCo Push} $\downarrow$ & \textsc{V\&T EE} $\downarrow$ & \textsc{Stocks-F\&B} $\downarrow$ & \textsc{Stocks-Health} $\downarrow$ & \textsc{Stocks-Tech} $\downarrow$ \\
\Xhline{0.5\arrayrulewidth}
Unimodal & $0.334 \pm 0.034$ & $0.202 \pm 0.022$ & $1.856 \pm 0.093$ & $0.541 \pm 0.010$ & $0.125 \pm 0.004$ \\
\Xhline{0.5\arrayrulewidth}
In-domain & $\mathbf{0.290 \pm 0.018}$ & $0.258 \pm 0.011$ & $1.856 \pm 0.093$ & $0.541 \pm 0.010$ & $0.125 \pm 0.004$ \\
Out-domain & $0.402 \pm 0.026$ & $\mathbf{0.185 \pm 0.011}$ & $\mathbf{1.820 \pm 0.138}$ & $\mathbf{0.526 \pm 0.017}$ & $\mathbf{0.120 \pm 0.008}$ \\
Improvement & - & \textcolor{gg}{$\mathbf{8.4\%}$} & \textcolor{gg}{$\mathbf{1.9\%}$} & \textcolor{gg}{$\mathbf{2.8\%}$} & \textcolor{gg}{$\mathbf{4.0\%}$}\\
\Xhline{3\arrayrulewidth}
\end{tabular}

\vspace{4mm}

\begin{tabular}{l|c|c|c|c|c}
\Xhline{3\arrayrulewidth}
Dataset & \textsc{ENRICO} $\uparrow$ & \textsc{MM-IMDb} $\uparrow$ & \textsc{AV-MNIST} $\uparrow$ & \textsc{Kinetics-S} $\uparrow$ & \textsc{Kinetics-L} $\uparrow$ \\
\Xhline{0.5\arrayrulewidth}
Unimodal & $47.0 \pm 1.6$ & $45.6 \pm 4.5$ & $65.1 \pm 0.2$ & $\mathbf{56.5}$ & $72.6$ \\
\Xhline{0.5\arrayrulewidth}
In-domain & $47.0 \pm 1.6$ & $49.8 \pm 1.7$ & $\mathbf{72.8 \pm 0.2}$ & $56.1$ & $\mathbf{74.7}$ \\
Out-domain & $\mathbf{51.0 \pm 1.4}$ & $\mathbf{50.2 \pm 0.9}$ & $72.3 \pm 0.2$ & $23.7$ & $71.7$ \\
Improvement & \textcolor{gg}{$\mathbf{8.5\%}$} & \textcolor{gg}{$\mathbf{0.8\%}$} & - & - & - \\
\Xhline{3\arrayrulewidth}
\end{tabular}

\vspace{-4mm}
\label{results:overall}
\end{table*}

\vspace{-1mm}
\textbf{Benefits of standardization:} From Table~\ref{results:overall}, simply applying methods proposed \textit{outside} of the same research area can improve the state-of-the-art performance on $9$ of the $15$ \names\ datasets, especially for relatively understudied domains and modalities (i.e., healthcare, finance, HCI).

\vspace{-1mm}
\textbf{Generalization across domains and modalities:} \names\ offers an opportunity to analyze algorithmic developments across a large suite of modalities, domains, and tasks. We summarize the following observations regarding performance across datasets and tasks (see details in Appendix~\ref{appendix:performance}):

\vspace{-1mm}
1. Many multimodal methods show their strongest performance on in-domain datasets and do not generalize across domains and modalities. For example, \textsc{MFAS}~\citep{perez2019mfas} works well on domains it was designed for (\textsc{AV-MNIST} and \textsc{MM-IMDb} in multimedia) but does not generalize to other domains such as healthcare (\textsc{MIMIC}). Similarly, \textsc{MulT}~\citep{tsai2019multimodal} performs extremely well on the affect recognition datasets it was designed for but struggles on other multimodal time-series data in the finance and robotics domains. Finally, \textsc{GradBlend}~\citep{wang2020makes}, an approach specifically designed to improve generalization in multimodal learning and tested on video and audio datasets (e.g., Kinetics), does not perform well on other datasets. In general, we observe high variance in the performance of multimodal methods across datasets in \names. Therefore, there still does not exist a one-size-fits-all model, especially for understudied modalities and tasks.

\vspace{-1mm}
2. There are methods that are surprisingly generalizable across datasets. These are typically general modality-agnostic methods such as \textsc{LF}. While simple, it is a strong method that balances simplicity, performance, and low complexity. However, it does not achieve the best performance on any dataset.

\vspace{-1mm}
3. Several methods such as \textsc{MFAS} and \textsc{CCA} are designed for only $2$ modalities (usually image and text), and \textsc{TF} and \textsc{MI} do not scale efficiently beyond $2/3$ modalities. We encourage the community to generalize these approaches across datasets and modalities on \names.

\textbf{Tradeoffs between modalities:} How far can we go with unimodal methods? Surprisingly far! From Table~\ref{results:overall}, we observe that decent performance can be obtained with the best performing modality. Further improvement via multimodal models may come at the expense of around $2-3\times$ the parameters.

\begin{figure*}[tbp]
\centering
    \vspace{-0mm}
    \begin{minipage}{0.4\textwidth}
        \centering
        \subfloat[\centering All datasets]{{\includegraphics[width=\textwidth]{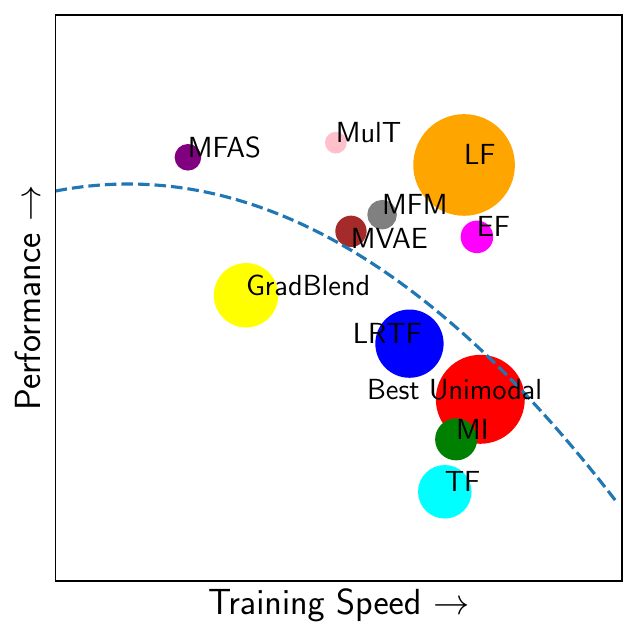}}}
    \end{minipage}%
    \begin{minipage}{0.4\textwidth}
        \centering
        \subfloat[\centering Datasets with $>6$ approaches]{{\includegraphics[width=\textwidth]{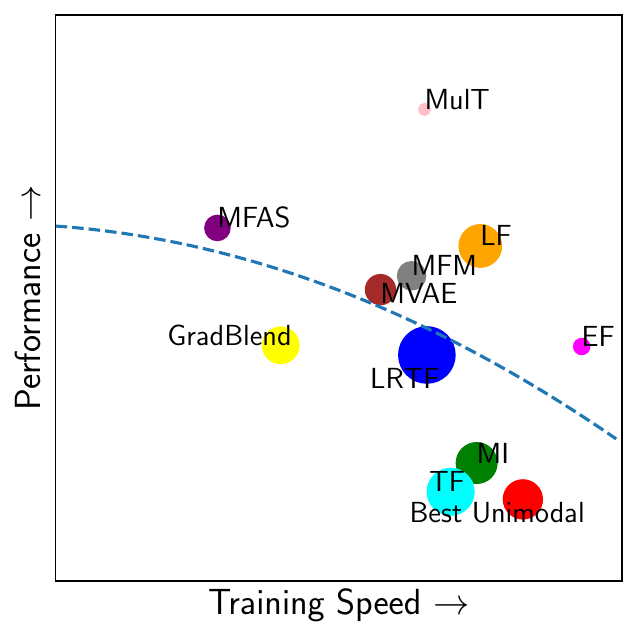}}}
    \end{minipage}
\caption{\textbf{Tradeoff between performance and complexity}. Size of circles shows variance in performance across (a) all datasets and (b) datasets on which we tested $>6$ approaches. We plot a dotted {\color{blue}{blue line}} of best quadratic fit to show the Pareto frontier. These strong tradeoffs should encourage future work in lightweight multimodal models that generalize across datasets, as well as in adapting several possibly well-performing methods (such as \textsc{MFAS} or \textsc{MulT}) to new datasets and domains.\vspace{-2mm}}
\label{figs:tradeoff_complexity}
\end{figure*}

\textbf{Tradeoffs between performance and complexity:} In Figure~\ref{figs:tradeoff_complexity}(a), we summarize the performance of all methods in terms of performance and complexity. We find a strong tradeoff between these two desiderata: simple fusion techniques (e.g., \textsc{LF}) are actually appealing choices which score high on both metrics, especially when compared to complex (but slightly better performing) methods such as architecture search (\textsc{MFAS}) or Multimodal Transformers (\textsc{MulT}). While \textsc{LF} is the easiest to adapt to new datasets and domains, we encountered difficulties in adapting several possibly well-performing methods (such as \textsc{MFAS} or \textsc{MulT}) to new datasets and domains. Therefore, while their average performance is only slightly better than \textsc{LF} on all datasets (see Figure~\ref{figs:tradeoff_complexity}(a)), they perform much better on well-studied datasets (see Figure~\ref{figs:tradeoff_complexity}(b)). We hope that the release of \names\ will greatly accelerate research in adapting complex methods on new datasets (see full results in Appendix~\ref{appendix:results_complexity}).

\begin{figure*}[tbp]
\centering
    \vspace{-0mm}
    \begin{minipage}{0.4\textwidth}
        \centering
        \subfloat[\centering Relative robustness]{{\includegraphics[width=\textwidth]{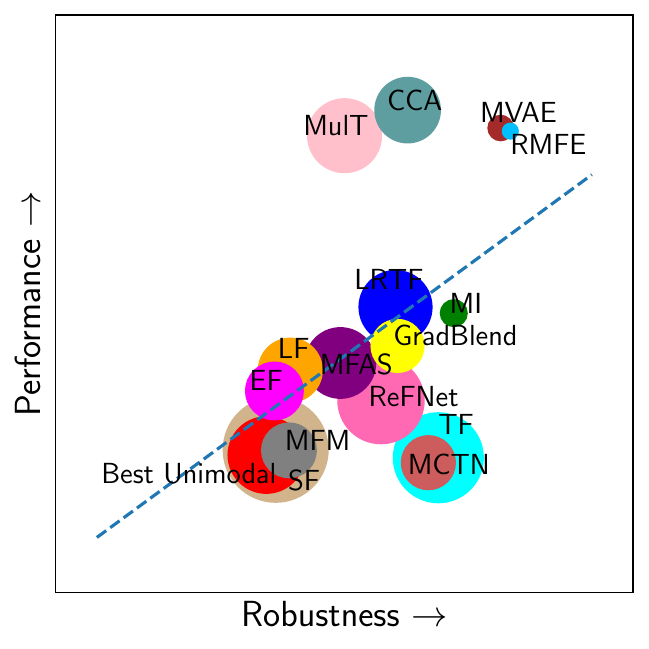}}}
    \end{minipage}%
    \begin{minipage}{0.4\textwidth}
        \centering
        \subfloat[\centering Effective robustness]{{\includegraphics[width=\textwidth]{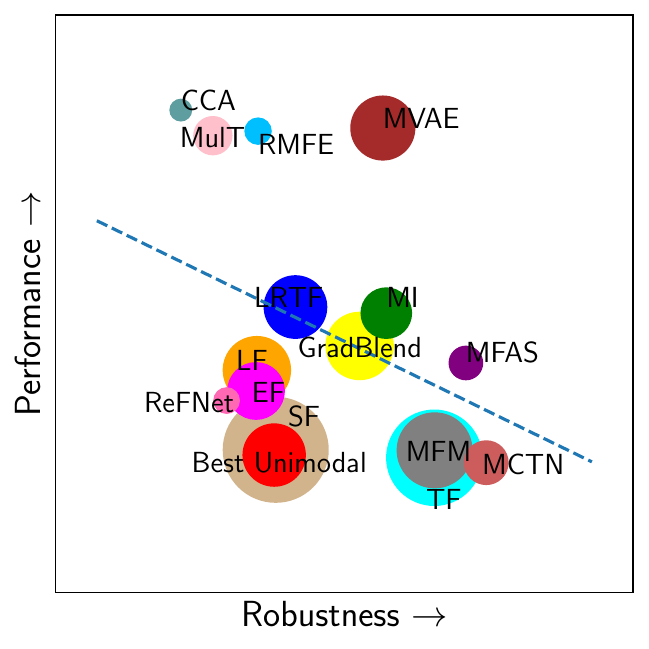}}}
    \end{minipage}
\caption{\textbf{Tradeoff between performance and robustness}. Size of circles shows variance in robustness across datasets. We show the line of best linear fit in dotted {\color{blue}{blue}}. While better performing methods show better \textit{relative} robustness (a), some suffer in \textit{effective} robustness since performance \textit{drops off faster} (b). Few models currently achieve both relative and effective robustness, which suggests directions for future research.\vspace{-2mm}}
\label{figs:tradeoff_robustness}
\end{figure*}

\textbf{Tradeoffs between performance and robustness:} In Figure~\ref{figs:tradeoff_robustness}, we plot a similar tradeoff plot between accuracy and (relative \& effective) robustness. As a reminder, relative robustness directly measures accuracy under imperfections while effective robustness measures the rate at which accuracy drops after equalizing for initial accuracy on clean test data (see Appendix~\ref{appendix:robustness} for details). We observe a positive correlation between performance and relative robustness (see Figure~\ref{figs:tradeoff_robustness}(a)), implying that models starting off with higher accuracy tend to stay above other models on the performance-imperfection curve. However, we observe a negative best fit between performance and effective robustness (see Figure~\ref{figs:tradeoff_robustness}(b)) because several well-performing methods such as \textsc{MulT}, \textsc{CCA}, and \textsc{MVAE} tend to \textit{drop off faster} after equalizing for initial accuracy on clean test data. Furthermore, very few models currently achieve both positive relative and effective robustness, which is a crucial area for future multimodal research (see full results in Appendix~\ref{appendix:results_robustness}).

\vspace{-3mm}
\section{Related Work}
\vspace{-3mm}

We review related work on standardizing datasets and methods in multimodal learning.

\vspace{-1mm}
\textbf{Comparisons with related benchmarks:} To the best of our knowledge, \names\ is the first multimodal benchmark with such a large number of datasets, modalities, and tasks. Most previous multimodal benchmarks have focused on a single research area such as within affective computing~\citep{gkoumas2021makes}, human multimodal language~\citep{mmsdk}, language and vision-based question answering~\citep{ferraro-etal-2015-survey,sharif2020vision}, text classification with external multimodal information~\citep{multimodaltoolkit}, and multimodal learning for education~\citep{datacollection2021}. \names\ is specifically designed to go beyond the commonly studied language, vision, and audio modalities to encourage the research community to explore relatively understudied modalities (e.g., tabular data, time-series, sensors, graph and set data) and build general multimodal methods that can handle a diverse set of modalities.

\vspace{-1mm}
Our work is also inspired by recent progress in better evaluation benchmarks for a suite of important tasks in ML such as language representation learning~\citep{wang2019superglue,wang-etal-2018-glue}, long-range sequence modeling~\citep{tay2020long}, multilingual representation learning~\citep{hu2020xtreme}, graph representation learning~\citep{hu2020open}, and robustness to distribution shift~\citep{koh2020wilds}. These well-crafted benchmarks have accelerated progress in new algorithms, evaluation, and analysis in their respective research areas.

\vspace{-1mm}
\textbf{Standardizing multimodal learning:} There have also been several attempts to build a single model that works well on a suite of multimodal tasks~\citep{li2019visualbert,lu2019vilbert,Su2020VLBERT}. However, these are limited to the language and vision space, and multimodal training is highly tailored for text and images. Transformer architectures have emerged as a popular choice due to their suitability for both language and image data~\citep{chen2020uniter,hu2021transformer} and a recent public toolkit was released for incorporating multimodal data on top of text-based Transformers for prediction tasks~\citep{multimodaltoolkit}. By going beyond Transformers and text data, \names\ opens the door to important research questions involving a much more diverse set of modalities and tasks while holistically evaluating performance, complexity, and robustness.

\vspace{-1mm}
\textbf{Analysis of multimodal representations:} Recent work has begun to carefully analyze and challenge long-standing assumptions in multimodal learning. They have shown that certain models do not actually learn cross-modal interactions but rather rely on ensembles of unimodal statistics~\citep{hessel2020emap} and that certain datasets and models are biased to the most dominant modality~\cite{cadene2019rubi,goyal2017making}, sometimes ignoring others completely~\citep{agrawal-etal-2016-analyzing}. These observations are currently only conducted on specific datasets and models without testing their generalization to others, a shortcoming we hope to solve using \names\ which enables scalable analysis over modalities, tasks, and models.

\vspace{-3mm}
\section{Conclusion}
\vspace{-3mm}

\textbf{Limitations:} While \names\ can help to accelerate research in multimodal ML, we are aware of the following possible limitations (see detailed future directions in Appendix~\ref{appendix:future}):

\vspace{-1mm}
1. \textit{Tradeoffs between generality and specificity:} While it is desirable to build models that work across modalities and tasks, there is undoubtedly merit in building modality and task-specific models that can often utilize domain knowledge to improve performance and interpretability (e.g., see neuro-symbolic VQA~\citep{vedantam2019probabilistic}, or syntax models for the language modality~\citep{cirik2018using}). \names\ is not at odds with research in this direction: in fact, by easing access to data, models, and evaluation, we hope that \names\ will challenge researchers to design interpretable models leveraging domain knowledge for many multimodal tasks. It remains an open question to define ``interpretability'' for other modalities beyond image and text, a question we hope \names\ will drive research in.

\vspace{-1mm}
2. \textit{Scale of datasets, models, and metrics:} We plan for \names\ to be a continuously-growing community effort with regular maintenance and expansion. While \names\ currently does not include several important research areas outside of multimodal fusion (e.g., question answering~\citep{agrawal2017vqa,hannan2020manymodalqa}, retrieval~\citep{zhen2019deep}, grounding~\citep{cirik-etal-2018-visual}, and reinforcement learning~\citep{luketina2019survey}), and is also limited by the models and metrics it supports, we outline our plan to expand in these directions in Appendix~\ref{appendix:future}.

\textbf{Projected expansions of \names:} In this subsection, we describe concrete ongoing and future work towards expanding \names\ (see details in Appendix~\ref{appendix:future}).

\vspace{-1mm}
1. \textit{Other multimodal research problems:} We are genuinely committed to building a community around these resources and continue improving it over time. While we chose to focus on multimodal fusion by design for this first version to have a more coherent way to standardize and evaluate methods across datasets, we acknowledge the breadth of multimodal learning and are looking forward to expanding it in other directions in collaboration with domain experts. We have already included $2$ datasets in captioning (and more generally for non-language outputs, retrieval): (1) Yummly-28K of paired videos and text descriptions of food recipes~\citep{min2016being}, and (2) Clotho dataset for audio-captioning~\citep{drossos2020clotho} as well as a language-guided RL environment Read to Fight Monsters (RTFM)~\citep{zhong2019rtfm} and are also working towards more datasets in QA, retrieval, and multimodal RL.

\vspace{-1mm}
To help in scalable expansion, we plan for an open call to the community for suggestions and feedback about domains, datasets, and metrics. As a step in this direction, we have concrete plans to use \names\ as a theme for future workshops and competitions (building on top of the multimodal workshops we have been organizing at \href[pdfnewwindow=true]{http://multicomp.cs.cmu.edu/naacl2021multimodalworkshop}{NAACL 2021}, \href[pdfnewwindow=true]{http://multicomp.cs.cmu.edu/acl2020multimodalworkshop}{ACL 2020}, and \href[pdfnewwindow=true]{http://multicomp.cs.cmu.edu/acl2018multimodalchallenge}{ACL 2019}, and in multimodal learning courses (starting with the \href[pdfnewwindow=true]{https://cmu-multicomp-lab.github.io/mmml-course/fall2020}{course taught annually at CMU}). Since \names\ is public and will be regularly maintained, the existing benchmark, code, evaluation, and experimental protocols can greatly accelerate any dataset and modeling innovations added in the future. In our public GitHub, we have included a section on contributing through task proposals or additions of datasets and algorithms. The authors will regularly monitor new proposals through this channel.

\vspace{-1mm}
2. \textit{New evaluation metrics:} We also plan to include evaluation for distribution shift, uncertainty estimation, tests for fairness and social biases, as well as labels/metrics for interpretable multimodal learning. In the latter, we plan to include the EMAP score~\citep{hessel2020emap} as an interpretability metric assessing whether cross-modal interactions improve performance.

\vspace{-1mm}
3. \textit{Multimodal transfer learning and co-learning:} Can training data in one dataset help learning on other datasets? \names\ enables easy experimentation of such research questions: our initial experiments on transfer learning found that pre-training on larger datasets in the same domain can improve performance on smaller datasets when fine-tuned on a smaller dataset: performance on the smaller \textsc{CMU-MOSI} dataset improved from $75.2$ to $75.8$ using the same late fusion model with transfer learning from the larger \textsc{UR-FUNNY} and \textsc{CMU-MOSEI} datasets. Furthermore, recent work has shown that multimodal training can help improve unimodal performance as well~\citep{socher2013zero,NIPS2019_8731,zadeh2020foundations}. While previous experiments were on a small scale and limited to a single domain, we plan to expand significantly on this phenomenon (multimodal co-learning) in future versions of \names.

\vspace{-1mm}
4. \textit{Multitask learning across modalities:} Multitask learning across multimodal tasks with a shared set of input modalities is a promising direction that can enable statistical strength sharing across datasets and efficiency in training a single model. Using \names, we also ran an extra experiment on multi-dataset multitask learning. We used the $4$ datasets in the affective computing domain and trained a single model across all $4$ of them with adjustable input embedding layers if the input features were different and separate classification heads for each dataset’s task. We found promising initial results with performance on the largest \textsc{CMU-MOSEI} dataset improving from $79.2$ to $80.9$ for a late fusion model and from $82.1$ to $82.9$ using a multimodal transformer model, although performance on the smaller \textsc{CMU-MOSI} dataset decreased from $75.2$ to $70.8$. We believe that these potential future studies in co-learning, transfer learning, and multi-task learning are strengths of \names\ since it shows the potential of interesting experiments and usage.

\textbf{In conclusion}, we present \names, a large-scale benchmark unifying previously disjoint efforts in multimodal research with a focus on ease of use, accessibility, and reproducibility, thereby paving the way towards a deeper understanding of multimodal models. Through its unprecedented range of research areas, datasets, modalities, tasks, and evaluation metrics, \names\ highlights several future directions in building more generalizable, lightweight, and robust multimodal models.

\vspace{-2mm}
\section*{Acknowledgements}
\vspace{-2mm}

This material is based upon work partially supported by the National Science Foundation (Awards \#1722822 and \#1750439), National Institutes of Health (Awards \#R01MH125740, \#R01MH096951, \#U01MH116925, and \#U01MH116923), BMW of North America, and SquirrelAI. PPL is supported by a Facebook PhD Fellowship and a Center for Machine Learning and Health Fellowship. RS is supported in part by NSF IIS1763562 and ONR Grant N000141812861.
Any opinions, findings, conclusions, or recommendations expressed in this material are those of the author(s) and do not necessarily reflect the views of the National Science Foundation, National Institutes of Health, Facebook, CMLH, Office of Naval Research, BMW of North America, and SquirrelAI, and no official endorsement should be inferred. We are extremely grateful to Amir Zadeh, Chaitanya Ahuja, Volkan Cirik, Murtaza Dalal, Benjamin Eysenbach, Tiffany Min, and Devendra Chaplot for helpful discussions and feedback, as well as Ziyin Liu and Chengfeng Mao for providing tips on working with financial time-series data. Finally, we would also like to acknowledge NVIDIA's GPU support.

\unhidefromtoc

{\small
\bibliography{main}
\bibliographystyle{plain}
}

\clearpage
\appendix

\section*{Appendix}

{
  \hypersetup{linkcolor=black}
  \tableofcontents
}

\clearpage

\vspace{-2mm}
\section{Broader Impact Statement}
\label{appendix:broader_impact}
\vspace{-2mm}

Multimodal data and models are ubiquitous in a range of real-world applications. \names\ and \codes\ is our aim to systematically categorize the plethora of datasets and models currently in use. While these contributions will accelerate research towards multimodal datasets and models as well as their real-world deployment, we believe that special care must be taken in the following regard to ensure that these models are safely deployed for real-world benefit:

\textbf{Time \& space complexity}: Modern multimodal datasets and models are large, especially when building on already large pretrained unimodal datasets and models such as BERT or ResNets. The increasing time and space complexity of these models can cause financial impacts resulting from the cost of hardware, electricity, and computation, as well as environmental impacts resulting from the carbon footprint required to fuel modern hardware. Therefore, there has been much recent interest in building lightweight machine learning models~\citep{strubell2019energy}.

\names\ also provides several efforts in this direction: 
\begin{enumerate}
    \item Firstly, \names\ alleviates the need for separate research groups to repeat preprocessing efforts when beginning to work on a new multimodal dataset, which often takes significant time when large video \& audio datasets and feature extractors are involved.
    \item Secondly, our standardized implementation of core approaches in \codes\ prevents duplicate efforts in adapting approaches to new datasets. We found that while many authors of these multimodal methods released their code publicly on GitHub, there was still some effort needed to adapt their code and tune their models to achieve the best performance on our standardized implementation in \codes. By standardizing these experimentation efforts, we can facilitate the sharing of code and trained models, ensure reproducibility across implementations, and save time and effort in the future.
    \item Finally, \names\ explicitly tests for complexity and encourages researchers to build lightweight models. While this has been less studied in multimodal research, we hope that our efforts will pave the way for greener multimodal learning.
\end{enumerate}

\textbf{Privacy and security:} There may be privacy risks associated with making predictions from multimodal data of recorded human behaviors. The datasets potentially in question might include those in affective computing (recorded video data labeled for sentiment, emotions, and personality attributes), and healthcare (health data labeled for disease and mortality rate). Therefore, it is crucial to obtain user consent before collecting device data. In our experiments with real-world data where people are involved (i.e., healthcare and affective computing), the creators of these datasets have taken the appropriate steps to only access public data which participants/content creators have consented for released to the public (see details in Appendix~\ref{appendix:dataset_details}). We only use these datasets for research purposes. All data was anonymized and stripped of all personal (e.g., personally identifiable information) and protected attributes (e.g., race, gender).

To deploy these algorithms at scale in the real world, it is also important to keep data and features private on each device without sending it to other locations using techniques such as federated learning~\citep{DBLP:journals/corr/abs-1812-06127,liang2020think}, differential privacy~\cite{geyer2017differentially}, or encryption~\cite{dankar2013practicing}.

\textbf{Social biases:} We acknowledge that there is a risk of exposure bias due to imbalanced datasets, especially when human-centric data and possibly sensitive labels are involved. For example, will models trained on imbalanced data disproportionately classify videos of a particular gender as displaying a particular emotion? Models trained on biased data have been shown to amplify the underlying social biases especially when they correlate with the prediction targets~\citep{DBLP:journals/corr/abs-1809-07842}. This leaves room for future work in exploring methods tailored for specific scenarios such as mitigating social biases in words~\citep{bolukbasi2016man}, sentences~\citep{liang2020fair}, images~\citep{10.1145/3209978.3210094}, and other modalities. Future research in multimodal models should also focus on quantifying the trade-offs between fairness and performance~\citep{DBLP:journals/corr/abs-1906-08386}. \names\ enables the large-scale study of these crucial research questions and we outline some of our ongoing and future efforts in expanding the evaluation metrics in \names\ to take these into account in Appendix~\ref{appendix:future}.

\textbf{Possible biases within each dataset:} In this section, we expand upon the previous two points regarding privacy and social biases by describing the possible biases in each domain/dataset included in \names. 
\begin{enumerate}
    \item \textit{Affective computing:} Analysis of sentiment, emotions, and personality might carry biases if care is not taken to appropriately anonymize the video data used. In \names, all models trained on affect recognition datasets use only pre-extracted non-invertible features that rely on general visual or audio features such as the presence of a smile or magnitude of voice. Therefore the features used in this paper cannot be used to identify the speaker~\citep{zadeh2016mosi,zadeh2018multimodal}. Furthermore, videos within the datasets all follow the creative commons license and follow fair use guidelines of YouTube. This license allows is the standard way for content creators to grant someone else permission to use and redistribute their work. We use no information regarding gender, ethnicity, identity, or video identifier in online sources. We emphasize that the models trained to perform automated affect recognition should not in any way be used to harm individuals and should only be used as a scientific study.
    
    In addition to privacy issues, we also studied the videos collected in these affective computing datasets and found no offensive content. While there are clearly expressions of highly negative sentiment or strong displays of anger and disgust, there are no offensive words used or personal attacks recorded in the video. All videos are related to movie or product reviews, TED talks, and TV shows. 
    \item \textit{Healthcare:} The \textsc{MIMIC} dataset~\citep{MIMIC} has been rigorously de-identified in accordance with Health Insurance Portability and Accountability Act (HIPAA) such that all possible personal information has been removed from the dataset. Removed personal information include patient name, telephone number, address, and dates. Dates of birth for patients aged over $89$ were shifted to obscure their true age. Please refer to Appendix~\ref{appendix:health_data} for de-identification details. Again, we emphasize that any multimodal models trained to perform prediction should only be used for scientific study and should not in any way be used for real-world prediction.
    \item \textit{Finance:} There is no personal/human data included and there is no risk of personally identifiable information and offensive content.
    \item \textit{Robotics:} There is no personal/human data included and there is no risk of personally identifiable information and offensive content.
    \item \textit{HCI:} There is no personal/human data included and there is no risk of personally identifiable information and offensive content.
    \item \textit{Multimedia:} For MM-IMDb and AV-MNIST, there is no personal/human data included and there is no risk of personally identifiable information and offensive content. For Kinetics, all videos within the dataset are obtained from public YouTube videos that follow the creative commons license which allows content creators to grant permission to use and redistribute their work. We use no information regarding gender, ethnicity, identity, or video identifier in online sources. We emphasize that the models trained to perform action recognition should not in any way be used to harm individuals and should only be used as a scientific study.
    
    We also checked to make sure that these videos do not contain offensive content. All videos are related to human actions and do not contain any offensive words/audio.
\end{enumerate}

Overall, \names\ offers opportunities to study these potential issues at scale across modalities, tasks, datasets, and domains. We plan to continue expanding this benchmark to rigorously test for these social impacts to improve the safety and reliability of multimodal models. For example, in Appendix~\ref{appendix:fairness}, we describe some concrete extensions to include evaluations for fairness and privacy of multimodal models trained on the datasets in \names. Our holistic evaluation metrics will also encourage the research community to quantify the tradeoffs between performance, complexity, robustness, fairness, and privacy in human-centric multimodal models.

\clearpage

\vspace{-2mm}
\section{Background: Multimodal Representation Learning}
\label{appendix:background}
\vspace{-2mm}

We first provide background focusing on multimodal representation learning and several core technical challenges in this area.

\vspace{-2mm}
\subsection{Problem Statement}
\vspace{-2mm}

We define a modality as a single particular mode in which a signal is expressed or experienced. Multiple modalities then refer to a combination of multiple signals each expressed or experienced in heterogeneous manners~\citep{baltruvsaitis2018multimodal}. We distinguish between the possible \textit{temporal resolution} of modalities that will impact the types of approaches used:

\vspace{-1mm}
1. \textit{Static} modalities include inputs without a time dimension such as images, tabular data (i.e., a table of numerical data).

\vspace{-1mm}
2. \textit{Temporal} modalities include those coming in a sequence with a time-dimension such as language (a sequence of tokens), video (a sequence of frames/audio features/optical flow features), or time-series data (a sequence of data points indexed by time).

The first version of \names\ focuses on benchmarks and algorithms for \textit{multimodal fusion}, where the main challenge is to join information from two or more modalities to perform a prediction. Classic examples include audio-visual speech recognition where visual lip motion is fused with speech signals to predict spoken words~\citep{dupont2000audio}. Note that in fusion problems, it should be well-defined to predict the label with a single modality only, which marks an important distinction to tasks in question answering and grounding where one modality is used to query information in another (e.g., visual question answering~\citep{agrawal2017vqa} using a text question to query information in the image). We outline our plans to extend future versions of \names\ to include more multimodal challenges such as question answering, retrieval, and grounding in Appendix~\ref{appendix:future}.

Formally, the multimodal fusion problem is defined as follows. We suppose there is a set of $M$ modalities drawn from a joint distribution $p(X_1, ..., X_M, Y)$ where $X_m$ is a random variable denoting data distributed according to modality $m$ and $Y$ is a random variable representing the label. If modality $m$ is a static modality, $X_m$ is a random vector without the time dimension. If modality $m$ is a temporal modality, $X_m$ is a random vector with a time dimension which can be represented as follows: $X_m = (X_m^1, X_m^2, ..., X_m^T)$ where $T$ is the number of time-steps in the temporal modality.

In multimodal fusion, a set of $M$ modalities is drawn from a joint distribution $p(X_1, ..., X_M, Y)$ where $X_m$ is a random variable denoting data distributed according to modality $m$ and $Y$ is a random variable representing the label. A multimodal dataset is a collection of draws of (data, label) pairs from the joint distribution $p(X_1, ..., X_M, Y)$. We denote a dataset as $\{ (\mathbf{x}_{i1}, ..., \mathbf{x}_{iM}, y_i) \}_{i=1}^n$. These draws from the true distribution are possibly biased (e.g., across individuals, topics, or labels) and noisy (e.g., due to noisy or missing modalities). A multimodal model is a set of functions $\{f_m : m \in [M] \} \cup f_{\textrm{mm}}$ where each of the $f_m$'s are unimodal encoders, one for each modality, and $f_{\textrm{mm}}$ is a multimodal fusion network. The unimodal encoders are specially designed with domain knowledge to learn representations from each modality (e.g., convolutional networks for images, temporal models for time-series data) resulting in unimodal representations $\mathbf{z}_1, ..., \mathbf{z}_M$. The multimodal network is designed to capture information across unimodal representations and summarize it in a multimodal representation $\mathbf{z}_\textrm{mm}$ that can be used to predict the label $y$. The goal of multimodal fusion is to learn a model with the lowest prediction error as measured on a held-out test set, while also balancing other potential objectives such as low complexity and robustness to imperfect data.

\vspace{-1mm}
\subsection{Technical Challenges}
\label{appendix:challenges}
\vspace{-1mm}

\names\ tests for the following holistic desiderata in multimodal fusion:

\begin{enumerate}
    \item \textit{Performance:} We summarize the following core challenges across all prediction tasks for multimodal learning with reference to Baltrusaitis et al.,~\citep{baltruvsaitis2018multimodal}. Solving these challenges is essential in any multimodal prediction problem, regardless of domain and task.
    \begin{enumerate}
        \item \textit{Unimodal structure and granularity}: The information coming from each modality follows a certain underlying structure and invariance, which needs to be processed by suitable unimodal encoders. While there are certain generally adopted unimodal encoders for commonly studied modalities such as images and text, there remain challenges in designing unimodal encoders with the right types of inductive biases for other less-studied modalities such as tabular and time-series data. Representations extracted from unimodal encoders should contain task-relevant information from that modality, expressed at the right granularity.
        \item \textit{Multimodal complementarity}: The information coming from different modalities have varying predictive power by themselves and also when complemented by each other. We refer to these as \textit{higher-order interactions}: first-order interactions define a predictive signal from a single granular unit of information in one modality to the label (e.g., the presence of a smile indicating positive sentiment); second-order interactions define a predictive signal from a pair of granular units of information across two modalities to the label (e.g., the presence of an eye-roll together with a positive word indicating sarcasm); and $n$th-order interactions extend the above definition to $n$ modalities. There are many possible interactions that explain the labels in a dataset, out of which only some may generalize to unseen data. It remains a challenge to discover these higher-order interactions using suitably expressive models. At the same time, the space of possible interactions is too large which requires suitable inductive biases in model design (see challenges regarding complexity in model design below).
        \item \textit{Multimodal alignment}: Information from different modalities often comes in different granularities. In order to learn predictive signals from higher-order interactions, there is a need to first identify the relations between granular units from two or more different modalities. This challenge requires a measure of the relationship between different modalities, which we call \textit{cross-modal alignment}.
    
        When dealing with temporal data, it also requires capturing possible long-range dependencies across time, which we call \textit{temporal alignment}. For example, it requires aligning the presence of an eye-roll together with a positive word to recognize sarcasm even when both signals happen at different times. This challenge extends cross-modal alignment to the temporal dimension.
    \end{enumerate}

    \item \textit{Complexity:} The space of possible interactions is very large which requires suitable inductive biases in model design. While more expressive models may perform better, these often come at the cost of time and space complexity during training and inference. To enable real-world deployment of multimodal models in a variety of settings~\citep{strubell2019energy}, there is a need to build lightweight models with cheap training and inference.

    \item \textit{Robustness:} Information from different modalities often display different noise topologies, and real-world multimodal signals possibly suffer from missing or noisy data in at least one of the modalities~\cite{baltruvsaitis2018multimodal}. While most methods are trained on carefully curated and cleaned datasets, there is a need to benchmark their robustness in realistic scenarios. The core challenge here is to build models that still perform well despite the presence of unimodal-specific or multimodal imperfections.
\end{enumerate}

\clearpage

\vspace{-2mm}
\section{\names\ Datasets}
\label{appendix:data}
\vspace{-2mm}

\begin{figure*}[]
\centering
\vspace{-0mm}
\includegraphics[width=0.9\linewidth]{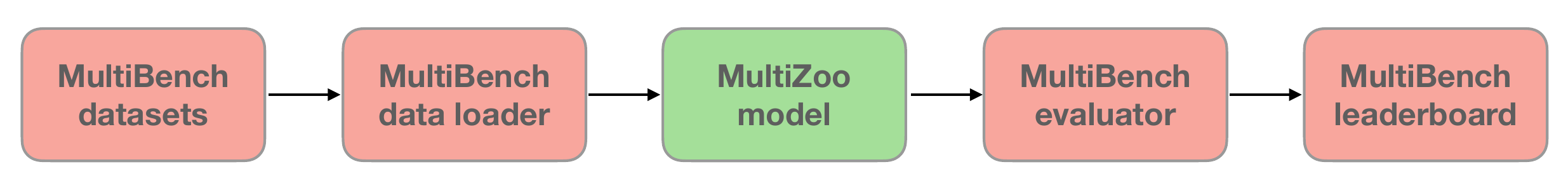}
\vspace{-0mm}
\caption{\names\ provides a standardized machine learning pipeline across data processing, data loading, multimodal models, evaluation metrics, and a public leaderboard to encourage future research in multimodal representation learning. \names\ aims to present a milestone in unifying disjoint efforts in multimodal machine learning research and paves the way towards a better understanding of the capabilities and limitations of multimodal models, all the while ensuring ease of use, accessibility, and reproducibility.\vspace{-0mm}}
\label{figs:pipeline}
\end{figure*}

\names\ provides a standardized machine learning pipeline that starts from data loading to running multimodal models, providing evaluation metrics, and a public leaderboard to encourage future research in multimodal representation learning (see Figure~\ref{figs:pipeline}). 

In this section, we provide additional details on the distribution, release, and maintenance of each of the datasets in \names\ as well as the maintenance of \names\ as a whole.

\vspace{-1mm}
\subsection{Dataset Selection}
\vspace{-1mm}

In this section, we discuss our choices of datasets in \names. We select each dataset based on its data collection method, input modalities, evaluation tasks, evaluation metric, and train/test splits that reflect real-world multimodal applications. We consulted with domain experts in each of the application areas to select datasets that satisfy the following properties:
\begin{enumerate}
    \item \textit{Realism in data collection, input modalities, preprocessing, and task:} Each of the datasets in \names\ reflect a subset of real-world sensory modalities collected in the wild. Realism is important since it brings natural noise topologies in each modality and in the prediction task. It is crucial that these datasets reflect real-world data such that capturing these imperfections through machine learning models can potentially bridge the gap towards real-world deployment.
    
    \item \textit{Diversity in research area:} We chose these research areas through a survey of recent research papers in multimodal learning across conferences in machine learning and beyond (e.g., HCI, NLP, vision, robotics conferences). Furthermore, we consulted with domain experts in applying multimodal learning to their respective application areas to determine areas of large potential. Through engaging with domain experts we were able to select research areas and datasets that reflected realism in data collection, input modalities, preprocessing, and tasks which present challenges for machine learning models and potential for real-world transfer of learned algorithms. These research areas that are designed to span both human-centric and data-centric machine learning. In the former, we selected HCI, healthcare, and robotics since these are fast-growing research areas with increasingly specialized tracks in machine learning conferences dedicated to them. In the latter, financial data analysis is an area with an inherently low signal-to-noise ratio reflecting extremely noisy, imperfect, and uncertain real-world datasets which provide challenges for current multimodal models. We also included several multimedia datasets due to the large resources publicly available on the internet which results in multimodal datasets of the largest scale.
    
    \item \textit{Diversity in modalities:} We started with a set of commonly studied modalities such as language, image, video, and audio. For each of the following research areas, we consulted with domain experts to choose datasets that are established, but not overstudied. More importantly, we aimed for diversity in modalities to truly test the generalization capabilities of modern multimodal models outside of commonly studied domains and modalities. For example, while there is much work in HCI involving images and text, we chose a modality representing a set of mobile applications for coverage. Similarly, in robotics, we consulted with domain experts to obtain datasets with high-frequency force and proprioception sensors that provide a unique challenge to machine learning researchers.
    
    \item \textit{Challenging for ML models:} We aim to choose datasets where the current state-of-the-art performance via machine learning models is still far from human performance (if human performance is provided, otherwise judged by a domain expert). This is to ensure that there is room for improvement through community involvement in this research area.
    
    \item \textit{Community expansion:} Finally, we would like to emphasize that we heavily encourage and actively seek out community participation in expanding \names\ to keep up with the incredible pace in multimodal machine learning research. We describe our plans for an open call for proposals of new research areas, datasets, and prediction tasks in section~\ref{appendix:future}.
\end{enumerate}

\vspace{-1mm}
\subsection{Dataset Details}
\label{appendix:dataset_details}
\vspace{-1mm}

We provide details for each of the research areas and datasets selected in \names. In each of the categories, we describe the research area, the datasets and their associated data collection process, their access restrictions and licenses, and any data preprocessing or feature extraction we used following current work in each of these domains.

\begin{figure*}[]
\centering
\vspace{-0mm}
\includegraphics[width=\linewidth]{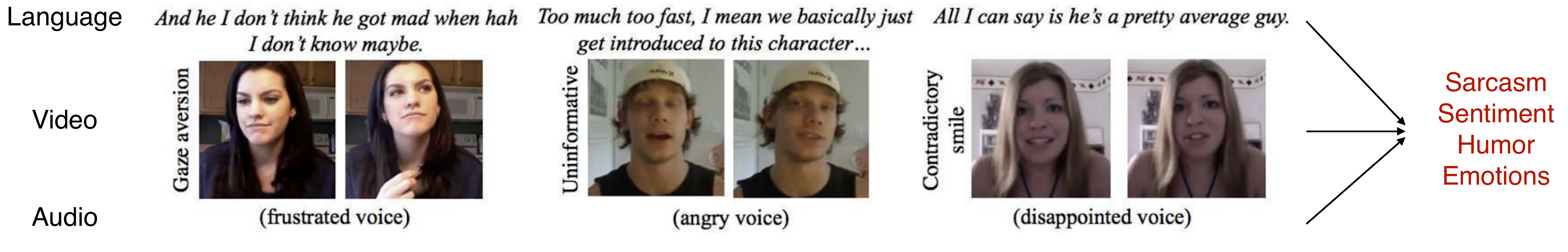}
\vspace{-0mm}
\caption{\textbf{Affective computing} studies the perception of human affective states (emotions, sentiment, and personalities) from our natural display of multimodal signals spanning language (spoken words), visual (facial expressions, gestures), and acoustic (prosody, speech tone)~\citep{picard2000affective}. \names\ contains $4$ datasets in this category involving fusing \textit{language}, \textit{video}, and \textit{audio} time-series data to predict sentiment (\textsc{CMU-MOSI}~\citep{zadeh2016mosi} and \textsc{CMU-MOSEI}~\citep{zadeh2018multimodal}), emotions (\textsc{CMU-MOSEI}~\citep{zadeh2018multimodal}), humor (\textsc{UR-FUNNY}~\cite{hasan2019ur}), and sarcasm (\textsc{MUStARD}~\citep{castro2019towards}).\vspace{-0mm}}
\label{figs:affect}
\end{figure*}

\vspace{-1mm}
\subsubsection{Affective Computing}
\vspace{-1mm}

\textbf{1. \textsc{MUStARD}} is a multimodal video corpus for research in automated sarcasm discovery~\citep{castro2019towards}. The dataset is compiled from popular TV shows including Friends, The Golden Girls, The Big Bang Theory, and Sarcasmaholics Anonymous. \textsc{MUStARD} consists of audiovisual utterances annotated with sarcasm labels. Each utterance is accompanied by its context, which provides additional information on the scenario where the utterance occurs, thereby providing a further challenge in the long-range modeling of multimodal information. Sarcasm is specifically chosen as an annotation task since it requires careful modeling of complementary information, particularly when the semantic information from each modality \textit{do not} agree with each other. 

\textbf{Data collection:} According to Castro et al.,~\citep{castro2019towards}, they conducted web searches on YouTube using keywords such as Friends sarcasm, Chandler sarcasm, Sarcasm 101, and Sarcasm in TV shows to obtain videos with sarcastic content from three main TV shows: Friends, The Golden Girls, and Sarcasmaholics Anonymous. To obtain non-sarcastic videos, they used a subset of $400$ videos from MELD, a multimodal emotion recognition dataset derived from the Friends TV series~\citep{poria-etal-2019-meld}. Videos from The Big Bang Theory were also collected by segmenting episodes using laughter cues from its audience.

\textbf{Access restrictions:} While we do not have the license to this dataset, it is a public dataset free to download by the research community from \url{https://github.com/soujanyaporia/MUStARD}.

\textbf{Licenses:} MIT, see \url{https://github.com/soujanyaporia/MUStARD/blob/master/LICENSE}

\textbf{Dataset preprocessing:} We followed these preprocessing steps for each modality as suggested in the original paper~\citep{castro2019towards}:
\begin{enumerate}
    \item \textit{Language:} Textual utterances are represented using pretrained BERT representations ~\citep{devlin2019bert} as well as Common Crawl pre-trained 300-dimensional GloVe word vectors~\citep{pennington2014glove} for each token.
    \item \textit{Visual:} Visual features are extracted for each frame using a pool5 layer of an ImageNet~\cite{deng2009imagenet} pretrained ResNet-152~\citep{he2016resnet} model. Every frame is first preprocessed by resizing, center-cropping, and normalizing it. We also use the OpenFace facial behavioral analysis tool~\citep{baltruvsaitis2016openface} to extract facial expression features.
    \item \textit{Audio:} Low-level features from the audio data stream are extracted using the speech processing library Librosa~\citep{mcfee2015librosa}. We also extract COVAREP~\cite{degottex2014covarep} features as is commonly used for the other datasets in the affective computing domain (see below).
\end{enumerate}

\textbf{Train, validation, and test splits:} there are $414$, $138$, and $138$ video segments in train, valid, and test data respectively, which gives a total of $690$ data points.

\textbf{2. \textsc{CMU-MOSI}} is a collection of $2,199$ opinion video clips each rigorously annotated with labels for subjectivity, sentiment intensity, per-frame, and per-opinion annotated visual features, and per-milliseconds annotated audio features~\cite{zadeh2016mosi}. Sentiment intensity is annotated in the range $[-3,+3]$ which enables fine-grained prediction of sentiment beyond the classical positive/negative split. Each video is collected from YouTube with a focus on video blogs, or vlogs which reflect the real-world distribution of speakers expressing their behaviors through monologue videos. \textsc{CMU-MOSI} is a realistic real-world multimodal dataset for affect recognition and is regularly used in competitions and workshops.

\textbf{Data collection:} According to Zadeh et al.,~\cite{zadeh2016mosi}, videos were collected from YouTube with a focus on video blogs indexed by \#vlog. A total of $93$ videos were randomly selected. The final set of videos contained $89$ distinct speakers, including $41$ female and $48$ male speakers. Most of the speakers were approximately between the ages of $20$ and $30$ from different backgrounds (e.g., Caucasian, African-American, Hispanic, Asian). All speakers expressed themselves in English and the videos originated from either the United States of America or the United Kingdom.

\textbf{Access restrictions:} The authors are part of the team who collected the \textsc{CMU-MOSI} dataset~\cite{zadeh2016mosi} so we have the license and right to redistribute this dataset. \textsc{CMU-MOSI} was originally downloaded from \url{https://github.com/A2Zadeh/CMU-MultimodalSDK}.

\textbf{Licenses:} Permission is hereby granted, free of charge, to any person obtaining a copy of this software and associated documentation files (the "Software"), to deal in the Software without restriction, including without limitation the rights to use, copy, modify, merge, publish, distribute, sublicense, and/or sell copies of the Software, and to permit persons to whom the Software is furnished to do so, subject to the conditions in \url{https://raw.githubusercontent.com/A2Zadeh/CMU-MultimodalSDK/master/LICENSE.txt}

\textbf{Train, validation, and test splits:} Each dataset contains several videos, and each video is further split into short segments (roughly $10-20$ seconds) that are annotated. We split the data at the level of videos so that segments from the same video will not appear across train, valid, and test splits. This enables us to train user-independent models instead of having a model potentially memorizing the average affective state of a user. There are $52$, $10$, and $31$ videos in train, valid, and test data respectively. Splitting up these videos gives a total of $1,284$, $229$, and $686$ segments respectively for a total of $2,199$ data points. 

\textbf{Dataset preprocessing:} We follow current work~\citep{liang2018computational,zadeh2018multimodal} and apply the following preliminary feature extraction for the \textsc{CMU-MOSI} dataset:
\begin{enumerate}
    \item \textit{Language:} Glove word embeddings~\cite{pennington2014glove} were used to embed a sequence of individual words from video segment transcripts into a sequence of word vectors that represent spoken text. The Glove word embeddings used are 300-dimensional word embedding trained on 840 billion tokens from the common crawl dataset, resulting in a sequence of dimension $T \times 300$ after alignment. The timing of word utterances is extracted using P2FA forced aligner~\cite{P2FA}. This extraction enables alignment between text, audio, and video.
    \item \textit{Visual:} We use the library Facet~\cite{emotient} to extract a set of visual features including facial action units, facial landmarks, head pose, gaze tracking, and HOG features. These visual features are extracted from the full video segment at 30Hz to form a sequence of facial gesture changes throughout time, resulting in a sequence of dimension $T \times 35$. In addition to Facet, OpenFace facial behavioral analysis tool~\citep{baltruvsaitis2016openface} is used to extract the facial expression features which include facial Action Units (AU) based on the Facial Action Coding System (FACS)~\citep{ekman1997universal}.
    \item \textit{Audio:} The software COVAREP~\cite{degottex2014covarep} is used to extract acoustic features including 12 Mel-frequency cepstral coefficients, pitch tracking and voiced/unvoiced segment features~\cite{drugman2011joint}, glottal source parameters~\cite{childers1991vocal}, peak slope parameters and maxima dispersion quotients~\cite{kane2013wavelet}. These visual features are extracted from the full audio clip of each segment at $100$Hz to form a sequence that represents variations in tone of voice over an audio segment, resulting in a sequence of dimension $T \times 74$.
\end{enumerate}

\textbf{3. \textsc{UR-FUNNY}} is the first large-scale multimodal dataset of humor detection in human speech~\citep{hasan2019ur}. UR-FUNNY is a realistic representation of multimodal language (including text, visual and acoustic modalities). This dataset opens the door to understanding and modeling humor in a multimodal framework, which is crucial since humor is an inherently multimodal communicative tool involving the effective use of words (text), accompanying gestures (visual), and prosodic cues (acoustic). UR-FUNNY consists of more than $16,000$ video samples from TED talks which are among the most diverse idea-sharing channels covering speakers from various backgrounds, ethnic groups, and cultures discussing a variety of topics from discoveries in science and arts to motivational speeches and everyday events. The diversity of speakers, topics, and unique annotation targets make it a realistic dataset for multimodal language modeling.

\textbf{Data collection:} According to Hasan et al.,~\citep{hasan2019ur} $1,866$ videos and their transcripts in English were collected from the TED portal, chosen from $1,741$ different speakers and across $417$ topics. The laughter markup is used to filter out $8257$ humorous punchlines from the transcripts. The context is extracted from the prior sentences to the punchline (until the previous humor instances or the beginning of the video is reached). Using a similar approach, $8,257$ negative samples are chosen at random intervals where the last sentence is not immediately followed by a laughter marker. After this negative sampling, there is a homogeneous $50\%$ split in the dataset between positive and negative humor examples.

\textbf{Access restrictions:} This is a public dataset free to download by the research community from \url{https://github.com/ROC-HCI/UR-FUNNY}. The authors of the dataset also note that videos on \url{www.ted.com} are publicly available for download~\citep{hasan2019ur}.

\textbf{Licenses:} No license was provided with this dataset.

\textbf{Dataset preprocessing:} We follow current work~\citep{liang2018computational,zadeh2018multimodal} and apply the same preliminary feature extraction as the \textsc{CMU-MOSI} dataset described above.

\textbf{Train, validation, and test splits:} Each dataset contains several videos, and each video is further split into short segments (roughly $10-20$ seconds) that are annotated. We split the data at the level of videos so that segments from the same video will not appear across train, valid, and test splits. This enables us to train user-independent models instead of having a model potentially memorizing the average affective state of a user. There are $1,166$, $300$, and $400$ videos in train, valid, and test data respectively. Splitting up these videos gives a total of $10,598$, $2,626$, and $3,290$ segments respectively for a total of $16,514$ data points,

\textbf{4. \textsc{CMU-MOSEI}} is the largest dataset of sentence-level sentiment analysis and emotion recognition in real-world online videos~\citep{liang2018multimodal,zadeh2018multimodal}. \textsc{CMU-MOSEI} contains more than $65$ hours of annotated video from more than $1,000$ speakers and $250$ topics. Each video is annotated for sentiment as well as the presence of 9 discrete emotions (angry, excited, fear, sad, surprised, frustrated, happy, disappointed, and neutral) as well as continuous emotions (valence, arousal, and dominance). The diversity of prediction tasks makes \textsc{CMU-MOSEI} a valuable dataset to test multimodal models across a range of real-world affective computing tasks. The dataset has been continuously used in workshops and competitions revolving around human multimodal language.

\textbf{Data collection:} According to Zadeh et al.,~\citep{zadeh2018multimodal}, videos from YouTube are automatically analyzed for the presence of one speaker in the frame using face detection to ensure the video is a monologue and rejecting videos that have moving cameras. A diverse set of $250$ frequently used topics in online videos is used as the seed for acquisition. The authors restrict the number of videos acquired from each channel to a maximum of $10$ and limit the videos to have manual and properly punctuated transcriptions. After manual quality inspection, they also performed automatic checks on the quality of video and transcript using facial feature extraction confidence and forced alignment confidence, before balancing the gender in the dataset using the data provided by annotators ($57\%$ male to $43\%$ female). 

\textbf{Access restrictions:} The authors are part of the team who collected the \textsc{CMU-MOSEI} dataset~\cite{zadeh2018multimodal} so we have the license and right to redistribute this dataset. \textsc{CMU-MOSEI} was originally downloaded from \url{https://github.com/A2Zadeh/CMU-MultimodalSDK}.

\textbf{Licenses:} Permission is hereby granted, free of charge, to any person obtaining a copy of this software and associated documentation files (the``"Software''), to deal in the Software without restriction, including without limitation the rights to use, copy, modify, merge, publish, distribute, sublicense, and/or sell copies of the Software, and to permit persons to whom the Software is furnished to do so, subject to the conditions in \url{https://raw.githubusercontent.com/A2Zadeh/CMU-MultimodalSDK/master/LICENSE.txt}

\textbf{Dataset preprocessing:} We follow current work~\citep{liang2018computational,zadeh2018multimodal} and apply the same preliminary feature extraction as the \textsc{CMU-MOSI} and \textsc{UR-FUNNY} datasets described above.

\textbf{Train, validation, and test splits:} Each dataset contains several videos, and each video is further split into short segments (roughly $10-20$ seconds) that are annotated. We split the data at the level of videos so that segments from the same video will not appear across train, valid, and test splits. This enables us to train user-independent models instead of having a model potentially memorizing the average affective state of a user. There are a total of $16,265$, $1,869$, and $4,643$ segments in train, valid, and test datasets respectively for a total of $22,777$ data points.

\vspace{-1mm}
\subsubsection{Healthcare}
\label{appendix:health_data}
\vspace{-1mm}

\begin{figure*}[]
\centering
\vspace{-0mm}
\includegraphics[width=\linewidth]{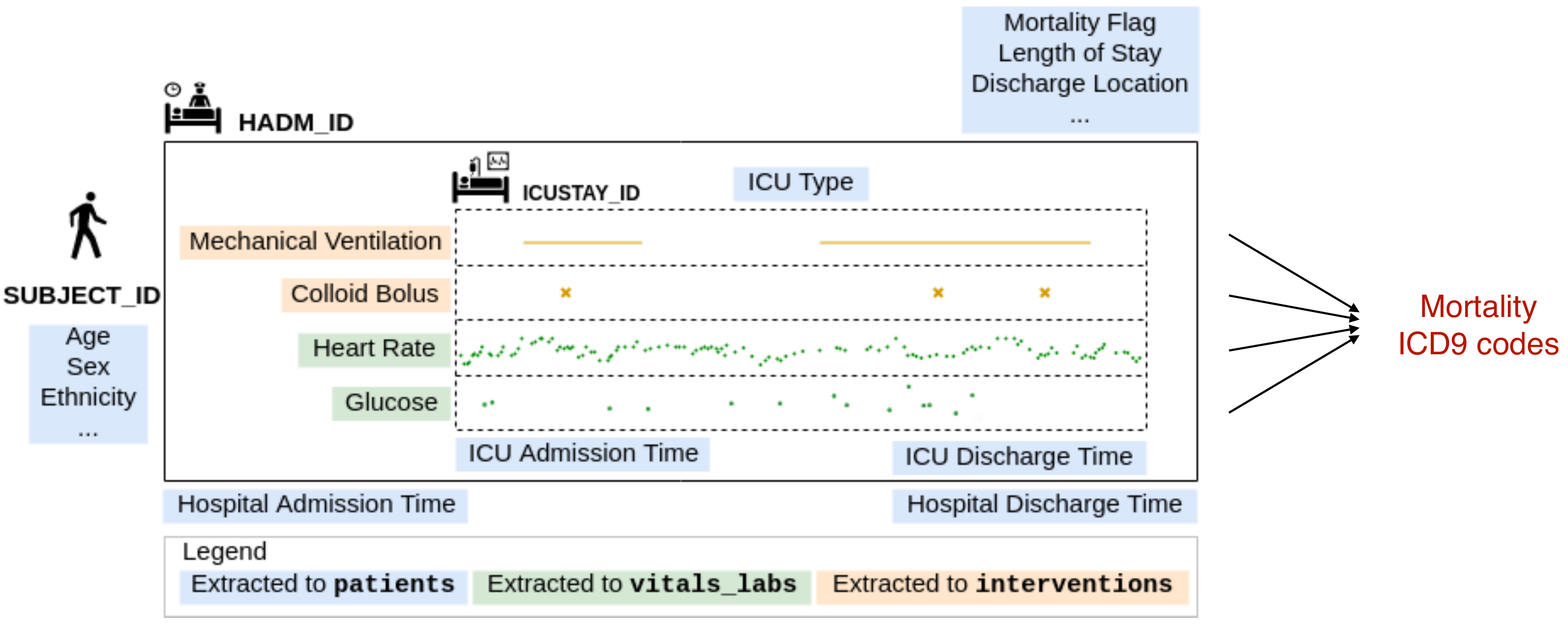}
\vspace{-0mm}
\caption{\textbf{Healthcare:} Medical decision-making often involves integrating complementary signals from several sources such as lab tests, imaging reports, and patient-doctor conversations. Multimodal models can help doctors make sense of high-dimensional data and assist them in the diagnosis process~\cite{amisha2019overview}. \names\ includes the \textsc{MIMIC} dataset~\citep{MIMIC} which records ICU patient data including \textit{time-series} data measured every hour and other \textit{tabular numerical} data about the patient (e.g., age, gender, ethnicity) to predict mortality rate and the disease ICD-$9$-code. Figure adapted from~\citep{wang2020mimic}.\vspace{-0mm}}
\label{figs:healthcare}
\end{figure*}

\textbf{1. \textsc{MIMIC-III}} (Medical Information Mart for Intensive Care III)~\citep{MIMIC} is a large, freely-available database comprising de-identified health-related data associated with over $40,000$ patients who stayed in critical care units of the Beth Israel Deaconess Medical Center between $2001$ and $2012$. Following~\cite{PURUSHOTHAM2018112}, we organized numerous patient data into two major modalities (using the $17$ features in feature set A in~\citep{PURUSHOTHAM2018112}): time series modality, which is a set of medical measurements of the patient taken every $1$ hour in a period of $24$ hours where each measurement is a vector of size $12$ ($12$ different measured numerical values); static modality, which is a set of medical information about the patient, represented in a vector of size $5$. We use these modalities for $3$ tasks: mortality prediction ($6$-class prediction on whether the patient dies in $1$ day, $2$ day, $3$ day, $1$ week, $1$ year, or longer than $1$ year), and $2$ ICD-$9$ code predictions (binary classification on whether the patient fits any ICD-$9$ code in group 1 ($140-239$) and binary classification on whether the patient fits any ICD-$9$ code in group 7 $460-519$).

\textbf{Data collection:} According to Johnson et al.,~\citep{MIMIC}, \textsc{MIMIC} contains data associated with $53,423$ distinct hospital admissions for adult patients (aged $16$ years or above) admitted to critical care units between $2001$ and $2012$, as well as $7,870$ neonates admitted between $2001$ and $2008$. The data covers $38,597$ distinct adult patients and $49,785$ hospital admissions. Data was also downloaded from several sources, including archives from critical care information systems, hospital electronic health record databases, and Social Security Administration Death Master File.

\textbf{Privacy:} Before data was incorporated into the \textsc{MIMIC}-III database, it was first de-identified in accordance with Health Insurance Portability and Accountability Act (HIPAA) standards using structured data cleansing and date shifting. The de-identification process removed all eighteen identifying data elements listed in HIPAA, such as patient name, date of birth (for patients over $89$ of age), telephone number, address, and dates. Protected health information was also removed from text fields, such as diagnostic reports and physician notes. We refer the reader to~\citep{PURUSHOTHAM2018112} for full de-identification details.

\textbf{Access restrictions:} We do not have the license and right to redistribute this dataset. Accessing \textsc{MIMIC} requires the completion of a training course and approval for access on PhysioNet (\url{https://physionet.org/about/database/}). However, we provide our own data preprocessing scripts for \textsc{MIMIC}, which transform the raw data into the standardized format for multimodal data and perform standardized splitting into the train, validation, and test splits. For a new user getting started with \textsc{MIMIC} data, all they would need to do is to complete the training course and obtain approval of access for scientific research from PhysioNet before they can use our public code to load all extracted features from the raw dataset in a version that can directly be used for machine learning studies. 

\textbf{Licenses:} MIT,  see \url{https://github.com/mit-lcp/mimic-code/blob/main/LICENSE}

\textbf{Dataset preprocessing:} We followed the instructions on \url{https://mimic.physionet.org/gettingstarted/access/} to download the dataset in the form of raw tables, then generated preprocessed data following the steps described in \url{https://github.com/USC-Melady/Benchmarking\_DL\_MIMICIII} (which takes $1-2$ weeks running time) to get the data used for experiments. Specifically, we will use data in the file \texttt{24hrs/series/imputed-normed-ep\_1\_24-stdized.npz}. When accessing this data from our code repo, set the \texttt{imputed\_path} of the npz file above in the \texttt{get\_data.py} and the script will generate the PyTorch data loader for the tasks (where we will normalize the data).

\textbf{Train, validation, and test splits:}  We split the data into train/valid/test sets randomly (using a fixed random seed) in a $80:10:10$ ratio (so $28,970$ train, $3,621$ valid, and $3,621$ test data points) for a total of $36,212$ data points.

\vspace{-1mm}
\subsubsection{Robotics}
\vspace{-1mm}

\begin{figure*}[]
\centering
\vspace{-0mm}
\includegraphics[width=0.7\linewidth]{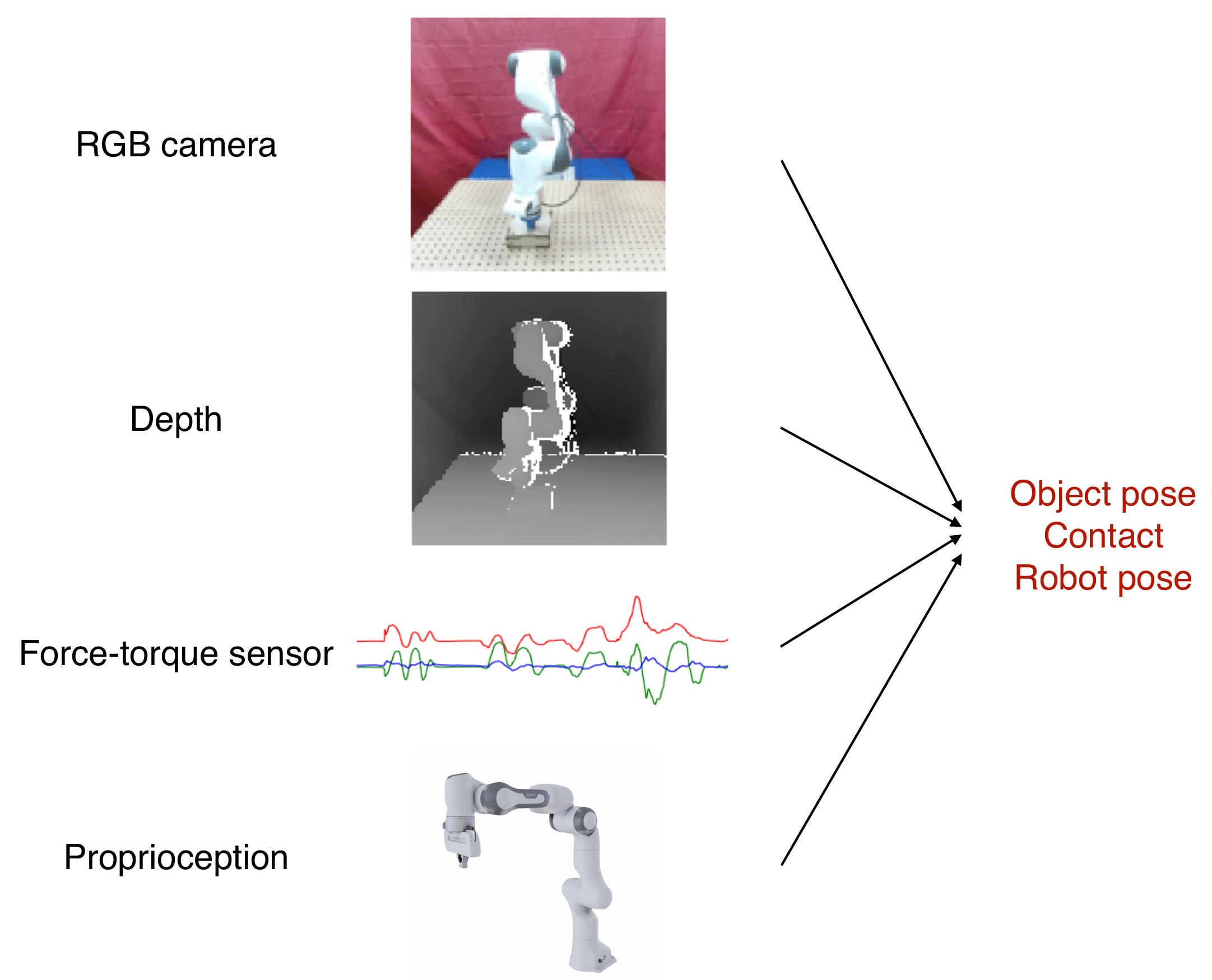}
\vspace{-0mm}
\caption{\textbf{Robotics:} Modern robot systems are equipped with multiple sensors to aid in their decision-making. We include the large-scale \textsc{MuJoCo Push}~\citep{lee2020multimodal} and \textsc{Vision\&Touch}~\citep{lee2019making_tro} datasets which record the manipulation of real and simulated robotic arms equipped with visual (RGB and depth), force, and proprioception sensors. In \textsc{MuJoCo Push}, the goal is to predict the pose of the object being pushed by the robot end-effector. In \textsc{Vision\&Touch}, the goal is to predict action-conditional learning objectives that capture forward dynamics of the different modalities (contact prediction and robot end-effector pose). Figure adapted from~\citep{lee2019making}.\vspace{-0mm}}
\label{figs:robotics}
\end{figure*}

\textbf{1. \textsc{MuJoCo Push}} is a planar pushing task, in which a $7$-DoF Panda Franka robot is pushing a circular puck with its end-effector in simulation. We estimate the 2D position of the unknown object on a table surface, while the robot intermittently interacts with the object. Similar to \textsc{Vision\&Touch}, planar pushing is a contact-rich task. However, instead of estimating robot states, this dataset is estimating the state of the object the robot is currently interacting with. While other robotics datasets have also studied planar pushing~\citep{bauza2019omnipush,yu2016more}, Yu et al.,~\citep{yu2016more} use a Vicon tracker (instead of raw RGB images) while Bauza et al.,~\citep{bauza2019omnipush} only collect visual and proprioceptive data.

\textbf{Data collection:} According to Lee et al.~\citep{lee2020multimodal}, this dataset consists of $1000$ trajectories with $250$ steps at \SI{10}{\hertz}, of a simulated Franka Panda robot arm pushing a circular puck in MuJoCo~\citep{todorov2012mujoco}. The pushing actions are generated by a heuristic controller that tries to move the end-effector to the center of the object. The multimodal inputs are gray-scaled images $(1 \times 32 \times 32)$ from an RGB camera, forces (and binary contact information) from a force/torque sensor, and the 3D position of the robot end-effector. The task is to predict the 2-D planar object pose which we measure by MSE. 

\textbf{Access restrictions:} While we do not have the license to this dataset, it is a public dataset free to download by the research community from \url{https://github.com/brentyi/multimodalfilter/}.

\textbf{Licenses:} MIT, see \url{https://github.com/brentyi/multimodalfilter/blob/master/LICENSE}.

\textbf{Dataset preprocessing:} Training, validation, and test data are each in their own files and can be used directly after downloading. Data is normalized using mean and variance from the train set. 

\textbf{Train, validation, and test splits:} This dataset contains $1000$ training data, $10$ validation data, and $300$ test data. Each data point is split into $29$ time-series sequences of length $16$. The total number of data points for training, validation, and test are $29,000$, $290$, and $8,700$ for a total of $37990$ data points.

\textbf{2. \textsc{Vision\&Touch}} is a real-world robot manipulation dataset that collects visual, force, and robot proprioception data (as well as the robot actions) for a peg insertion task. The robot is a $7$-DoF, torque-controlled Franka Panda robot, which has a triangle peg attached to its end-effector. Rigidly attached to the table in front of the robot is a box with a triangle hole. The robot attempts to insert the peg into the hole, a contact-rich manipulation task that has been studied for decades due to its relevance in manufacturing. Vision, force, and proprioception are feedback modalities shown to be complementary and concurrent during contact-rich manipulation~\cite{blake2004neural}.

\textbf{Data collection:} According to Lee et al.,~\citep{lee2019making_tro}, the data is collected by running on the robot a random policy (that takes random actions) as well as a heuristic policy (that attempts peg insertion). Four sensor modalities are available, including robot proprioception, an RGB-D camera, and a force-torque sensor. The proprioceptive input is the robot end-effector pose as well as linear and angular velocity. They are computed using forward kinematics. RGB images and depth maps are recorded from a fixed camera (Kinect v2 camera) pointed at the robot. Input images to our model are down-sampled to $128\times 128$. The force sensor provides $6$-axis feedback on the forces and moments along the $x, y, z$ axes. The OptoForce force sensor is mounted between the last joint and the peg. The robot action data is also collected at every timestep. The robot action is the Cartesian end-effector position displacement and $z$-axis roll rotation of the end-effector. There are $150$ trajectories collected, each with $1000$ timesteps of data collected. While the dataset originally was intended for representation learning for reinforcement learning, We use $2$ tasks from the \textsc{Vision\&Touch} datasets: (1) predicting binary contact in the next time step and (2) predicting end-effector position measured in MSE. 

\textbf{Access restrictions:} While we do not have the license to this dataset, it is a public dataset free to download by the research community from \url{https://github.com/stanford-iprl-lab/multimodal_representation/}.

\textbf{Licenses:} MIT, see \url{https://github.com/stanford-iprl-lab/multimodal_representation/blob/master/LICENSE}.

\textbf{Dataset preprocessing:} Dataset has already been pre-processed and can be downloaded directly at \url{https://github.com/stanford-iprl-lab/multimodal_representation/}. The dataset comes as a zipped file with $3000$ hdf5 files, each with $50$ timesteps of data. In order to get action-conditional contact as well as robot end-effector position, the dataset uses the contact and end-effector position data from the next timestep. Since the data from the first time step cannot be used, only $49$ of $50$ timesteps of data per file can be used.

\textbf{Train/validation split:} This dataset uses a $80:20$ training and validation split. There are $117600$ training data points and $29400$ validation data points. Since the original dataset does not contain test data, we report validation performance instead of test performance for this dataset.

\vspace{-1mm}
\subsubsection{Finance}
\vspace{-1mm}

We created the following financial datasets which consist of historical stock data retrieved from publicly available online financial databases. We record the opening price of each stock from \mbox{2000-06-01} to \mbox{2021-02-28}, which creates a total of $5218$ time steps. Details of each dataset are described in its own section below.

\begin{figure*}[]
\centering
\vspace{-0mm}
\includegraphics[width=0.7\linewidth]{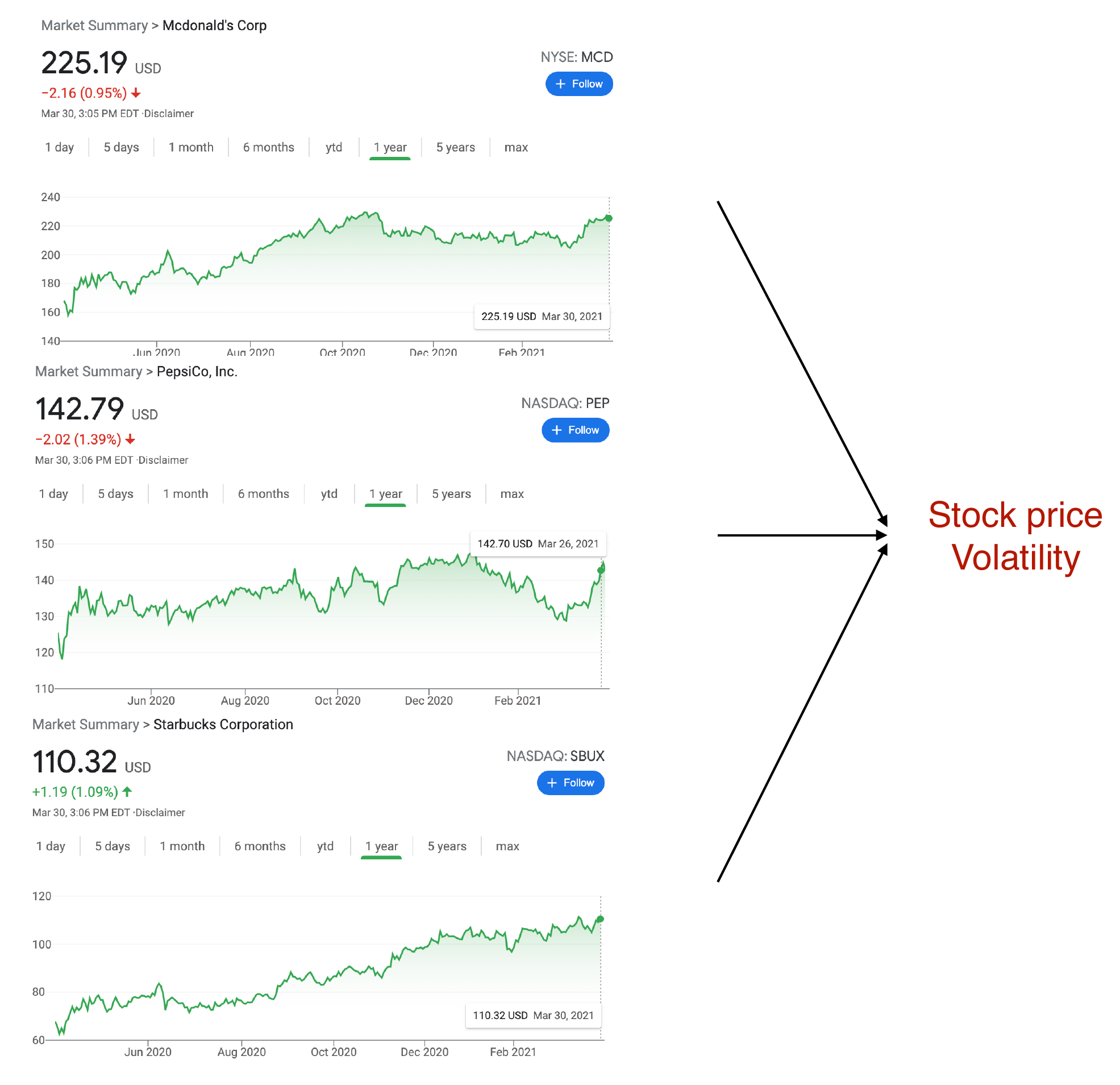}
\vspace{-0mm}
\caption{\textbf{Finance:} We scrape historical stock data from the internet and create our own dataset for financial time-series prediction across $3$ groups of correlated stocks: \textsc{Stocks-F\&B}, \textsc{Stocks-Health}, and \textsc{Stocks-Tech}. Within each group, the previous stock prices of a set of stocks are used as multimodal input to predict the squared return of a related stock (e.g., using Apple, Google, and Microsoft historical data to predict future prices of Microsoft).\vspace{-0mm}}
\label{figs:finance}
\end{figure*}

\textbf{1. \textsc{Stocks-F\&B}} consists of $18$ selected stocks from S\&P 500 stocks categorized by GICS as {Restaurants} or {Packaged Foods \& Meats}. We select \texttt{MCD}, \texttt{SBUX}, \texttt{HSY}, and \texttt{HRL} for initial experiments on this dataset, record their opening prices, and preprocess the data following the preprocessing procedures below.

\textbf{2. \textsc{Stocks-Health}} consists of $63$ selected stocks from S\&P $500$ stocks categorized by GICS as {Health Care}. We select \texttt{MRK}, \texttt{WST}, \texttt{CVS}, \texttt{MCK}, \texttt{ABT}, \texttt{UNH}, and \texttt{TFX} for initial experiments on  this dataset, record their opening prices, and preprocess the data following the preprocessing procedures below.

\textbf{3. \textsc{Stocks-Tech}} consists of $100$ selected stocks from S\&P 500 stocks categorized by GICS as {Information Technology} or {Communication Services}. We select \texttt{AAPL}, \texttt{MSFT}, \texttt{AMZN}, \texttt{INTC}, \texttt{AMD}, and \texttt{MSI} for initial experiments on this dataset, record their opening prices, and preprocess the data following the preprocessing procedures below.

\textbf{Access restrictions:} The datasets were collected from Yahoo Finance, which is publicly available but does not allow redistribution of their data. We provide automated download and preprocessing scripts for this dataset.

\textbf{Licenses:} We could not find a finance dataset with a free redistribution license that includes historical financial data. As such, we provide automated download and preprocessing scripts as part of this project, which utilizes the open-source \texttt{pandas-datareader} to download raw finance data. We used the open-source code at \url{https://github.com/pydata/pandas-datareader/blob/master/pandas_datareader/yahoo/components.py}. The automated scripts we provide are licensed under an MIT License.

\textbf{Dataset preprocessing:} Data is downloaded, converted to returns, and normalized. Labels are converted to squared returns. Each time series is split in chronological order, where the test split corresponds to the latest prices. For each data point, $500$ previous returns are used to predict the squared return of the next day. The first $500$ time steps are not predicted since they do not have $500$ previous steps. We consider each stock as a modality; unimodal datasets have the input stock identical to the target stock. To keep memory usage practical for \textsc{MulT}~\cite{tsai2019multimodal} models, we evenly separate the stocks into $3$ groups and use each group as a modality when preprocessing for \textsc{MulT}~\cite{tsai2019multimodal}.

\textbf{Train, validation, and test splits:} We split the data according to time. There are $3200$ continuous days of stock prices in the train data (\mbox{2002-06-04} start to \mbox{2015-02-18} end date), $500$ continuous days of stock prices in the valid data (\mbox{2015-02-19} start to \mbox{2017-02-10} end date), and $1017$ continuous days of stock prices in the test data (\mbox{2017-02-13} start to \mbox{2021-02-26} end date).

\vspace{-1mm}
\subsubsection{HCI}
\vspace{-1mm}

\begin{figure*}[]
\centering
\vspace{-0mm}
\includegraphics[width=0.7\linewidth]{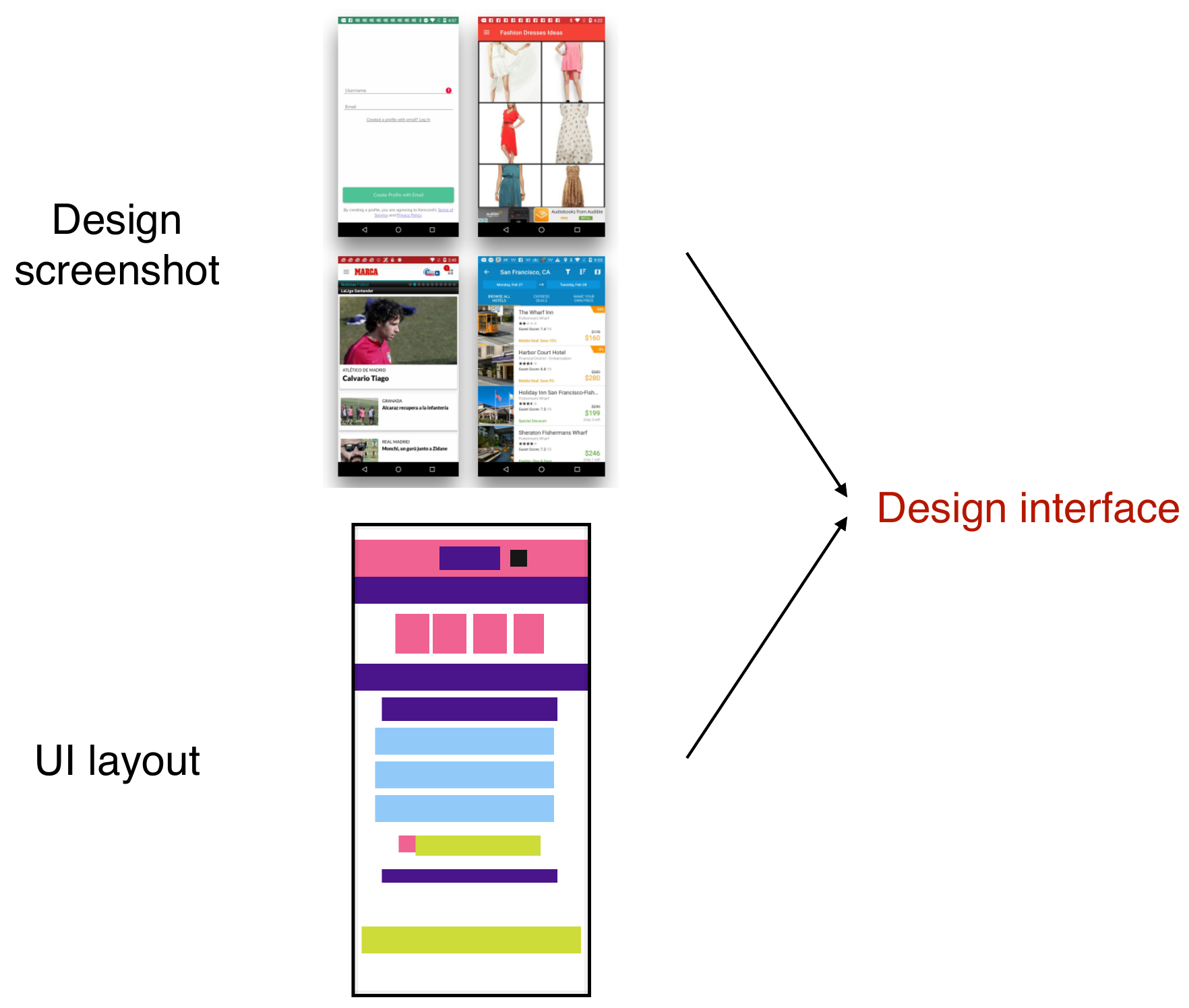}
\vspace{-0mm}
\caption{\textbf{Human Computer Interaction (HCI)} studies the design and use of computer technology with a focus on the interactive interfaces between humans and computers. We use the \textsc{Enrico} (Enhanced Rico) dataset~\cite{deka2017rico,leiva2020enrico} of Android app screens (consisting of an image as well as a set of apps and their locations) categorized by their design motifs and collected for data-driven design applications such as design search, user interface (UI) layout generation, UI code generation, and user interaction modeling. Figure adapted from~\citep{deka2017rico,leiva2020enrico}.\vspace{-0mm}}
\label{figs:hci}
\end{figure*}

\textbf{1. \textsc{ENRICO}} (Enhanced Rico)~\cite{leiva2020enrico} is a dataset of Android app screens categorized by their design motifs. ENRICO was collected to help data-driven design applications such as design search, UI layout generation, UI code generation, and user interaction modeling. ENRICO is a subset of RICO~\cite{deka2017rico}, which is a large dataset of app screens collected by the automated and semi-automated ``crawling'' of Android apps available on the Google Play Store.

The RICO and ENRICO datasets have been used as benchmarks for data-driven models of design in scaffolding the creation of mobile apps. These constitute a set of relevant examples that help designers understand best practices and trends in building human-centered interfaces. Building multimodal models on these examples will enable systems that can predict whether a UI design will achieve its targeted goals even before it is deployed to millions of people. In the long run, this will enable the large-scale creation of personalized UI designs that can automatically adapt to diverse users and contexts.

The authors of ENRICO employed two main modalities for app classification: (1) the app screenshot and (2) the view hierarchy. The app screenshot is given in the form of an image. The view hierarchy is a type of metadata associated with some UI screens that describe the spatial and structural layout of UI elements. This view hierarchy can be treated as a \textit{set} since it contains an unordered collection of UI elements each containing metadata and their spatial and structural layout.

\textbf{Data collection:}
The original RICO dataset was collected using a combination of manual (\textit{i.e.,} crowdworkers) and automated (\textit{i.e.,} app crawler) methods.
More information about how the apps were downloaded and captured is available in the RICO paper \cite{deka2017rico}.
The ENRICO dataset is a subset of RICO that was created by first randomly sampling $10000$ screens from RICO and labeling a high-quality subset ($1460$ screens) that can be categorized into $20$ design categories.
More information about the collection and annotation process is available in the ENRICO paper~\cite{leiva2020enrico}.

\textbf{Access restrictions:} While we do not have the license to this dataset, it is a public dataset free to download by the research community from \url{https://github.com/luileito/enrico}.

\textbf{Licenses:} MIT, see \url{https://github.com/luileito/enrico/blob/master/LICENSE}

\textbf{Dataset preprocessing:} We extract the following features from each modality: 
\begin{enumerate}
    \item \textit{Image:} The authors of ENRICO used a VGG-16 network (augmented with batch normalization and dropout) to encode app screenshots. To reduce overfitting on the relatively small dataset ($1460$ examples), we use a VGG-11 network pre-trained on ImageNet, with a frozen feature extraction network and a slimmed-down classifier network.
    \item \textit{Set:} We followed prior modeling approaches~\cite{deka2017rico,leiva2020enrico} to represent the view hierarchy as a set of UI elements spatially rendered as a ``wireframe'' (similar to a semantic map).
The wireframe was then fed into the same VGG-11 network used to encode the screenshot.
Another possibility, which we briefly explored, is to use a set encoder~\cite{zaheer2017deep} to use a permutation invariant function to compute a pooled representation of the set of mobile applications.
We found that the CNN-based approach resulted in better performance, as it allowed the network to be initialized from a pre-trained checkpoint, although our experiments were initial and there is still ample room for future work to explore better encoders for this set modality.
\end{enumerate}

\textbf{Train, validation, and test splits:} The original paper doesn't provide official splits for training, validation, and testing. We used a known seed to deterministically shuffle the dataset and create splits for training ($65\%$, $947$ examples), validation ($15\%$, $219$ examples), and testing ($20\%$, $292$ examples).

\vspace{-1mm}
\subsubsection{Multimedia}
\vspace{-1mm}

\textbf{1. \textsc{AV-MNIST}} is a multimodal dataset created by paring audio of a human reading digits from the \textsc{FSDD} dataset~\citep{fsdd} with written digits in the \textsc{MNIST} dataset~\citep{mnist} with a task to predict the digit into one of $10$ classes ($0-9$). Since existing models can already complete the digit recognition task from either modality quite well, one common practice in previous work~\citep{vielzeuf2018centralnet} is to increase the difficulty by removing $75\%$ of energy in the visual modality via PCA and adding noise from \textsc{ESC-50}~\citep{piczak2015dataset} to the audio modality, such that models have to leverage information from both modalities to make accurate predictions. \textsc{ESC-50} is a realistic dataset collected from real-world audio of various everyday objects. Therefore, \textsc{AV-MNIST} serves as a good starting point of a relatively simple multimodal dataset but with underlying challenges of complementarity and noisy data. In fact, the method of injecting real-world background noises into the audio modality also inspired more tests for robustness included in \names. \textsc{AV-MNIST} has served as a popular benchmark for evaluating the effectiveness of multimodal fusion models~\citep{perez2019mfas,vielzeuf2018centralnet}.

\textbf{Data collection:} According to Vielzeuf et al.,~\cite{vielzeuf2018centralnet}, \textsc{AV-MNIST} starts with the entirety of the \textsc{MNIST} image and \textsc{FSDD} audio datasets. The audio samples are augmented by adding randomly chosen `noise' samples from the \textsc{ESC-50} dataset~\citep{piczak2015dataset}, to reach the same number of examples as in \textsc{MNIST} ($55000$ training, $5000$ validation, and $10000$ testing examples).

\textbf{Access restrictions:} This dataset is programmatically generated by combining $2$ unimodal datasets: \textsc{MNIST} and \textsc{FSDD} (with the additional audio signal from \textsc{ESC-50}). While we do not have the license to these datasets, they are public datasets free to download by the research community.

\textbf{Licenses:} \textsc{MNIST} is released with a Creative Commons Attribution-Share Alike 3.0. \textsc{FSDD} is released with a Creative Commons Attribution-ShareAlike 4.0 International license. \textsc{ESC-50} is released with a Creative Commons Attribution Non-Commercial license. All of these licenses allow redistribution of the datasets.

\textbf{Dataset preprocessing:} To create the dataset, we downloaded \textsc{MNIST} from \url{http://yann.lecun.com/exdb/mnist/}, \textsc{FSDD} from \url{https://github.com/Jakobovski/free-spoken-digit-dataset}, \textsc{ESC-50} from \url{https://github.com/karolpiczak/ESC-50}, and generated \textsc{AV-MNIST} with the scripts provided in \url{https://github.com/slyviacassell/\_MFAS/blob/master/datasets/avmnist\_gen.py}. Note that since the official implementation of the preprocessing is not released, our preprocessing, as well as all other existing preprocessing scripts, may differ from the original preprocessing in some details (such as keeping at most or at least $25\%$ of energy in the image modality, and some parameters in adding noise to audio), so the performance of models in our version of \textsc{AV-MNIST} should not be compared directly with the performance of models on \textsc{AV-MNIST} in other papers.

No preprocessing is done for the image modality. For audio, it is converted to a $112x112$ Spectogram. See the code in \url{https://github.com/slyviacassell/\_MFAS/blob/master/datasets/avmnist\_gen.py} for details.

\textbf{Train, validation, and test splits:} Data splits for \textsc{AV-MNIST} follow that of the \textsc{MNIST} dataset, with $55000$ training, $5000$ validation, and $10000$ testing examples. 

\textbf{2. \textsc{MM-IMDb}} is the largest publicly available multimodal dataset for genre prediction on movies~\citep{arevalo2017gated}. \textsc{MM-IMDb} starts from the movies of the MovieLens $20$M dataset and expands this dataset by collecting genre, poster, and plot information for each movie. The final dataset contains ratings for $25,959$ movies. \textsc{MM-IMDb} is a realistic real-world multimodal dataset and is a popular benchmark for multimodal learning~\citep{arevalo2017gated,kiela2019supervised,perez2019mfas}.

\textbf{Data collection:} According to Arevalo et al.,~\citep{arevalo2017gated}, \textsc{MM-IMDb} dataset is built with the IMDb ids provided by the Movielens $20$M dataset that contains ratings of $27,000$ movies. Using the IMDbPY 3 library, movies that do not contain their poster image were filtered out. The resulting dataset comprises $25,959$ movies along with their plot, poster, genres, and other $50$ additional metadata fields such as year, language, writer, director, aspect ratio, etc. The task is to perform multilabel classification into one of $23$ movie genres.

\textbf{Access restrictions:} While we do not have the license to this dataset, it is a public dataset free to download by the research community from \url{http://lisi1.unal.edu.co/mmimdb/} and \url{https://github.com/johnarevalo/gmu-mmimdb/}.

\textbf{Licenses:} MIT, see \url{https://github.com/johnarevalo/gmu-mmimdb/blob/master/LICENSE}

\textbf{Dataset preprocessing:} We used the same method as \citep{arevalo2017gated} to extract features from texts and images.
\begin{enumerate}
    \item \textit{Text:} We used the pretrained Google Word2vec\footnote{\hyperref[googelw2v]{https://code.google.com/archive/p/word2vec/}} to extract text features. The final vocabulary contains $41,612$, which is the intersection of Google word2vec words and the \textsc{MM-IMDb} plots. We converted all text to lowercase following existing work.
    \item \textit{Image:} All images were scaled, and cropped when required, to $160 \times 256$ pixels keeping the aspect ratio. A VGG-16 model~\citep{Simonyan15} is applied as the image feature extractor. This CNN consists of $5$ convolutional layers of $5, 3, 3, 3, 3$ squared filters and $2 \times 2$ pool sizes. Each convolutional layer has $16$ hidden units. The convolutional layers are connected with a MaxoutMLP on top. 
\end{enumerate}

\textbf{Train, validation, and test splits:} The MM-IMDb dataset is split by genre into train, valid, and test datasets containing $15552$, $2608$, and $7799$. The split was performed so that training, valid and test sets comprise $60\%$, $10\%$, $30\%$ samples of each genre respectively. 

\begin{figure*}[]
\centering
\vspace{-0mm}
\includegraphics[width=\linewidth]{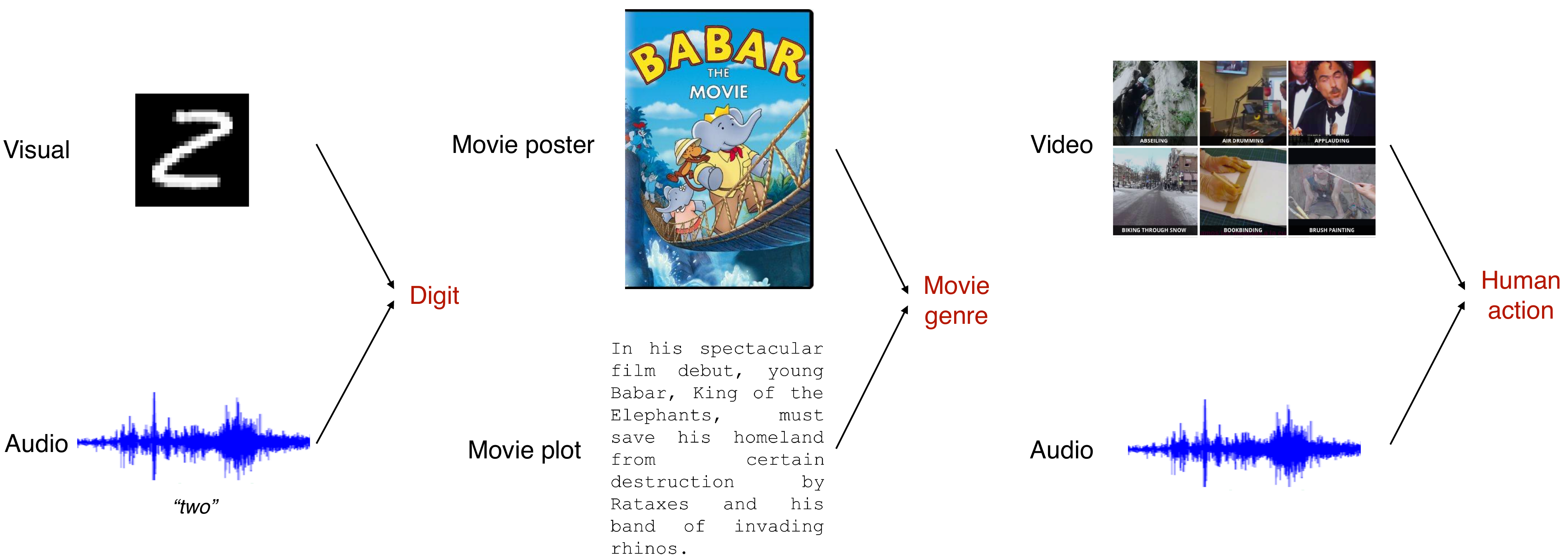}
\vspace{-0mm}
\caption{\textbf{Multimedia:} A significant body of research in multimodal learning has been fueled by the large availability of multimedia data (language, image, video, and audio) on the internet. \names\ includes $3$ popular large-scale multimedia datasets with varying sizes and levels of difficulty: (1) Audio-Visual MNIST (\textsc{AV-MNIST})~\citep{vielzeuf2018centralnet} is assembled from images of handwritten digits~\cite{mnist} and audio samples of spoken digits~\cite{tidigits}, (2) Multimodal IMDb (\textsc{MM-IMDb})~\citep{arevalo2017gated} uses movie titles, metadata, and movie posters to perform multi-label classification to a set of movie genres, and (3) \textsc{Kinetics}~\citep{kay2017kinetics} contains video and audio of $306,245$ video clips annotated for $400$ human actions. To ease experimentation, we split \textsc{Kinetics} into small and large partitions (see Appendix~\ref{appendix:data}). Figure adapted from~\citep{arevalo2017gated,kay2017kinetics}.\vspace{-0mm}}
\label{figs:multimedia}
\end{figure*}

\textbf{3. \textsc{Kinetics}} is a series of large-scale curated video clips covering a diverse range of human actions. We use the original Kinetics-$400$ dataset~\citep{kay2017kinetics} which contains $400$ human action classes, with at least $400$ video clips for each action. Each clip lasts around $10$s and is taken from a different YouTube video. This is one of the largest publicly available multimodal datasets with a total of $306,245$ video clips spanning $400$ human actions. Therefore, \textsc{Kinetics} is suitable for testing the scalability of multimodal models to extremely large datasets. Furthermore, recognizing human actions is a core challenge in a variety of applications such as human-AI interaction, robotics, and human behavior analysis.

The sheer scale of the \textsc{Kinetics} dataset means that even the simplest models take up to several weeks to finish training. To enable multimodal learning from video and audio while also increasing access across researchers with limited computing resources, we subsample the \textsc{Kinetics} dataset into small and large partitions:

\textsc{Kinetics-S:} We subsampled $5$ human actions: \textit{archery}, \textit{breakdancing}, \textit{crying}, \textit{dining}, \textit{singing} and retained all video clips annotated for these $5$ actions. We selected these actions randomly out of the 400 actions in Kinetics-$400$. This gave us a total of $2624$ video clips in the small version of the dataset. Training a basic supervised learning model on \textsc{Kinetics-S} takes roughly $2$ hours on a single GPU.

\textsc{Kinetics-L:} This represents the entire \textsc{Kinetics}-$400$ dataset with $306,245$ video clips spanning $400$ human actions. Training a basic supervised learning model on \textsc{Kinetics-L} takes roughly $2$ weeks on a single GPU.

\textbf{Data collection:} We refer the reader to Kay et al.,~\citep{kay2017kinetics} for a detailed description of the dataset collection process. Briefly, the authors (1) started with a list of human actions from sources spanning existing action datasets, motion capture, and crowdsourcing, (2) obtained candidate clips from YouTube and extracted temporal positions within a video, (3) performed manual labeling for human actions with Amazon’s Mechanical Turk, and (4) cleaning up and de-noising the selected videos.

\textbf{Access restrictions:} While we do not have the license to this dataset, it is a public dataset free to download by the research community from \url{https://deepmind.com/research/open-source/kinetics}.

\textbf{Licenses:} Creative Commons Attribution 4.0 International, so we are free to share, copy, and redistribute the material in any medium or format, see \url{https://deepmind.com/research/open-source/kinetics}.

\textbf{Dataset preprocessing:} We downloaded links from \url{https://deepmind.com/research/open-source/kinetics} and preprocessed them with the torchvision Kinetics scripts.

We processed the video and audio modalities as follows:
\begin{enumerate}
    \item \textit{Video:} We use $150 \times 224 \times 224 \times 3$ input clips, created with a frame skip of $2$, a center crop with shape $(224, 224)$, and the normalization step required for using torchvision.models.
    \item \textit{Audio:} We use log-scaled mel spectrograms with $763$ temporal frames by $40$ Mel filters, element-wise averaging $2$-channel waveforms to yield single channel ones.
\end{enumerate}

\textbf{Train, validation, and test splits}: We use the $80.5/6.5/13$ split provided by the original dataset, taking all the data points in our chosen classes. This yields $2112$, $171$, and $341$ data points in train, validation, and test splits respectively for \textsc{Kinetics-S} and $246527$, $19906$, and $39812$ data points in train, validation, and test splits respectively for \textsc{Kinetics-L}.

\vspace{-1mm}
\subsection{Documentation}
\vspace{-1mm}

We provide documentation for \names\ in the form of datasheets for datasets~\citep{gebru2018datasheets}:
\begin{enumerate}
    \item \textbf{Motivation}
    \begin{enumerate}
        \item \textit{For what purpose was the dataset created? Was there a specific task in mind? Was there a specific gap that needed to be filled? Please provide a description.}
        
        Learning multimodal representations involves integrating information from multiple heterogeneous sources of data. It is a challenging yet crucial area with numerous real-world applications in multimedia, affective computing, robotics, finance, and healthcare. Unfortunately, current research focuses primarily on a fixed set of modalities and tasks without a concrete understanding of generalization across domains and modalities, complexity during training and inference, and robustness to noisy and missing modalities. In order to standardize multimodal research and accelerate progress towards understudied modalities and tasks while ensuring real-world robustness, we release \names, a systematic and unified large-scale benchmark for multimodal learning spanning $15$ datasets, $10$ modalities, $20$ prediction tasks, and $6$ research areas. \names\ provides an automated end-to-end machine learning pipeline that simplifies and standardizes data loading, experimental setup, and model evaluation. To enable holistic evaluation, \names\ summarizes both performance as well as the potential drawbacks involving increased time and space complexity and risk of decreased robustness from other modalities. To accompany \names, we also provide a standardized implementation of $20$ core approaches in multimodal learning unifying innovations in fusion paradigms, optimization objectives, and training approaches.
        
        \names\ datasets present significant challenges of scalability to large-scale multimodal datasets and robustness to realistic imperfections, which present fruitful opportunities for future research. We hope that \names\ will present a milestone in unifying disjoint efforts in multimodal machine learning research and paves a way towards a better understanding of the capabilities and limitations of multimodal models, all the while ensuring ease of use, accessibility, and reproducibility. \names, our standardized implementation, and leaderboards are publicly available, will be regularly updated, and welcomes inputs from the community.
        
        \item \textit{Who created the dataset (e.g., which team, research group) and on behalf of which entity (e.g., company, institution, organization)?}
        
        \names\ is created primarily by the MultiComp Lab in the Language Technologies Institute and Machine Learning Department of the School of Computer Science at Carnegie Mellon University, in collaboration with several other researchers in the Human-Computer Interaction Institute and Computer Science Department at Carnegie Mellon University as well as at Johns Hopkins University, Stanford University, and UT Austin. The creation of \names\ is for purely research purposes only.
        
        \item \textit{Who funded the creation of the dataset? If there is an associated grant, please provide the name of the grantor and the grant name and number.}
        
        This material was based upon work partially supported by the National Science Foundation (Awards \#1722822 and \#1750439) and National Institutes of Health (Awards \#R01MH125740, \#R01MH096951, \#U01MH116925, and \#U01MH116923), NSF IIS1763562, and ONR Grant N000141812861. Any opinions, findings, and conclusions, or recommendations expressed in this material are those of the author(s) and do not necessarily reflect the views of the National Science Foundation or National Institutes of Health, and no official endorsement should be inferred.
        
        \item \textit{Any other comments?}
        
        No.
    \end{enumerate}
    
    \item \textbf{Composition}
    \begin{enumerate}
        \item \textit{What do the instances that comprise the dataset represent (e.g., documents, photos, people, countries)? Are there multiple types of instances (e.g., movies, users, and ratings; people and interactions between them; nodes and edges)? Please provide a description.}
        
        We describe each dataset in detail in Appendix~\ref{appendix:dataset_details}. \names\ provides a comprehensive suite of multimodal datasets to benchmark current and proposed approaches in multimodal representation learning. It covers a diverse range of research areas (affective computing, healthcare, robotics, finance, HCI, and multimedia), dataset sizes (small, medium, and large), input modalities (in the form of $\ell$: language, $i$: image, $v$: video, $a$: audio, $t$: time-series, $ta$: tabular, $o$: optical flow, $f$: force sensor, $p$: proprioception sensor, $s$: set), and prediction tasks (affect recognition, robot manipulation, stock prediction, design interface, action recognition, movie genre prediction, and digit prediction).
    
        \item \textit{How many instances are there in total (of each type, if appropriate)?}

        We describe each dataset's statistics in detail in Appendix~\ref{appendix:dataset_details}. We chose datasets to span small, medium, and large sizes. The smallest dataset contains $1,460$ instances (and training a model takes roughly a few minutes on a single GPU) while the largest one contains $306,245$ instances (and training a model takes roughly $2$ weeks on a single GPU). This enables accessibility for researchers with limited computational resources, while also allowing for large-scale studies of multimodal datasets and models.

        \item \textit{Does the dataset contain all possible instances or is it a sample (not necessarily random) of instances from a larger set? If the dataset is a sample, then what is the larger set? Is the sample representative of the larger set (e.g., geographic coverage)? If so, please describe how this representativeness was validated/verified. If it is not representative of the larger set, please describe why not (e.g., to cover a more diverse range of instances, because instances were withheld or unavailable).}
        
        Each of the datasets is collected in different ways that we detail in Appendix~\ref{appendix:dataset_details}. To summarize, each dataset consists of samples from a larger set since it is impossible to include all videos/stock data/medical data/robotics data in the world. Each dataset is collected with the aim to be representative of the entire population.

        \item \textit{What data does each instance consist of? ``Raw'' data (e.g., unprocessed text or images) or features? In either case, please provide a description.}
        
        We describe in detail the raw data and processed features for each dataset in Appendix~\ref{appendix:dataset_details}. To summarize, \names\ contains both raw modality data as well as processed data with predefined feature extractors following current work.

        \item \textit{Is there a label or target associated with each instance? If so, please provide a description.}
        
        We describe in detail the labels for each dataset in Appendix~\ref{appendix:dataset_details}. To summarize, \names\ contains $6$ research areas with a total of $15$ prediction tasks spanning affect recognition, robot manipulation, stock prediction, design interface, action recognition, movie genre prediction, and digit prediction.

        \item \textit{Is any information missing from individual instances? If so, please provide a description, explaining why this information is missing (e.g., because it was unavailable). This does not include intentionally removed information, but might include, e.g., redacted text.}
        
        No, all datasets are provided in full. For robustness tests, we do inject noise and imperfections into each dataset to simulate the performance of machine learning models on real-world imperfections (see Appendix~\ref{appendix:robustness} for details).
        
        \item \textit{Are relationships between individual instances made explicit (e.g., users’ movie ratings, social network links)? If so, please describe how these relationships are made explicit.}
        
        We describe in detail the relationships between modalities for each dataset in Appendix~\ref{appendix:dataset_details}.
        
        \item \textit{Are there recommended data splits (e.g., training, development/validation, testing)? If so, please provide a description of these splits, explaining the rationale behind them.}
        
        Yes, \names\ provides a data loading pipeline that directly loads train, validation, and test splits according to current work. We provide these details for each dataset in Appendix~\ref{appendix:dataset_details}.
    
        \item \textit{Are there any errors, sources of noise, or redundancies in the dataset? If so, please provide a description.}
        
        We do not know of any errors in each of the datasets included in \names. However, we will always be on the lookout for potential issues and update them via \weburl\ and \dataurl.

        \item \textit{Is the dataset self-contained, or does it link to or otherwise rely on external resources (e.g., websites, tweets, other datasets)? If it links to or relies on external resources, a) are there guarantees that they will exist, and remain constant, over time; b) are there official archival versions of the complete dataset (i.e., including the external resources as they existed at the time the dataset was created); c) are there any restrictions (e.g., licenses, fees) associated with any of the external resources that might apply to a future user? Please provide descriptions of all external resources and any restrictions associated with them, as well as links or other access points, as appropriate.}
        
        Most of the datasets in \names\ have been collected, stored, processed, and are self-contained. There are some datasets that depend on external resources which we explain below:
        \begin{enumerate}
            \item MIMIC: We depend on the original dataset to be hosted on \url{https://mimic.physionet.org/gettingstarted/access/}. Unfortunately, since we are not allowed to redistribute the raw data and users need to complete training to access the raw data, we are unable to provide a self-contained version of the MIMIC dataset. We are currently planning to add several new multimodal datasets in the healthcare domain that can be self-contained after appropriate de-identification.
            \item Finance: Yahoo Finance prohibits the redistribution of their data. We depend on the original data to be hosted on Yahoo Finance and provide automated downloading and preprocessing scripts for the datasets based on \texttt{pandas-datareader}, which has original code at \url{https://github.com/pydata/pandas-datareader/blob/master/pandas_datareader/yahoo/components.py}
        \end{enumerate}
        
        \item \textit{Does the dataset contain data that might be considered confidential (e.g., data that is protected by legal privilege or by doctor-patient confidentiality, data that includes the content of individuals’ non-public communications)? If so, please provide a description.}
        
        From the authors of MIMIC~\citep{MIMIC}: ``The project was approved by the Institutional Review Boards of Beth Israel Deaconess Medical Center (Boston, MA) and the Massachusetts Institute of Technology (Cambridge, MA). Requirement for individual patient consent was waived because the project did not impact clinical care and all protected health information was de-identified.''
        
        To the best of our knowledge, all other datasets do not contain confidential data and are publicly available for research purposes.

        \item \textit{Does the dataset contain data that, if viewed directly, might be offensive, insulting, threatening, or might otherwise cause anxiety? If so, please describe why.}
        
        We reviewed the datasets and found no offensive content. While there are clearly expressions of highly negative sentiment or strong displays of anger and disgust in the affective computing videos, there are no offensive words used or personal attacks recorded in the video. All videos are related to movie or product reviews, TED talks, and TV shows.

        \item \textit{Does the dataset relate to people? If not, you may skip the remaining questions in this section.}
        
        Yes, the healthcare, affective computing, and Kinetics (multimedia) datasets relate to people. The other datasets in \names\ do not.

        \item \textit{Does the dataset identify any subpopulations (e.g., by age, gender)? If so, please describe how these subpopulations are identified and provide a description of their respective distributions within the dataset.}
        
        The following datasets relate to people:
        \begin{enumerate}
            \item Affective computing: These datasets do not identify any subpopulations in their modeling decisions. However, the raw data comes in the form of videos publicly available and free to download from YouTube. Sub-population and demographic information can be inferred from these raw videos.
            \item MIMIC: According to the authors~\citep{MIMIC}: ``The median age of adult patients is $65.8$ years and $55.9\%$ patients are male.''
            \item Kinetics: This dataset does not identify any subpopulations. However, the raw data comes in the form of videos publicly available and free to download from YouTube. Sub-population and demographic information can be inferred from these raw videos.
        \end{enumerate}

        \item \textit{Is it possible to identify individuals (i.e., one or more natural persons), either directly or indirectly (i.e., in combination with other data) from the dataset? If so, please describe how.}
        
        The following datasets relate to people:
        \begin{enumerate}
            \item Affective computing: One can see the person in the raw video, but the dataset contains no personal information. We do not explicitly use information regarding gender, ethnicity, identity, or video identifier in online sources. All pre-extracted features are non easily invertible and only rely on general visual or audio features such as the presence of a smile or magnitude of voice~\citep{zadeh2016mosi,zadeh2018multimodal}.
        
            \item MIMIC: The \textsc{MIMIC} dataset has been rigorously de-identified in accordance with Health Insurance Portability and Accountability Act (HIPAA) such that all possible personal information has been removed from the dataset. Removed personal information includes patient name, telephone number, address, and dates. Dates of birth for patients aged over $89$ were shifted to obscure their true age. Please refer to Appendix~\ref{appendix:health_data} for de-identification details. Again, we emphasize that any multimodal models trained to perform prediction should only be used for scientific study and should not in any way be used for real-world prediction.
        
            \item Kinetics: One can see the person in the raw video, but the dataset does not contain direct personal information. We do not explicitly use information regarding gender, ethnicity, identity, or video identifier in online sources.
        \end{enumerate}

        \item \textit{Does the dataset contain data that might be considered sensitive in any way (e.g., data that reveals racial or ethnic origins, sexual orientations, religious beliefs, political opinions or union memberships, or locations; financial or health data; biometric or genetic data; forms of government identification, such as social security numbers; criminal history)? If so, please provide a description.}
        
        \names\ contains datasets with financial and healthcare data. However, all these datasets are publicly available for research purposes. Healthcare data (\textsc{MIMIC}) has been rigorously de-identified in accordance with the Health Insurance Portability and Accountability Act (HIPAA) such that all possible personal information (patient name, telephone number, address, and dates, date of birth) has been removed from the dataset. Please refer to Appendix~\ref{appendix:health_data} for de-identification details.

        \item \textit{Any other comments?}
        
        No.
    \end{enumerate}
    
    \item \textbf{Collection Process}
    \begin{enumerate}
        \item \textit{How was the data associated with each instance acquired? Was the data directly observable (e.g., raw text, movie ratings), reported by subjects (e.g., survey responses), or indirectly inferred/derived from other data (e.g., part-of-speech tags, model-based guesses for age or language)? If data was reported by subjects or indirectly inferred/derived from other data, was the data validated/verified? If so, please describe how.}
        
        We include the collection process for each dataset in Appendix~\ref{appendix:dataset_details}.

        \item \textit{What mechanisms or procedures were used to collect the data (e.g., hardware apparatus or sensor, manual human curation, software program, software API)? How were these mechanisms or procedures validated?}
        
        We include these details in Appendix~\ref{appendix:dataset_details}.
        
        \item \textit{If the dataset is a sample from a larger set, what was the sampling strategy (e.g., deterministic, probabilistic with specific sampling probabilities)?}
        
        We include sampling methods for each dataset in Appendix~\ref{appendix:dataset_details}.
        
        \item \textit{Who was involved in the data collection process (e.g., students, crowdworkers, contractors) and how were they compensated (e.g., how much were crowdworkers paid)?}
        
        We include annotation details for each dataset in Appendix~\ref{appendix:dataset_details}.
        
        \item \textit{Over what timeframe was the data collected? Does this timeframe match the creation timeframe of the data associated with the instances (e.g., recent crawl of old news articles)? If not, please describe the timeframe in which the data associated with the instances was created.}
        
        We include timeframes for each dataset in Appendix~\ref{appendix:dataset_details}.

        \item \textit{Were any ethical review processes conducted (e.g., by an institutional review board)? If so, please provide a description of these review processes, including the outcomes, as well as a link or other access point to any supporting documentation.}
        
        From the authors of MIMIC~\citep{MIMIC}: ``The project was approved by the Institutional Review Boards of Beth Israel Deaconess Medical Center (Boston, MA) and the Massachusetts Institute of Technology (Cambridge, MA). Requirement for individual patient consent was waived because the project did not impact clinical care and all protected health information was de-identified.''
        
        \item \textit{Does the dataset relate to people? If not, you may skip the remainder of the questions in this section.}
        
        Yes, the healthcare, affective computing, and Kinetics (multimedia) datasets relate to people. The other datasets in \names\ do not.

        \item \textit{Did you collect the data from the individuals in question directly, or obtain it via third parties or other sources (e.g., websites)?}
        
        Affective computing and Kinetics datasets are collected from YouTube videos that follow the creative commons license and follow fair use guidelines of YouTube. According to the authors for the MIMIC dataset~\citep{MIMIC}: ``Data was downloaded from several sources, including archives from critical care information systems, hospital electronic health record databases, and Social Security Administration Death Master File.''

        \item \textit{Were the individuals in question notified about the data collection? If so, please describe (or show with screenshots or other information) how the notice was provided, and provide a link or other access point to, or otherwise reproduce, the exact language of the notification itself.}
        
        Affective computing and Kinetics datasets are collected from YouTube videos that follow the creative commons license and follow fair use guidelines of YouTube. This is the standard way for content creators to grant someone else permission to use and redistribute their work. According to the authors for the MIMIC dataset~\citep{MIMIC}: ``The project was approved by the Institutional Review Boards of Beth Israel Deaconess Medical Center (Boston, MA) and the Massachusetts Institute of Technology (Cambridge, MA). Requirement for individual patient consent was waived because the project did not impact clinical care and all protected health information was de-identified.''

        \item \textit{Did the individuals in question consent to the collection and use of their data? If so, please describe (or show with screenshots or other information) how consent was requested and provided, and provide a link or other access point to, or otherwise reproduce, the exact language to which the individuals consented.}
        
        Affective computing and Kinetics datasets are collected from YouTube videos that follow the creative commons license and follow fair use guidelines of YouTube which allows content creators to grant someone else permission to use and redistribute their work. According to the authors for the MIMIC dataset~\citep{MIMIC}: ``Requirement for individual patient consent was waived because the project did not impact clinical care and all protected health information was de-identified.''
        
        \item \textit{If consent was obtained, were the consenting individuals provided with a mechanism to revoke their consent in the future or for certain uses? If so, please provide a description, as well as a link or other access point to the mechanism (if appropriate).}
        
        N/A.

        \item \textit{Has an analysis of the potential impact of the dataset and its use on data subjects (e.g., a data protection impact analysis) been conducted? If so, please provide a description of this analysis, including the outcomes, as well as a link or other access point to any supporting documentation.}
        
        N/A.
        
        \item \textit{Any other comments?}
        
         N/A.
    \end{enumerate}
    
    \item \textbf{Preprocessing/cleaning/labeling}
    \begin{enumerate}
        \item \textit{Was any preprocessing/cleaning/labeling of the data done (e.g., discretization or bucketing, tokenization, part-of-speech tagging, SIFT feature extraction, removal of instances, processing of missing values)? If so, please provide a description. If not, you may skip the remainder of the questions in this section.}
        
        Yes, we followed the convention in prior research for any preprocessing done to the datasets. We explain these steps in Appendix~\ref{appendix:dataset_details}.
        
        \item \textit{Was the ``raw'' data saved in addition to the preprocessed/cleaned/labeled data (e.g., to support unanticipated future uses)? If so, please provide a link or other access point to the ``raw'' data.}
        
        Yes, we include the raw data in \names\ in addition to the preprocessed features. The raw data (usually in the form of raw text, videos, audio, time series etc) are useful for users to perform their own feature extraction and also for robustness tests on raw data itself (e.g., imperfections in the raw text through spelling errors and missing words). There are certain cases where we are not allowed to distribute the raw data: for MIMIC where users must undergo training to download the raw data, and for finance datasets where Yahoo Finance is publicly available but does not allow redistribution of raw data. For both of these datasets, we provide automated download and preprocessing scripts once the raw data is downloaded through the correct procedure by each user (see details in Appendix~\ref{appendix:dataset_details}).
        
        \item \textit{Is the software used to preprocess/clean/label the instances available? If so, please provide a link or other access point.}
        
        Yes, we provided all links and references to preprocessing steps in Appendix~\ref{appendix:dataset_details}.
        
        \item \textit{Any other comments?}
        
        No.
    \end{enumerate}
    
    \item \textbf{Uses}
    \begin{enumerate}
        \item \textit{Has the dataset been used for any tasks already? If so, please provide a description.}
        
        Yes, \names\ contains several datasets that have been used in the multimodal ML community. We provide links to the original repositories of each dataset and their original citations in Appendix~\ref{appendix:dataset_details}.
        
        \item \textit{Is there a repository that links to any or all papers or systems that use the dataset? If so, please provide a link or other access point.}
        
        We provide links to the original repositories of each dataset and their original citations in Appendix~\ref{appendix:dataset_details}. We also include references to general multimodal methods implemented in \codes\ in Appendix~\ref{appendix:algos}. Many of these methods have been tested by their original authors on a small subset of datasets in \names. In addition to these references, the leading authors maintain a reading list on topics in multimodal ML at~\citep{readinglist} which contains links to papers, datasets, code, academic courses, conferences, and workshops relevant to the multimodal ML community.
        
        \item \textit{What (other) tasks could the dataset be used for?}
        
        In addition to building multimodal models for the prediction tasks, datasets in \names\ can also be used for:
        \begin{enumerate}
            \item Unsupervised learning across multimodal data/unsupervised pre-training of multimodal models.
            \item Interpreting relationships between modalities.
            \item Designing models for robustness to noisy and missing modalities.
            \item Investigating alignment between modalities.
            \item Other multimodal tasks including but not limited to: co-learning, translation, retrieval, and grounding~\citep{baltruvsaitis2018multimodal}.
        \end{enumerate}
        
        \item \textit{Is there anything about the composition of the dataset or the way it was collected and preprocessed/cleaned/labeled that might impact future uses? For example, is there anything that a future user might need to know to avoid uses that could result in unfair treatment of individuals or groups (e.g., stereotyping, quality of service issues) or other undesirable harms (e.g., financial harms, legal risks) If so, please provide a description. Is there anything a future user could do to mitigate these undesirable harms?}
        
        We are careful to outline all possible risks associated with each dataset in Appendix~\ref{appendix:dataset_details} and also in our broader impact statement (Appendix~\ref{appendix:broader_impact}). We acknowledge that there could be risks regarding the privacy and security of data, as well as the real-world deployment of these methods whenever human-centric data is involved (e.g., in healthcare, affective computing, and multimedia). We discussed data demographics in the previous section and it should be taken into consideration when making claims regarding the generalization of models to new users. We also emphasize that these multimodal datasets and methods should only be used for research purposes and not for actual real-world deployment until research can sufficiently verify their safety. Finally, we are carefully working with domain experts towards better understanding biases in these multimodal datasets and models as well as their real-world safety.
        
        \item \textit{Are there tasks for which the dataset should not be used? If so, please provide a description.}
        
        Yes, we emphasize that all multimodal models trained to perform prediction on these datasets should not in any way be used to harm individuals and should only be used as a scientific study. They should not be deemed safe for real-world deployment. In particular, the models used to make predictions of affective states, human actions, health indicators, and financial indicators are particularly sensitive and should not be used to inform any real-world decisions. All results must only be used as a scientific study of machine learning methods. See more details in Appendix~\ref{appendix:broader_impact}.

        \item \textit{Any other comments?}
        
        No.
    \end{enumerate}
    
    \item \textbf{Distribution}
    \begin{enumerate}
        \item \textit{Will the dataset be distributed to third parties outside of the entity (e.g., company, institution, organization) on behalf of which the dataset was created? If so, please provide a description.}
        
        Yes, the benchmark will be distributed to the public research community for theoreticians and practitioners to experiment on multimodal data.
        
        \item \textit{How will the dataset be distributed (e.g., tarball on website, API, GitHub)? Does the dataset have a digital object identifier (DOI)?}
        
        We plan to distribute \names\ via our public GitHub: \dataurl. We also include a landing website page on \weburl\ that includes an introduction to the benchmark, links to the relevant papers on multimodal datasets and algorithms, and a public leaderboard to keep track of current progress on these multimodal tasks.
        
        \item \textit{When will the dataset be distributed?}
        
        The dataset is currently available for use. 
        
        \item \textit{Will the dataset be distributed under a copyright or other intellectual property (IP) license, and/or under applicable terms of use (ToU)? If so, please describe this license and/or ToU, and provide a link or other access point to, or otherwise reproduce, any relevant licensing terms or ToU, as well as any fees associated with these restrictions.}
        
        We release the benchmark and code under an MIT license: see \url{https://github.com/pliang279/MultiBench/blob/main/LICENSE}, which allows for sharing and distribution of the code for research purposes. Each of the datasets in \names\ has their own licenses which we detail in Appendix~\ref{appendix:dataset_details}.
        
        \item \textit{Have any third parties imposed IP-based or other restrictions on the data associated with the instances? If so, please describe these restrictions, and provide a link or other access point to, or otherwise reproduce, any relevant licensing terms, as well as any fees associated with these restrictions.}
        
        Yes, \names\ brings together a collection of several existing datasets in the multimodal research that were built by their individual authors who have original licenses for these datasets. We only included the datasets with licenses that allow for redistribution (MIT or Creative Commons license) and are freely downloadable for research purposes. We detailed all dataset licenses in Appendix~\ref{appendix:dataset_details}.
        
        \item \textit{Do any export controls or other regulatory restrictions apply to the dataset or to individual instances? If so, please describe these restrictions, and provide a link or other access point to, or otherwise reproduce, any supporting documentation.}
        
        We are not aware of any such restrictions.
        
        \item \textit{Any other comments?}
        
        No.
    \end{enumerate}
    
    \item \textbf{Maintenance}
    \begin{enumerate}
        \item \textit{Who is supporting/hosting/maintaining the dataset?}
        
        The dataset is supported and hosted by the team of authors at CMU. The team will also lead the maintenance and expansion of \names. The team will also work with the other collaborators on the paper who are domain experts in each research area \names\ covers, such as robotics, HCI, healthcare, and finance.
        
        \item \textit{How can the owner/curator/manager of the dataset be contacted (e.g., email address)?}
        
        We provide all contact addresses at \weburl.
        
        \item \textit{Is there an erratum? If so, please provide a link or other access point.}
        
        All erratum and updates to the dataset will be tracked via GitHub commit histories at \dataurl. We will also provide updates via our landing page \weburl.
        
        \item \textit{Will the dataset be updated (e.g., to correct labeling errors, add new instances, delete instances)? If so, please describe how often, by whom, and how updates will be communicated to users (e.g., mailing list, GitHub)?}
        
        Yes, we plan for long-term maintenance and expansion of the dataset. All erratum and updates to the dataset will be tracked via GitHub commit histories at \dataurl. We will also provide updates via our landing page \weburl. Please refer to Appendix~\ref{appendix:hosting_maintenance} for details.
        
        \item \textit{If the dataset relates to people, are there applicable limits on the retention of the data associated with the instances (e.g., were individuals in question told that their data would be retained for a fixed period of time and then deleted)? If so, please describe these limits and explain how they will be enforced.}
        
        The individuals in question were not notified about the data collection. For YouTube videos, they are released under a creative commons license which is the standard way for content creators to grant someone else permission to use and redistribute their work. According to the authors for the MIMIC dataset~\citep{MIMIC}: ``The project was approved by the Institutional Review Boards of Beth Israel Deaconess Medical Center (Boston, MA) and the Massachusetts Institute of Technology (Cambridge, MA). Requirement for individual patient consent was waived because the project did not impact clinical care and all protected health information was de-identified.''
        
        \item \textit{Will older versions of the dataset continue to be supported/hosted/maintained? If so, please describe how. If not, please describe how its obsolescence will be communicated to users.}
        
        Yes, we will maintain a GitHub history for all updates and older versions of datasets and code in \names.
        
        \item \textit{If others want to extend/augment/build on/contribute to the dataset, is there a mechanism for them to do so? If so, please provide a description. Will these contributions be validated/verified? If so, please describe how. If not, why not? Is there a process for communicating/distributing these contributions to other users? If so, please provide a description.}
        
        Yes, we will create a system where users can create pull requests on GitHub to include their datasets and models. The authors will verify that the additions are in the scope of multimodal learning and do not break the current experimental code. We will work with these authors to ensure that their data and algorithms can be included in \names.
        
        \item \textit{Any other comments?}
        
        No.
    \end{enumerate}
\end{enumerate}

\vspace{-1mm}
\subsection{Benchmark Distribution}
\label{appendix:distribution}
\vspace{-1mm}

We plan to distribute the \names\ benchmark via our public GitHub: \dataurl. We also include a landing website page on \weburl\ that includes an introduction to the benchmark, links to the relevant papers on multimodal datasets and algorithms, and a public leaderboard to keep track of current progress on these multimodal tasks.

The GitHub and webpage will also allow feedback from the research community in suggesting and adding new datasets and algorithms. Finally, we plan to include a list of planned future updates to \names\ on the webpage along with their target release dates.

\vspace{-1mm}
\subsection{Hosting and Maintenance}
\label{appendix:hosting_maintenance}
\vspace{-1mm}

We have a long-term plan to continue the expansion and maintenance of \names. Here we summarize the main directions we plan to expand towards and leave details and other areas of future work to Appendix~\ref{appendix:future}.
\begin{itemize}
    \item Maintenance: \names\ will be continuously hosted via GitHub which provides stable access to code and a landing page website. We guarantee that \names\ will be available for a long time through our distribution channels. The authors themselves are also actively working on multimodal learning in affective computing, robotics, healthcare, human-computer interaction, and multimedia. The authors are also involved in efforts in applying multimodal machine learning to finance. As a result of these long-term collaborative research efforts, the authors will continue to maintain and expand on the datasets and code provided in \names.
    \item Expansion of datasets: We plan to include more datasets for multimodal fusion as well as more research areas in multimodal learning such as retrieval, question answering, grounding, and reinforcement learning. While these research areas are very different, we hope that insights in multimodal representations can be shared across them.
    \item Expansion of evaluation: To enable holistic evaluation, we plan to build on top of our metrics by adding robustness to distribution shift, uncertainty measures, tests for fairness and social biases, as well as labels/metrics for interpretable multimodal learning.
    \item Expansion of datasets: We plan to encourage students taking the multimodal machine learning course at CMU (\url{https://cmu-multicomp-lab.github.io/mmml-course/fall2020/}) to use the benchmark and add their proposed datasets and models to it.
    \item Expansion of methods: The authors currently collect a very up-to-date reading list of core multimodal papers \url{https://github.com/pliang279/awesome-multimodal-ml} and plan to continuously update \codes\ with new multimodal methods proposed by the community.
\end{itemize}

\vspace{-1mm}
\subsection{Author Statement}
\vspace{-1mm}

The authors carefully reviewed the information present in this document. To the best of our knowledge, the datasets in \names\ can be used for research purposes, following the methodology and licenses described in the dataset section (Appendix~\ref{appendix:dataset_details}).

\vspace{-1mm}
\subsection{License}
\vspace{-1mm}

Each of the datasets included in \names\ includes their own licenses which we detail in Appendix~\ref{appendix:dataset_details}.. We release all preprocessing code across all datasets using the MIT license. All other codes for multimodal algorithms in \codes\, as well as evaluation scripts, are also released via an MIT license: see \url{https://github.com/pliang279/MultiBench/blob/main/LICENSE}, which allows for sharing and distribution of the code for research purposes.

\vspace{-1mm}
\subsection{Metadata}
\vspace{-1mm}

We have included structured metadata for \names\ on our landing page: \weburl.

\vspace{-1mm}
\subsection{Persistence of \names}
\vspace{-1mm}

\names\ is publicly hosted on \dataurl. For larger datasets that cannot be uploaded to GitHub, we plan to upload the processed dataset to CMU Box. We are still exploring the best options for sharing large datasets. Users need to download these processed datasets, place them into a correct folder, and run the \names\ data loader and machine learning pipeline.

\clearpage

\vspace{-2mm}
\section{\names\ Evaluation Protocol}
\label{appendix:eval}
\vspace{-2mm}

To enable holistic evaluation, \names\ offers a comprehensive evaluation methodology to assess (1) generalization across domains and modalities, (2) complexity during training and inference, and (3) robustness to noisy and missing modalities: We describe the evaluation protocol for each desiderata in detail in each of the following subsections:

\vspace{-1mm}
\subsection{Performance}
\vspace{-1mm}

\names\ provides standardized evaluation using metrics designed for each dataset, ranging from MSE and MAE for regression to accuracy, micro \& macro F1-score, and AUPRC for classification on each dataset. To assess for generalization, we compute the variance of a particular model's performance across all datasets in \names\ on which it is tested. We split these results on multiple datasets into \textit{in-domain} datasets and \textit{out-domain} datasets. \textit{In-domain} datasets refer to model performance on datasets that it was initially proposed and tested on, while \textit{out-domain} datasets refer to model performance on the remaining datasets. Comparing out-domain vs in-domain performance, as well as variance in performance across datasets as a whole, allow us to summarize the generalization statistics of each multimodal model.

\vspace{-1mm}
\subsection{Complexity}
\label{appendix:complexity}
\vspace{-1mm}

Modern ML research, unfortunately, causes significant impacts to energy consumption~\citep{strubell2019energy}, a phenomenon often exacerbated in processing high-dimensional multimodal data. As a step towards quantifying energy complexity and recommending lightweight multimodal models, \names\ records the amount of information taken in bits (i.e., data size), number of model parameters, and time and memory resources required during the entire training process. To enforce consistency, the training time measured for all models on each dataset is run on the same CPUs and GPUs. We report training memory by measuring peak memory usage of the python process during the entire training process using python \texttt{memory\_profiler} toolkit (\url{https://pypi.org/project/memory-profiler/}). When counting the number of parameters when training a model, we only count the parameters in persistent modules during training and does not count the ephemeral networks or modules created in the middle of the training process (such as the networks trained for determining weights in \textsc{GradBlend} or the fusion architectures created as part of the architecture search process in \textsc{MFAS}).

In addition to training time and resources, real-world models may need to be small and compact to run on mobile devices~\cite{radu2016towards}. To account for this, \names\ also records \textit{inference time and parameters}. We report inference time by measuring the time it takes for the trained model to complete inference on the entire test set of the dataset. In some cases, only parts of the parameters used in training are counted towards the inference parameters (for example, the parameters in decoders of \textsc{MVAE} and \textsc{MFM} are part of training parameters but not part of inference parameters).

\vspace{-1mm}
\subsection{Robustness to Imperfect Data}
\label{appendix:robustness}
\vspace{-1mm}

Real-world multimodal data is often imperfect as a result of missing entries, noise corruption, or missing modalities entirely. For example, multimodal dialogue systems trained on acted TV shows are susceptible to poor performance when deployed in the real world where users might be less expressive in using facial gestures. This calls for robust models that can still make accurate predictions despite only having access to a (possibly noisy~\citep{liang2019tensor}) subset of signals~\citep{pham2019found}. To standardize efforts in evaluating the robustness of multimodal models, \names\ includes the following robustness tests as part of the evaluation:

\vspace{-1mm}
\subsubsection{Modality-specific Imperfections}
\vspace{-1mm}

Modality-specific imperfections are independently applied to each modality taking into account the unique noise topologies in that source of data (i.e., flips and crops of images, natural misspellings in text, abbreviations in spoken audio). We describe all the modality-specific imperfections we implement in \names\ in the following:

\textbf{Language:} Imperfections in the language modality can occur at various granularities spanning the character, word, phrase, and sentence levels. With reference to~\cite{belinkov2018synthetic}, many of these imperfections occur at the raw text data level and are usually results of spelling errors on a QWERTY keyboard as well as abbreviations in written, typed, and spoken text. Given a word $w$ of length $n$ and a fixed probability $p\in(0,1)$, we implement the following language-specific imperfections:
\begin{enumerate}
    \item Spelling errors: note that spelling mistakes are different from intentionally changed word forms (e.g. abbreviation used in instant messaging service) since they are unintentional~\cite{10.1145/1568296.1568315}. We simulate typos by replacing each letter with a letter having an adjacent position on a QWERTY keyboard with probability $p$.
    \item Short message noise: Short Message Service (SMS) data usually include intentional corruptions of words and phrases like abbreviations, phonetic substitutions, omission of characters and words, and dialectal and informal usages~\cite{10.1145/1568296.1568315}. We implement the following:
    \begin{enumerate}
        \item Simulate sticky keys: given a number $m$, choosing $m$ letters of a word randomly to repeat with probability $p$.
        \item Simulate quick typing: given a number $m$, choosing $m$ letters of a word randomly to omit with probability $p$.
    \end{enumerate}
    \item Random permutation of letters: swapping adjacent two letters is a common natural noise when typing quickly \cite{belinkov2018synthetic}. Random permutation of the entire word or the majority of letters is a form of synthetic noise. We implement the following:
    \begin{enumerate}
        \item Swap two random adjacent letters (except for the first and the last letter) with probability $p$.
        \item Permute the middle chunk of a word: denote the middle chunk (all letters except the first and the last letter) as $w[1:n]$, with probability $p$, produce a permutation $f$ with the first and last letter fixed, i.e. $f(0)=0, f(n)=n$. The shuffled word is $w'$ with $w'[f(i)]=w[i]$ for all $i\in[n]$.
    \end{enumerate}
\end{enumerate}

\textbf{Image:} Given a RGB image $\textbf{X}\in\mathbb{Z}^{W\times H\times 3}$ where $W$ and $H$ are the height and width of the image, let $R,G,B$ be the $W\times H$ matrices of three color channels. We implement the following robustness tests in the image modality:
\begin{enumerate}
    \item Noises in digital images: various noises are naturally prevalent in digital images during image acquisition, coding, transmission, and processing steps~\cite{boyat2015review}. We implement the following:
    \begin{enumerate}
        \item Gaussian/electronic noise that normalizes histogram with respect to the gray values. We add Gaussian noise as a $W\times H$ matrix with each entry following Gaussian distribution $\mathcal{N}(0, p)$.
        \item Impulse valued/salt-and-pepper noise that has dark pixels in bright regions and bright pixels in dark regions. To add salt-and-pepper noise, for each pixel $x \in \textbf{X}$, we convert $x=0$ (white) or $x=255$ (black) into a dead pixel with uniform distribution with probability $p$.
        \item Periodic noise such that it looks like some repeating patterns are exposed on top of the affected image. We add periodic noise by exposing the original image to periodic patterns with probability $p$.
    \end{enumerate}
    \item Color errors:
    \begin{enumerate}
        \item Convert the image to grayscale: $0.3R + 0.59G + 0.11B$ with probability $p$.
        \item Decrease the contrast with probability $p$.
        \item Negate the color: let $\textbf{X'}$ be the inverted image then $\forall i\in[W], j\in[H], k\in[3], \textbf{X}'(i,j,k)=255-\textbf{X}[i,j,k]$ with probability $p$.
        \item Change the white-balance by increasing/decreasing the temperature with probability $p$.
        \item Colorize the image with probability $p$.
    \end{enumerate}
    \item Flips, crops, and rotations:
    \begin{enumerate}
        \item Horizontal flipping with probability $p$.
        \item Color space transformation - isolating a single color channel and changing brightness etc with probability $p$.
        \item Random cropping changes with probability $p$.
        \item Rotate the image by random angle $\in[20,40]$ with probability $p$.
        \item Translation of images to the left, right, up, or down with probability $p$.
    \end{enumerate}
\end{enumerate}
Most of these transformations are achieved with the Python Imaging Library (PIL).

\textbf{Video:} We treat video data as a time series of images. For each image in the video, we apply the image-specific robustness tests as described above. In addition, we also apply the following tests to simulate imperfections in time-series data:
\begin{enumerate}
    \item Random drop: dropping the datapoint at random time step with probability $p$.
    \item Structured drop: given a time step $t$, $m$ consecutive time steps with at least one nonzero signal are dropped with probability $p$.
\end{enumerate}

\textbf{Audio:} Audio is typically represented as a time-series signal. Noises are primarily caused by imperfections in the recording device, which can cause static Gaussian noise to be added to the recorded temporal waveform at random time steps, background noise to be picked up at higher magnitudes, and certain time steps (or consecutive time steps) to be dropped from the recording. We implement the following unimodal noises in the audio modality:
\begin{enumerate}
    \item Additive white Gaussian noise: given an array of length $N$ of a sampled audio segment, we add white gaussian noise, which is an array of $N$ with each entry following a normal distribution with mean $0$ and standard deviation $p$.
\end{enumerate}
In addition to these imperfections applied at a single time step, we also apply the following across the entire time-series signal:
\begin{enumerate}
    \item Random drop: dropping the datapoint at random time step with probability $p$.
    \item Structured drop: given a time step $t$, $m$ consecutive time steps with at least one nonzero signal are dropped with probability $p$.
\end{enumerate}

\textbf{Time-series} data consists of a sequence with a time-dimension (a sequence of data points indexed by time). Following Liang et al.,~\cite{liang2019tensor}, we implement the following types of noise and missing values in time-series data:
\begin{enumerate}
    \item White noise added independently at every time step (noise sampled from zero-mean Gaussian with standard deviation $p$).
    \item Random drop: dropping the datapoint at random time step with probability $p$.
    \item Structured drop: given a time step $t$, $m$ consecutive time steps across modalities are dropped with probability $p$.
\end{enumerate}

\textbf{Optical flow:} We treat optical flow in a similar manner as time-series data and implement the same robustness tests.

\textbf{Force and proprioception sensors:} We also treat these sensors in robotics as time-series data with a key difference - we add noise/drop time steps at a higher frequency since force and proprioception sensors often record data at a higher frequency.

\textbf{Tabular} data takes the form of rows, each of which contains information about some feature (e.g., age, ). We define the following robustness tests on tabular data:
\begin{enumerate}
    \item Random drops of elements from the table with probability $p$.
    \item Random swaps elements in the table with probability $p$.
\end{enumerate}

\textbf{Sets} are data instances where the collection of input elements satisfy permutation invariance, which is in contrast to fixed dimensional vectors that are commonplace in machine learning on images, text, and audio. The key difference between sets and tabular data is that each element in the set is often assumed to be from the same distribution (e.g., a point cloud is a set of 3D coordinates). We define the following types of noise on an input set modality:
\begin{enumerate}
    \item Random dropping of elements from the set with probability $p$.
    \item Adding noise to elements of the set with noise sampled from zero-mean Gaussian with standard deviation $p$.
\end{enumerate}

\vspace{-1mm}
\subsubsection{Multimodal Imperfections}
\vspace{-1mm}

Multimodal imperfections capture correlations in imperfections across modalities (e.g., missing modalities~\cite{pham2019found}, or a chunk of time missing in multimodal time-series data~\citep{liang2019tensor}). These represent settings where data collection across modalities is correlated rather than independent.
\begin{enumerate}
    \item Correlated noise: adding noise to all modalities with probability $p$, where noise is defined according to the aforementioned modality-specific noises.
    \item Correlated drop: dropping all modalities with probability $p$, where dropping patterns are defined according to the aforementioned modality-specific drops.
    \item Temporal drop: in the case of temporal modalities recorded in parallel (e.g., video, audio, and text recorded across time; financial time-series data recorded across days), we perform correlated drops across all modalities at random time steps with probability $p$.
    \item Structured temporal drop: we extend temporal drop such that given a time step $t$, we perform temporal drop on $m$ consecutive time steps with probability $p$.
    \item Missing modalities: dropping an entire modality with probability $p$.
\end{enumerate}

\vspace{-1mm}
\subsubsection{Robustness Measure}
\vspace{-1mm}

We train the model on clean training data and evaluate it under increasing levels of noise added only to test data. To simulate realistic noise and imperfections in test data, we follow the modality-specific and multimodal imperfections as described above. Given a multimodal dataset with $M$ modalities, this allows us to create $M+1$ partitions of imperfect test datasets: one partition of increasing noise levels for modality-specific imperfections within each modality (which gives a total of $M$ partitions) and one partition of multimodal imperfections across all modalities. For datasets where it is not possible to create multimodal imperfections due to the lack of a shared dimension (e.g., image and text datasets typically do not share any correlated dimension, but multimodal time-series datasets share an underlying time dimension), we implement the first $M$ modality-specific imperfections which results in $M$ imperfect data partitions.

\textbf{A qualitative visualization:} Given each test partition, we take a unimodal or multimodal model trained on clean data and plot model performance on the $y$-axis as increasing levels of noise is added to the test data, on a range of $0$ (no noise) to $1$ (complete noise) along the $x$-axis. This allows us to visually inspect the robustness of each model as increasing imperfections are added to the test data. Visually, a robust model should maintain high accuracy (or low MSE) as much as possible despite increasing levels of noise.

\textbf{A quantitative metric:} While the visualization technique above allows one to compare the robustness of several multimodal models across the same dataset, it does not allow us to aggregate robustness performance across the broad range of datasets and tasks in \names. To design such a metric, we extend the quantitative robustness measures proposed in Taori et al.,~\citep{taori2020measuring} to deal with multimodal imperfections across a range of imperfection levels $\sigma \in [0.0, 1.0]$. 

We begin by reviewing the example proposed in Taori et al.,~\citep{taori2020measuring}: suppose we are given two models $f_1$ and $f_2$, where accuracy $\textrm{acc}_\textrm{clean}(f_1) = 0.8$, $\textrm{acc}_\textrm{noisy}(f_2) = 0.75$ (i.e., a $5\%$ drop in accuracy from the imperfections), and $\textrm{acc}_\textrm{clean}(f_2) = 0.9$, $\textrm{acc}_\textrm{noisy}(f_2) = 0.76$ (a $14\%$ drop). Model $f_2$ has higher accuracy on the noisy test set, but overall sees a drop of $14\%$ from the clean to the noisy test set. In contrast, $f_1$ starts off with a lower accuracy but sees only a $5\%$ drop. To capture both these desiderata (i.e., having higher accuracy at \textit{all} levels and \textit{lower drops} in accuracy), Taori et al.,~\citep{taori2020measuring} introduce two notions of robustness: relative and effective robustness.

\textbf{Relative robustness} directly measures accuracy under imperfection. A model with higher relative robustness would display higher accuracy at \textit{all} levels of imperfection compared to a baseline model. We measure the relative robustness of all multimodal models as compared to a baseline \textsc{LF} (simple late fusion with concatenation) method since that is the most basic method tested on all datasets. We compute relative robustness of a model $f$ using the formula
\begin{equation}
    \tau(f) = \int_\sigma \textrm{acc}_\sigma (f) - \textrm{acc}_\sigma (\textsc{LF}) \ \textrm{d} \sigma,
\end{equation}
which essentially measures the area between two performance-imperfection curves as imperfection levels $\sigma$ increase from $0.0$ to $1.0$ (we compute a discrete approximation to the integral).

\textbf{Effective robustness} measures the rate of accuracy drops as imperfection levels increase. However, to reliably measure the rate of accuracy drops, one must remove the confounding variable brought by differences in initial accuracies on clean test data. Taori et al.,~\citep{taori2020measuring} therefore propose to measure whether a model can offer higher accuracy on the noisy test set \textit{beyond what is expected from having higher accuracy on the original test set}. Taori et al., use a log-linear fit on the set of (accuracy on noisy test data, accuracy on clean test data) points across a range of models trained on ImageNet to measure the expected accuracy on noisy test data given a new model's performance on clean test data. Graphically, effective robustness then corresponds to a model's performance on noisy test data lying above the linear trendline. Similar to relative robustness, we measure the effective robustness of multimodal models relative to the accuracy trend of the \textsc{LF} baseline, which we denote as $\beta_{\textrm{LF}}$. We compute effective robustness of a model $f$ using the formula
\begin{equation}
    \rho(f) = \int_\sigma \textrm{acc}_\sigma (f) - \beta_{\textrm{LF}} \left( \textrm{acc}_{0.0} (f) \right) \ \textrm{d} \sigma,
\end{equation}
which essentially measures the area between the performance-imperfection curve of model $f$ and a \textit{shifted} performance-imperfection curve of the \textsc{LF} baseline (shifted to match the initial accuracy of model $f$ at imperfection level $0.0$). A model with higher effective robustness should lie above this shifted accuracy curve at all imperfection levels $\sigma$. Again, we compute a discrete approximation to the integral.

Overall, a robust multimodal model should obtain both high relative and effective robustness.

\vspace{-1mm}
\subsection{Aggregating Measures Across Datasets and Tasks}
\vspace{-1mm}

\names\ benefits from benchmarking multimodal models across a diverse set of datasets, modalities, and tasks. While it is useful to analyze methods on a single dataset in isolation, it is also useful to assess the generalization and failure modes of methods across multiple datasets. Therefore, we need a way to reliably summarize the above metrics (performance, complexity, and robustness) across datasets despite their being on vastly different scales (e.g., accuracy for different numbers of categories) and orders (e.g., accuracy vs RMSE). We find that min-max normalization of results per dataset into a $0-1$ scale (where min and max are appropriately reversed for RMSE/MSE metrics) before averaging across datasets gives a reliable indicator of overall performance across multiple datasets.

\clearpage

\vspace{-2mm}
\section{\codes: A Zoo of Multimodal Algorithms}
\label{appendix:algos}
\vspace{-2mm}

In this section, we provide more details into our choice of standardizing multimodal representation learning as well as the implementation of our standardized library. In each category, we carefully describe the algorithm, motivate its effect in tackling one of the core challenges in section~\ref{appendix:challenges}, and provide references to the original code that we adapted to include in \codes.

\vspace{-1mm}
\subsection{Selection of Algorithms in \codes}
\vspace{-1mm}

We begin by discussing our choices of algorithms in \codes. We consulted with domain experts in each of the application areas to select methods that satisfy the following properties:
\begin{enumerate}
    \item \textit{Diversity in areas:} We chose algorithms that present novel perspectives across a suite of machine learning research domains spanning data preprocessing, fusion paradigms, optimization objectives, and training procedures.  
    \item \textit{Coverage of technical challenges:} Each of the algorithms selected in \codes\ are chosen because they provide unique perspectives to the technical challenges in multimodal learning as elucidated in Appendix~\ref{appendix:challenges}. In Table~\ref{data:models}, we provided a coarse attempt in categorizing each of the technical challenges in multimodal learning. As a result, we did not include too many methods in any category (e.g., multiple methods that are based on model architectures that tackle similar challenges of learning complementary information). Even within the same category and within those tackling the same technical challenge, we attempted to select ones that were fundamentally different (e.g., architectures based on domain knowledge, general-purpose Transformers, and architecture search).
    \item \textit{SOTA on a particular dataset:} For each dataset chosen in \names, we aim to include the model that currently achieves state-of-the-art performance on that dataset. This allows us to assess the best performing model \textit{within} the same domain of the dataset, as well as the best performing model \textit{outside} the domain of the dataset.
    \item \textit{Community expansion:} Any set of initial methods that we will choose will represent only a small sample of the powerful multimodal methods out there. We will encourage community participation in expanding the methods in \codes\ and encourage researchers to implement new methods using a similar modular structure to reduce confounding factors, enable standardized sharing of code, and ensure reproducibility in results.
\end{enumerate}

\begin{table*}[]
\fontsize{9}{11}\selectfont
\setlength\tabcolsep{3.0pt}
\caption{\codes\ provides a standardized implementation of the following multimodal methods to enable accessibility for new researchers and reproducibility of results. These approaches span advances in data processing, fusion paradigms, optimization objectives, and training procedures. We choose these approaches since they offer complementary perspectives towards tacking the fundamental challenges in multimodal fusion: (1) aligning signals across modalities at the right granularity, (2) learning complementary information across aligned signals, and (3) maintaining robustness in the presence of noisy and missing modalities.}
\centering
\footnotesize
\vspace{-0mm}

\begin{tabular}{c|c|c|c|c}
\Xhline{3\arrayrulewidth}
Category & Method & Alignment & Complementarity & Robustness \\
\Xhline{0.5\arrayrulewidth}
\multirow{1}{*}{Data} & \textsc{WordAlign}~\cite{chen2017multimodal} & \textcolor{gg}\cmark & \textcolor{rr}\xmark & \textcolor{rr}\xmark \\
\Xhline{0.5\arrayrulewidth}
\multirow{6}{*}{Model} & \textsc{EF}, \textsc{LF}~\cite{baltruvsaitis2018multimodal} & \textcolor{rr}\xmark & \textcolor{gg}\cmark & \textcolor{rr}\xmark \\
& \textsc{TF}~\citep{zadeh2017tensor}, \textsc{LRTF}~\citep{liu2018efficient} & \textcolor{rr}\xmark & \textcolor{gg}\cmark & \textcolor{rr}\xmark \\
& \textsc{MI-Matrix}, \textsc{MI-Vector}, \textsc{MI-Scalar}~\citep{Jayakumar2020Multiplicative} & \textcolor{rr}\xmark & \textcolor{gg}\cmark & \textcolor{rr}\xmark \\
& \textsc{NL Gate}~\cite{wang2020makes} & \textcolor{rr}\xmark & \textcolor{gg}\cmark & \textcolor{rr}\xmark \\
& \textsc{MulT}~\citep{tsai2019multimodal} & \textcolor{gg}\cmark & \textcolor{gg}\cmark & \textcolor{rr}\xmark \\
& \textsc{MFAS}~\citep{perez2019mfas} & \textcolor{rr}\xmark & \textcolor{gg}\cmark & \textcolor{rr}\xmark \\
\Xhline{0.5\arrayrulewidth}
\multirow{5}{*}{Objective} & \textsc{CCA}~\citep{andrew2013deep} & \textcolor{gg}\cmark & \textcolor{rr}\xmark & \textcolor{rr}\xmark \\
& \textsc{ReFNet}~\cite{sankaran2021multimodal} & \textcolor{gg}\cmark & \textcolor{rr}\xmark & \textcolor{rr}\xmark \\
& \textsc{MFM}~\cite{tsai2019learning} & \textcolor{rr}\xmark & \textcolor{gg}\cmark & \textcolor{rr}\xmark \\
& \textsc{MVAE}~\citep{wu2018multimodal} & \textcolor{rr}\xmark & \textcolor{gg}\cmark & \textcolor{rr}\xmark \\
& \textsc{MCTN}~\citep{pham2019found} & \textcolor{rr}\xmark & \textcolor{rr}\xmark & \textcolor{gg}\cmark \\
\Xhline{0.5\arrayrulewidth}
\multirow{2}{*}{Training} & \textsc{GradBlend}~\citep{wang2020makes} & \textcolor{rr}\xmark & \textcolor{gg}\cmark & \textcolor{gg}\cmark \\
& \textsc{RMFE}~\cite{gat2020removing} & \textcolor{rr}\xmark & \textcolor{gg}\cmark & \textcolor{gg}\cmark \\
\Xhline{0.5\arrayrulewidth}
\Xhline{3\arrayrulewidth}
\end{tabular}

\vspace{-4mm}
\label{data:models}
\end{table*}


\vspace{-1mm}
\subsection{Data Preprocessing}
\vspace{-1mm}

\textbf{Temporal alignment:} As a preprocessing step, performing temporal alignment~\cite{chen2017multimodal} has been shown to help tackle the multimodal alignment problem in the case of time-series data. This approach makes an implicit assumption on the temporal granularity of the modalities (e.g., at the level of words for text) and aligns information from the remaining modalities to the same temporal granularity. We call this approach \textsc{WordAlign}~\cite{chen2017multimodal} and apply it to temporal data with text being one of the modalities. We use the temporal alignment provided in \url{https://github.com/A2Zadeh/CMU-MultimodalSDK}. Specifically, it performs alignment at the granularity of words. Given a sentence with words $w_1, ..., w_T$ each annotated with their start and end times $(s_1, e_1), (s_2, e_2), ..., (s_T, e_T)$, word-level alignment takes the non-text modality features (which are typically extracted at a higher frequency) and averages them during the intervals $e_1-s_1, e_2-s_2, ..., e_T-s_T$. This results in a text sequence of $T$ words alongside aligned non-text modality sequences of $T$ time-steps as well.

\vspace{-1mm}
\subsection{Fusion Paradigms}
\vspace{-1mm}

\textbf{Early and late fusion} have been the de-facto first-approach when tackling new multimodal problems. Early fusion performs concatenation at the input data level before using a suitable prediction model (i.e., $\mathbf{z}_\textrm{mm} = \left[ \mathbf{x}_1, \mathbf{x}_2\right]$) and late fusion applies suitable unimodal models to each modality to obtain their feature representations, concatenates these features, and defines a classifier to the label (i.e., $\mathbf{z}_\textrm{mm} = \left[ \mathbf{z}_1, \mathbf{z}_2\right]$)~\citep{baltruvsaitis2018multimodal}. \codes\ includes their implementations denoted as \textsc{EF} and \textsc{LF} respectively. Since these are basic building blocks in the multimodal learning field, we implement them ourselves.

\textbf{Tensors} are specifically designed to tackle the multimodal complementarity challenge by explicitly capturing higher-order interactions across modalities~\cite{zadeh2017tensor}. Given unimodal representations $\mathbf{z}_1, \mathbf{z}_2$, a multimodal tensor representation is defined as $\mathbf{z}_\textrm{mm} = \begin{bmatrix} \mathbf{z}_{1} \\ 1 \end{bmatrix} \otimes \begin{bmatrix} \mathbf{z}_{2} \\ 1 \end{bmatrix}$ where $\otimes$ denotes an outer product. However, computing tensor products is expensive since their dimension scales exponentially with the number of modalities. Several efficient variants have been proposed to approximate expensive full tensor products with cheaper variants while maintaining performance~\cite{hou2019deep,liang2019tensor,liu2018efficient}. \codes\ includes Tensor Fusion (\textsc{TF})~\citep{zadeh2017tensor} as well as approximate Low-rank Tensor Fusion (\textsc{LRTF})~\citep{liu2018efficient}.

We use the Tensor Fusion implementation in \url{https://github.com/Justin1904/TensorFusionNetworks} and the Low-rank Tensor Fusion implementation in \url{https://github.com/Justin1904/Low-rank-Multimodal-Fusion}. As future work, we also plan to include more expressive higher-order tensor fusion methods~\cite{hou2019deep}.

\textbf{Multiplicative Interactions (MI)} further generalize tensor products to include learnable parameters that capture the interactions between streams of information~\citep{Jayakumar2020Multiplicative}. In its most general form, MI defines a bilinear product $\mathbf{z}_\textrm{mm} = \mathbf{z}_1 \mathbb{W} \mathbf{z}_2 + \mathbf{z}_1^\top \mathbf{U} + \mathbf{V} \mathbf{z}_2 + \mathbf{b}$ where $\mathbb{W}, \mathbf{U}, \mathbf{Z}$, and $\mathbf{b}$ are trainable parameters. By appropriately constraining the rank and structure of these parameters, MI recovers HyperNetworks~\citep{ha2016hypernetworks} (unconstrained parameters resulting in a matrix output), Feature-wise linear modulation (FiLM)~\cite{perez2018film,zhong2019rtfm} (diagonal parameters resulting in vector output), and 
Sigmoid units~\cite{dauphin2017language} (scalar parameters resulting in scalar output). \codes\ includes all $3$ as \textsc{MI-Matrix}, \textsc{MI-Vector}, and \textsc{MI-Scalar} respectively.

Since code was not released for the Multiplicative Interactions paper~\citep{Jayakumar2020Multiplicative}, we implemented the \textsc{MI} layer ourselves. We also referred to the implementation of Feature-wise linear modulation (FiLM)~\cite{perez2018film} from \url{https://github.com/ethanjperez/film} and added it as a module in \names, which we call \textsc{FiLM}. While \textsc{MI-Vector} (i.e., diagonal parameters in a MI layer which results in a vector output) corresponds to the most basic implementation of \textsc{FiLM}, the original \textsc{FiLM} layer uses multiple non-linear layers instead of a single linear transformation in \textsc{MI-Vector} which has been shown to improve performance~\cite{perez2018film}.

\textbf{Gated attention models} are prevalent in learning combinations of two representations that dynamically change for every input~\cite{chaplot2017gated,wang2020makes,xu2015show}. Its general form can be written as $\mathbf{z}_\textrm{mm} = \mathbf{z}_1 \odot h(\mathbf{z}_2)$, where $h$ represents a function with sigmoid activation and $\odot$ denotes the element-wise product. The output $h(\mathbf{z}_2)$ is commonly referred to as ``attention weights'' learned from $\mathbf{z}_2$ used to attend on $\mathbf{z}_1$.

We implement the Query-Key-Value mechanism as \textsc{NL Gate} as proposed in~\cite{wang2020makes} by referring to the implementation of in \url{https://github.com/facebookresearch/VMZ}. This attention mechanism is conceptually similar to the \textsc{MI-Vector} case above but recent work has explored more expressive forms of $h$ such as using a Query-Key-Value mechanism~\cite{wang2020makes} or several fully-connected layers~\citep{chaplot2017gated} rather than a linear transformation in \textsc{MI-Vector}.

\textbf{Temporal attention models} are useful in tackling the challenge of multimodal alignment and complementarity. Transformer models~\citep{vaswani2017attention} have been shown to be useful for temporal multimodal data by automatically aligning and capturing complementary features at different time-steps~\cite{tsai2019multimodal,yao-wan-2020-multimodal}. We include the Multimodal Transformer (\textsc{MulT})~\citep{tsai2019multimodal} which uses a Crossmodal Transformer block that uses $\mathbf{z}_1$ to attend to $\mathbf{z}_2$ (and vice-versa), before concatenating both representations to obtain $\mathbf{z}_\textrm{mm} = \left[ \mathbf{z}_{1 \rightarrow 2}, \mathbf{z}_{2 \rightarrow 1} \right] = \left[ \textsc{CM}(\mathbf{z}_1, \mathbf{z}_2), \textsc{CM}(\mathbf{z}_2, \mathbf{z}_1) \right]$.

We use the public implementation available at \url{https://github.com/yaohungt/Multimodal-Transformer} which includes a basic crossmodal transformer block designed for $2$ modalities. To extend this to $3$ modalities, the crossmodal transformer block is repeated across all $3$ sets of modality pairs (i.e., $\mathbf{z}_\textrm{mm} = \left[ \mathbf{z}_{1 \rightarrow 2}, \mathbf{z}_{2 \rightarrow 1}, \mathbf{z}_{1 \rightarrow 3}, \mathbf{z}_{3 \rightarrow 1}, \mathbf{z}_{2 \rightarrow 3}, \mathbf{z}_{3 \rightarrow 2} \right]$). While this is still computationally feasible for $3$ modalities such as the language, video, and audio datasets that \textsc{MulT} was originally designed for, this quickly becomes intractable for problems involving more than $3$ modalities. To adapt \textsc{MulT} for the financial prediction task involving more than $10$ modalities, we cluster all modalities into $3$ groups based on similarities in their data and perform early fusion on the data within each cluster before applying \textsc{MulT} only on the $3$ clusters of modalities. While \textsc{MulT} is a strong model based on performance, it poses scalability issues that should be the subject of future work (i.e., since the number of cross-modal attention blocks grows quadratically with the number of modalities).

\textbf{Architecture search:} Finally, instead of hand-designing multimodal architectures, several approaches define a set of atomic neural operations (e.g., linear transformation, activation, attention, etc.) and use architecture search to automatically learn the best order of these operations for a given multimodal task~\cite{perez2019mfas,xu2021mufasa}. We focus on the more general approach, \textsc{MFAS}~\citep{perez2019mfas}, designed for language and vision datasets.

We adapt the implementation from \url{https://github.com/juanmanpr/mfas}. While this approach is categorized under innovations in model architecture (since it primarily targets better architectures for multimodal fusion), its code in the \codes\ toolkit is implemented under training structures, since architecture search requires an outer loop to learn model architectures over multiple inner supervised learning loops that train an individual model architecture. Therefore, we are unable to integrate \textsc{MFAS} directly with the basic supervised learning training loops like we do for the other fusion paradigms described above.

\vspace{-1mm}
\subsection{Optimization Objectives}
\vspace{-1mm}

In addition to the standard supervised losses (e.g., cross-entropy for classification, MSE/MAE for regression), several proposed methods have proposed new optimization objectives based on:

\textbf{Prediction-level alignment:} There has been extensive research in defining objective functions to tackle the challenge of multimodal alignment: capturing a representation space where semantically similar concepts from different modalities are close together. While primarily useful for cross-modal retrieval~\cite{liang2020cross,zhen2019deep}, recent work has also shown its utility in learning representations for prediction~\cite{bachman2019learning,cui2020unsupervised,lee2019making,tian2019contrastive}. These alignment objectives have been applied at both prediction and feature levels. In the former, we implement Canonical Correlation Analysis (\textsc{CCA})~\cite{andrew2013deep,wang2015deep}, which computes $\mathcal{L}_\textrm{CCA} = \textrm{corr} \left( g_1(\mathbf{z}_{1}), g_2(\mathbf{z}_{2}) \right)$ where $g_1,g_2$ are auxiliary classifiers mapping each unimodal representation to the label. This method corresponds to prediction-level alignment since they aim to learn representations of each modality that agree on the label, as measured by the correlation of label predictions made by each modality across a batch of samples.

We refer to the paper that most closely implements CCA-based alignment for multimodal data (specifically directly testing on the CMU-MOSI dataset)~\citep{sun2020learning}. Since the authors did not release their code, we implemented it from scratch with reference to CCA implementations from \url{https://github.com/Michaelvll/DeepCCA} and \url{https://github.com/VahidooX/DeepCCA}.

\textbf{Feature-level alignment:} In the latter, contrastive learning has emerged as a popular approach that brings similar concepts close in feature space and different concepts far away~\cite{cui2020unsupervised,lee2019making,tian2019contrastive}. \codes\ includes \textsc{ReFNet}~\cite{sankaran2021multimodal} which includes a self-supervised contrastive loss between unimodal representations $\mathbf{z}_{1}, \mathbf{z}_{2}$ and the multimodal representation $\mathbf{z}_\textrm{mm}$, i.e., $\mathcal{L}_\textrm{contrast} = 1 - \textrm{cos} (\mathbf{z}_\textrm{mm}, g_1(\mathbf{z}_{1})) + 1 - \textrm{cos} (\mathbf{z}_\textrm{mm}, g_2(\mathbf{z}_{2})) $ where $g_1,g_2$ is an auxiliary layer mapping each modality's representation into the joint multimodal space. The intuition here is that the unimodal representations $\mathbf{z}_{1}, \mathbf{z}_{2}$ and the multimodal representation $\mathbf{z}_\textrm{mm}$ should be aligned in the multimodal feature space as measured by cosine similarity. While the original \textsc{ReFNet} method does not use negative samples, closely related work in multi-view contrastive learning has extended this idea to use negative samples which is more closely in line with recent work in contrastive learning~\cite{tian2019contrastive}.

Since they did not release code, we implement \textsc{ReFNet} ourselves on top of current supervised learning modules in \codes.

\textbf{Reconstruction objectives:} Methods based on generative-discriminative models (e.g., VAEs) include an objective to reconstruct the input (or some part of the input)~\cite{lee2019making,tsai2019learning}. These have been shown to better preserve task-relevant information learned in the representation, especially in settings with sparse supervised signals such as robotics~\cite{lee2019making} and long videos~\cite{tsai2019learning}. We include the Multimodal Factorized Model (\textsc{MFM})~\citep{tsai2019learning} which is a general approach that learns a representation $\mathbf{z}_\textrm{mm}$ that can reconstruct input data $\mathbf{x}_{1}, \mathbf{x}_{2}$ while also predicting the label. The multimodal representation is a concatenation of factorized representations $\mathbf{z}_1$, $\mathbf{z}_2$, ..., $\mathbf{z}_M$, and $\mathbf{z}_y$.

Since \textsc{MFM} optimizes a variational lower-bound to the log likelihood, the overall objective consists of $3$ terms - generative, discriminative, and prior regularization:
\begin{equation}
\label{eq:approxmulti}
\begin{split}
&\underset{f_i, f_\textrm{mm}, g_i, g_y}{\mathrm{min}}\,\,\mathbf{E}_{{P}_{\mathbf{x}_{1:M},\mathbf{y}}}\mathbf{E}_{{f_1}({\mathbf{z}_{1}|\mathbf{x}_{1}})} \cdots \mathbf{E}_{{f_M}({\mathbf{z}_{M}| \mathbf{x}_{M}})} \mathbf{E}_{{f_{\textrm{mm}}}({\mathbf{z}_y|\mathbf{x}_{1:M}})}\\
&\left[\sum_{i=1}^M \norm{ \mathbf{x}_i, g_{i} (\mathbf{z}_{i}, \mathbf{z_y}) }_2 + \ell \left(\mathbf{y}, g_y( \mathbf{z_y}) \right) \right]  + \lambda \textrm{MMD}({Q}_{\mathbf{z}}, {P}_{\mathbf{z}}),
\end{split}
\end{equation}
where $f_i$'s are encoders from each modality to representations, $f_\textrm{mm}$ is a multimodal encoder to the joint representation $\mathbf{z}_y$, $g_i$'s are decoders from latent representations back into input data, and $g_y$ is a classification head to the label. The final $\textrm{MMD}$ term is a regularizer to bring the representations close to a unit Gaussian prior. The multimodal encoder $f_\textrm{mm}$ in \textsc{MFM} can be instantiated with any multimodal model from section~\ref{model_design} (e.g., learning $\mathbf{z}_y$ via tensors and adding a term to reconstruct input data). We use the public implementation in \url{https://github.com/pliang279/factorized}, which uses a temporal attention model as $f_\textrm{mm}$ for multimodal time-series data. For the remaining experiments we replace $f_\textrm{mm}$ with a simple late fusion but also run some experiments with multimodal methods that are state-of-the-art in each domain.

\textbf{Improving robustness:} These approaches modify the objective function to account for robustness to noisy~\cite{liang2019tensor} or missing~\cite{lee2020detect,ma2021smil,pham2019found} modalities. \codes\ includes \textsc{MCTN}~\cite{pham2019found} which uses cycle-consistent translation to predict the noisy/missing modality from present ones. The key insight is that a joint representation between modalities $\mathbf{x}_{1}$ and $\mathbf{x}_{2}$ can be learned by using $\mathbf{x}_{1}$ to predict $\mathbf{x}_{2}$, in a vein similar to machine translation or image/text style transfer. \textsc{MCTN} defines a cyclic translation path $\mathbf{x}_{1} \rightarrow \mathbf{z}_\textrm{mm} \rightarrow \hat{\mathbf{x}}_{2} \rightarrow \mathbf{z}_\textrm{mm} \rightarrow \hat{\mathbf{x}}_{1}$ and adds additional reconstruction losses $\mathcal{L}_\textrm{rec} = \norm{ \mathbf{x}_{1} - \hat{\mathbf{x}}_{1} }_2 + \norm{ \mathbf{x}_{2} - \hat{\mathbf{x}}_{2} }_2$ on top of the supervised learning loss. The representations $\mathbf{z}_\textrm{mm}$ learned via translation are then used to predict the label. Surprisingly, the model needs to take in only $\mathbf{x}_{1}$ at test time and is therefore robust to all levels of noise or missingness in $\mathbf{x}_{2}$.

\vspace{-1mm}
\subsection{Training Procedures}
\vspace{-1mm}

\textbf{Improving generalization:} Recent work has found that directly training a multimodal model with all modalities using supervised learning is sub-optimal since different modalities overfit and generalize at different rates. \codes\ includes an approach to solve this, called Gradient Blending (\textsc{GradBlend}), that computes generalization statistics for each modality to determine their weights during multimodal fusion~\citep{wang2020makes}. We use the implementation in \url{https://github.com/facebookresearch/VMZ} and modify it to be part of the \codes\ training structures.

We also include a similar work, Regularization by Maximizing Functional Entropies (\textsc{RMFE}), which uses functional entropy to balance the contribution of each modality to the classification result~\cite{gat2020removing}. We use the public implementation from \url{https://github.com/itaigat/removing-bias-in-multi-modal-classifiers}.

\vspace{-1mm}
\subsection{Domain-specific Methods}
\vspace{-1mm}

Finally, we also implemented several domain-specific methods that had been applied to each domain. These include sensor fusion~\citep{lee2019making} and Kalman filtering~\citep{lee2020multimodal} for robotics, and the multimodal Refiner network~\cite{sankaran2021multimodal} for multimedia experiments. We refer the reader to the respective papers for algorithmic details.

\clearpage

\vspace{-2mm}
\section{Integrating \names\ and \codes: A Brief Tutorial}
\label{appendix:code}
\vspace{-2mm}

\names\ is available via our public GitHub: \dataurl. We also include a landing website page on \weburl\ that includes an introduction to the benchmark, links to the relevant papers on multimodal datasets and algorithms, and a public leaderboard to keep track of current progress on these multimodal tasks. In this section, we provide more details for the loading of datasets ML pipeline provided by \names. We also describe the modular implementation of multimodal models in \codes\ and provide several code examples to illustrate its usage.

\vspace{-1mm}
\subsection{Reading the Dataset}
\vspace{-1mm}

We provide scripts for reading each dataset supported by \codes\ at \texttt{dataset/[dataset\_name]/get\_data.py} in the repository. For each dataset, the user will need to first follow downloading and preprocessing instructions documented in Section~\ref{appendix:dataset_details} or in the comments of the \texttt{get\_data.py}. The python script contains a function (usually called \texttt{get\_dataloader}) that takes in required arguments (such as the location of the preprocessed dataset or compressed data, etc) and it will output a tuple of three PyTorch Dataloader objects for train, valid, and test split of the dataset respectively. You can feed these dataloaders directly into training structures in \codes.

\vspace{-1mm}
\subsection{Unimodal Models}
\vspace{-1mm}

In addition to the multimodal models described in Appendix~\ref{appendix:algos} that are the main subject of study in this area, each dataset and modality typically also requires an initial processing stage either through feature extraction (see Appendix~\ref{appendix:dataset_details} for initial feature extraction done on each dataset) and/or unimodal models on raw data/extracted features.

To standardize the implementation of unimodal models, \codes\ includes an implementation of several standard unimodal models that we encountered when running experiments on the diverse range of datasets and modalities in \names. Each unimodal model is implemented as a function class that takes in either raw data or extracted features from a modality and returns a unimodal representation tensor after applying the function. \codes\ includes the following unimodal methods:
\begin{enumerate}
    \item \textsc{Multi-layer Perceptrons} form the building blocks of many deep learning methods and are generally suitable for any modality that has undergone feature extraction into a vector that does not require any more processing with inductive biases. Their general structure means that they can be flexibly adapted for the tabular, set, and image, and text (e.g., see Deep Averaging Network~\citep{iyyer2015deep}) modalities. They have also been used as a starting point for force and proprioception sensors in robotics if data does not come in the form of time-series~\citep{lee2019making}.
    \item \textsc{Convolutional Networks}~\citep{lecun1995convolutional} are typically used over the image modality. They are also used on the audio modality if an initial preprocessing step of converting raw audio to spectrograms is used.
    \item \textsc{ResNets}~\citep{he2016resnet} are an improvement over ConvNets to enable training of deeper models and have been used extensively for images and audio spectrograms.
    \item \textsc{Recurrent Networks}~\citep{rumelhart1985learning}, \textsc{GRUs}~\citep{chung2014empirical}, and \textsc{LSTMs}~\citep{hochreiter1997long} are suitable for temporal data in the form of text, video, audio, and time-series modalities.
    \item \textsc{Transformers}~\citep{vaswani2017attention} have recently emerged as a strong alternative to recurrent models by using self-attention rather than an accumulative memory. They are also suitable for text, video, audio, and time-series modalities. We also implemented recently proposed \textsc{Vision Transformers}~\citep{dosovitskiy2020image} that adapt Transformer models for image classification as well.
    \item \textsc{Deep Sets}~\citep{zaheer2017deep} was proposed as a permutation-invariant method for machine learning on sets, and was shown to outperform prior methods such as MLPs that are sensitive to the permutation of elements.
    \item Finally, we also included several domain-specific methods that we encountered as we were accumulating the datasets in \names. Some of these methods include \textsc{MaxOut} networks~\citep{pmlr-v28-goodfellow13} used for \textsc{MM-IMDb}~\citep{arevalo2017gated} and \textsc{Causal Convolution}~\cite{oord2016wavenet} for the high-frequency force sensors used in robotics datasets~\citep{lee2019making,lee2020multimodal}.
\end{enumerate}

\vspace{-1mm}
\subsection{Multimodal Models}
\vspace{-1mm}

\codes\ includes an implementation of all multimodal methods described in Appendix~\ref{appendix:algos}. Each multimodal method (i.e., fusion paradigm) is implemented as a Pytorch Module class taking in unimodal tensors and returning final multimodal representation vectors. We implemented several common fusion modules, such as Concatenation, Early-Concatenation (i.e., concatenate in input space), Stack, FilM, Multiplicative-Interactions (MI), Tensor Fusion, LRTF, NL-gate, and more described in Appendix~\ref{appendix:algos}. When the training algorithm requires non-standard multimodal representations (e.g., more than one vector output from fusion module) or the unimodal encoders produce non-standard unimodal representations (i.e., not a single vector representation), special fusion modules will be needed in these situations. For example, we wrote a \texttt{roboticsConcat} module that performs concatenation for the \textsc{Vision\&Touch} dataset due to its non-standard unimodal encoder output. We also have special fusion modules for optimization objectives or training structures such as \textsc{MVAE}, \textsc{MFAS}, and \textsc{GradBlend}. The design of modular fusion modules gives flexibility in model design, as users can reuse a previous fusion module directly in most cases but can also write their own special fusion modules easily.

\vspace{-1mm}
\subsection{Classification Head}
\vspace{-1mm}

Finally, \codes\ includes flexible implementations of classification heads that take in the multimodal representation and return a label either directly (perhaps with some activation) for regression or a softmax over classes for classification.

\vspace{-1mm}
\subsection{Optimization Objectives}
\vspace{-1mm}

The optimization objectives are modules that take in the classification or regression result produced by the model and the ground-truth (as well as other necessary inputs if applicable) and return a loss that can be used to optimize the model based on the desired objective. In most methods we simply use \texttt{torch.nn.CrossEntropyLoss} as the objective for classification tasks and \texttt{torch.nn.MSELoss} as the objective for regression tasks. However, in certain training structures, special objectives are required. For example, \codes\ includes implementations of objective functions such as weighted reconstruction loss and ELBO loss used in reconstruction-based methods \textsc{MFM} and \textsc{MVAE}, and there are also implementations of alignment-based objectives such as CCA and contrastive learning. The final optimization objective returns a weighted sum of these prediction objectives and auxiliary objectives, where the user is free to specify these weights as hyperparameters.

\vspace{-1mm}
\subsection{Training Structures}
\vspace{-1mm}

Training Structures are the main body of \codes\ programs. All other modules (unimodal models, fusion paradigms, optimization objectives, classification heads, etc) can be seen as exchangeable plugins to these training structures. The training structure determines the main training algorithm, with the most common one being \texttt{supervised\_learning} (training unimodal, multimodal, and classification parameters directly for a task-specific supervised learning objective).

More advanced methods may change this training structure either through additional optimization objectives (\textsc{MVAE}~\citep{wu2018multimodal}, \textsc{MFM}~\cite{tsai2019learning}) or via extensions of supervised learning through dynamic weighting of modalities (\textsc{GradBlend}~\citep{wang2020makes}) or an outer architecture search training loop (\textsc{MFAS}~\citep{perez2019mfas}). Each of these methods, therefore, have their own training structure module.

These interchangeable plugin modules give a lot of flexibility in adapting each training structure to new tasks. For example, for the experiments described in Section~\ref{appendix:setup}, the methods that are primarily based on different fusion paradigms (i.e., \textsc{EF}, \textsc{LF}, \textsc{TF}, \textsc{LRTF}, \textsc{MI}, \textsc{NL-Gate}, \textsc{MulT} etc all use the same training structure (\texttt{supervised\_learning}) with different plugin fusion modules (and different unimodal encoders and heads based on datasets and tasks). Similarly, while most of these more advanced training structures were originally paired with a simple \textsc{LF} model in their original papers, our modular implementation makes it possible to combine advances in fusion paradigms with training structures in future work.

\vspace{-1mm}
\subsection{Performance Evaluation}
\vspace{-1mm}

We standardize evaluation using metrics designed for each dataset, ranging from MSE and MAE for regression to accuracy, micro \& macro F1-score, and AUPRC for classification. We use the standard PyTorch and scikit-learn implementations of these performance metrics.

\vspace{-1mm}
\subsection{Complexity Evaluation}
\vspace{-1mm}

We report training memory by measuring peak memory usage of the python process during the entire training process using python \texttt{memory\_profiler} toolkit (\url{https://pypi.org/project/memory-profiler/}). When counting the number of parameters when training a model, we only count the parameters in persistent modules during training and does not count the ephemeral networks or modules created in the middle of the training process (such as the networks trained for determining weights in \textsc{GradBlend} or the fusion architectures created as part of the architecture search process in \textsc{MFAS}).

\vspace{-1mm}
\subsection{Robustness Evaluation}
\vspace{-1mm}

For robustness experiments, modality-specific and multimodal imperfections are implemented as modules. A separate version of data loader is created for each dataset to test robustness, which adds custom unimodal or multimodal imperfections of increasing noise levels $\sigma \in [0,1]$ to the original clean test set. A testing module is also provided specifically for robustness experiments, which evaluates the model on increasing levels of noisy test datasets and prints out the metrics for visualization. In this way, \textsc{MultiZoo} allows highly modular data loading and robustness evaluation that requires minimal modification to the regular training and testing workflow.

\textsc{MultiZoo} includes evaluation protocols summarizing these robustness results. It includes visualization functions of the performance-imperfection curves across datasets in \names. We also implemented relative and effective robustness as two quantitative metrics for robustness evaluation. For relative robustness, we approximate the area under the performance-imperfection curves for each model across \names\ datasets. For effective robustness, we take the performance-imperfection curve of \textsc{LF} evaluated on the same dataset equalized for initial accuracy on clean test data. For both metrics, we normalized performance across all models evaluated on the same dataset.

\vspace{-1mm}
\subsection{Code Snippets}
\vspace{-1mm}

\begin{algorithm}[tb]
   \caption{PyTorch code integrating \names\ datasets and \codes\ models.}
   \label{alg:code_supp}
   
    \definecolor{codeblue}{rgb}{0.25,0.5,0.5}
    \lstset{
      basicstyle=\fontsize{7.2pt}{7.2pt}\ttfamily\bfseries,
      commentstyle=\fontsize{7.2pt}{7.2pt}\color{codeblue},
      keywordstyle=\fontsize{7.2pt}{7.2pt},
    }
\begin{lstlisting}[language=python]
from datasets.get_data import get_dataloader
from unimodals.common_models import ResNet, Transformer
from fusions.common_fusions import MultInteractions
from training_structures.gradient_blend import train, test

# loading Multimodal IMDB dataset
traindata, validdata, testdata = get_dataloader('multimodal_imdb')
out_channels = 3
# defining ResNet and Transformer unimodal encoders
encoders = [ResNet(in_channels=1, out_channels, layers=5),
            Transformer(in_channels=1, out_channels, layers=3)]
# defining a Multiplicative Interactions fusion layer
fusion = MultInteractions([out_channels*8, out_channels*32], out_channels*32, 'matrix')
classifier = MLP(out_channels*32, 100, labels=23)
# training using Gradient Blend algorithm
model = train(encoders, fusion, classifier, traindata, validdata, 
        epochs=100, optimtype=torch.optim.SGD, lr=0.01, weight_decay=0.0001)
# testing
performance, complexity, robustness = test(model, testdata)
\end{lstlisting}
\vspace{-2mm}
\end{algorithm}

In Algorithm~\ref{alg:code_supp}, we show a sample code snippet in Python that loads a dataset from \names\ (Appendix~\ref{appendix:dataset_details}), defines the unimodal and multimodal architectures, optimization objectives, and training procedures (Appendix~\ref{appendix:algos}), before running the evaluation protocol (Appendix~\ref{appendix:eval}). Our \codes\ toolkit is easy to use and trains entire multimodal models in less than $10$ lines of code. By standardizing the implementation of each module and disentangling the individual effects of models, optimizations, and training, \codes\ ensures accessibility and reproducibility of its multimodal algorithms.

\clearpage

\vspace{-2mm}
\section{Experimental Setup}
\label{appendix:setup}
\vspace{-2mm}

In this section, we provide additional details of the experimental setup. All experiments were conducted on a server with $4 \times$ Nvidia GTX $980$ Ti GPUs, $5 \times$ Nvidia Tesla P$40$ GPUs, $2 \times$ Nvidia Tesla K$40$c GPUs, $4 \times$ Nvidia TITAN X GPUs, $1 \times$ Tesla T$4$ GPU, and $1 \times$ Tesla V$100$ GPU. The server also contained $32 \times$ Intel(R) Xeon(R) CPU (E$5-2670$, $2.60$GHz).

\vspace{-1mm}
\subsection{Affective Computing}
\vspace{-1mm}

\begin{table}[t]
\fontsize{8.5}{11}\selectfont
\centering
\caption{Table of hyperparameters for prediction on affective computing dataset.\vspace{2mm}}
\setlength\tabcolsep{3.5pt}
\begin{tabular}{l | l | c | c}
\Xhline{3\arrayrulewidth}
\textbf{Component} & \textbf{Model} & \textbf{Parameter} & \textbf{Value} \\
\Xhline{0.5\arrayrulewidth}
\multirow{4}{*}{GRU Encoder} & \multirow{4}{*}{GRU}
& Input sizes & $[5, 20, 35, 74, 300, 704]$ \\
& & Hidden sizes & $[32, 32, 64, 128, 512, 1024]$ \\
& & Num of layers & $1$ or $2$ \\
& & Dropout & $0.0$ or $0.1$ \\
\Xhline{0.5\arrayrulewidth}
\multirow{4}{*}{Transformer Encoder~\citep{vaswani2017attention}} & \multirow{4}{*}{\shortstack[l]{Transformer~\citep{vaswani2017attention}}}
& Input sizes & $[5, 20, 35, 74, 300, 704]$ \\
& & Hidden sizes & $[5, 10, 20, 40, 40, 50]$ \\
& & Num heads & $2$ or $3$ \\
& & Dropout & $0.2$ \\
\Xhline{0.5\arrayrulewidth}
\multirow{4}{*}{Head} & \multirow{4}{*}{MLP}
& Input sizes & $[5, 20, 32, 64, 128, 256]$ \\
& & Hidden sizes & $[5, 20, 32, 64, 128, 256]$ \\
& & Num layers & $2$ \\
& & Dropout & $0.2$ \\
\Xhline{0.5\arrayrulewidth}
\multirow{4}{*}{\textsc{MCTN}~\cite{pham2019found} Encoder} & \multirow{4}{*}{GRU}
& Input sizes & $300$ \\
& & Hidden sizes & $[32, 64]$ \\
& & Num of layers & $1$ or $2$ \\
& & Dropout & $0.0$ or $0.1$ \\
\Xhline{0.5\arrayrulewidth}
\multirow{4}{*}{\textsc{MCTN}~\cite{pham2019found} Decoder} & \multirow{4}{*}{GRU}
& Input sizes & $[32, 64]$ \\
& & Hidden sizes & $300$ \\
& & Num of layers & $1$ or $2$ \\
& & Dropout & $0.0$ or $0.1$ \\
\Xhline{0.5\arrayrulewidth}
\multirow{3}{*}{\textsc{MCTN}~\cite{pham2019found} Seq2Seq} & \multirow{3}{*}{GRU+GRU}
& teaching ratio & $0.5$ \\
& & Embed sizes & $32$ \\
& & ${\mu}_{t_1}, {\mu}_{c}, {\mu}_{t_2}$ & $0.01$ \\
\Xhline{0.5\arrayrulewidth}
\multirow{5}{*}{Fusion} & \multirow{2}{*}{\textsc{LRTF}~\cite{liu2018efficient}}
& Num ranks & $64$ \\
& & Output sizes & $128$ \\
\Xcline{2-4}{0.5\arrayrulewidth}
& \multirow{1}{*}{\textsc{MI-Matrix}~\cite{Jayakumar2020Multiplicative}}
& Hidden size & $128$ \\
\Xcline{2-4}{0.5\arrayrulewidth}
& \multirow{2}{*}{\textsc{MulT}~\cite{tsai2020multimodal}}
& Hidden size & $40$ \\
& & Num heads & $8$ or $10$ \\
\Xhline{0.5\arrayrulewidth}
\multirow{10}{*}{Training}
& & Loss & MAE or Cross Entropy \\
& & Batch size & $32$ \\
& & Seq Length & $50$ or $20$ \\
& & Num epochs & $100$ or $300$ \\
& & Early stop & True \\
& & Patience & $[8, 20]$ \\
& & Activation & ReLU \\
& & Optimizer & AdamW \\
& & Weight Decay & $1 \times 10^{-4}$ \\
& & Learning rate & $1 \times 10^{-4}$ \\
\Xhline{3\arrayrulewidth}
\end{tabular}
\vspace{-4mm}
\label{affect_params}
\end{table}

\textbf{Hyperparameters:} We show the hyperparameters used for models on datasets in the Affective Computing domain in Table~\ref{affect_params}. For each dataset we tune the following hyperparameters selected from the following ranges: the learning rate is selected between $0.00001$ to $0.001$ and set to be $0.0001$ in the beginning; Early stopping is applied with patience 8 to $20$ before overfitting happens; The input sizes and hidden sizes vary according to the different modalities and datasets. The ${\mu}_{t_0}$, ${\mu}_{c}$, and ${\mu}_{t_1}$ hyperparameters in \textsc{MCTN}~\cite{pham2019found} is tuned between $0.005$ to $0.1$. The sequence length varies from $20$ to $50$. Only punchline sentences (target sentences) are used in \textsc{UR-Funny}~\citep{hasan2019ur} and \textsc{MUsTARD}~\citep{castro2019towards} following the original papers.

Hyperparameters were selected based on performance on the validation set. For models that had been previously proposed and tested on these datasets, we use the same hyperparameters as those reported in their paper or public code.

All experiments were repeated $10$ times and a mean and standard deviation was computed.

\vspace{-1mm}
\subsection{Healthcare}
\vspace{-1mm}

\begin{table}[t]
\fontsize{8.5}{11}\selectfont
\centering
\caption{Table of hyperparameters for prediction on MIMIC dataset in the healthcare domain.\vspace{2mm}}
\setlength\tabcolsep{3.5pt}
\begin{tabular}{l | l | c | c}
\Xhline{3\arrayrulewidth}
\textbf{Component} & \textbf{Model} & \textbf{Parameter} & \textbf{Value} \\
\Xhline{0.5\arrayrulewidth}
\multirow{2}{*}{Static Encoder} & \multirow{2}{*}{2-layer MLP}
& Hidden sizes & $[10,10]$ \\
& & Activation & LeakyReLU(0.2) \\
\Xhline{0.5\arrayrulewidth}
\multirow{2}{*}{Static Decoder} & \multirow{2}{*}{2-layer MLP}
& Layer sizes & $[200,40,5]$ \\
& & Activation & LeakyReLU(0.2) \\
\Xhline{0.5\arrayrulewidth}
\multirow{1}{*}{Time Series Encoder} & \multirow{1}{*}{GRU}
& Hidden dim & $30$ \\\Xhline{0.5\arrayrulewidth}
\multirow{1}{*}{Time Series Decoder} & \multirow{1}{*}{GRU}
& Hidden dim & $30$ \\
\Xhline{0.5\arrayrulewidth}
\multirow{2}{*}{Classification Head} & \multirow{2}{*}{2-Layer MLP} & Hidden size & $40$ \\
& & Activation & LeakyReLU($0.2$) \\
\Xhline{0.5\arrayrulewidth}
\multirow{6}{*}{Fusion} & \multirow{2}{*}{LRTF~\cite{liu2018efficient}} & Output dim & $100$ \\
& & Ranks & $40$ \\
\cline{2-4}
& \multirow{3}{*}{NL-Gate~\cite{wang2020makes}} & thw-dim/c-dim/tf-dim & $24/30/10$ \\
& & key linear & $[10,300]$ \\
& & value linear & $[10,300]$ \\
\cline{2-4}
& \multirow{1}{*}{MI-Matrix~\cite{Jayakumar2020Multiplicative}} & output dim & $100$ \\

\Xhline{0.5\arrayrulewidth}
\multirow{40}{*}{Training}
& \multirow{5}{*}{\begin{tabular}[c]{@{}l@{}}Unimodal, LF, LRTF,\\ MI-Matrix, NL-gate\end{tabular}} & Loss & Cross Entropy \\
& & Batch size & $40$ \\
& & Num epochs & $20$ \\
& & Optimizer & RMSprop \\
& & Learning rate & $0.001$ \\
\cline{2-4}
& \multirow{8}{*}{\textsc{GradBlend}~\cite{wang2020makes}} & Loss & Cross Entropy \\
& & Batch size & $40$ \\
& & Num epochs & $300$ \\
& & Optimizer & SGD \\
& & Learning Rate & $0.005$ \\
& & GB-epoch & $20$ \\
& & v-rate & $0.8$ \\
& & finetune epoch & $25$ \\
\cline{2-4}
& \multirow{7}{*}{MVAE~\cite{wu2018multimodal}} & Loss & Cross Entropy + ELBO\\
& & Batch size & $40$ \\
& & Num epochs & $30$ \\
& & Optimizer & Adam \\
& & Learning Rate & $0.001$ \\
& & Cross Entropy Weight & $2.0$ \\
& & Latent Representation Fusion & ProductOfExpert \\
\cline{2-4}
& \multirow{8}{*}{MFM~\cite{tsai2019learning}} & Loss 
& {\begin{tabular}[c]{@{}c@{}} Cross Entropy \\\ + Reconstruction(MSE) \end{tabular}} \\
& & Batch size & $40$ \\
& & Num epochs & $30$ \\
& & Optimizer & Adam \\
& & Learning Rate & $0.001$ \\
& & Recon Loss Modality Weights & $[1,1]$ \\
& & Cross Entropy Weight & $2.0$ \\
& & Intermediate Modules & {\begin{tabular}[c]{@{}c@{}} MLPs $[200,100,100]$, \\\ $[200,100,100],[400,100,100]$ \end{tabular}} \\
\cline{2-4}
& \multirow{12}{*}{MFAS~\citep{perez2019mfas}} & Batch size & $32$ \\
& & Epochs/search iters & $3/3/6$ \\
& & Num samples/surrogates per epoch & $15/50$ \\
& & $\eta$ max/min/Ti/Tm & $10^{-3}/10^{-6}/1/2$ \\
& & Temperature init/final/decay & $10.0/0.2/4.0$ \\
& & Max progression level & $4$ \\
& & Surrogate learning rate & $0.001$ \\
& & Surrogate hidden size & $100$ \\
& & Surrogate embedding size & $100$ \\
& & Search space & $(3,3,2)$ \\
& & Optimizer & Adam \\
& & Representation Size & $16$ \\
\Xhline{3\arrayrulewidth}
\end{tabular}
\vspace{-4mm}
\label{health_params}
\end{table}

We show the hyperparameters used for models on datasets in the Healthcare domain in Table~\ref{health_params}. The unimodal architectures follow the original paper that created this partition of \textsc{MIMIC}~\citep{PURUSHOTHAM2018112}, then we tune the following hyperparameters selected from the following ranges: Learning rate is tuned between $0.1$ and $0.0001$; the number of epochs is selected based on when overfitting happens; for hyperparameters specific to architectures or training structures (such as \textsc{GradBlend}, \textsc{MFAS}), we followed the same configuration as the original papers where these methods are proposed.
 
All experiments were repeated $10$ times and a mean and standard deviation was computed.

\vspace{-1mm}
\subsection{Robotics}
\vspace{-1mm}

\begin{table}[t]
\fontsize{8.5}{11}\selectfont
\centering
\caption{Table of hyperparameters for prediction on \textsc{MuJoCo Push} dataset in the robotics domain.\vspace{2mm}}
\setlength\tabcolsep{3.5pt}
\begin{tabular}{l | l | c | c}
\Xhline{3\arrayrulewidth}
\textbf{Component} & \textbf{Model} & \textbf{Parameter} & \textbf{Value} \\
\Xhline{0.5\arrayrulewidth}
\multirow{1}{*}{Pos Encoder} & \multirow{1}{*}{Linear}
& Hidden sizes & $[64, 64, 64\text{ (residual)}]$ \\
\Xhline{0.5\arrayrulewidth}
\multirow{1}{*}{Sensors Encoder} & \multirow{1}{*}{Linear}
& Hidden sizes & $[64, 64, 64\text{ (residual)}]$ \\
\Xhline{0.5\arrayrulewidth}
\multirow{4}{*}{Image Encoder} & \multirow{4}{*}{CNN}
& Filter sizes & $[5, 3, 3, 3, 3]$ \\
& & Num filters & $[32, 32, 32, 16, 8]$ \\
& & Filter strides & $1$ \\
& & Filter padding & $[2, 1, 1, 1, 1]$ \\
\Xhline{0.5\arrayrulewidth}
\multirow{1}{*}{Control Encoder} & \multirow{1}{*}{Linear}
& Hidden sizes & $[64, 64, 64\text{ (residual)}]$ \\
\Xhline{0.5\arrayrulewidth}
\multirow{6}{*}{Fusion} & \multirow{2}{*}{\shortstack[l]{\\Early Fusion \&\\ Unimodal LSTM}}
& Hidden size & $512$ \\
& & Num layers & $2$ \\
\Xcline{2-4}{0.5\arrayrulewidth}
& \multirow{2}{*}{Late Fusion LSTM}
& Hidden size & $256$ \\
& & Num layers & $1$ \\
\Xcline{2-4}{0.5\arrayrulewidth}
& \multirow{2}{*}{\textsc{MulT}~\cite{tsai2020multimodal}}
& Embed size & $64$ \\
& & Num heads & $4$ \\
\Xhline{0.5\arrayrulewidth}
\multirow{1}{*}{Classification Head} & \multirow{1}{*}{Linear} & \multirow{1}{*}{Hidden size} & \multirow{1}{*}{$64$} \\
\Xhline{0.5\arrayrulewidth}
\multirow{6}{*}{Training}
& & Loss & Mean Squared Error \\
& & Batch size & $32$ \\
& & Num epochs & $20$ \\
& & Activation & ReLU \\
& & Optimizer & Adam \\
& & Learning rate & $10^{-5}$ \\
\Xhline{3\arrayrulewidth}
\end{tabular}
\vspace{-2mm}
\label{robotics_params_gentle_push}
\end{table}

\begin{table}[t]
\fontsize{8.5}{11}\selectfont
\centering
\caption{Table of hyperparameters for prediction on \textsc{Vision\&Touch} dataset in the robotics domain.\vspace{2mm}}
\setlength\tabcolsep{3.5pt}
\begin{tabular}{l | l | c | c}
\Xhline{3\arrayrulewidth}
\textbf{Component} & \textbf{Model} & \textbf{Parameter} & \textbf{Value} \\
\Xhline{0.5\arrayrulewidth}
\multirow{4}{*}{Image Encoder} & \multirow{4}{*}{CNN}
& Filter sizes & $[7, 5, 5, 3, 3, 3]$ \\
& & Num filters & $[16, 32, 64, 64, 128, 128]$ \\
& & Filter strides & $[2, 2, 2, 2, 2, 2]$ \\
& & Filter padding & Same \\
\Xhline{0.5\arrayrulewidth}
\multirow{4.2}{*}{Force Encoder} & \multirow{4.2}{*}{\shortstack[l]{Causal Convolution\\\cite{oord2016wavenet}}}
& Filter sizes & $[2, 2, 2, 2, 2]$ \\
& & Num filters & $[16, 32, 64, 128, 256]$ \\
& & Filter strides & $[2, 2, 2, 2, 2]$ \\
& & Filter padding & $1$ \\
\Xhline{0.5\arrayrulewidth}
\multirow{1}{*}{Proprio Encoder} & \multirow{1}{*}{Linear}
& Hidden sizes & $[32, 64, 128, 256]$ \\
\Xhline{0.5\arrayrulewidth}
\multirow{4}{*}{Depth Encoder} & \multirow{4}{*}{CNN}
& Filter sizes & $[3, 3, 4, 3, 3, 3]$ \\
& & Num filters & $[32, 64, 64, 64, 128, 128]$ \\
& & Filter strides & $[2, 2, 2, 2, 2, 2]$ \\
& & Filter padding & Same \\
\Xhline{0.5\arrayrulewidth}
\multirow{1}{*}{Action Encoder} & \multirow{1}{*}{Linear}
& Hidden sizes & $[32, 32]$ \\
\Xhline{0.5\arrayrulewidth}
\multirow{2}{*}{Classification Head} & \multirow{2}{*}{2-Layer MLP} & Hidden size & $128$ \\
& & Activation & LeakyReLU($0.2$) \\
\Xhline{0.5\arrayrulewidth}
\multirow{3}{*}{Fusion} & \multirow{2}{*}{LRTF~\cite{liu2018efficient}} & Output dim & $200$ \\
& & Ranks & $40$ \\
\Xcline{2-4}{0.5\arrayrulewidth}
& \multirow{1}{*}{Sensor Fusion~\cite{lee2019making}} & z-dim & $128$ \\
\Xhline{0.5\arrayrulewidth}
\multirow{13}{*}{Training}
& & Loss & {\begin{tabular}[c]{@{}c@{}} Contact: Cross Entropy \\\ End-Effector: MSE  \end{tabular}} \\
& & Batch size & $64$ \\
& & Num epochs & {\begin{tabular}[c]{@{}c@{}} Sensor Fusion: $50$ \\\ LRTF: $35$; Others: $15$ \end{tabular}} \\
& & Optimizer & Adam \\
& & Learning rate & {\begin{tabular}[c]{@{}c@{}} Contact: $10^{-4}$ \\\ End-Effector: $5\times 10^{-4}$ \end{tabular}} \\
\cline{2-4}
& \multirow{4}{*}{\textsc{RefNet} ~\cite{sankaran2021multimodal}} & Loss & Cross Entropy + Contrast \\
& & Batch size & $40$ \\
& & Optimizer/Learning Rate & Adam / $0.0005$ \\
& & Refiner & MLP($1056,2000,65760$)\\
& & Self Loss Weight & $0.0001$\\
\Xhline{3\arrayrulewidth}
\end{tabular}
\vspace{-2mm}
\label{robotics_params_vision_touch}
\end{table}

We show the hyperparameters used for \textsc{MuJoCo Push} in Table~\ref{robotics_params_gentle_push} and \textsc{Vision\&Touch} in Table~\ref{robotics_params_vision_touch}.

For \textsc{MuJoCo Push}, we follow hyperparameters and preprocessing in the original paper~\citep{lee2020multimodal}. Unimodal modules follow the original hyperparameters assigned to the input modality.

For \textsc{Vision\&Touch}, we follow hyperparameters in the original paper~\citep{lee2019making} for all unimodal modules as well as Sensor Fusion (which is the method proposed in~\citep{lee2019making}).

All other hyperparameters were selected based on performance on the validation set. For models that had been previously proposed and tested on these datasets, we use the same hyperparameters as those reported in their paper or public code. The original \textsc{Vision\&Touch} dataset did not have a unique test dataset, so we report their best performance on the validation set instead.

All experiments were repeated $10$ times and a mean and standard deviation was computed.

\vspace{-1mm}
\subsection{Finance}
\vspace{-1mm}

\begin{table}[t]
\fontsize{8.5}{11}\selectfont
\centering
\caption{Table of hyperparameters for stock prediction on finance datasets (we use the same hyperparameters on all $3$ datasets: \textsc{Stocks-F\&B}, \textsc{Stocks-Health}, and \textsc{Stocks-Tech}).\vspace{2mm}}
\setlength\tabcolsep{3.5pt}
\begin{tabular}{l | c | c}
\Xhline{3\arrayrulewidth}
\textbf{Model} & \textbf{Parameter} & \textbf{Value} \\
\Xhline{0.5\arrayrulewidth}
\multirow{1}{*}{\shortstack[l]{\\Unimodal \&\\ Early Fusion LSTM}} & \multirow{2}{*}{Hidden dim} & \multirow{2}{*}{$128$} \\
& & \\
\Xhline{0.5\arrayrulewidth}
Late Fusion LSTM & Hidden dim & $16$ \\
\Xhline{0.5\arrayrulewidth}
\multirow{3}{*}{\textsc{Transformer~\citep{vaswani2017attention}}}
& Embed dim & $9$ \\
& Num heads & $3$ \\\
& Layers & $3$ \\
\Xhline{0.5\arrayrulewidth}
\multirow{3}{*}{\textsc{MulT~\cite{tsai2019multimodal}}}
& Embed dim & $9$ \\
& Num heads & $3$ \\\
& Layers & $3$ \\
\Xhline{0.5\arrayrulewidth}
\textsc{GradBlend~\cite{wang2020makes}} LSTM & Hidden dim & $128$ \\
\Xhline{0.5\arrayrulewidth}
\multirow{8.5}{*}{Training}
& Loss & Mean Squared Error \\
& Batch size & $16$ \\
& Max seq length & $500$ \\
& Activation & ReLU \\
& Optimizer & Adam \\
& Learning rate & $10^{-3}$ \\
\Xcline{2-3}{0.5\arrayrulewidth}
& Num epochs &
    \begin{tabular}{l|c}
        Unimodal, EF & $2$ \\
        \Xhline{0.5\arrayrulewidth}
        \multirow{1}{*}{\shortstack[l]{\\LF, Transformer,\\\textsc{MulT}, \textsc{GradBlend}}} & \multirow{2}{*}{$4$} \\
        & \\
    \end{tabular} \\
\Xhline{3\arrayrulewidth}
\end{tabular}
\vspace{-4mm}
\label{finance_params}
\end{table}

We show the hyperparameters used for models on datasets in the Finance domain in Table~\ref{finance_params}. For each dataset, we tune the following hyperparameters selected from the following ranges: Hidden/embed dim ($4-512$), Transformer/\textsc{MulT} layers ($1-4$), Transformer/\textsc{MulT} heads ($1-4$), epochs ($1-32$), and batch size ($4-128$). Hyperparameters were selected based on performance on the validation set. Note that this dataset overfits quickly when model complexity is increased; several hyperparameters are kept small for this reason.

All experiments were repeated $10$ times and a mean and standard deviation was computed.

\vspace{-1mm}
\subsection{HCI}
\vspace{-1mm}


\begin{table}[t]
\fontsize{8.5}{11}\selectfont
\centering
\caption{Table of hyperparameters for prediction on \textsc{ENRICO} dataset in the HCI domain.\vspace{2mm}}
\setlength\tabcolsep{3.5pt}
\begin{tabular}{l | c | c}
\Xhline{3\arrayrulewidth}
\textbf{Model} & \textbf{Parameter} & \textbf{Value} \\
\Xhline{0.5\arrayrulewidth}
\multirow{1}{*}{\shortstack[l]{\\Unimodal}} & \multirow{2}{*}{Hidden dim} & \multirow{2}{*}{$16$} \\
& & \\
\Xhline{0.5\arrayrulewidth}
Late Fusion & Hidden dim & $32$ \\
\Xhline{0.5\arrayrulewidth}
\textsc{GradBlend~\cite{wang2020makes}} & Hidden dim & $32$ \\
\Xhline{0.5\arrayrulewidth}
\textsc{RefNet} ~\cite{sankaran2021multimodal} & Hidden dim & $32$ \\
\Xhline{0.5\arrayrulewidth}
\multirow{2}{*}{MI-Matrix~\cite{Jayakumar2020Multiplicative}}
& Hidden dim & $32$ \\
& Input dims & $16, 16$ \\
\Xhline{0.5\arrayrulewidth}
\multirow{2}{*}{Tensor Matrix}
& Hidden dim & $32$ \\
& Input dims & $16, 16$ \\
\Xhline{0.5\arrayrulewidth}
\multirow{3}{*}{LRTF~\cite{liu2018efficient}}
& Hidden dim & $32$ \\
& Input dims & $16, 16$ \\
& Rank & $20$ \\
\Xhline{0.5\arrayrulewidth}
CCA~\citep{sun2020learning} & Hidden dim & $32$ \\
\Xhline{0.5\arrayrulewidth}
\multirow{8.5}{*}{Training}
& Loss & Class-weighted Cross Entropy \\
& Batch size & $32$ \\
& Activation & ReLU \\
& Dropout & $0.2$ \\
& Optimizer & Adam \\
& Learning rate & $10^{-5}$ \\
\Xcline{2-3}{0.5\arrayrulewidth}
& Num epochs & $50$ \\
\Xhline{3\arrayrulewidth}
\end{tabular}
\vspace{-4mm}
\label{hci_params}
\end{table}

We show the hyperparameters used for models on the \textsc{ENRICO} dataset in the HCI domain in Table~\ref{hci_params}. We tune the learning rate by starting from $10^{-4}$, the value reported in the original paper~\cite{leiva2020enrico}. We searched in a range between $10^{-2}$ and $10^{-6}$ and found that $10^{-5}$ led to the best performance.
We tested hidden dimension sizes from $8$ to $128$ and found that a size of $16$ was sufficient for the unimodal encoders.
Note that this dataset is small and overfits quickly when model complexity is increased.
We minimized the risk of overfitting by keeping several hyperparameters (\textit{e.g.,} hidden dim) small.
For more information, refer to the dataset preprocessing section for \textsc{ENRICO}.

All experiments were repeated $10$ times and a mean and standard deviation was computed.

\vspace{-1mm}
\subsection{Multimedia}
\vspace{-1mm}

\begin{table}[t]
\fontsize{8.5}{11}\selectfont
\centering
\vspace{-20mm}
\caption{Table of hyperparameters for prediction on \textsc{AV-MNIST} dataset in the multimedia domain.\vspace{1mm}}
\setlength\tabcolsep{3.5pt}
\begin{tabular}{l | l | c | c}
\Xhline{3\arrayrulewidth}
\textbf{Component} & \textbf{Model} & \textbf{Parameter} & \textbf{Value} \\
\Xhline{0.5\arrayrulewidth}
\multirow{4}{*}{Image Encoder} & \multirow{4}{*}{LeNet-3}
& Filter Sizes & $[5,3,3,3]$ \\
& & Num Filters & $[6,12,24,48]$ \\
& & Filter Strides / Filter Paddings & $[1,1,1,1]$ /$[2,1,1,1]$  \\
& & Max Pooling & $[2,2,2,2]$ \\
\Xhline{0.5\arrayrulewidth}
\multirow{3}{*}{Image Decoder} & \multirow{3}{*}{DeLeNet-3}
& Filter Sizes & $[4,4,4,8]$ \\
& & Num Filters & $[24,12,6,3]$ \\
& & Filter Strides / Filter Paddings & $[2,2,2,4]$/$[1,1,1,1]$\\
\Xhline{0.5\arrayrulewidth}
\multirow{4}{*}{Audio Encoder} & \multirow{4}{*}{LeNet-5}
& Filter Sizes & $[5,3,3,3,3,3]$ \\
& & Num Filters & $[6,12,24,48,96,192]$ \\
& & Filter Strides / Filter Paddings & $[1,1,1,1,1,1]$/$[2,1,1,1,1,1]$ \\
& & Max Pooling & $[2,2,2,2,2,2]$ \\
\Xhline{0.5\arrayrulewidth}
\multirow{3}{*}{Audio Decoder} & \multirow{3}{*}{DeLeNet-5}
& Filter Sizes & $[4,4,4,4,4,8]$ \\
& & Num Filters & $[96,48,24,12,6,3]$ \\
& & Filter Strides / Filter Paddings & $[2,2,2,2,2,4]$/$[1,1,1,1,1,1]$ \\
\Xhline{0.5\arrayrulewidth}
\multirow{2}{*}{Classification Head} & \multirow{2}{*}{2-Layer MLP} & Hidden size & $100$ \\
& & Activation & LeakyReLU($0.2$) \\
\Xhline{0.5\arrayrulewidth}
\multirow{3}{*}{Fusion} & \multirow{2}{*}{LRTF~\cite{liu2018efficient}} & Output dim & $120$ \\
& & Ranks & $40$ \\
\cline{2-4}
& \multirow{1}{*}{MI-Matrix~\cite{Jayakumar2020Multiplicative}} & output dim & $240$ \\

\Xhline{0.5\arrayrulewidth}
\multirow{41}{*}{Training}
& \multirow{4}{*}{\begin{tabular}[c]{@{}l@{}}Unimodal, LF, \\\ LRTF, MI-Matrix\end{tabular}} & Loss & Cross Entropy \\
& & Batch size & $40$ \\
& & Num epochs & LRTF: $30$, Others: $25$ \\
& & Optimizer/Learning rate/weight decay & SGD/$0.05$/$0.0001$ \\
\cline{2-4}
& \multirow{6}{*}{\textsc{GradBlend}~\cite{wang2020makes}} & Loss & Cross Entropy \\
& & Batch size & $40$ \\
& & Num epochs & $300$ \\
& & Optimizer/Learning rate & SGD/$0.05$ \\
& & GB-epoch/finetune-epoch & $10$/$25$ \\
& & v-rate & $0.8$ \\
\cline{2-4}
& \multirow{6}{*}{MVAE~\cite{wu2018multimodal}} & Loss & Cross Entropy + ELBO\\
& & Batch size & $40$ \\
& & Num epochs & $20$ \\
& & Optimizer/Learning rate & Adam/$0.001$ \\
& & Cross Entropy Weight & $2.0$ \\
& & Latent Representation Fusion & ProductOfExpert \\
\cline{2-4}
& \multirow{7}{*}{MFM~\cite{tsai2019learning}} & Loss & {\begin{tabular}[c]{@{}c@{}} Cross Entropy \\\ + Reconstruction(MSE) \end{tabular}} \\
& & Batch size & $40$ \\
& & Num epochs & $25$ \\
& & Optimizer/Learning rate & Adam/$0.001$ \\
& & Recon Loss Modality Weights & $[1,1]$ \\
& & Cross Entropy Weight & $2.0$ \\
& & Intermediate Modules & {\begin{tabular}[c]{@{}c@{}} MLPs $[200,100,100]$, \\\ $[200,100,100],[400,100,100]$ \end{tabular}} \\
\cline{2-4}
& \multirow{11}{*}{MFAS~\citep{perez2019mfas}} & Batch size & $32$ \\
& & Main epochs/search iters/epochs per model & $3/3/6$\\
& & Num samples/surrogates per epoch & $15/50$ \\
& & $\eta$ max/min/Ti/Tm & $10^{-3}$/$10^{-6}$/ $1/2$ \\
& & Temperature init/final/decay & $10.0/0.2/4.0$ \\
& & Max progression level & $4$ \\
& & Surrogate learning rate & $0.001$ \\
& & Surrogate hidden/embedding size & $100/100$ \\
& & Search space & $(3,5,2)$ \\
& & Optimizer & Adam \\
& & Representation Size & $16$ \\
\cline{2-4}
& \multirow{3}{*}{CCA~\citep{sun2020learning}} & Batch size & $800$ \\
& & Loss & CCALoss \\
& & Optimizer/Learning Rate/Weight Decay & AdamW/ $0.01/0.01$ \\
\cline{2-4}
& \multirow{5}{*}{\textsc{RefNet} ~\cite{sankaran2021multimodal}} & Loss & Cross Entropy + Contrast \\
& & Batch size & $40$ \\
& & Optimizer/Learning Rate & SGD / $0.05$ \\
& & Refiner & MLP($384,1000,13328$)\\
& & Self Loss Weight & $0.1$\\
\Xhline{3\arrayrulewidth}
\end{tabular}
\vspace{-4mm}
\label{multimedia_params1}
\end{table}

\begin{table}[t]
\fontsize{8.5}{11}\selectfont
\centering
\vspace{-10mm}
\caption{Table of hyperparameters for prediction on \textsc{MM-IMDb} dataset in the multimedia domain.\vspace{2mm}}
\setlength\tabcolsep{3.5pt}
\begin{tabular}{l | l | c | c}
\Xhline{3\arrayrulewidth}
\textbf{Component} & \textbf{Model} & \textbf{Parameter} & \textbf{Value} \\
\Xhline{0.5\arrayrulewidth}
\multirow{3}{*}{Text Encoder} &  \multirow{3}{*}{2-Layer MaxoutMLP} & Hidden size & $512$\\
& & Output dim & $128/256/512$\\
& & MLP num & $2$\\
\hline
\multirow{3}{*}{Image Encoder} &  \multirow{3}{*}{2-Layer MaxoutMLP} & Hidden size & $1024$\\
& & Output dim & $128/256/512$\\
& & MLP num & $2$\\
\hline
\multirow{5}{*}{Classification Head} & Linear & &\\
\cline{2-4}
& \multirow{2}{*}{2-Layer MLP} & Hidden size & $512$\\
& & Activation& ReLU\\
\cline{2-4}
& \multirow{2}{*}{2-Layer Maxout\_Linear} & Hidden size & $512$ \\
& & MLP num & $2$ \\
\hline
\multirow{4}{*}{Fusion} & Concatenate & & \\
\cline{2-4}
& \multirow{2}{*}{LRTF~\cite{liu2018efficient}} & Output dim & $512$\\
& & Ranks & $128$\\
\cline{2-4}
& \multirow{1}{*}{MI-Matrix~\cite{Jayakumar2020Multiplicative}} & output dim & $1024$ \\
\hline
\multirow{43}{*}{Training} & \multirow{8}{*}{\begin{tabular}[c]{@{}c@{}}Unimodal, EF, LF, \\LRTF, MI-Matrix\end{tabular}} & Loss & Binary Cross Entropy\\
& & Batch size & $128$\\
& & Num epochs & \begin{tabular}[c]{@{}l@{}}Text: $125$, Image: $25$, LF:$5$, \\EF/LRTF:$15$, MI-Matrix:$20$ \end{tabular}\\
& & Optimizer & AdamW \\
& & Learning rate & \begin{tabular}[c]{@{}l@{}}Unimodal: $0.0001$, EF: $0.04$, \\LF/LRTF/MI-Matrix: $0.008$ \end{tabular}\\
& & Weight decay & $0.01$\\
\cline{2-4}
& \multirow{7}{*}{CCA~\citep{sun2020learning}} & Loss & Binary Cross Entropy + CCA\\
& & CCA weight & $0.001$ \\
& & Batch size & $800$\\
& & Num epochs & $20$\\
& & Optimizer & AdamW \\
& & Learning rate & $0.01$\\
& & Weight decay & $0.01$\\
\cline{2-4}
& \multirow{8}{*}{RMFE~\cite{gat2020removing}} & Loss &\begin{tabular}[c]{@{}c@{}} Binary Cross Entropy \\+ Regularization\end{tabular}\\
& & Regularization weight & $1e-10$ \\
& & Batch size & $128$\\
& & Num epochs & $10$\\
& & Optimizer & AdamW \\
& & Learning rate & $0.01$\\
& & Weight decay & $0.01$ \\
\cline{2-4}
& \multirow{9}{*}{\textsc{RefNet}~\cite{sankaran2021multimodal}} & Loss &\begin{tabular}[c]{@{}c@{}} Binary Cross Entropy \\+ Contrast + Self-supervised\end{tabular}\\
& & Contrast weight & $0.0001$\\
& & Self-supervised weight & $0.1$\\
& & Batch size & $128$ \\
& & Num epochs & $10$\\
& & Optimizer & AdamW \\
& & Learning rate & $0.01$ \\
& & Weight decay & $0.01$ \\
\cline{2-4}
& \multirow{10}{*}{MFM~\cite{tsai2019learning}} & Loss &\begin{tabular}[c]{@{}c@{}} Binary Cross Entropy \\+ Reconstruction(MSE)\end{tabular}\\
& & Batch size & $128$ \\
& & Num epochs & $10$\\
& & Optimizer & Adam\\
& & Learning rate & $0.005$\\
& & Recon Loss Modality Weight & $[1,1]$\\
& & Cross Entropy Weight & $2.0$\\
& & Intermediate Modules & \begin{tabular}[c]{@{}c@{}} MLP $[512,256,256]$ \\ MLP $[512,256,256]$\\MLP $[1024,512,256]$\end{tabular}\\
\Xhline{3\arrayrulewidth}
\end{tabular}
\vspace{-4mm}
\label{multimedia_params2}
\end{table}

\begin{table}[t]
\fontsize{8.5}{11}\selectfont
\centering
\caption{Table of hyperparameters for prediction on \textsc{Kinetics} dataset in the multimedia domain.\vspace{2mm}}
\setlength\tabcolsep{3.5pt}
\begin{tabular}{l | l | c | c}
\Xhline{3\arrayrulewidth}
Component &Model & Parameter & Value \\
\Xhline{0.5\arrayrulewidth}
\multirow{2}{*}{Video Encoder} &  \multirow{2}{*}{ResNet \citep{he2016resnet} + LSTM}
& ResNet Version & $18$-layer \\
& & LSTM Hidden size & $64$\\
\hline
\multirow{4}{*}{Audio Encoder} &  \multirow{4}{*}{ResNet \citep{he2016resnet} + $2$-Layer MLP}
& ResNet Version & $50$-layer\\
& & MLP hidden size & $200$ \\
& & MLP output size & $64$ \\
& & MLP activation & ReLU \\
\hline
\multirow{3}{*}{Classification Head} & Linear & &\\
\cline{2-4}
& \multirow{2}{*}{2-Layer MLP} & Hidden size & $200$\\
& & Activation & ReLU\\
\hline
Fusion & Concatenate & & \\
\hline
\multirow{5}{*}{Training} & \multirow{5}{*}{Unimodal, LF} & Loss & Cross Entropy\\
& & Batch size & $16$\\
& & Num epochs & $15$ \\
& & Optimizer & Adam \\
& & Learning rate & $0.0001$ \\
\Xhline{3\arrayrulewidth}
\end{tabular}
\vspace{-4mm}
\label{multimedia_params3}
\end{table}

We show the hyperparameters used for models on datasets in the Multimedia domain in Tables~\ref{multimedia_params1},~\ref{multimedia_params2},~\ref{multimedia_params3}.

For \textsc{AV-MNIST}, used the same LeNet unimodal encoders following current work~\citep{vielzeuf2018centralnet}. We tuned learning rates between $0.1$ and $0.001$. The default batch size is $40$, although it can be changed in some methods (such as \textsc{CCA}) to make sure the methods work as intended; the number of epochs is selected based on when overfitting happens; for hyperparameters specific to architectures or training structures (such as \textsc{GradBlend}, \textsc{MFAS}), we followed the same configuration as the original papers where these methods are proposed. 

For \textsc{MM-IMDb}, used the same MaxoutLinear unimodal encoders following current work~\citep{arevalo2017gated}. Learning rates were tuned between $0.1$ and $0.001$ except for unimodal training. The default batch size is $128$ while that for \textsc{CCA} is $800$ to make sure the methods work as intended. The number of epochs was selected based on early stopping with patience equal to $7$, which means if the macro F1 on the validation set did not improve for $7$ epochs, training was stopped early.

For \textsc{Kinetics}, we use a ResNet-LSTM for the visual modality encoder and the architectures described by Wang et al \citep{wang2020makes} for the rest of the models. We use a learning rate of $0.0001$, batch size of $16$, and $15$ epochs for the small dataset experiments. For the large dataset experiments, we used the setup described by Wang et al \citep{wang2020makes}.

Hyperparameters were selected based on performance on the validation set. For models that had been previously proposed and tested on these datasets, we use the same hyperparameters as those reported in their paper or public code.

All experiments were repeated $10$ times and a mean and standard deviation was computed.

\clearpage

\vspace{-2mm}
\section{Experimental Results}
\label{appendix:results}
\vspace{-2mm}

In this section, we provide additional experimental results and observations. For all experimental tables, we describe the accuracy metrics using Acc$(c)$ where $c$ is the number of classes. AUPRC stands for the area under the precision-recall curve which is a useful performance metric for imbalanced data in settings where one cares a lot about finding positive examples. MSE stands for mean squared error. We use up and down arrows ($\uparrow$ and $\downarrow$) to indicate metrics where higher is better (Acc, AUPRC) and metrics where lower is better (MSE) respectively.

\vspace{-1mm}
\subsection{Affective Computing}
\vspace{-1mm}

\begin{table*}[]
\fontsize{9}{11}\selectfont
\setlength\tabcolsep{6.0pt}
\caption{Results on multimodal datasets in the affective computing domain. \textbf{U}: unimodal models, \textbf{M}: multimodal fusion paradigms, \textbf{O}: optimization objectives, \textbf{T}: training structures. \textsc{MulT} is the best performing model on all these datasets, and is categorized as an in-domain method since it was originally proposed and tested on affect recognition datasets. Many out-domain methods struggle on these datasets.}
\centering
\footnotesize
\vspace{-0mm}
\begin{tabular}{l|l|c|c|c|c}
\Xhline{3\arrayrulewidth}
& Dataset & \multicolumn{1}{c|}{\textsc{MUStARD}} & \multicolumn{1}{c|}{\textsc{CMU-MOSI}} & \multicolumn{1}{c|}{\textsc{UR-FUNNY}} & \multicolumn{1}{c}{\textsc{CMU-MOSEI}} \\
& Metric & Acc$(2)$ $\uparrow$ & Acc$(2)$ $\uparrow$ & Acc$(2)$ $\uparrow$ & Acc$(2)$ $\uparrow$ \\
\Xhline{0.5\arrayrulewidth}
\multirow{3}{*}{\textbf{U}} & Unimodal ($\ell$) & $68.6\pm0.4$ & $74.2\pm0.5$ & $58.3\pm0.2$ & $ 78.8\pm1.5$ \\
& Unimodal ($a$) &  $ 64.9\pm0.4 $ & $65.5\pm0.2  $ & $ 57.2\pm0.9  $ & $ 66.4\pm0.7 $ \\
& Unimodal ($v$) & $ 65.7\pm0.7 $  & $ 66.3\pm0.3  $ & $ 57.3\pm0.5  $ & $ 67.2\pm0.4 $ \\
\Xhline{0.5\arrayrulewidth}
\multirow{8}{*}{\textbf{M}} & \textsc{EF-GRU} & $ 66.3\pm0.3 $ & $ 73.2\pm2.2 $ & $ 60.2\pm0.5 $ & $ 78.4\pm0.6 $ \\
& \textsc{LF-GRU} & $ 66.1\pm0.9 $  & $ 75.2\pm0.8 $ & $ 62.5\pm0.5 $ & $79.2\pm0.4$ \\
& \textsc{EF-Transformer} & $ 65.3\pm1.4 $  & $ 78.8\pm0.4 $ & $ 62.9\pm0.2 $ & $ 79.6\pm0.3 $ \\
& \textsc{LF-Transformer} & $ 66.1\pm0.9 $ & $ 79.6\pm0.4 $ & $ 63.4\pm0.3 $ & $ 80.6\pm 0.3$ \\
& \textsc{TF}~\cite{zadeh2017tensor} & $ 62.1\pm 2.2 $ & $ 74.4\pm0.2 $ &  $61.2\pm0.4 $ & $79.4\pm0.5$ \\
& \textsc{LRTF}~\cite{liu2018efficient} & $ 65.2\pm 1.5 $ & $ 76.3\pm0.3 $ &  $62.7\pm0.2 $ & $79.6\pm0.6$ \\
& \textsc{MI-Matrix}~\cite{Jayakumar2020Multiplicative} & $ 61.8\pm 0.3$ & $73.9\pm0.4 $ &  $61.9\pm0.3$ & $ 76.5\pm 0.4$\\
& \textsc{MulT}~\cite{tsai2019multimodal} &  $ \mathbf{71.8\pm0.3} $ & $ \mathbf{83.0\pm0.1} $ & $\mathbf{66.7\pm0.3}$ & $\mathbf{82.1\pm0.5} $ \\

\Xhline{0.5\arrayrulewidth}
\multirow{3}{*}{\textbf{O}} & \textsc{MFM}~\cite{tsai2019learning} & $ 66.3\pm0.3 $ & $78.1\pm0.9 $ & $62.4\pm1.1$ & $ 79.4\pm0.7 $\\
& \textsc{MVAE}~\cite{wu2018multimodal} & $ 64.5\pm0.4 $ & $77.2\pm0.3 $ &  $62.0\pm0.5$ & $ 79.1\pm0.2 $\\
& \textsc{MCTN}~\cite{pham2019found} & $ 63.2\pm1.4$  & $76.9\pm2.1 $ & $63.2\pm0.8$ & $ 76.4\pm0.4 $ \\

\Xhline{0.5\arrayrulewidth}
\multirow{1}{*}{\textbf{T}} & \textsc{GradBlend}~\cite{wang2020makes} & $ 66.1\pm0.3 $  & $75.5\pm0.5 $ & $62.3\pm0.3$ & $ 78.1\pm0.3 $ \\

\Xhline{3\arrayrulewidth}
\end{tabular}
\vspace{-2mm}
\label{results:affect_supp}
\end{table*}

\begin{table*}[]
\fontsize{9}{11}\selectfont
\setlength\tabcolsep{3.0pt}
\vspace{-20mm}
\caption{Complexity results for datasets in the affective computing domain.  \textbf{U}: unimodal models, \textbf{M}: multimodal fusion paradigms, \textbf{O}: optimization objectives, \textbf{T}: training structures.}
\centering
\footnotesize
\vspace{-1mm}

\begin{tabular}{l|l|cccccc}
\Xhline{3\arrayrulewidth}
& Dataset & \multicolumn{6}{c}{\textsc{MUStARD}} \\
& Metric & \begin{tabular}[c]{@{}c@{}}Epochs \\ trained\end{tabular} & \begin{tabular}[c]{@{}c@{}}Training \\ time (s)\end{tabular} & \begin{tabular}[c]{@{}c@{}}Training \\ params (M)\end{tabular} & \begin{tabular}[c]{@{}c@{}}Training peak \\ memory (MB)\end{tabular} & \begin{tabular}[c]{@{}c@{}}Inference \\ time (s)\end{tabular} & \begin{tabular}[c]{@{}c@{}}Inference \\ params (M) \end{tabular} \\
\Xhline{0.5\arrayrulewidth}
\multirow{3}{*}{\textbf{U}} & Unimodal ($\ell$) & $43$ & $381$ & $0.12$ & $2347$ & $0.33$ & $0.12$ \\
& Unimodal ($v$) & $48$ & $56$ & $0.01$ & $2288$ & $0.24$ & $0.01$ \\
& Unimodal ($a$) & $69$ & $288$ & $0.001$ & $2288$ & $0.25$ & $0.001$ \\
\Xhline{0.5\arrayrulewidth}
\multirow{7}{*}{\textbf{M}} & \textsc{EF-GRU} & $126$ & $168$ & $0.84$ & $2291$ & $0.34$ & $0.84$ \\
& \textsc{LF-GRU} & $74$ & $52$ & $1.52$ & $2307$ & $0.40$ & $1.52$ \\ 
& \textsc{EF-Transformer} & $30$ & $601$ & $1.86$ & $2423$ & $0.79$ & $1.86$ \\
& \textsc{LF-Transformer} & $42$ & $1868$ & $14.0$ & $2586$ & $1.02$ & $14.0$ \\
& \textsc{TF}~\citep{zadeh2017tensor} & $46$ & $1370$ & $14.7$ &  $2542$ &  $1.62$ &  $14.7$ \\
& \textsc{LRTF}~\citep{liu2018efficient} & $33$ & $49$ & $0.68$ & $2483$ & $0.50$ & $0.68$ \\
& \textsc{MulT}~\cite{tsai2019multimodal} & $31$ & $2414$ & $1.93$ & $3345$ & $3.01$ & $1.93$ \\

\Xhline{0.5\arrayrulewidth}
\multirow{3}{*}{\textbf{O}} & \textsc{MFM}~\cite{tsai2019learning} & $40$ & $2138$ & $4.85$ &  $2417$ & $1.48$ & $4.33$ \\
& \textsc{MVAE}~\citep{wu2018multimodal} & $33$ & $4645$ & $4.32$ & $2695$ & $2.11$ & $4.05$ \\
& \textsc{MCTN}~\cite{pham2019found} & $100$ & $1026$ & $0.19$ & $2359$ & $1.02$ & $0.19$ \\

\Xhline{0.5\arrayrulewidth}
\multirow{1}{*}{\textbf{T}} & \textsc{GradBlend}~\cite{wang2020makes} & $100$ & $6012$ &$1.95$ &$2406$ & $0.42$ & $1.58$ \\

\Xhline{2\arrayrulewidth}
\end{tabular}

\vspace{1mm}

\begin{tabular}{l|l|cccccc}
\Xhline{3\arrayrulewidth}
& Dataset & \multicolumn{6}{c}{CMU-MOSI} \\
& Metric & \begin{tabular}[c]{@{}c@{}}Epochs \\ trained\end{tabular} & \begin{tabular}[c]{@{}c@{}}Training \\ time (s)\end{tabular} & \begin{tabular}[c]{@{}c@{}}Training \\ params (M)\end{tabular} & \begin{tabular}[c]{@{}c@{}}Training peak \\ memory (MB)\end{tabular} & \begin{tabular}[c]{@{}c@{}}Inference \\ time (s)\end{tabular} & \begin{tabular}[c]{@{}c@{}}Inference \\ params (M) \end{tabular} \\
\Xhline{0.5\arrayrulewidth}
\multirow{3}{*}{\textbf{U}} & Unimodal ($\ell$) & $30$ & $590$ & $0.17$ & $2347$ & $0.49$ & $0.17$ \\
& Unimodal ($v$) & $35$ & $71$ & $0.01$ & $2288$ & $0.36$ & $0.01$\\
& Unimodal ($a$) & $188$ & $346$ & $0.001$ & $2288$ & $0.38$ & $0.001$ \\

\Xhline{0.5\arrayrulewidth}
\multirow{7}{*}{\textbf{M}} & \textsc{EF-GRU} & $106$ & $221$ & $1.42$ & $2291$ & $0.44$ & $1.42$ \\ 
& \textsc{LF-GRU} & $14$ & $60$ & $1.84$ & $2307$ & $0.58$ & $1.84$ \\ 
& \textsc{EF-Transformer} & $20$ & $635$ & $2.18$ & $2423$ & $1.07$ & $2.18$ \\
& \textsc{LF-Transformer} & $33$ & $2011$ & $15.1$ & $2586$ & $2.12$ & $15.1$ \\
& \textsc{TF}~\citep{zadeh2017tensor} & $35$ & $384$ & $12.2$ &  $2867$ &  $2.38$ &  $12.2$ \\
& \textsc{LRTF}~\citep{liu2018efficient} & $43$ & $172$ & $0.82$ & $2454$ & $0.59$ & $0.82$ \\
& \textsc{MulT}~\cite{tsai2019multimodal} & $22$ & $2414$ & $2.38$ & $3345$ & $4.30$ & $2.38$ \\

\Xhline{0.5\arrayrulewidth}
\multirow{3}{*}{\textbf{O}} & \textsc{MFM}~\cite{tsai2019learning} & $31$ & $1692$ & $5.53$ & $2455$ & $1.52$ & $4.98$ \\
& \textsc{MVAE}~\citep{wu2018multimodal} & $35$ & $3820$ & $5.31$ & $2564$ & $2.03$ & $4.69$ \\
& \textsc{MCTN}~\cite{pham2019found} & $100$ & $1149$ & $0.19$ & $2366$ & $0.98$ & $0.19$ \\

\Xhline{0.5\arrayrulewidth}
\multirow{1}{*}{\textbf{T}} & \textsc{GradBlend}~\cite{wang2020makes} & $300$ & $18869$ & $3.91$ & $2355$ & $0.59$ & $1.86$ \\
\Xhline{2\arrayrulewidth}

\end{tabular}

\vspace{1mm}

\begin{tabular}{l|l|cccccc}
\Xhline{3\arrayrulewidth}
& Dataset & \multicolumn{6}{c}{\textsc{UR-FUNNY}} \\
& Metric & \begin{tabular}[c]{@{}c@{}}Epochs \\ trained\end{tabular} & \begin{tabular}[c]{@{}c@{}}Training \\ time (s)\end{tabular} & \begin{tabular}[c]{@{}c@{}}Training \\ params (M)\end{tabular} & \begin{tabular}[c]{@{}c@{}}Training peak \\ memory (MB)\end{tabular} & \begin{tabular}[c]{@{}c@{}}Inference \\ time (s)\end{tabular} & \begin{tabular}[c]{@{}c@{}}Inference \\ params (M) \end{tabular} \\
\Xhline{0.5\arrayrulewidth}
\multirow{3}{*}{\textbf{U}} & Unimodal ($\ell$) & $32$ & $602$ & $1.99$ & $6524$ & $1.82$ & $1.99$ \\
& Unimodal ($v$) & $29$ & $70$ & $0.14$ & $6528$ & $1.61$ & $0.14$ \\
& Unimodal ($a$) & $40$ & $1039$ & $0.03$ & $6599$ & $1.66$ & $0.03$ \\

\Xhline{0.5\arrayrulewidth}
\multirow{7}{*}{\textbf{M}} & \textsc{EF-GRU} & $34$& $612$ & $3.58$ & $6535$ & $2.51$ & $3.58$ \\
& \textsc{LF-GRU} & $10$ & $498$ & $2.28$ & $6791$ & $3.25$ & $2.28$ \\
& \textsc{EF-Transformer} & $32$ & $2358$ & $4.87$ & $7086$ & $3.81$ & $4.87$ \\
& \textsc{LF-Transformer} & $33$ & $6024$ & $34.5$ & $7288$ & $6.75$ & $34.5$ \\
& \textsc{TF}~\citep{zadeh2017tensor} & $32$ & $2780$ & $21.3$ & $7165$ & $6.35$ & $21.3$ \\
& \textsc{LRTF}~\citep{liu2018efficient} & $25$ & $2057$ & $1.05$ & $6931$ & $3.32$ & $1.05$ \\
& \textsc{MulT}~\cite{tsai2019multimodal} & $30$ & $8096$ & $5.01$ & $9572$ & $12.1$ & $5.01$ \\

\Xhline{0.5\arrayrulewidth}
\multirow{2}{*}{\textbf{O}} & \textsc{MFM}~\cite{tsai2019learning} & $30$ & $5123$ & $6.89$ &  $6970$ & $10.3$ & $6.23$ \\
& \textsc{MVAE}~\citep{wu2018multimodal} & $32$ & $10670$ & $6.59$ & $7038$ & $12.1$ & $6.10$ \\
& \textsc{MCTN}~\cite{pham2019found} & $100$ & $10857$ & $0.19$ & $6578$ & $4.39$ & $0.19$ \\

\Xhline{0.5\arrayrulewidth}
\multirow{1}{*}{\textbf{T}} & \textsc{GradBlend}~\cite{wang2020makes} & $100$ & $19212$ &$4.12$ &$6832$ & $3.42$ & $2.31$ \\
\Xhline{2\arrayrulewidth}

\end{tabular}

\vspace{1mm}

\begin{tabular}{l|l|cccccc}
\Xhline{3\arrayrulewidth}
& Dataset & \multicolumn{6}{c}{\textsc{CMU-MOSEI}} \\
& Metric & \begin{tabular}[c]{@{}c@{}}Epochs \\ trained\end{tabular} & \begin{tabular}[c]{@{}c@{}}Training \\ time (s)\end{tabular} & \begin{tabular}[c]{@{}c@{}}Training \\ params (M)\end{tabular} & \begin{tabular}[c]{@{}c@{}}Training peak \\ memory (MB)\end{tabular} & \begin{tabular}[c]{@{}c@{}}Inference \\ time (s)\end{tabular} & \begin{tabular}[c]{@{}c@{}}Inference \\ params (M) \end{tabular} \\
\Xhline{0.5\arrayrulewidth}
\multirow{3}{*}{\textbf{U}} & Unimodal ($\ell$) & $23$ & $561$ & $1.80$ & $5830$ & $1.79$ & $1.80$ \\
& Unimodal ($v$) & $27$ & $647$ & $0.12$ & $5817$ & $1.46$ & $0.12$ \\
& Unimodal ($a$) & $39$ & $910$ & $0.03$ & $5818$ & $1.48$ & $0.03$ \\

\Xhline{0.5\arrayrulewidth}
\multirow{7}{*}{\textbf{M}} & \textsc{EF-GRU} & $22$ & $548$ & $3.23$ & $5835$ & $2.01$ & $3.23$ \\
& \textsc{LF-GRU} & $9$ & $443$ & $2.08$ & $5996$ & $2.55$ & $2.08$ \\
& \textsc{EF-Transformer} & $30$ & $1658$ & $4.49$ & $6082$ & $2.88$ & $4.49$ \\
& \textsc{LF-Transformer} & $35$ & $5504$ & $31.5$ & $6996$ & $5.65$ & $31.5$ \\
& \textsc{TF}~\citep{zadeh2017tensor} & $30$ & $2784$ & $22.6$ & $6337$ & $5.89$  & $22.6$ \\
& \textsc{LRTF}~\citep{liu2018efficient} & $22$ & $2057$ & $0.78$ & $6102$ & $2.45$ & $0.78$ \\
& \textsc{MulT}~\cite{tsai2019multimodal} & $32$ & $6033$ & $4.75$ & $7572$ & $10.1$ & $4.75$ \\

\Xhline{0.5\arrayrulewidth}
\multirow{2}{*}{\textbf{O}} & \textsc{MFM}~\cite{tsai2019learning} & $33$ & $5340$ & $6.65$ &  $6088$ & $9.42$ & $5.97$ \\
& \textsc{MVAE}~\citep{wu2018multimodal} & $40$ & $11673$ & $6.21$ & $6782$ & $12.0$ & $5.89$ \\
& \textsc{MCTN}~\cite{pham2019found} & $100$ & $12242$ & $0.19$ & $6526$ & $4.84$ & $0.19$ \\

\Xhline{0.5\arrayrulewidth}
\multirow{1}{*}{\textbf{T}} & \textsc{GradBlend}~\cite{wang2020makes} & $100$ & $18176$ &$3.89$ &$6042$ & $2.63$ & $2.25$ \\

\Xhline{3\arrayrulewidth}
\end{tabular}

\vspace{-2mm}
\label{results:affect_complexity}
\end{table*}

We show the full performance results in Table~\ref{results:affect_supp} and complexity results in Table~\ref{results:affect_complexity}. Here we list some observations regarding these results:
\begin{enumerate}
    \item Language is usually the best performing modality, especially on sentiment and emotion prediction. However, the improvement of language over audio and video on humor prediction and sarcasm prediction is much less. This follows our intuition that while language is primarily useful for sentiment and emotion prediction, audio and visual are strong predictors for humor and sarcasm.
    \item The best performing method over these datasets is consistently the Multimodal Transformer (\textsc{MulT}~\citep{tsai2019learning}), which was originally tested on predicting sentiment and emotions on the \textsc{CMU-MOSI} and \textsc{CMU-MOSEI} dataset. We find that it is a general method and generalizes to humor and sarcasm prediction as well.
    \item However, while it \textsc{MulT} achieves the best performance, it suffers in complexity, taking more than $12\times$ the inference time of unimodal models and $3-4\times$ several simpler early or late fusion multimodal baselines.
    \item Some methods that work well on humor, sentiment, and emotion prediction do not generalize to sarcasm detection, such as tensor fusion (\textsc{TF}) and reconstruction-based models (\textsc{MVAE} and \textsc{MFM}). It is not a surprise that this coincides with sarcasm being the least studied task as well. Furthermore, we believe that it is a task with extremely complementary information (e.g., sarcasm is usually displayed via text and video/audio features contradicting each other). We hope that \names\ can encourage further research in such multimodal tasks since current methods do not generalize to these tasks.
    \item Several out-of-domain methods, such as \textsc{GradBlend} do not work well. In fact we find that the variance of the \textsc{GradBlend} method is quite high and shows strong performance on several datasets but struggles on others.
    \item \textsc{MCTN} is designed for robustness and only uses the language modality at test time. While it was shown to work well for relatively easier fusion tasks in predicting sentiment, emotions, and humor~\citep{pham2019found}, we find that it struggles on the more challenging sarcasm prediction task. 
\end{enumerate}

\begin{figure*}[]
\centering
    \includegraphics[width=\textwidth]{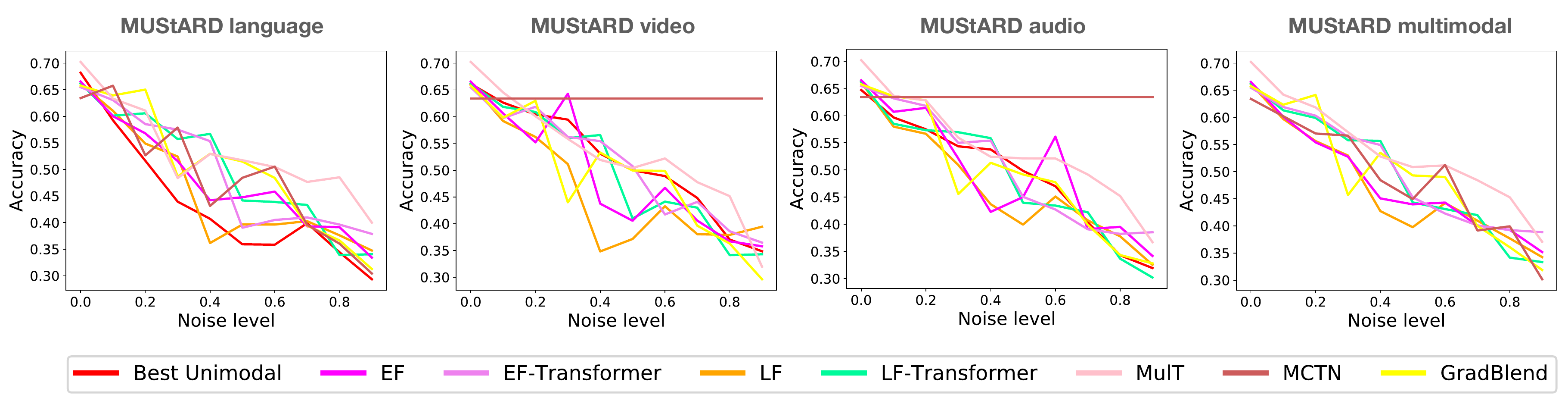}
\caption{Robustness of multimodal models with increasing levels of noise on the \textsc{MUStARD} dataset in the affective computing domain.\vspace{-2mm}}
\label{figs:robustness_sarcasm}
\end{figure*}

\begin{figure*}[]
\centering
    \includegraphics[width=\textwidth]{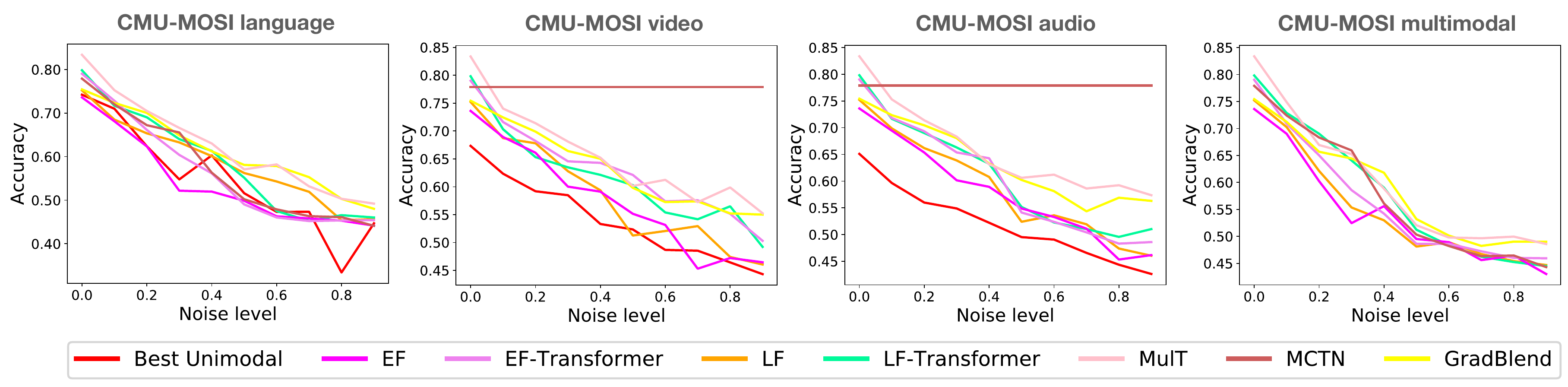}
\caption{Robustness of multimodal models with increasing levels of noise on the \textsc{CMU-MOSI} dataset in the affective computing domain.\vspace{-2mm}}
\label{figs:robustness_mosi}
\end{figure*}

\begin{figure*}[]
\centering
    \includegraphics[width=\textwidth]{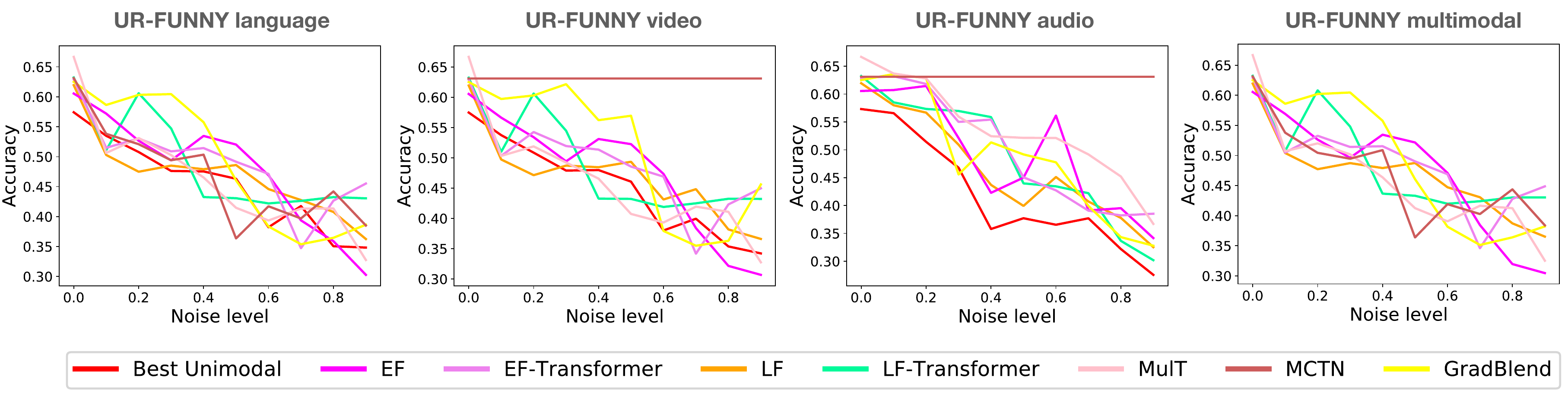}
\caption{Robustness of multimodal models with increasing levels of noise on the \textsc{UR-FUNNY} dataset in the affective computing domain.\vspace{-2mm}}
\label{figs:robustness_humor}
\end{figure*}

\begin{figure*}[]
\centering
    \includegraphics[width=\textwidth]{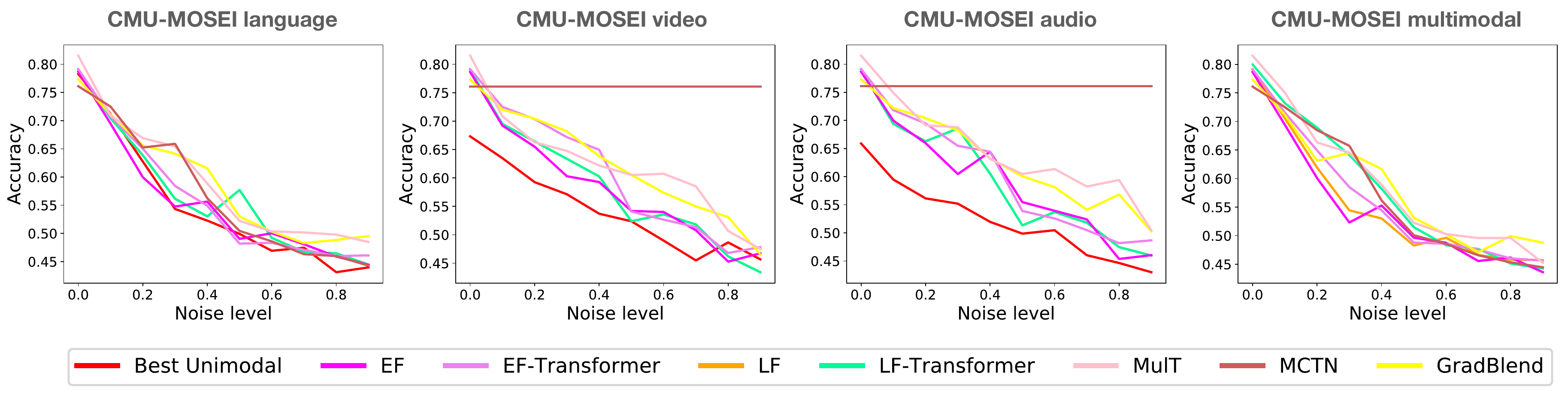}
\caption{Robustness of multimodal models with increasing levels of noise on the \textsc{CMU-MOSEI} dataset in the affective computing domain.\vspace{-2mm}}
\label{figs:robustness_mosei}
\end{figure*}

We show the robustness of multimodal models with increasing levels of noise on \textsc{MUStARD} in Figure~\ref{figs:robustness_sarcasm}, \textsc{CMU-MOSI} in Figure~\ref{figs:robustness_mosi}, \textsc{UR-FUNNY} in Figure~\ref{figs:robustness_humor}, and \textsc{CMU-MOSEI} in Figure~\ref{figs:robustness_mosei}. We highlight the following observations:
\begin{enumerate}
    \item Unimodal and multimodal models are in general not robust to increasing noise and imperfections in these datasets. Performance drops off very quickly towards random.
    \item We find that multimodal models are slightly more robust than unimodal models. For video and audio, the unimodal method is the least robust. However, for language, the unimodal model can actually be more robust than several multimodal models. In other words, multimodal models are more robust to video and audio while being less robust to language, which is the best performing modality. We believe that directly training multimodal models via supervised learning can be prone to overfitting on the most informative modality (in this case language) which causes the multimodal model to be even less robust than unimodal models in language. A similar observation was the motivation behind the \textsc{GradBlend} approach to balance overfitting and generalization across different modalities~\citep{wang2020makes}.
    \item \textsc{GradBlend}~\citep{wang2020makes} seems to be a surprisingly robust approach while also generalizing to several datasets. \textsc{GradBlend} was not in fact not initially designed for the affective computing domain, although it was designed for similar multimodal time-series data in the multimedia domain.
    \item \textsc{MCTN}~\citep{pham2019found} was designed as a robust alternative to multimodal models since it uses multimodal data at training time but only language data at test time. On imperfections to video and audio, \textsc{MCTN} therefore stays constant and can potentially be a viable alternative that learns a unimodal model from multimodal data during training but remains unimodal at testing.
\end{enumerate}

\clearpage

\vspace{-1mm}
\subsection{Healthcare}
\vspace{-1mm}

\begin{table*}[]
\fontsize{9}{11}\selectfont
\setlength\tabcolsep{3.0pt}
\caption{Results on the \textsc{MIMIC} dataset in the healthcare domain. \textbf{U}: unimodal models, \textbf{M}: multimodal fusion paradigms, \textbf{O}: optimization objectives, \textbf{T}: training structures. Several out-domain methods perform well on \textsc{MIMIC} and improve upon the current state-of-the-art performance on in-domain methods.}
\centering
\footnotesize
\vspace{-0mm}
\begin{tabular}{l|l|c|cc|cc}
\Xhline{3\arrayrulewidth}
& Dataset & \multicolumn{1}{c|}{\textsc{MIMIC Mortality}}& \multicolumn{2}{c|}{\textsc{MIMIC ICD-$9$ $10-19$}} & \multicolumn{2}{c}{\textsc{MIMIC ICD-$9$ $70-79$}}  \\
& Metric & Acc$(6)$ $\uparrow$ & Acc$(2)$ $\uparrow$ & AUPRC$(2)$ $\uparrow$ & Acc$(2)$ $\uparrow$ & AUPRC$(2)$ $\uparrow$ \\
\Xhline{0.5\arrayrulewidth}
& Most frequent & $ 76.1 $&  $ 83.1 $ & $ - $  & $ 52.5 $ & $ - $ \\ \hline
\multirow{2}{*}{\textbf{U}} & Unimodal ($t$) & $ 76.7 \pm 0.3 $&  $ 83.6 \pm 0.1 $ & $ 35.0 \pm 0.9 $ & $ 67.6 \pm 0.4 $ & $ 72.9 \pm 0.3 $  \\
& Unimodal ($ta$) & $ 76.4 \pm 0.2 $  &  $ 91.4 \pm 0.0 $ & $ 68.4 \pm 0.1 $& $ 56.3 \pm 0.3 $ & $ 54.6 \pm 0.4 $ \\
\Xhline{0.5\arrayrulewidth}
\multirow{5}{*}{\textbf{M}} & \textsc{LF} & $77.9 \pm 0.3$& $91.5 \pm 0.1$ & $74.2 \pm 0.7$ & $\mathbf{68.9 \pm 0.5}$ & $74.3 \pm 0.4$  \\
& \textsc{LRTF}~\citep{liu2018efficient} & $ 78.2\pm 0.3 $ &  $ 91.5 \pm 0.1 $ & $ \mathbf{75.1 \pm 0.3} $ & $ 68.5 \pm 0.4 $ & $  73.8 \pm 0.4  $ \\
& \textsc{MI-Matrix}~\citep{Jayakumar2020Multiplicative} & $ 77.6\pm 0.4 $  &  $ 91.5 \pm 0.1 $ & $ 74.2 \pm 0.6 $& $ 67.9 \pm 0.3 $ & $ 73.0 \pm 0.5 $ \\
& \textsc{NL Gate}~\cite{wang2020makes} & $78.1 \pm 0.2$ &  $\mathbf{91.6 \pm 0.1}$ & $73.8 \pm 0.7$& $68.7 \pm 0.5$ & $74.3 \pm 0.4$  \\
& \textsc{MFAS}~\citep{perez2019mfas} & $77.9 \pm 0.2 $ & $ 91.4 \pm 0.0 $ & $ 70.3 \pm 1.2 $& $ 68.5 \pm 0.4 $ & $ 73.7 \pm 0.4 $ \\
\Xhline{0.5\arrayrulewidth}

\multirow{2}{*}{\textbf{O}} & \textsc{MFM}~\cite{tsai2019learning} & $ 78.2 \pm 0.3 $ &  $ 91.5 \pm 0.1 $ & $ 75.0 \pm 0.5 $& $ 68.8 \pm 0.4 $ & $ \mathbf{74.4 \pm 0.4} $  \\
& \textsc{MVAE}~\citep{wu2018multimodal} & $ 78.0 \pm 0.3 $&  $ \mathbf{91.6 \pm 0.1} $ & $ 73.5 \pm 1.4 $  & $ 68.7 \pm 0.6 $ & $ 74.0 \pm 0.7 $ \\
\Xhline{0.5\arrayrulewidth}

\multirow{1}{*}{\textbf{T}} & \textsc{GradBlend}~\cite{wang2020makes} & $\mathbf{78.2 \pm 0.2}$&  $91.5 \pm 0.1$ & $74.1 \pm 0.4$ & $68.0 \pm 0.7$ & $73.2 \pm 0.5$  \\

\Xhline{3\arrayrulewidth}
\end{tabular}

\vspace{-2mm}
\label{results:healthcare_supp}
\end{table*}

\begin{table*}[]
\fontsize{9}{11}\selectfont
\setlength\tabcolsep{3.0pt}
\caption{Complexity results for datasets in the healthcare domain. ((*) This is the number of parameters in modules input to \textsc{MFAS} at the start of training, \textsc{MFAS} will generate more parameters during the architecture search process). \textbf{U}: unimodal models, \textbf{M}: multimodal fusion paradigms, \textbf{O}: optimization objectives, \textbf{T}: training structures.}
\centering
\footnotesize
\vspace{-0mm}

\begin{tabular}{l|l|cccccc}
\Xhline{3\arrayrulewidth}
& Dataset & \multicolumn{6}{c}{\textsc{MIMIC}} \\
& Metric & \begin{tabular}[c]{@{}c@{}}Epochs \\ trained\end{tabular} & \begin{tabular}[c]{@{}c@{}}Training \\ time (s)\end{tabular} & \begin{tabular}[c]{@{}c@{}}Training \\ params (M)\end{tabular} & \begin{tabular}[c]{@{}c@{}}Training peak \\ memory (MB)\end{tabular} & \begin{tabular}[c]{@{}c@{}}Inference \\ time (s)\end{tabular} & \begin{tabular}[c]{@{}c@{}}Inference \\ params (M) \end{tabular} \\
\Xhline{0.5\arrayrulewidth}
\multirow{2}{*}{\textbf{U}} & Unimodal ($t$) & $20$ & $46.4$ & $0.019$ & $2360$ & $0.41$ & $0.019$ \\
& Unimodal ($ta$) & $20$ & $34.6$ & $0.001$ & $2359$ & $0.39$ & $0.001$ \\
\Xhline{0.5\arrayrulewidth}

\multirow{5}{*}{\textbf{M}} & \textsc{LF} & $20$ & $49.4$ & $0.034$ & $2362$ & $0.41$ & $0.034$ \\
& \textsc{LRTF}~\citep{liu2018efficient} & $50$ & $261$ & $0.008$ & $2575$ & $0.41$ & $0.008$ \\
& \textsc{MI-Matrix}~\citep{Jayakumar2020Multiplicative} & $20$ & $56.6$ & $0.801$ & $2377$ & $0.39$ & $0.801$ \\
& \textsc{NL Gate}~\cite{wang2020makes} & $20$ & $51.4$ & $0.040$ & $2422$ & $0.43$ & $0.040$ \\
& \textsc{MFAS}~\citep{perez2019mfas} & $42\times 6$ & $3762$ & $0.086^*$ & $2360$ & $1.79$ & $0.016$ \\
\Xhline{0.5\arrayrulewidth}

\multirow{2}{*}{\textbf{O}} & \textsc{MFM}~\cite{tsai2019learning} & $25$ & $221$ & $0.323$ & $2438$ & $0.85$ & $0.315$ \\
& \textsc{MVAE}~\citep{wu2018multimodal} & $30$ & $486$ & $0.312$ & $2553$ & $0.89$ & $0.305$ \\
\Xhline{0.5\arrayrulewidth}

\multirow{1}{*}{\textbf{T}} & \textsc{GradBlend}~\cite{wang2020makes} & $300$ & $2785$ & $0.063$ & $2575$ & $0.45$ & $0.034$ \\
\Xhline{3\arrayrulewidth}
\end{tabular}

\vspace{-2mm}
\label{results:healthcare_complexity}
\end{table*}

We show the full results in Table~\ref{results:healthcare_supp} and complexity results in Table~\ref{results:healthcare_complexity}. Here we list some observations regarding these results:
\begin{enumerate}
    \item We find that results across all models show small variations on \textsc{MIMIC}, which suggests that many current multimodal approaches may not generalize that well to the input modalities and prediction tasks that \textsc{MIMIC} tests for.
    \item In particular, while \textsc{MFAS} (architecture search) is otherwise a pretty general solution that works well across quite a few datasets, it struggles on \textsc{MIMIC}. While there has been a recently proposed \textsc{MUFASA}~\citep{xu2021mufasa} method that adapts architecture search specifically for healthcare datasets, we were not able to test this method on our partition of \textsc{MIMIC}, and it is in our top priorities to implement that approach into \codes\ and accurately benchmark its performance on a suite of datasets.
    \item Late Fusion (\textsc{LF}) with simple concatenation was the best-performing model in the evaluations conducted by the previous paper that used the exact same partition as ours~\citep{PURUSHOTHAM2018112}. It actually works quite well compared to more complex models evaluated here, as it has the best performance on ICD-$9$ group 7 task and is quite close to the best performing models in the other two. This may suggest that simple multimodal models such as Late Fusion is worth being tried first on healthcare datasets.
    \item The reconstruction-based multimodal models such as \textsc{MVAE} and \textsc{MFM} have strong performance on this dataset, possibly due to the low dimensions of the input modalities. This suggests that reconstruction-based architectures and objectives might work well on datasets with simple or low-dimensional modalities which are easier to reconstruct.
\end{enumerate}

\begin{figure*}[]
\centering
    \includegraphics[width=\textwidth]{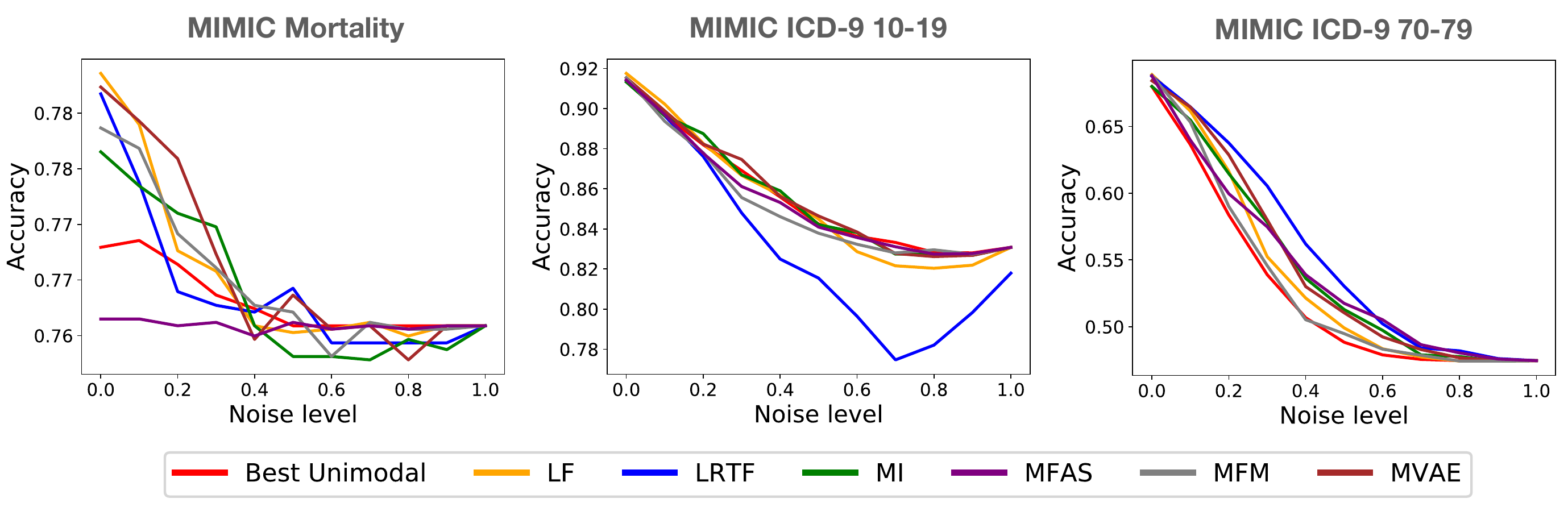}
\caption{Robustness of multimodal models with increasing levels of noise on the \textsc{MIMIC} dataset in the healthcare domain.\vspace{-2mm}}
\label{figs:robustness_health}
\end{figure*}

Finally, we show the robustness of multimodal models with increasing levels of noise on the \textsc{MIMIC} dataset in Figure~\ref{figs:robustness_health}. We highlight the following observations:
\begin{enumerate}
    \item Unimodal and multimodal models are in general not robust to increasing noise and imperfections in the table and time-series modalities. Performance drops off very quickly towards random.
    \item In general, multimodal models are slightly more robust than unimodal models. The behavior is best exhibited in the ICD-$9$ group 7 task where many models start off strong, but multimodal models remain more robust than the best unimodal model. This perhaps indicates that multimodal models do learn to use information from other sources when another one is noisy.
    \item There is high variance in the robustness of each multimodal model even within the same dataset and modalities but across different prediction tasks. We observe that \textsc{LRTF} is the most robust model on the ICD-$9$ group 7 task but the least robust model on the ICD-$9$ group 1 task. This high variance is a concern especially given the close similarity across both of these tasks.
\end{enumerate}

\clearpage

\vspace{-1mm}
\subsection{Robotics}
\vspace{-1mm}

\begin{table*}[]
\fontsize{9}{11}\selectfont
\setlength\tabcolsep{6.0pt}
\caption{Results on multimodal datasets in the robotics domain. \textbf{U}: unimodal models, \textbf{M}: multimodal fusion paradigms, \textbf{O}: optimization objectives, \textbf{T}: training structures.}
\centering
\footnotesize
\vspace{-0mm}

\begin{tabular}{l|l|c}
\Xhline{3\arrayrulewidth}
& Dataset & \multicolumn{1}{c}{\textsc{MuJoCo Push}} \\
& Metric & MSE $\downarrow$ \\
\Xhline{0.5\arrayrulewidth}
\multirow{4}{*}{\textbf{U}} & Unimodal ($i$) & $0.334 \pm 0.034$ \\
& Unimodal ($f$) & $4.266 \pm 0.085$ \\
& Unimodal ($p$) & $3.885 \pm 0.004$ \\
& Unimodal ($c$) & $3.804 \pm 0.005$ \\
\Xhline{0.5\arrayrulewidth}
\multirow{4}{*}{\textbf{M}} & \textsc{EF-LSTM} & $0.363 \pm 0.038$ \\
& \textsc{LF-LSTM} & $\mathbf{0.290 \pm 0.018}$ \\
& \textsc{TF}~\citep{zadeh2017tensor} & $0.574 \pm 0.059$ \\
& \textsc{MulT}~\cite{tsai2020multimodal} & $0.402 \pm 0.026$ \\

\Xhline{3\arrayrulewidth}
\end{tabular}

\vspace{4mm}

\begin{tabular}{l|l|c|c}
\Xhline{3\arrayrulewidth}
& Dataset & \multicolumn{1}{c|}{\textsc{Vision\&Touch Contact}} & \multicolumn{1}{c}{\textsc{Vision\&Touch End Effector}} \\
& Metric & Acc$(2)$ $\uparrow$ & MSE ($\times 10^{-4}$) $\downarrow$ \\
\Xhline{0.5\arrayrulewidth}
\multirow{3}{*}{\textbf{U}} & Unimodal ($i$) & $ 83.6 \pm 0.3 $ & $ 1.99 \pm 0.160 $ \\
& Unimodal ($f$) & $ \mathbf{93.6 \pm 0.1} $ & $ 87.2 \pm 0.477 $ \\
& Unimodal ($p$) & $ 85.6 \pm 0.6 $ & $ 0.202 \pm 0.022 $ \\
\Xhline{0.5\arrayrulewidth}
\multirow{3}{*}{\textbf{M}} & LF & $ \mathbf{93.6 \pm 0.1} $ & $ \mathbf{0.185 \pm 0.011} $ \\
& Sensor Fusion~\cite{lee2019making} & $ 93.4 \pm 0.1 $ & $ 0.258 \pm 0.011 $ \\
& \textsc{LRTF}~\citep{liu2018efficient} & $ 93.3 \pm 0.1$ & $ 0.232 \pm 0.031 $ \\
\Xhline{0.5\arrayrulewidth}
\textbf{O} & \textsc{RefNet}~\cite{sankaran2021multimodal} & $93.5 \pm 0.1$ & $ 0.203 \pm 0.025 $ \\

\Xhline{3\arrayrulewidth}
\end{tabular}

\vspace{-2mm}
\label{results:robotics_supp}
\end{table*}

We show the full results in Table~\ref{results:robotics_supp} and complexity results in Table~\ref{results:robotics_complexity}. Here we list some observations regarding these results:
\begin{enumerate}
    \item We find that in all robotics tasks, there exists one modality with extremely strong unimodal performance (force in \textsc{Vision\&Touch} contact task, proprioception in \textsc{Vision\&Touch} End Effector task, image in \textsc{MuJoCo Push}).
    \item On the \textsc{Vision\&Touch} dataset, we found that Late Fusion outperforms the method of choice in the original paper for the dataset~\citep{lee2019making} (Sensor Fusion) on both tasks, so Late Fusion seems to generalize well to this domain.
    \item In our experiments, as well as the baselines~\citep{lee2019making}, the action modality is typically treated as a general modality without specific modeling. Future work should explore whether this is the best way to encode action as a modality in these action-conditional prediction tasks, and possibly unify these datasets with those used in embodied multimodal learning~\citep{das2018embodied,li2019robust,luketina2019survey}.
    \item We plan to include several more reinforcement learning tasks for multimodal learning in robotics. It remains an open question where multimodal representations suitable for fusion-type prediction tasks are also suitable for reinforcement learning tasks. Adding such reinforcement learning tasks from multiple sensors to \names\ will enable more accurate benchmarking of the generalization capabilities of these multimodal models.
\end{enumerate}

\begin{table*}[]
\fontsize{9}{11}\selectfont
\setlength\tabcolsep{3.0pt}
\caption{Complexity results for datasets in the robotics domain. \textbf{U}: unimodal models, \textbf{M}: multimodal fusion paradigms, \textbf{O}: optimization objectives, \textbf{T}: training structures.}
\centering
\footnotesize
\vspace{-0mm}

\begin{tabular}{l|l|cccccc}
\Xhline{3\arrayrulewidth}
& Dataset & \multicolumn{6}{c}{\textsc{MuJoCo Push}} \\
& Metric & \begin{tabular}[c]{@{}c@{}}Epochs \\ trained\end{tabular} & \begin{tabular}[c]{@{}c@{}}Training \\ time (s)\end{tabular} & \begin{tabular}[c]{@{}c@{}}Training \\ params (M)\end{tabular} & \begin{tabular}[c]{@{}c@{}}Training peak \\ memory (MB)\end{tabular} & \begin{tabular}[c]{@{}c@{}}Inference \\ time (s)\end{tabular} & \begin{tabular}[c]{@{}c@{}}Inference \\ params (M) \end{tabular} \\
\Xhline{0.5\arrayrulewidth}
\multirow{4}{*}{\textbf{U}} & Unimodal ($i$) & $20$ & $738 \pm 133$ & $3.88$ & $3607 \pm 1$ & $3.46 \pm 0.02$ & $3.88$ \\
& Unimodal ($f$) & $20$ & $288 \pm 39$ & $3.33$ & $3595 \pm 2$ & $0.91 \pm 0.08$ & $3.33$ \\
& Unimodal ($p$) & $20$ & $252 \pm 6$ & $3.33$ & $3594 \pm 1$ & $0.87 \pm 0.04$ & $3.33$ \\ 
& Unimodal ($c$) & $20$ & $372 \pm 64$ & $3.33$ & $3594 \pm 1$ & $0.86 \pm 0.04$ & $3.33$ \\
\Xhline{0.5\arrayrulewidth}
\multirow{4}{*}{\textbf{M}} & \textsc{EF} & $20$ & $815 \pm 34$ & $3.92$ & $3654 \pm 1$ & $4.44 \pm 0.55$ & $3.92$ \\
& \textsc{LF-LSTM} & $20$ & $856 \pm 46$ & $1.90$ & $3636 \pm 1$ & $4.32 \pm 0.45$ & $1.90$ \\
& \textsc{TF-LSTM}~\citep{zadeh2017tensor} & $20$ & $1914 \pm 31$ & $23.5$ & $4530 \pm 9$ & $7.75 \pm 0.12$ & $23.5$ \\
& \textsc{MulT}~\cite{tsai2020multimodal} & $20$ & $4792 \pm 62$ & $14.6$ & $6530 \pm 16$ & $22.4 \pm 0.28$ & $14.6$ \\
\Xhline{3\arrayrulewidth}
\end{tabular}

\vspace{4mm}

\begin{tabular}{l|l|cccccc}
\Xhline{3\arrayrulewidth}
& Dataset & \multicolumn{6}{c}{\textsc{Vision\&Touch}} \\
& Metric & \begin{tabular}[c]{@{}c@{}}Epochs \\ trained\end{tabular} & \begin{tabular}[c]{@{}c@{}}Training \\ time (s)\end{tabular} & \begin{tabular}[c]{@{}c@{}}Training \\ params (M)\end{tabular} & \begin{tabular}[c]{@{}c@{}}Training peak \\ memory (MB)\end{tabular} & \begin{tabular}[c]{@{}c@{}}Inference \\ time (s)\end{tabular} & \begin{tabular}[c]{@{}c@{}}Inference \\ params (M) \end{tabular} \\
\Xhline{0.5\arrayrulewidth}
\multirow{3}{*}{\textbf{U}} & Unimodal ($i$) & $15$ & $2633$ & $1.00$ & $5530$ & $63.9$ & $1.00$ \\
& Unimodal ($f$) & $15$ & $2185$ & $0.13$ & $2426$ & $51.6$ & $0.13$ \\ 
& Unimodal ($p$) & $15$ & $2514$ & $0.08$ & $2389$ & $59.5$ & $0.08$ \\ 

\Xhline{0.5\arrayrulewidth}
\multirow{3}{*}{\textbf{M}} & \textsc{LF} & $15$ & $2672$ & $1.20$ & $5572$ & $64.4$ & $1.20$ \\
& Sensor Fusion~\cite{lee2019making} & $50$ & $11604$ & $1.10$ & $4467$ & $62.6$ & $1.10$ \\
& \textsc{LRTF}~\citep{liu2018efficient} & $35$ & $8366$ & $1.09$ & $4987$ & $64.4$ & $1.09$ \\
\Xhline{0.5\arrayrulewidth}
\textbf{O} & \textsc{RefNet}~\cite{sankaran2021multimodal} & $15$ & $3819$ & $135$ & $6067$ & $65.0$ & $1.20$ \\
\Xhline{3\arrayrulewidth}
\end{tabular}

\vspace{-2mm}
\label{results:robotics_complexity}
\end{table*}

\begin{figure*}[]
\centering
    \includegraphics[width=\textwidth]{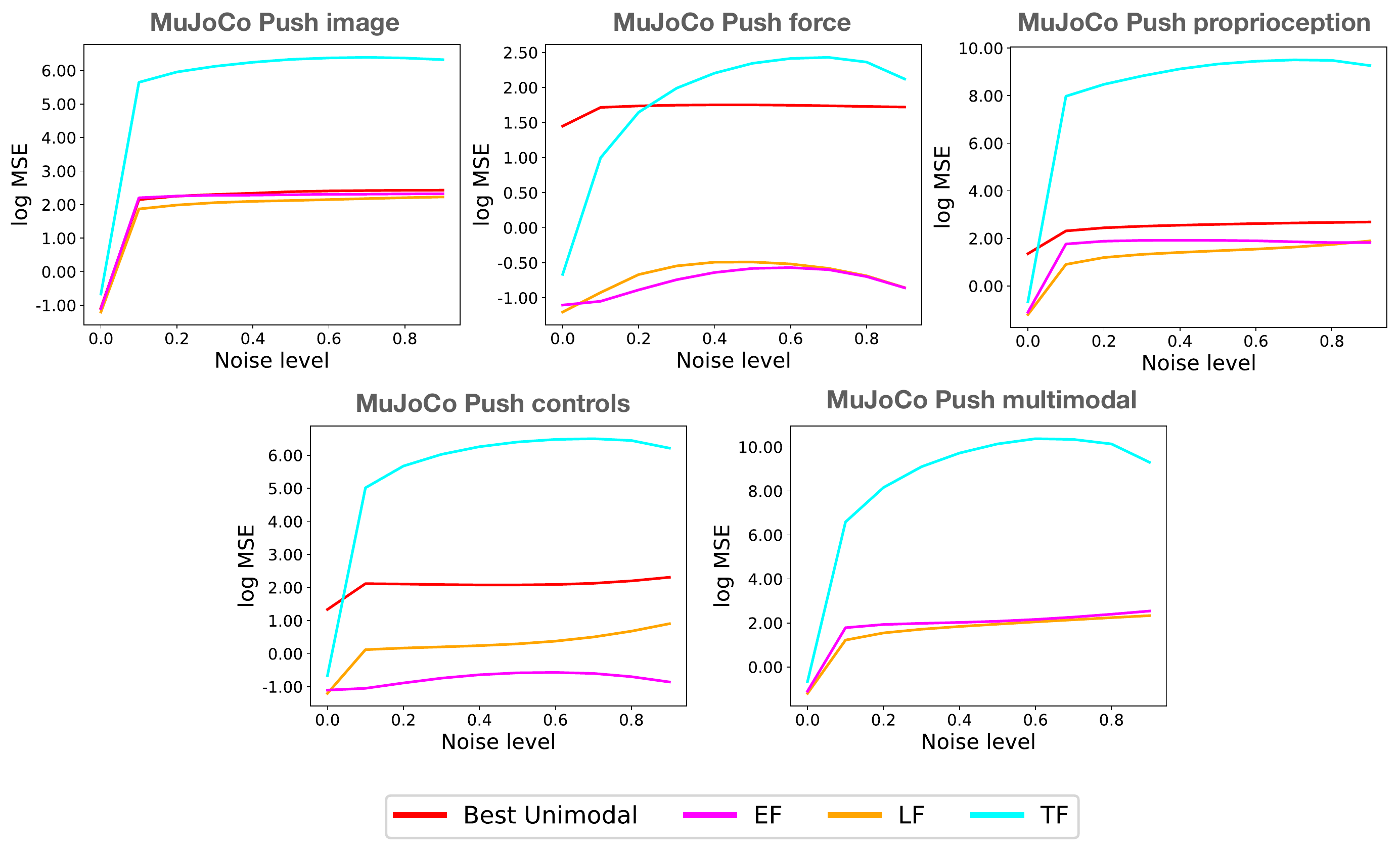}
\caption{Robustness of multimodal models with increasing levels of noise on the \textsc{MuJoCo Push} dataset in the robotics domain.\vspace{-2mm}}
\label{figs:robustness_push}
\end{figure*}

\begin{figure*}[]
\centering
    \includegraphics[width=\textwidth]{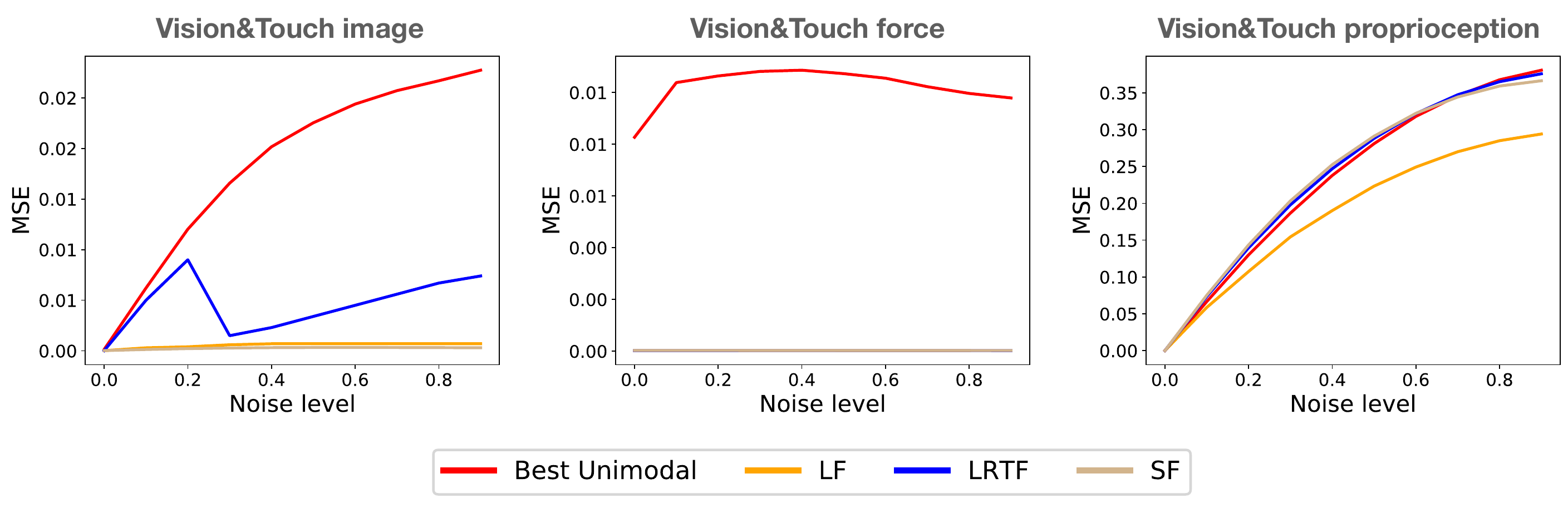}
\caption{Robustness of multimodal models with increasing levels of noise on the \textsc{Vision\&Touch} dataset in the robotics domain.\vspace{-2mm}}
\label{figs:robustness_vt}
\end{figure*}

Finally, we show the robustness of multimodal models with increasing levels of noise on \textsc{MuJoCo Push} in Figure~\ref{figs:robustness_push} and on \textsc{Vision\&Touch} in Figure~\ref{figs:robustness_vt}. We highlight the following observations:
\begin{enumerate}
    \item For \textsc{MuJoCo Push} we plot the MSE using a log scale on the $y$-axis since the error of the \textsc{TF} method blows up significantly much faster than the other methods.
    \item We observe that multimodal methods are much more robust than unimodal methods, which match the robustness results as reported in the paper~\citep{lee2019making} where the trained multimodal model is robust and able to recover from external forces on the force sensor or occlusions to the image sensor. This observation is true for both datasets.
    \item For \textsc{Vision\&Touch}, we observe that unimodal performance is especially bad for the object pose prediction task. The remaining multimodal models are relatively robust as compared to unimodal performance. The most robust models seem to be Sensor Fusion~\citep{lee2019making} (\textsc{SF}) and Late Fusion (\textsc{LF}).
\end{enumerate}

\clearpage

\vspace{-1mm}
\subsection{Finance}
\vspace{-1mm}

\begin{table*}[]
\fontsize{9}{11}\selectfont
\setlength\tabcolsep{6.0pt}
\caption{Results on multimodal datasets in the finance domain. \textbf{U}: unimodal models, \textbf{M}: multimodal fusion paradigms, \textbf{O}: optimization objectives, \textbf{T}: training structures. \textsc{MulT} struggles on these datasets even though it performs strongly on similar multimodal time-series datasets in the affective computing domain. Other methods also show high variance across different data partitions.}
\centering
\footnotesize
\vspace{-0mm}
\begin{tabular}{l|l|c|c|c}
\Xhline{3\arrayrulewidth}
& Dataset & \multicolumn{1}{c|}{\textsc{Stocks-F\&B}} & \multicolumn{1}{c|}{\textsc{Stocks-Health}} & \multicolumn{1}{c}{\textsc{Stock-Tech}} \\
& Metric & MSE $\downarrow$ & MSE $\downarrow$ & MSE $\downarrow$ \\
\Xhline{0.5\arrayrulewidth}
& Mean & $2.140$ & $0.575$ & $0.140$ \\
\Xhline{0.5\arrayrulewidth}
\multirow{2}{*}{\textbf{U}} & ARIMA & $2.199$ & $0.620$ & $0.152$ \\
& Unimodal & $1.856 \pm 0.093$ & $0.541 \pm 0.010$ & $0.125 \pm 0.004$ \\
\Xhline{0.5\arrayrulewidth}
\multirow{5}{*}{\textbf{M}} & \textsc{EF-LSTM} & $1.835 \pm 0.098$ & $\mathbf{0.526 \pm 0.017}$ & $0.121 \pm 0.003$ \\
& \textsc{LF-LSTM} & $1.893 \pm 0.106$ & $0.541 \pm 0.018$ & $\mathbf{0.120 \pm 0.008}$ \\
& \textsc{EF-Transformer} & $2.144 \pm 0.014$ & $0.573 \pm 0.006$ & $0.143 \pm 0.003$ \\
& \textsc{LF-Transformer} & $2.155 \pm 0.023$ & $0.573 \pm 0.006$ & $0.143 \pm 0.004$ \\
& \textsc{MulT}~\cite{tsai2020multimodal} & $2.053 \pm 0.022$ & $0.555 \pm 0.005$ & $0.135 \pm 0.003$ \\

\Xhline{0.5\arrayrulewidth}
\multirow{1}{*}{\textbf{T}} & \textsc{GradBlend}~\citep{wang2020makes} & $\mathbf{1.820 \pm 0.138}$ & $0.537 \pm 0.011$ & $0.138 \pm 0.030$ \\

\Xhline{3\arrayrulewidth}
\end{tabular}

\vspace{-2mm}
\label{results:finance_supp}
\end{table*}

We show the full results in Table~\ref{results:finance_supp} and complexity results in Table~\ref{results:finance_complexity}. Here we list some observations regarding these results:
\begin{enumerate}
    \item We do observe better performance using multimodal models as compared to unimodal ones, which suggests that multiple financial signals do help in stock prediction. Several multimodal models do generalize to this more challenging area which presents scalability challenges due to a large number of modalities ($18/63/100$ as compared to $2/3$ in most datasets), as well as robustness challenges arising from real-world data with an inherently low signal-to-noise ratio.
    \item There has been very little research in multimodal models in this area, and no public implementations of multimodal models on actual finance data. By adapting current models on this dataset, we observe decent performance of several out of domain methods. Specifically, early fusion (\textsc{EF}) works well which we believe to be due to the little heterogeneity in data origins (i.e., all data comes in the form of time-series data, which is much less heterogeneous as compare to image and text datasets).
    \item There remains high variance in the performance of multimodal models even within the same domain: we observe that the best multimodal is not consistent across the $3$ partitions of finance datasets, which suggests that current multimodal models remain highly sensitive to the task at hand.
    \item Perhaps surprisingly, our experiments on using a Transformer found that they performed worse off than LSTM models. We hypothesize that these large Transformer models might be prone to overfitting on these small and noisy datasets.
    \item These datasets present scalability issues to a large number of modalities. We find that we had to adapt several methods such as Tensor Fusion (\textsc{TF}) and Multimodal Transformer (\textsc{MulT}) since they scale exponentially and quadratically with the number of modalities respectively, which does not scale to these finance datasets with more than $10$ modalities. We had to adapt these models by performing an initial clustering over the modalities to form $2/3$ groups, performing early fusion by concatenating the data within each group and forming $2/3$ `modalities' before applying methods such as Tensor Fusion (\textsc{TF}) and Multimodal Transformer (\textsc{MulT}). This might explain their slightly worse performance, especially \textsc{MulT} given its strong performance and generalization to different datasets in the affective computing domain. Future research should focus on more scalable multimodal methods to a large number of modalities. Unfortunately, the bulk of multimodal research being in language and vision means that this question is relatively unexplored.
\end{enumerate}

\begin{table*}[]
\fontsize{9}{11}\selectfont
\setlength\tabcolsep{3.0pt}
\caption{Complexity results for datasets in the finance domain. \textbf{U}: unimodal models, \textbf{M}: multimodal fusion paradigms, \textbf{O}: optimization objectives, \textbf{T}: training structures.}
\centering
\footnotesize
\vspace{-0mm}
\begin{tabular}{l|l|cccccc}
\Xhline{3\arrayrulewidth}
& Dataset & \multicolumn{6}{c}{\textsc{Stocks-F\&B}} \\
& Metric & \begin{tabular}[c]{@{}c@{}}Epochs \\ trained\end{tabular} & \begin{tabular}[c]{@{}c@{}}Training \\ time (s)\end{tabular} & \begin{tabular}[c]{@{}c@{}}Training \\ params (M)\end{tabular} & \begin{tabular}[c]{@{}c@{}}Training peak \\ memory (MB)\end{tabular} & \begin{tabular}[c]{@{}c@{}}Inference \\ time (s)\end{tabular} & \begin{tabular}[c]{@{}c@{}}Inference \\ params (M) \end{tabular} \\
\Xhline{0.5\arrayrulewidth}
\multirow{1}{*}{\textbf{U}} & Unimodal ($t$) & 2 & $9.5 \pm 0.1$ & $0.067$ & $3028 \pm 3$ & $0.50 \pm 0.01$ & $0.067$ \\
\Xhline{0.5\arrayrulewidth}
\multirow{5}{*}{\textbf{M}} & EF-LSTM & 2 & $9.7 \pm 0.1$ & $0.069$ & $3067 \pm 21$ & $0.51 \pm 0.01$ & $0.069$ \\
& LF-LSTM & 4 & $62 \pm 0.4$ & $0.005$ & $2433 \pm 4$ & $1.74 \pm 0.02$ & $0.005$ \\
& EF-Transformer & 4 & $25 \pm 0.3$ & $0.118$ & $2434 \pm 3$ & $0.62 \pm 0.01$ & $0.118$ \\
& LF-Transformer & 4 & $88 \pm 0.3$ & $0.472$ & $2468 \pm 1$ & $1.70 \pm 0.00$ & $0.472$ \\
& MulT~\cite{tsai2020multimodal} & 4 & $160 \pm 1$ & $0.125$ & $3313 \pm 1$ & $4.82 \pm 0.06$ & $0.125$ \\

\Xhline{0.5\arrayrulewidth}
\multirow{1}{*}{\textbf{T}} & \textsc{GradBlend}~\citep{wang2020makes} & 4 & $409 \pm 2$ & $0.338$ & $3102 \pm 1$ & $0.44 \pm 0.01$ & $0.069$ \\
\Xhline{3\arrayrulewidth}
\end{tabular}

\vspace{4mm}

\begin{tabular}{l|l|cccccc}
\Xhline{3\arrayrulewidth}
& Dataset & \multicolumn{6}{c}{\textsc{Stocks-Health}} \\
& Metric & \begin{tabular}[c]{@{}c@{}}Epochs \\ trained\end{tabular} & \begin{tabular}[c]{@{}c@{}}Training \\ time (s)\end{tabular} & \begin{tabular}[c]{@{}c@{}}Training \\ params (M)\end{tabular} & \begin{tabular}[c]{@{}c@{}}Training peak \\ memory (MB)\end{tabular} & \begin{tabular}[c]{@{}c@{}}Inference \\ time (s)\end{tabular} & \begin{tabular}[c]{@{}c@{}}Inference \\ params (M) \end{tabular} \\
\Xhline{0.5\arrayrulewidth}
\multirow{1}{*}{\textbf{U}} & Unimodal ($t$) & 2 & $9.6 \pm 0.1$ & $0.067$ & $3032 \pm 15$ & $0.51 \pm 0.01$ & $0.067$ \\
\Xhline{0.5\arrayrulewidth}
\multirow{5}{*}{\textbf{M}} & EF-LSTM & 2 & $9.6 \pm 0.1$ & $0.070$ & $3083 \pm 2$ & $0.51 \pm 0.02$ & $0.070$ \\
& LF-LSTM & 4 & $108 \pm 1$ & $0.009$ & $2464 \pm 7$ & $2.89 \pm 0.04$ & $0.009$ \\
& EF-Transformer & 4 & $25 \pm 0.4$ & $0.118$ & $2466 \pm 4$ & $0.65 \pm 0.02$ & $0.118$ \\
& LF-Transformer & 4 & $159 \pm 1$ & $0.826$ & $2524 \pm 1$ & $2.93 \pm 0.01$ & $0.826$ \\
& MulT~\cite{tsai2020multimodal} & 4 & $162 \pm 1$ & $0.125$ & $3315 \pm 1$ & $4.88 \pm 0.04$ & $0.125$ \\

\Xhline{0.5\arrayrulewidth}
\multirow{1}{*}{\textbf{T}} & \textsc{GradBlend}~\citep{wang2020makes} & 4 & $582 \pm 4$ & $0.541$ & $3141 \pm 2$ & $0.49 \pm 0.01$ & $0.070$ \\
\Xhline{3\arrayrulewidth}
\end{tabular}

\vspace{4mm}

\begin{tabular}{l|l|cccccc}
\Xhline{3\arrayrulewidth}
& Dataset & \multicolumn{6}{c}{\textsc{Stock-Tech}} \\
& Metric & \begin{tabular}[c]{@{}c@{}}Epochs \\ trained\end{tabular} & \begin{tabular}[c]{@{}c@{}}Training \\ time (s)\end{tabular} & \begin{tabular}[c]{@{}c@{}}Training \\ params (M)\end{tabular} & \begin{tabular}[c]{@{}c@{}}Training peak \\ memory (MB)\end{tabular} & \begin{tabular}[c]{@{}c@{}}Inference \\ time (s)\end{tabular} & \begin{tabular}[c]{@{}c@{}}Inference \\ params (M) \end{tabular} \\
\Xhline{0.5\arrayrulewidth}
\multirow{1}{*}{\textbf{U}} & Unimodal ($t$) & 2 & $9.5 \pm 0.1$ & $0.067$ & $3023 \pm 1$ & $0.51 \pm 0.01$ & $0.067$ \\
\Xhline{0.5\arrayrulewidth}
\multirow{5}{*}{\textbf{M}} & EF-LSTM & 2 & $9.6 \pm 0.1$ & $0.070$ & $3075 \pm 4$ & $0.53 \pm 0.01$ & $0.070$ \\
& LF-LSTM & 4 & $92 \pm 0.5$ & $0.007$ & $2453 \pm 4$ & $2.51 \pm 0.04$ & $0.007$ \\
& EF-Transformer & 4 & $25 \pm 0.4$ & $0.118$ & $2453 \pm 1$ & $0.63 \pm 0.01$ & $0.118$ \\
& LF-Transformer & 4 & $135 \pm 1$ & $0.708$ & $2506 \pm 1$ & $2.52 \pm 0.00$ & $0.708$ \\
& MulT~\cite{tsai2020multimodal} & 4 & $161 \pm 1$ & $0.125$ & $3315 \pm 2$ & $4.79 \pm 0.03$ & $0.125$ \\

\Xhline{0.5\arrayrulewidth}
\multirow{1}{*}{\textbf{T}} & \textsc{GradBlend}~\citep{wang2020makes} & 4 & $500 \pm 3$ & $0.473$ & $3167 \pm 1$ & $0.44 \pm 0.01$ & $0.070$ \\
\Xhline{3\arrayrulewidth}
\end{tabular}

\vspace{-2mm}
\label{results:finance_complexity}
\end{table*}

\begin{figure*}[]
\centering
    \includegraphics[width=\textwidth]{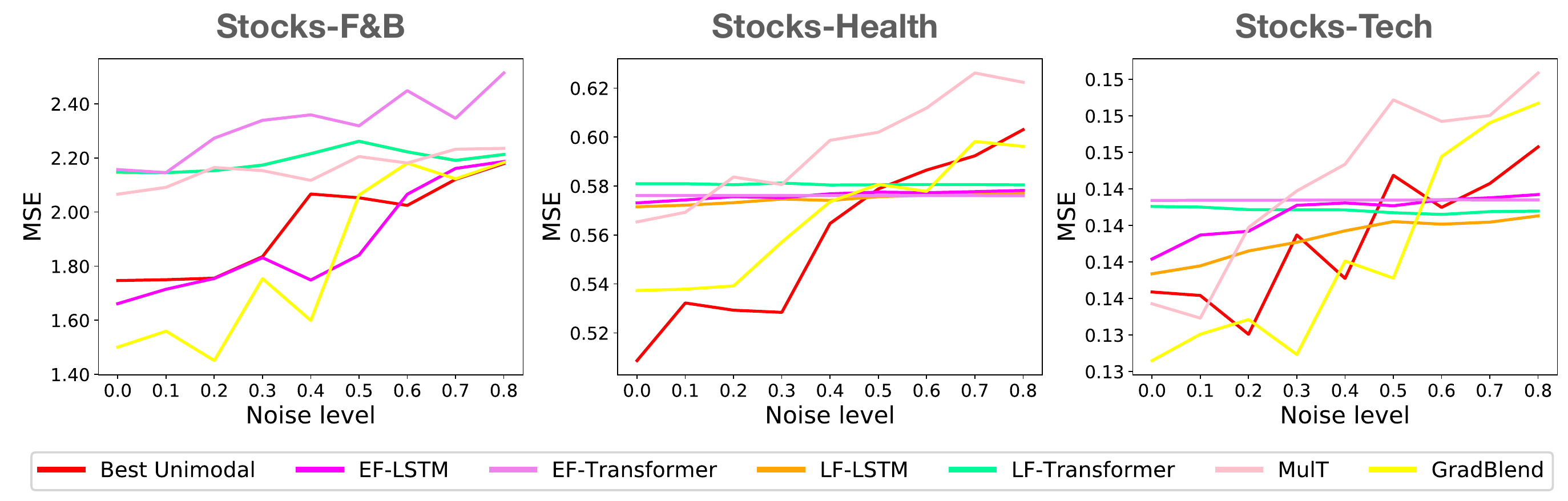}
\caption{Robustness of multimodal models with increasing levels of noise on the stock prediction datasets in the finance domain.\vspace{-2mm}}
\label{figs:robustness_finance}
\end{figure*}

Finally, we show the robustness of multimodal models with increasing levels of noise on the finance datasets in Figure~\ref{figs:robustness_finance}. We highlight the following observations:
\begin{enumerate}
    \item We again observe a similar trend where the best multimodal models (\textsc{MulT} and sometimes \textsc{EF}) are more robust than the best unimodal model. However, different from other datasets, we find that certain multimodal models can be worse in performance and robustness than the best unimodal model. \textsc{LF} in particular is not very robust and performs worse than the best unimodal method.
    \item The Gradient Blend (\textsc{GradBlend}) method is interesting since it starts off with the best (lowest) MSE but is the least robust -- its error increases really quickly and ends up worse than several models that it was initially outperforming on $0$ noise levels.
    \item We find that several approaches might be underfitting the data on \textsc{Stocks-Health} and \textsc{Stocks-Tech}. These methods do not start off with a good MSE and are also not affected significantly at increasing noise levels, showing a roughly straight horizontal line in Figure~\ref{figs:robustness_finance}.
\end{enumerate}

\clearpage

\vspace{-1mm}
\subsection{HCI}
\vspace{-1mm}

\begin{table*}[]
\fontsize{9}{11}\selectfont
\setlength\tabcolsep{6.0pt}
\caption{Results on the \textsc{ENRICO} dataset in the HCI domain. \textbf{U}: unimodal models, \textbf{M}: multimodal fusion paradigms, \textbf{O}: optimization objectives, \textbf{T}: training structures. Several out-domain methods perform well on \textsc{MIMIC} and improve upon the current state-of-the-art performance on in-domain methods.}
\centering
\footnotesize
\vspace{-0mm}
\begin{tabular}{l|l|c}
\Xhline{3\arrayrulewidth}
& Dataset & \multicolumn{1}{c}{\textsc{ENRICO}} \\
& Metric & Acc$(20)$ $\uparrow$ \\
\Xhline{0.5\arrayrulewidth}
\multirow{2}{*}{\textbf{U}} & Unimodal ($i$) & $47.0 \pm 1.6$ \\
& Unimodal ($s$) & $46.1 \pm 1.3$ \\
\Xhline{0.5\arrayrulewidth}
\multirow{4}{*}{\textbf{M}} & \textsc{LF} & $50.8 \pm 2.0$ \\
& \textsc{TF}~\citep{zadeh2017tensor} & $46.6 \pm 1.9$\\
& \textsc{LRTF}~\citep{liu2018efficient} & $47.1 \pm 2.9$\\
& \textsc{MI-Matrix}~\citep{Jayakumar2020Multiplicative} & $46.7 \pm 2.4$ \\

\Xhline{0.5\arrayrulewidth}
\multirow{2}{*}{\textbf{O}} & \textsc{CCA}~\citep{sun2020learning} & $50.1 \pm 1.4$\\
& \textsc{RefNet}~\cite{sankaran2021multimodal} & $44.4 \pm 2.2$\\

\Xhline{0.5\arrayrulewidth}
\multirow{1}{*}{\textbf{T}} & \textsc{GradBlend}~\citep{wang2020makes} & $\mathbf{51.0 \pm 1.4}$ \\

\Xhline{3\arrayrulewidth}
\end{tabular}

\vspace{-2mm}
\label{results:hci_supp}
\end{table*}

We show the full results in Table~\ref{results:hci_supp} and results on complexity in Table~\ref{results:hci_complexity}. Here we list some observations regarding these results:
\begin{enumerate}
    \item The \textsc{ENRICO} paper \cite{leiva2020enrico} does not include code or provide many details about their experiments (e.g., data splits, hyperparameters). Compared to their reported results, our reproduction resulted in better performance for the set modality and worse performance for the screenshot modality.
    \item Using multiple modalities can help prediction on \textsc{ENRICO}, boosting performance over the best unimodal model by $4\%$.
    \item Similar to finance, there has been very little research in multimodal models for HCI. We observe decent performance of several out of domain methods, especially \textsc{GradBlend} which offers a slight improvement over a standard \textsc{LF} model.
    \item Certain more complex methods, unfortunately, do not work that well on this dataset. On the architecture side, more expressive methods such as \textsc{TF}, \textsc{LRTF} and \textsc{MI} do not offer improvements over a simple \textsc{LF} model. We hypothesize that these more complex models have a larger number of trainable parameters which make them more prone to overfitting to small and noisy datasets.
\end{enumerate}

\begin{table*}[]
\fontsize{9}{11}\selectfont
\setlength\tabcolsep{3.0pt}
\caption{Complexity results for datasets in the HCI domain.  \textbf{U}: unimodal models, \textbf{M}: multimodal fusion paradigms, \textbf{O}: optimization objectives, \textbf{T}: training structures.}
\centering
\footnotesize
\vspace{-0mm}
\begin{tabular}{l|l|cccccc}
\Xhline{3\arrayrulewidth}
& Dataset & \multicolumn{6}{c}{\textsc{ENRICO}} \\
& Metric & \begin{tabular}[c]{@{}c@{}}Epochs \\ trained\end{tabular} & \begin{tabular}[c]{@{}c@{}}Training \\ time (s)\end{tabular} & \begin{tabular}[c]{@{}c@{}}Training \\ params (M)\end{tabular} & \begin{tabular}[c]{@{}c@{}}Training peak \\ memory (MB)\end{tabular} & \begin{tabular}[c]{@{}c@{}}Inference \\ time (s)\end{tabular} & \begin{tabular}[c]{@{}c@{}}Inference \\ params (M) \end{tabular} \\
\Xhline{0.5\arrayrulewidth}
\multirow{2}{*}{\textbf{U}} & Unimodal ($i$) & $50$ & $ 1601$ & $9.6$ & $2796$ & $7.3$ & $19.3$ \\
& Unimodal ($s$) & $50$ & $1644$ & $9.6$ &$2771$ & $8.1$ & $19.3$ \\

\Xhline{0.5\arrayrulewidth}
\multirow{4}{*}{\textbf{M}} & \textsc{LF} & $50$ & $1714$ &$19.3$ & $2730$ & $8.7$ & $19.3$ \\
& \textsc{TF}~\citep{zadeh2017tensor} & $50$ & $2012$ &  $19.3$ & $2718$ & $10.9$ & $19.3$\\
& \textsc{LRTF}~\citep{liu2018efficient} & $50$ & $1853$ &  $19.3$ & $2717$ & $9.7$ & $19.3$\\
& \textsc{MI-Matrix}~\citep{Jayakumar2020Multiplicative} & $50 $ & $1604$ &  $19.3$ & $2730$ & $8.5$ & $19.3$ \\

\Xhline{0.5\arrayrulewidth}
\multirow{2}{*}{\textbf{O}} & \textsc{CCA}~\citep{sun2020learning} & $50$ & $2945$ &  $19.3$ & $2923$ & $9.1$ & $19.3$ \\
& \textsc{RefNet}~\cite{sankaran2021multimodal} & $50$ & $1747$ &  $25.7$ & $2757$ & $13.8$ & $25.7$ \\
\Xhline{0.5\arrayrulewidth}
\multirow{1}{*}{\textbf{T}} & \textsc{GradBlend}~\citep{wang2020makes} & $50$ & $2618$ &  $19.3$ & $2610$ & $12.1$ & $19.3$\\

\Xhline{3\arrayrulewidth}
\end{tabular}

\vspace{-2mm}
\label{results:hci_complexity}
\end{table*}

\begin{figure*}[]
\centering
    \includegraphics[width=\textwidth]{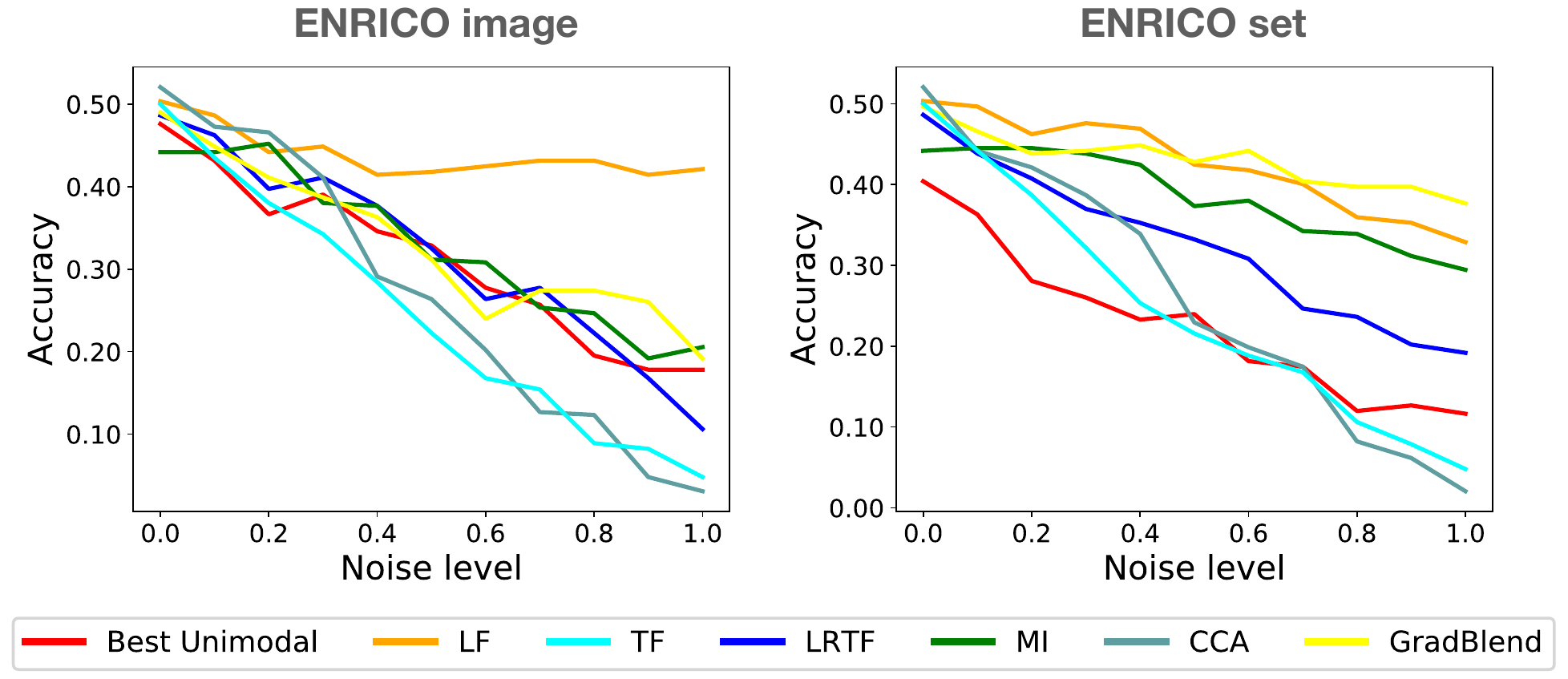}
\caption{Robustness of multimodal models with increasing levels of noise on the \textsc{ENRICO} dataset in the HCI domain. \vspace{-2mm}}
\label{figs:robustness_hci}
\end{figure*}

We show robustness results with increasing levels of noise in Figure~\ref{figs:robustness_hci}. We highlight the following observations:
\begin{enumerate}
    \item We again observe a similar trend where the best multimodal models (\textsc{LF} and sometimes \textsc{GradBlend}) are more robust than the best unimodal model. However, different from other datasets, we find that certain multimodal models can be worse in performance and robustness than the best unimodal model. \textsc{TF} in particular is not robust and performs worse than the best unimodal method.
    \item \textsc{LF} is surprisingly robust to imperfections in the image modality and shows a very stable trend despite high levels of noise, implying that the model has learned to rely on the set modality instead when the image is imperfect.
    \item Multimodal models show a high variance in robustness at high noise levels -- performance can range from $5\%$ to $40\%$ at the highest noise levels.
\end{enumerate}

\clearpage

\vspace{-1mm}
\subsection{Multimedia}
\vspace{-1mm}

\begin{table*}[]
\fontsize{9}{11}\selectfont
\setlength\tabcolsep{6.0pt}
\caption{Results on multimodal datasets in the multimedia domain. \textbf{U}: unimodal models, \textbf{M}: multimodal fusion paradigms, \textbf{O}: optimization objectives, \textbf{T}: training structures. We observe high variance in model performance across datasets with no method showing consistently strong performance.}
\centering
\footnotesize
\vspace{-0mm}
\begin{tabular}{l|l|cc}
\Xhline{3\arrayrulewidth}
& Dataset & \multicolumn{2}{c}{\textsc{MM-IMDb}} \\
& Metric & Micro F1$(23)$ $\uparrow$ & Macro F1$(23)$ $\uparrow$ \\
\Xhline{0.5\arrayrulewidth}
\multirow{2}{*}{\textbf{U}} & Unimodal ($\ell$) & $ 58.6\pm1.3 $ & $ 45.6\pm4.5 $ \\
& Unimodal ($i$) & $ 40.1\pm1.3 $ & $ 25.3\pm0.6 $ \\

\Xhline{0.5\arrayrulewidth}
\multirow{4}{*}{\textbf{M}} & \textsc{EF} & $ 58.9\pm2.6 $ & $ 49.8\pm1.7 $\\
& \textsc{LF} & $ 58.8\pm1.6 $ & $ 49.2\pm2.0 $\\
& \textsc{LRTF}~\citep{liu2018efficient} & $ 59.2\pm0.5 $ & $49.2\pm0.6$\\
& \textsc{MI-Matrix}~\citep{Jayakumar2020Multiplicative} & $ 58.3\pm1.0 $ & $48.0\pm1.1$ \\

\Xhline{0.5\arrayrulewidth}
\multirow{3}{*}{\textbf{O}} & \textsc{CCA}~\citep{sun2020learning} & $ \mathbf{59.3\pm1.2} $ & $ \mathbf{50.2\pm0.9} $ \\
& \textsc{RefNet}~\cite{sankaran2021multimodal} & $ 59.2\pm2.7 $ & $50.2\pm1.4 $ \\
& \textsc{MFM}~\cite{tsai2019learning} & $ 38.4\pm1.6 $ & $22.3\pm1.3$\\

\Xhline{0.5\arrayrulewidth}
\multirow{1}{*}{\textbf{T}}
& \textsc{RMFE} & $ 58.6\pm2.3 $ & $47.1\pm2.0$ \\

\Xhline{3\arrayrulewidth}
\end{tabular}

\vspace{4mm}

\begin{tabular}{l|l|c}
\Xhline{3\arrayrulewidth}
& Dataset & \multicolumn{1}{c}{\textsc{AV-MNIST}} \\
& Metric & Acc$(10)$ $\uparrow$ \\
\Xhline{0.5\arrayrulewidth}
\multirow{2}{*}{\textbf{U}} & Unimodal ($i$) & $65.1 \pm 0.2$ \\
& Unimodal ($a$) & $42.0 \pm 0.2$ \\
\Xhline{0.5\arrayrulewidth}
\multirow{4}{*}{\textbf{M}} & \textsc{LF} & $71.7 \pm 0.4$ \\
& \textsc{LRTF}~\citep{liu2018efficient} & $71.5 \pm 0.5$ \\
& \textsc{MI-Matrix}~\citep{Jayakumar2020Multiplicative} & $71.2 \pm 0.5$ \\
& \textsc{MFAS}~\citep{perez2019mfas} & $\mathbf{72.8 \pm 0.2}$ \\

\Xhline{0.5\arrayrulewidth}
\multirow{4}{*}{\textbf{O}} & \textsc{CCA}~\citep{sun2020learning} & $ 71.9 \pm 0.4 $ \\
& \textsc{RefNet}~\cite{sankaran2021multimodal} & $ 70.9 \pm 0.6 $ \\
& \textsc{MFM}~\cite{tsai2019learning} & $ 71.8 \pm 0.4 $ \\
& \textsc{MVAE}~\citep{wu2018multimodal} & $72.3 \pm 0.2$ \\

\Xhline{0.5\arrayrulewidth}
\multirow{1}{*}{\textbf{T}} & \textsc{GradBlend}~\citep{wang2020makes} & $68.5 \pm 0.5$ \\

\Xhline{3\arrayrulewidth}
\end{tabular}

\vspace{4mm}

\begin{tabular}{l|l|c|c}
\Xhline{2\arrayrulewidth}
& Dataset & \multicolumn{1}{c|}{\textsc{Kinetics-S}}  & \multicolumn{1}{c}{\textsc{Kinetics-L}} \\
& Metric & Acc$(5)$ $\uparrow$ & Acc$(400)$ $\uparrow$ \\
\Xhline{0.5\arrayrulewidth}
\multirow{2}{*}{\textbf{U}} & Unimodal ($v$) & $\mathbf{56.5}$ & $72.6$ \\
& Unimodal ($a$) & $39.7$ & $19.7$ \\
\Xhline{0.5\arrayrulewidth}
\multirow{1}{*}{\textbf{M}} & \textsc{LF} & $56.1$ & $71.7$ \\
\Xhline{0.5\arrayrulewidth}
\multirow{1}{*}{\textbf{T}} & \textsc{GradBlend}~\citep{wang2020makes} & $23.7$ & $\mathbf{74.7}$ \\

\Xhline{3\arrayrulewidth}
\end{tabular}

\vspace{-2mm}
\label{results:multimedia_supp}
\end{table*}

We show the full results in Table~\ref{results:multimedia_supp} and results on complexity in Table~\ref{results:multimedia_complexity}. Here we list some observations regarding these results:
\begin{enumerate}
    \item The current SOTA on \textsc{AV-MNIST} is based on architecture search: \textsc{MFAS}~\cite{perez2019mfas}. Amongst all the methods we evaluated, \textsc{MFAS} is still the best performing method and beats the second best method (\textsc{MVAE}) by $0.5\%$. Meanwhile, Gradient Blend (\textsc{GradBlend}) does not seem to generalize well to this dataset, as it performs worse than all other multimodal methods.
    \item On \textsc{MM-IMDb}, we attempted several methods on the objective function side. We found that using contrastive learning (\textsc{RefNet})~\cite{sankaran2021multimodal} or canonical correlation analysis (\textsc{CCA}) were quite useful in improving performance, with both outperforming purely architectural baselines without alignment as an optimization objective. In particular, while the \textsc{CCA} approach for multimodal fusion was originally proposed for affect recognition datasets~\citep{sun2020learning}, we find that they also generalize to the multimedia domain.
    \item On \textsc{Kinetics}, Gradient Blend (\textsc{GradBlend})~\citep{wang2020makes} was shown to work really well in their original paper. However, we found that this approach does not generalize well to other datasets such as \textsc{AV-MNIST}. We also created a smaller version of Kinetics called \textsc{Kinetics-S} to enable quick prototyping of multimodal models. Unfortunately, we found that \textsc{GradBlend} also struggles on the smaller partition of Kinetics.
    \item For \textsc{Kinetics-S}, we also observed that the visual unimodal model slightly outperformed the late fusion model despite the latter using more modalities. This reflects the observations by Wang et al.,~\citep{wang2020makes} on the original full version of the \textsc{Kinetics} dataset.
    \item Therefore, we find that multimodal models still struggle on the \textsc{Kinetics} dataset with multimodal performance on simple models (\textsc{LF}) unable to outperform unimodal methods. While \textsc{GradBlend} can improve multimodal performance, it comes at the expense of $\sim 3\times$ the training time. Future research should explore building lightweight and effective multimodal models on \textsc{Kinetics} as well as other datasets in \names.
\end{enumerate}

\begin{table*}[]
\fontsize{9}{11}\selectfont
\setlength\tabcolsep{3.0pt}
\caption{Complexity results for datasets in the multimedia domain. ((*) This is the number of params in modules input to \textsc{MFAS} at the start of training, \textsc{MFAS} will generate more params during the architecture search process). \textbf{U}: unimodal models, \textbf{M}: multimodal fusion paradigms, \textbf{O}: optimization objectives, \textbf{T}: training structures.}
\centering
\footnotesize
\vspace{-0mm}

\begin{tabular}{l|l|cccccc}
\Xhline{3\arrayrulewidth}
& Dataset & \multicolumn{6}{c}{\textsc{MM-IMDb}} \\
& Metric & \begin{tabular}[c]{@{}c@{}}Epochs \\ trained\end{tabular} & \begin{tabular}[c]{@{}c@{}}Training \\ time (s)\end{tabular} & \begin{tabular}[c]{@{}c@{}}Training \\ params (M)\end{tabular} & \begin{tabular}[c]{@{}c@{}}Training peak \\ memory (MB)\end{tabular} & \begin{tabular}[c]{@{}c@{}}Inference \\ time (s)\end{tabular} & \begin{tabular}[c]{@{}c@{}}Inference \\ params (M) \end{tabular} \\
\Xhline{0.5\arrayrulewidth}
\multirow{2}{*}{\textbf{U}} & Unimodal ($\ell$) & $125$ & $622$ & $0.55$ & $2146$ & $2.07$ & $0.55$ \\
& Unimodal ($i$) & $25$ & $127$ & $4.86$ & $2176$ & $2.14$ & $4.86$ \\
\Xhline{0.5\arrayrulewidth}
\multirow{8}{*}{\textbf{M}} & \textsc{EF} & $15$ & $117$ & $5.05$ & $2010$ & $3.24$ & $5/05$\\
& \textsc{LF} & $5$ & $45$ & $10.3$ & $2016$ & $3.44$ & $10.3$ \\
& \textsc{LRTF}~\citep{liu2018efficient} & $15$ & $741$ & $10.3$ & $2448$ & $5.57$ & $10.3$\\
& \textsc{MI-Matrix}~\citep{Jayakumar2020Multiplicative} & $20$ & $735$ & $280$ & $4036$ & $3.59$ & $280$\\
& \textsc{MFM}~\cite{tsai2019learning} & $10$ & $78$ & $21.3$ & $2038$ & $3.36$ & $10.9$\\
& \textsc{CCA}~\citep{sun2020learning} & $20$ & $1025$ & $9.51$ & $2273$ & $3.33$ & $9.51$\\
& \textsc{RMFE}~\cite{gat2020removing} & $10$ & $104$ & $8.78$ & $22297$ & $3.46$ & $8.78$ \\
& \textsc{RefNet}~\cite{sankaran2021multimodal} & $10$ & $2207$ & $27.0$ & $2899$ & $3.47$ & $10.3$ \\ 
\Xhline{3\arrayrulewidth}
\end{tabular}

\vspace{4mm}

\begin{tabular}{l|l|cccccc}
\Xhline{3\arrayrulewidth}
& Dataset & \multicolumn{6}{c}{\textsc{AV-MNIST}} \\
& Metric & \begin{tabular}[c]{@{}c@{}}Epochs \\ trained\end{tabular} & \begin{tabular}[c]{@{}c@{}}Training \\ time (s)\end{tabular} & \begin{tabular}[c]{@{}c@{}}Training \\ params (M)\end{tabular} & \begin{tabular}[c]{@{}c@{}}Training peak \\ memory (MB)\end{tabular} & \begin{tabular}[c]{@{}c@{}}Inference \\ time (s)\end{tabular} & \begin{tabular}[c]{@{}c@{}}Inference \\ params (M) \end{tabular} \\
\Xhline{0.5\arrayrulewidth}
\multirow{2}{*}{\textbf{U}} & Unimodal ($i$) & $25$ & $106$  & $0.02$ & $9549$ & $0.95$ & $0.02$ \\
& Unimodal ($a$) & $25$ & $158$ & $0.24$ & $11895$ & $1.35$ & $0.24$ \\

\Xhline{0.5\arrayrulewidth}
\multirow{4}{*}{\textbf{M}} & \textsc{LF} & $25$ & $260$ & $0.26$ & $11917$ & $1.20$ & $0.26$ \\
& \textsc{MI-Matrix}~\citep{Jayakumar2020Multiplicative} & $25$ & $289$ & $2.53$ & $11509$ & $1.21$ & $2.53$ \\
& \textsc{LRTF}~\citep{liu2018efficient} & $30$ & $470$ & $0.25$ & $11610$ & $1.25$ & $0.25$  \\
& \textsc{MFAS}~\citep{perez2019mfas} & $172\times 6$ & $17648$ & $0.14^*$ & $9444$ & $4.39$ & $0.07$ \\

\Xhline{0.5\arrayrulewidth}
\multirow{4}{*}{\textbf{O}} & \textsc{CCA}~\citep{sun2020learning} & $25$ & $310$ & $0.25$ & $9548$ & $1.42$ & $0.25$ \\
& \textsc{RefNet}~\cite{sankaran2021multimodal} & $15$ & $1179$ & $14.01$ & $15931$ & $4.39$ & $0.28$ \\
& \textsc{MFM}~\cite{tsai2019learning} & $25$ & $544$ & $0.92$ & $9570$ & $4.76$ & $0.45$ \\
& \textsc{MVAE}~\citep{wu2018multimodal} & $20$ & $679$ & $0.81$ & $9755$ & $4.98$ & $0.34$ \\

\Xhline{0.5\arrayrulewidth}
\multirow{1}{*}{\textbf{T}} & \textsc{GradBlend}~\cite{wang2020makes} & $300$ & $12539$ & $0.29$ & $12029$ & $1.51$ & $0.26$ \\

\Xhline{3\arrayrulewidth}
\end{tabular}

\vspace{4mm}

\begin{tabular}{l|l|cccccc}
\Xhline{3\arrayrulewidth}
& Dataset & \multicolumn{6}{c}{\textsc{Kinetics-Small}} \\
& Metric & \begin{tabular}[c]{@{}c@{}}Epochs \\ trained\end{tabular} & \begin{tabular}[c]{@{}c@{}}Training \\ time (s)\end{tabular} & \begin{tabular}[c]{@{}c@{}}Training \\ params (M)\end{tabular} & \begin{tabular}[c]{@{}c@{}}Training peak \\ memory (MB)\end{tabular} & \begin{tabular}[c]{@{}c@{}}Inference \\ time (s)\end{tabular} & \begin{tabular}[c]{@{}c@{}}Inference \\ params (M) \end{tabular} \\
\Xhline{0.5\arrayrulewidth}
\multirow{2}{*}{\textbf{U}} & Unimodal ($v$) & $15$ & $6702$ & $12.0$ & $12151$ & $13.7$ & $12.0$ \\
& Unimodal ($a$) & $15$ & $46767$ & $25.8$ & $8533$ & $60.9$ & $25.8$ \\
\Xhline{0.5\arrayrulewidth}
\multirow{1}{*}{\textbf{M}} & \textsc{LF} & $15$ & $20283$ & $37.8$ & $9525$ & $13.9$ & $37.8$ \\
\Xhline{0.5\arrayrulewidth}
\multirow{1}{*}{\textbf{T}} & \textsc{GradBlend}~\cite{wang2020makes} & $15$ & $20283$ & $37.8$ & $9525$ & $13.9$ & $37.8$ \\
\Xhline{3\arrayrulewidth}
\end{tabular}

\vspace{4mm}

\begin{tabular}{l|l|cccccc}
\Xhline{3\arrayrulewidth}
& Dataset & \multicolumn{6}{c}{\textsc{Kinetics-Large}} \\
& Metric & \begin{tabular}[c]{@{}c@{}}Epochs \\ trained\end{tabular} & \begin{tabular}[c]{@{}c@{}}Training \\ time (s)\end{tabular} & \begin{tabular}[c]{@{}c@{}}Training \\ params (M)\end{tabular} & \begin{tabular}[c]{@{}c@{}}Training peak \\ memory (MB)\end{tabular} & \begin{tabular}[c]{@{}c@{}}Inference \\ time (s)\end{tabular} & \begin{tabular}[c]{@{}c@{}}Inference \\ params (M) \end{tabular} \\
\Xhline{0.5\arrayrulewidth}
\multirow{2}{*}{\textbf{U}} & Unimodal ($v$) & $45$ & $938280$ & $12.0$ & $12151$ & $1918$ & $12.0$ \\
& Unimodal ($a$) & $45$ & $947380$ & $33.5$ & $8533$ & $8526$ & $33.5$ \\
\Xhline{0.5\arrayrulewidth}
\multirow{1}{*}{\textbf{M}} & \textsc{LF} & $45$ & $2839620$ & $45.5$ & $9525$ & $1946$ & $45.5$ \\
\Xhline{0.5\arrayrulewidth}
\multirow{1}{*}{\textbf{T}} & \textsc{GradBlend}~\cite{wang2020makes} & $45$ & $2839620$ & $45.5$ & $9525$ & 1946 & $45.5$ \\
\Xhline{3\arrayrulewidth}
\end{tabular}

\vspace{-2mm}
\label{results:multimedia_complexity}
\end{table*}

\begin{figure*}[]
\centering
    \includegraphics[width=\textwidth]{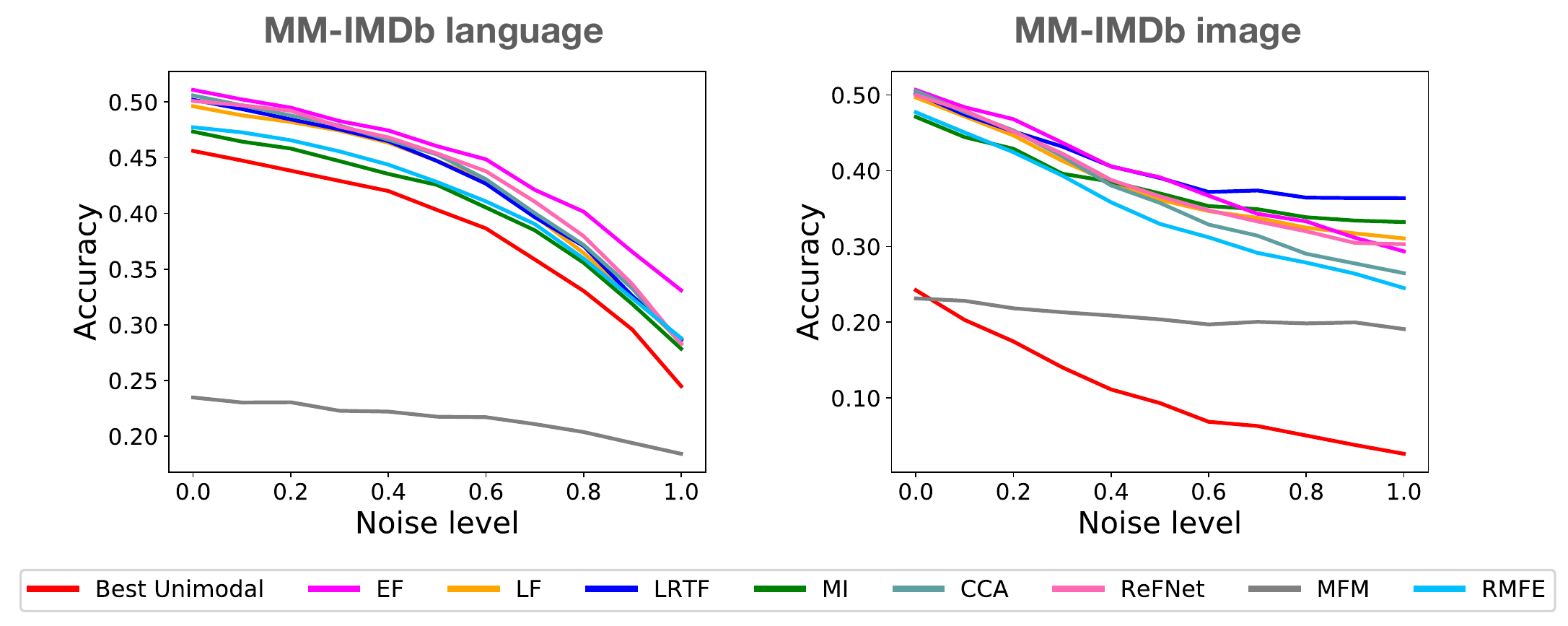}
\caption{Robustness of multimodal models with increasing levels of noise on the \textsc{MM-IMDb} dataset in the multimedia domain. \vspace{-2mm}}
\label{figs:robustness_imdb}
\end{figure*}

We show robustness results with increasing levels of noise on the \textsc{MM-IMDb} datasets in Figure~\ref{figs:robustness_imdb}. We highlight the following observations:
\begin{enumerate}
    \item Multimodal models outperform unimodal models when it comes to robustness (and initial performance). This is especially true for imperfections in the image modality. We believe that multimodal models are able to successfully rely on the other modality when one is imperfect. We find that the gap between multimodal and unimodal performance is very significant on the image modality. However, the gap is much smaller on the text modality.
    \item \textsc{MFM} was a method tested initially for affective computing datasets but we found it did not generalize to \textsc{MM-IMDb}, giving poor initial performance and poor robustness. We believe the high-dimensionality of image and text input means that reconstruction of input modalities is difficult, which causes reconstruction-based objectives in \textsc{MFM} to suffer.
    \item On the whole, multimodal models are more robust to imperfections on the image modality as compared to the language modality. However, unimodal performance is much better on language than on image, which implies that the language modality is more informative. Similar to the observations on the affective computing datasets, we also find that multimodal models tend to overfit to the more informative modality (language) and are therefore less robust to imperfections in the more informative modality.
\end{enumerate}

\clearpage

\vspace{-1mm}
\subsection{Performance}
\label{appendix:performance}
\vspace{-1mm}

In this subsection, we summarize several general observations regarding the performance of multimodal models across domains, modalities, and tasks.

In the following analysis, we aggregate the performance of models by first assigning each task a weight of $\frac{1}{n}$ where $n$ is the number of tasks in a dataset (e.g., there are $3$ tasks in the \textsc{MIMIC} dataset: mortality, ICD-$9$ group 1, and ICD-$9$ group 7 prediction). Then, we compute the scaled performance of each model on each task by min-max normalization -- setting the best-performing model's performance to $1$ and worst-performing model's performance to $0$, and scaling the performance of all remaining models linearly between $0$ and $1$. Note that we only take the best unimodal performance into account when determining the best and worst-performing models. Then, the final performance score for each model is computed by a weighted average of its scaled performances on all tasks that model was evaluated on.

\textbf{Benefits of standardization:} Simply applying methods in a research different area achieves state-of-the-art performance on $9$ out of the $15$ tasks. We find that this is especially true for domains and modalities that have been relatively less studied in multimodal research (i.e., healthcare, finance, HCI). Performance gains can be obtained using multimodal methods \textit{outside} of that research area. Therefore, this motivates the benefits of standardizing and unifying areas of research in multimodal machine learning. We believe that the ever-expanding diversity of datasets in \names\ can greatly accelerate research in multimodal learning.

\begin{figure}
\centering
\includegraphics[width=\textwidth]{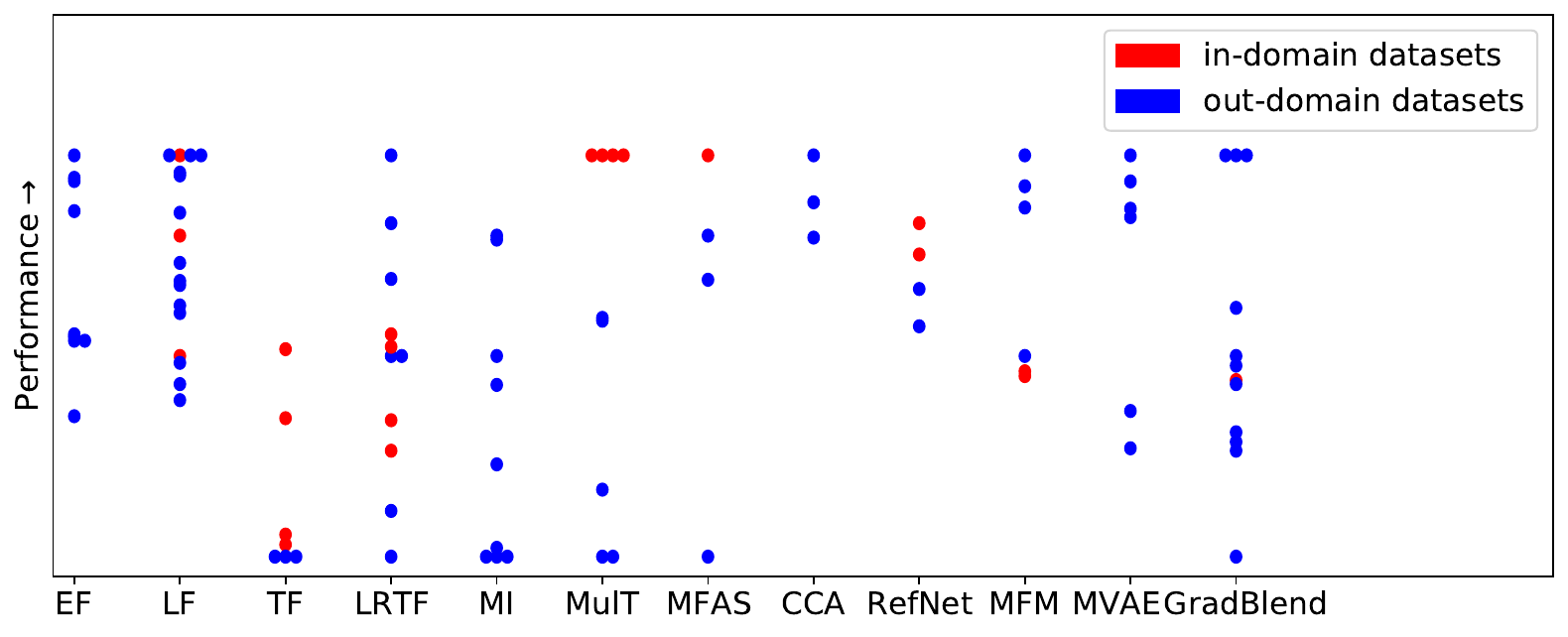}
\caption{Relative performance of each model across in-domain ({\color{red}{red}} dots) and out-domain datasets ({\color{blue}{blue}} dots). \textit{In-domain} refers to the performance on datasets that the method was previously proposed for and \textit{out-domain} shows performance on the remaining datasets. We find that many methods show strongest performance on in-domain datasets which drop when tested on different domains, modalities, and tasks. In general, we also observe high variance in the performance of multimodal methods across datasets in \names, which suggest open questions in building more generalizable models.\vspace{-2mm}}
\label{fig:perfonly1}
\end{figure}

\begin{figure}
\centering
\includegraphics[width=\textwidth]{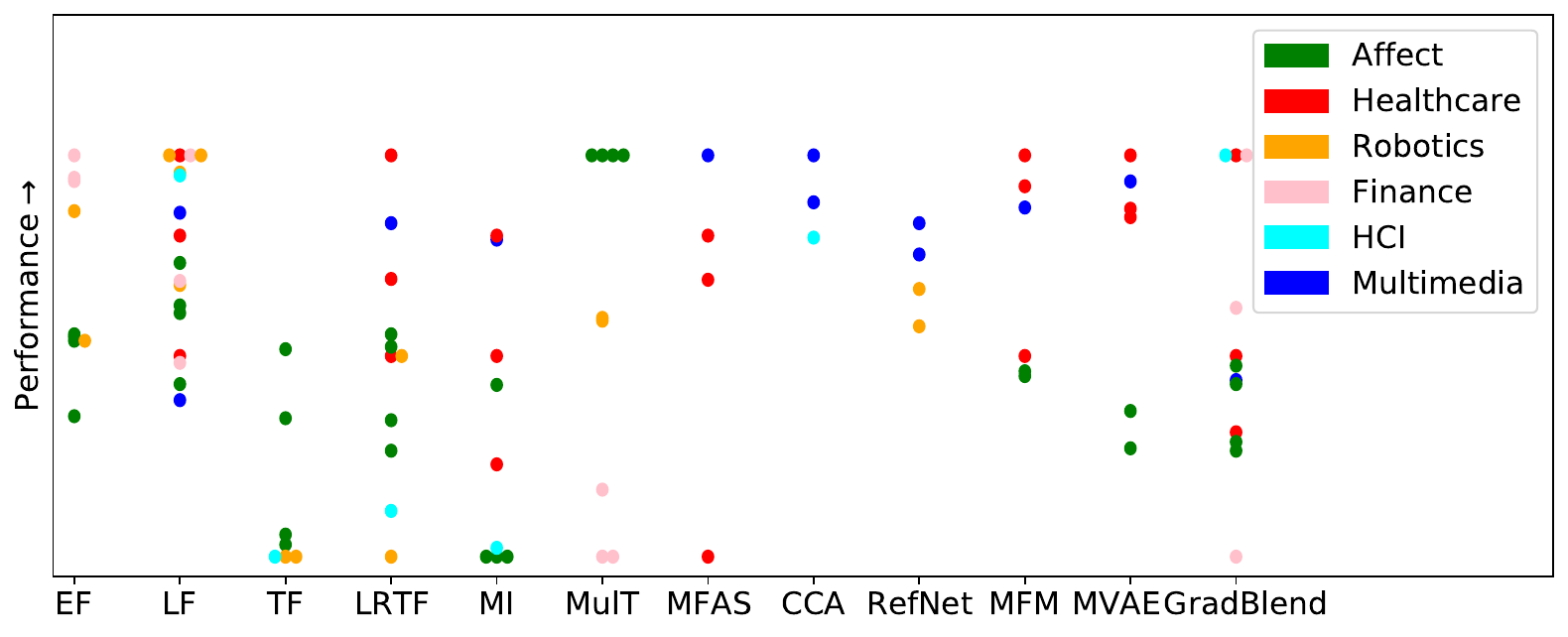}
\caption{Relative performance of each model across different domains. We find that the performance of multimodal models varies significantly across datasets spanning different research areas and modalities. Similarly, the best-performing methods on each domain are also different. Therefore, there still does not exist a one-size-fits-all model, especially for understudied modalities and tasks.\vspace{-2mm}}
\label{fig:perfonly2}
\end{figure}

\textbf{Generalization across domains and modalities:} \names\ offers an opportunity to analyze algorithmic developments across a large suite of modalities, domains, and tasks. We illustrate these observations through $2$ summary plots of the generalization performance of multimodal models. Firstly, in Figure~\ref{fig:perfonly1}, we plot the performance of each multimodal method across all datasets that it is tested on, using the color {\color{red}{red}} to indicate performance on datasets that it was initially proposed and tested on (which we label as \textit{in-domain}), and {\color{blue}{blue}} to indicate its performance on the remaining datasets (which we label as \textit{out-domain}). Secondly, in Figure~\ref{fig:perfonly2}, we color-code the performance on each dataset depending on which research area the dataset belongs to (one of the $6$ research areas covered in \names).

We summarize several observations regarding generalization across domains and modalities below:
\begin{enumerate}
    \item Many multimodal methods still do not generalize across domains and datasets. For examples, \textsc{MFAS}~\citep{perez2019mfas} works well on domains it was designed for (\textsc{AV-MNIST} and \textsc{MM-IMDb} in the multimedia domain), but does not generalize to other domains such as healthcare (\textsc{MIMIC}). Similarly, the method designed for robustness, \textsc{MCTN}~\citep{pham2019found}, does not generalize to datasets within the affective computing domain (\textsc{UR-FUNNY} and \textsc{MUStARD}). Finally, \textsc{GradBlend}~\citep{wang2020makes}, an approach specifically designed to improve generalization in multimodal learning and tested on video and audio datasets (e.g., Kinetics), does not perform well on other datasets. Therefore, there still does not exist a one-size-fits-all model, especially on understudied modalities and tasks.
    \item From Figure~\ref{fig:perfonly1}, we find that many methods show their strongest performance on in-domain datasets, and their performance drops when tested on different domains, modalities, and tasks. Some interesting observations are that \textsc{MulT} performs extremely well on the affect recognition datasets it was designed for but struggles on other multimodal time-series in the finance and robotics domains. On the other hand, \textsc{MFM} shows an impressive performance in generalizing to new domains, although its in-domain performance has been exceeded by several other methods.
    \item From Figure~\ref{fig:perfonly1}, we also observe high variance in the performance of multimodal methods across datasets in \names, which suggest open questions in building more generalizable models. We find that \textsc{LF} is quite stable and always achieves above-average performance.
    \item There are methods that are surprisingly generalizable across datasets. These are typically general modality-agnostic methods such as \textsc{LF}. While simple, it is a strong method that balances simplicity, performance, and low complexity. However, it does not achieve the best performance on any dataset, which suggests that it is a good starting point but perhaps not the best eventual method.
    \item From Figure~\ref{fig:perfonly2}, we find that performance also varies significantly across research areas.
    \item Several methods such as \textsc{MFAS} and \textsc{CCA} are designed for only $2$ modalities (usually image and text), and \textsc{TF} and \textsc{MI} do not scale efficiently beyond $2/3$ modalities. Therefore, we were unable to directly adapt these approaches to other datasets. We encourage the community to generalize these approaches across datasets and modalities on \names.
\end{enumerate}

\textbf{Tradeoffs between modalities:} How far can we go with unimodal methods? Surprisingly far! From each of the individual tables, we observe that decent performance can be obtained with the best performing modality. Further improvement via multimodal models may come at the expense of around $2-3\times$ the parameters.

\clearpage

\vspace{-1mm}
\subsection{Complexity}
\label{appendix:results_complexity}
\vspace{-1mm}

We aggregate the complexity of each model by taking the weighted average of the relative complexity of the model across tasks on which it is evaluated. The weights are assigned in the same way as performance weights described in the subsection above (i.e., performing min-max normalization across models within each task and averaging across the normalized performance across all datasets that the model was tested on). The relative complexity of each model on each task is computed by dividing its training time by the best unimodal model's training time and taking the negative $\log10$ of this value (we take negative log because some more complex methods can take hundreds of times the training time of simpler methods). Thus, the higher the value of aggregated complexity, the faster the model trains.

\begin{figure*}[]
\centering
    \begin{minipage}{0.5\textwidth}
        \centering
        \subfloat[\centering All datasets]{{\includegraphics[width=\textwidth]{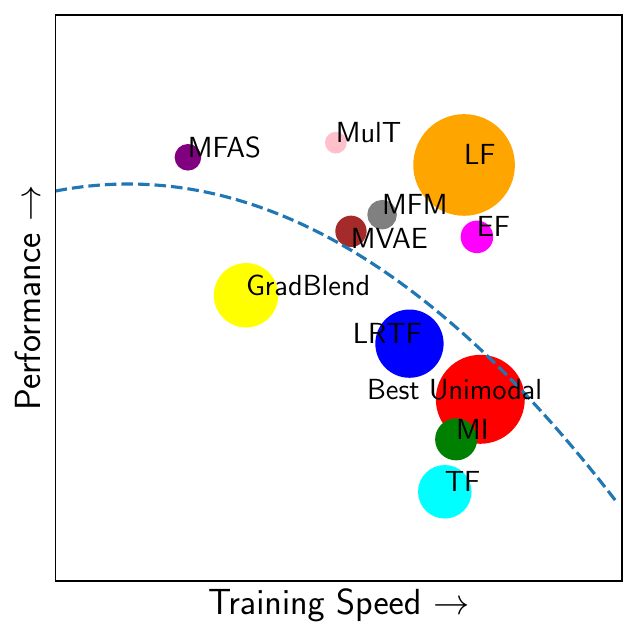}}}
    \end{minipage}%
    \begin{minipage}{0.5\textwidth}
        \centering
        \subfloat[\centering Datasets with $>6$ approaches]{{\includegraphics[width=\textwidth]{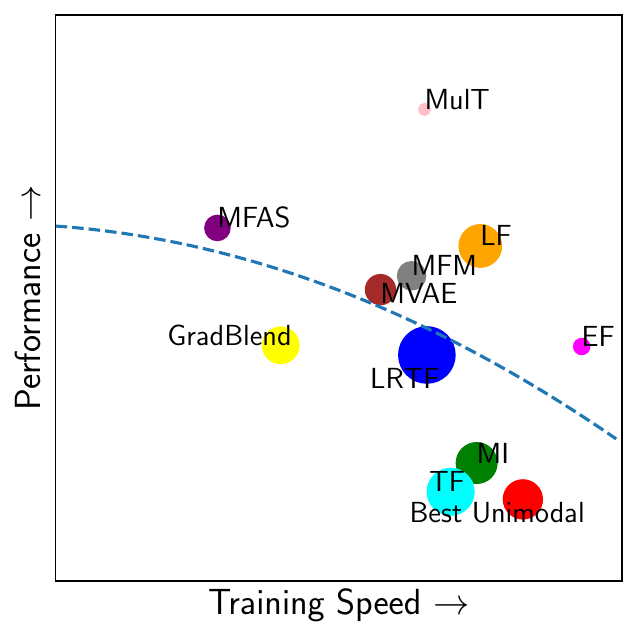}}}
    \end{minipage}
\caption{\textbf{Tradeoff between performance and complexity}. Size of circles shows variance in performance and complexity across (a) all datasets and (b) datasets on which we tested $>6$ approaches. We plot a dotted {\color{blue}{blue}} line of best quadratic fit to show the Pareto frontier between performance and complexity. These strong tradeoffs suggest that future work should focus on lightweight multimodal models that generalize throughout datasets in \names. It remains an open question whether several well-performing methods (such as \textsc{MFAS} or \textsc{MulT}) can be successfully adapted to new domains and tasks, since they work much better on more-studied datasets (b) as compared to over all datasets (a).\vspace{-4mm}}
\label{figs:tradeoff_complexity_supp}
\end{figure*}

Based on the full results above, we summarize the overall tradeoff between performance and complexity in Figure~\ref{figs:tradeoff_complexity_supp}(a). We aggregate performance and complexity statistics by first performing min-max normalization within each data to a scale of $0-1$ for performance and complexity separately. Note that for metrics where lower is better (i.e., MSE or RMSE) we reverse the direction of min-max normalization. We then aggregate normalized statistics across all datasets and plot the tradeoff between performance and complexity. We highlight the following observations:
\begin{enumerate}
    \item In Figure~\ref{figs:tradeoff_complexity_supp}, we plot a dotted {\color{blue}{blue}} line of best quadratic fit to show the Pareto frontier between performance and complexity. We choose a quadratic fit since it is common to fit a curve rather than a straight line when considering the tradeoff frontier between $2$ variables (related to the law of diminishing returns in economics). Using this plot, we find a strong tradeoff between these two desiderata: simple fusion techniques (e.g., early fusion \textsc{EF} and late fusion \textsc{LF}) are actually appealing choices that score high on both metrics, especially when compared to complex (but slightly better performing) methods such as architecture search \textsc{MFAS} or Multimodal Transformers \textsc{MulT}.
    \item Using this quadratic curve, we find that the best unimodal model is under the curve (i.e., worse-off than the Pareto front). This implies that while unimodal models train the fastest, several multimodal methods can outperform them despite being slightly slower, and is an overall better choice when taking both performance and complexity into account. \textsc{LF} is an appealing choice that lies above the curve.
    \item While \textsc{LF} is the easiest to adapt to new datasets and domains, we encountered difficulties in adapting several possibly well-performing methods (such as \textsc{MFAS} or \textsc{MulT}) to new datasets and domains. \textsc{MFAS} is designed with a specific set of atomic architectural elements in mind which makes it most suitable for convolutional networks. \textsc{MulT} is suitable for multimodal time-series data and it is unclear how to adapt its fusion paradigm to modalities without a temporal dimension. For a more fair comparison, in Figure~\ref{figs:tradeoff_complexity_supp}(b), we plot the accumulated performance for methods only on the most commonly studied datasets where we experimented with more than $6$ methods. We find that these well-performing methods (\textsc{MFAS} or \textsc{MulT}) show only slightly better than \textsc{LF} on all datasets (see Figure~\ref{figs:tradeoff_complexity_supp}(a)), they (see Figure~\ref{figs:tradeoff_complexity_supp}(b)). Therefore, it is important for future research to focus on models that \textit{generalize} to multiple domains, modalities, and tasks, since our results suggest that many methods currently do not satisfy these desiderata.
    \item These plots do not completely capture the picture since complexity is measured via total training time (training speed), which can be prohibitively high for methods such as \textsc{MFAS}, \textsc{MVAE}, and \textsc{GradBlend}. However, these methods are primarily slow due to extra parameters or training procedures during training, and once the model is trained, test-time inference is fast and cheap. Plotting a performance-complexity tradeoff using a different complexity metric will likely result in different observations. Overall, \names\ enables a holistic evaluation of training and test-time space and memory complexity so practitioners can choose the most suitable model for their real-world application setting.
\end{enumerate}

\clearpage

\vspace{-1mm}
\subsection{Robustness}
\label{appendix:results_robustness}
\vspace{-1mm}

\begin{figure*}[]
\centering
    \begin{minipage}{0.5\textwidth}
        \centering
        \subfloat[\centering Relative robustness]{{\includegraphics[width=\textwidth]{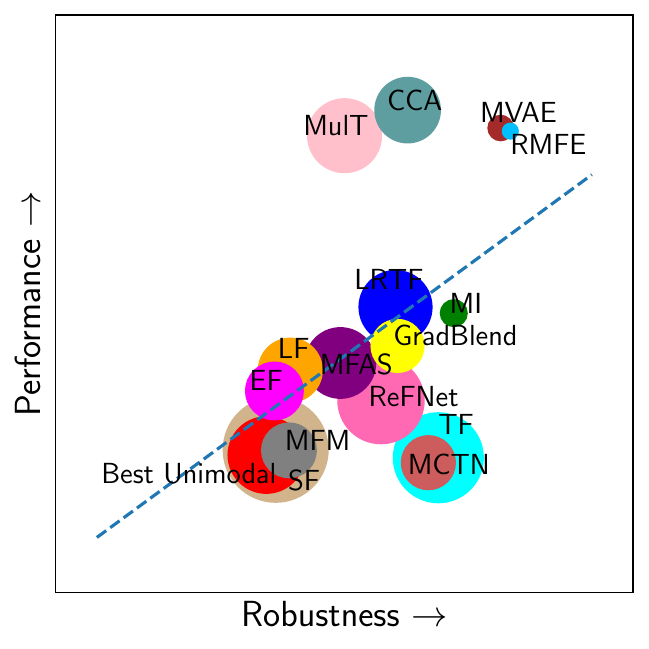}}}
    \end{minipage}%
    \begin{minipage}{0.5\textwidth}
        \centering
        \subfloat[\centering Effective robustness]{{\includegraphics[width=\textwidth]{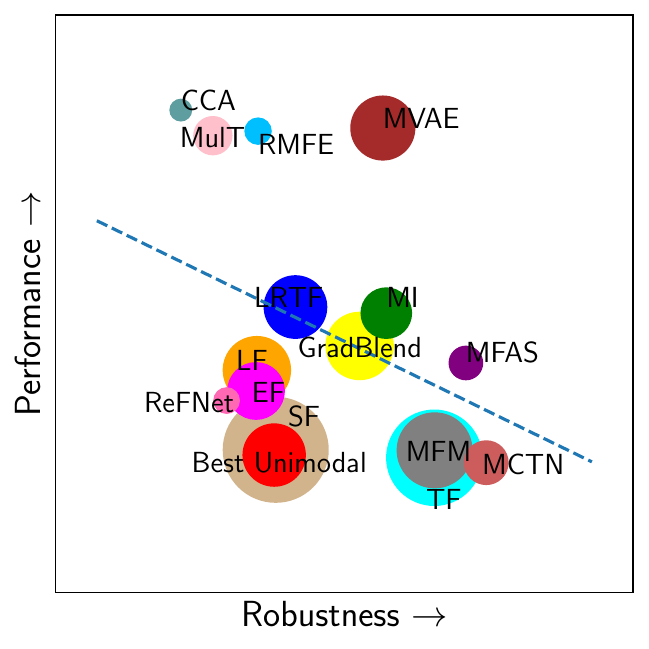}}}
    \end{minipage}
\caption{\textbf{Tradeoff between performance and robustness}. Size of circles shows variance in performance and robustness across datasets. We show the line of best linear fit in dotted {\color{blue}{blue}}. While better performing methods display better \textit{relative} robustness (as shown by the positive trend in the best linear fit in (a)), some of these methods suffer in \textit{effective} robustness. In other words, their performance \textit{drops off faster} as shown by the negative linear fit in (b). Few models currently achieve both relative and effective robustness, which suggests a crucial area for future multimodal research.\vspace{-4mm}}
\label{figs:tradeoff_robustness_supp}
\end{figure*}

\begin{figure*}[]
\centering
    \includegraphics[width=0.5\textwidth]{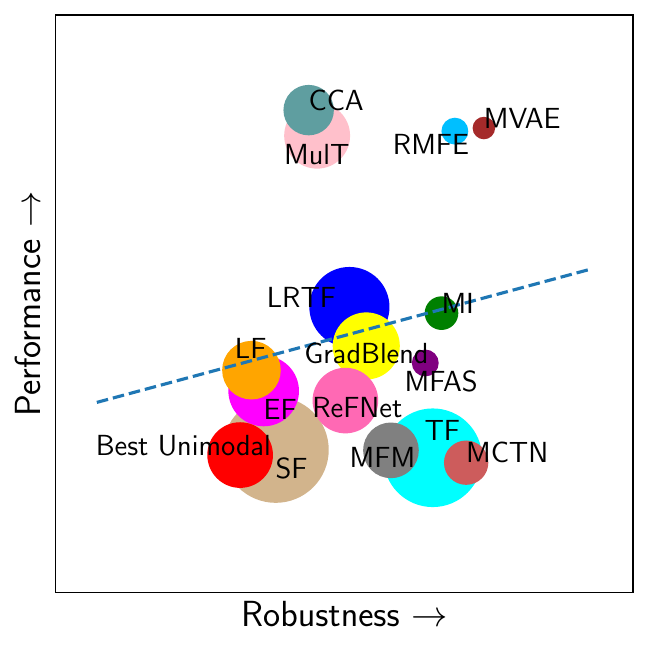}
\caption{\textbf{Overall tradeoff between performance and robustness} obtained by averaging the relative and effective robustness values in Figure~\ref{figs:tradeoff_robustness_supp}. We show the line of best linear fit in dotted {\color{blue}{blue}}. There is only a slight positive trend between performance and overall robustness of these multimodal models. Therefore, few well-performing models currently achieve both relative and effective robustness, which is a crucial area for future multimodal research.\vspace{-4mm}}
\label{figs:tradeoff_robustness2_supp}
\end{figure*}

In this section, we summarize our observations regarding the tradeoffs between accuracy and robustness, where we use the quantitative metrics for relative and effective robustness as described in Appendix~\ref{appendix:robustness}. As a reminder, relative robustness directly measures accuracy under imperfections while effective robustness measures the rate of accuracy drops with imperfection after equalizing for initial accuracy on clean test data. In Figure~\ref{figs:tradeoff_robustness_supp}, we plot a similar tradeoff plot between accuracy and (relative \& effective) robustness. Again, we aggregate performance and complexity statistics by first performing min-max normalization within each data to a scale of $0-1$ for performance and robustness separately. We aggregate normalized statistics across all datasets and plot the tradeoff between performance and robustness. We highlight the following observations:
\begin{enumerate}
    \item We show the line of best linear fit for relative and effective robustness in dotted {\color{blue}{blue}} in Figure~\ref{figs:tradeoff_robustness_supp}. We observe a slight positive correlation between performance and relative robustness, which implies that models starting off with higher accuracy tend to stay above other models on the performance-imperfection curve. In particular, several methods such as \textsc{MVAE} and \textsc{RMFE} show strong performance and robustness.
    \item However, we observe a slightly negative correlation for effective robustness. Unfortunately, several well-performing methods such as \textsc{MulT}, \textsc{CCA}, and \textsc{MVAE} tend to \textit{drop off faster} after equalizing for initial accuracy on clean test data.
    \item Finally, we plot an average of relative and effective robustness in Figure~\ref{figs:tradeoff_robustness2_supp} as an overall quantitative measure of robustness. We observe that very few models currently achieve both relative and effective robustness, which prompts an area for future multimodal research.
\end{enumerate}

\vspace{-1mm}
\subsection{Summary of Takeaway Messages}
\vspace{-1mm}

From these results, we emphasize the main take-away messages and motivate several directions for future work:
\begin{enumerate}
    \item Benefits of standardization: Applying methods in a research different area achieves state-of-the-art performance on $9$ out of the $15$ datasets, especially those relatively less studied in multimodal research (i.e., healthcare, finance, HCI). This motivates the benefits of standardizing and unifying areas of research in multimodal learning. We hope that \names\ and \codes\ can be a step in this direction.
    
    \item Generalization across domains and modalities:
    \begin{enumerate}
        \item Many multimodal methods still do not generalize across domains and datasets, showing high variance across datasets in \names. Some of these methods perform worse on out-of-domain datasets than in-domain datasets while other methods are designed in a specific manner for certain modalities and domains which makes them unable to be adapted to other datasets in straightforward ways.
        \item Certain simple methods (e.g., \textsc{LF}) are surprisingly generalizable. However, it does not achieve the best performance on any dataset, which suggests that it is a good starting point but perhaps not the best method.
    \end{enumerate}
    
    \item Decent performance can be obtained with the best performing modality, which motivates the need for new datasets that offer challenges and opportunities in multimodal modeling not achievable from unimodal methods.

    \item There is a strong tradeoff between performance and complexity which suggests that future work should also focus on lightweight multimodal models that generalize throughout datasets in \names.
    
    \item Tradeoffs between performance and robustness:
    \begin{enumerate}
        \item Models starting off with higher accuracy tend to stay above other models on the performance-imperfection curve.
        \item However, several well-performing methods also tend to \textit{drop off faster} after equalizing for initial accuracy on clean test data.
        \item Overall, very few models currently achieve both relative and effective robustness, which prompts an area for future multimodal research.
    \end{enumerate}
\end{enumerate}

\clearpage

\vspace{-2mm}
\section{Future Directions}
\label{appendix:future}
\vspace{-2mm}

We plan to ensure the continual availability, maintenance, and expansion of \names. Several immediate future directions include expansions in the datasets provided, algorithms implemented in \codes, and broadening the holistic evaluation of multimodal models.

\vspace{-1mm}
\subsection{Datasets}
\vspace{-1mm}

One main area of expansion lies in the datasets supported by \names. We first describe the categories of multimodal datasets in the fusion domain that we plan to add in the following months. We also plan to include several new application areas where multimodal fusion is useful, such as cross-modal retrieval, multimodal question answering, and grounding across modalities, which we will detail in the following subsections. Finally, we explain our plan for community-based expansion of datasets and models based on user feedback that will happen in parallel.

\vspace{-1mm}
\subsubsection{Fusion}
\vspace{-1mm}

Within the same category of multimodal fusion, we plan to add datasets within the same application domains as well as to expand to new application domains. Within the current domains, we plan to include (1) the hateful memes challenge~\citep{kiela2020hateful} as a core challenge in multimedia to ensure safer learning from ubiquitous text and images from the internet, (2) more datasets in the robotics and HCI domains where there are many opportunities for multimodal modeling, and (3) several datasets which are of broad interest but are released via licenses that restrict redistribution such as dyadic emotion recognition on IEMOCAP~\citep{busso2008iemocap}, deception prediction on from real-world Trial Data~\cite{perez2015deception}, and multilingual affect recognition on CMU-MOSEAS~\cite{zadeh2020moseas} which was only just recently released. We are currently working with the authors to integrate some of these datasets into \names\ in the near future. These new datasets will benchmark multimodal modeling in human-centric areas where privacy and fairness can be important desiderata. Furthermore, it will enable benchmarking of multimodal learning in languages other than English which is important towards building more accessible multimodal models that include the language modality.

\vspace{-1mm}
\subsubsection{Retrieval}
\vspace{-1mm}

Another area of great interest lies in cross-modal retrieval~\citep{liang2020cross,zhen2019deep}. In this area, the goal is to retrieve semantically similar data from a new modality using a modality as a query (e.g., given a phrase, retrieve the closest image describing that phrase). The core challenge is to perform alignment of representations across both modalities. Retrieval has been studied primarily in the multimedia space (e.g., retrieving images, video, and audio given a text query) and we hope to add some of these datasets as well as to expand datasets for cross-modal retrieval using different combinations of query and retrieved modalities.

\vspace{-1mm}
\subsubsection{Question Answering}
\vspace{-1mm}

Within the domain of language and vision, there has been growing interest in language-based question answering (i.e., ``query'' modality) of entities in the visual, video, or embodied domain (i.e., ``queried'' modality). Datasets such as Visual Question Answering~\citep{agrawal2017vqa}, Social IQ~\citep{zadeh2019social}, and Embodied Question Answering~\citep{das2018embodied} have been proposed to benchmark the performance of multimodal models in these settings. A core challenge lies in aligning words asked in the question with entities in the queried modalities, which can take the form of visual entities in images or videos, and actions in embodied environments. We plan to add these datasets as soon as possible, and also plan to add QA over multiple queried modalities such as text, images, and tables as proposed in recent work~\citep{hannan2020manymodalqa,talmor2021multimodalqa}.

\vspace{-1mm}
\subsubsection{Grounding}
\vspace{-1mm}

Grounding is the task of linking entities (often at their most granular level) in one modality with entities in another modality. As an example, in the domain of language and vision, a well-studied grounding task is visual referring expressions - the task of localizing an object in an image referred to by a natural language expression (e.g., \textit{half of a sandwich on the right side of a plate nearest a coffee mug})~\citep{cirik-etal-2018-visual}. Grounding can be seen as a more fine-grained version of retrieval where the retrieved modality of interest is at the level of sub-patches of an image. We currently do not include tasks in the grounding area since there are no datasets outside using language to query images (and their subregions). We plan to include grounding datasets in the language and vision domain but also encourage research in extending this research problem to other modalities (e.g., using language to query video/audio/sets/tables).

\vspace{-1mm}
\subsubsection{Reinforcement Learning}
\vspace{-1mm}

Learning from multiple modalities in an interactive setting is an area of interest as a step towards building more intelligent embodied agents that can perceive the visual world, language instructions, auditory feedback, and other sensor modalities. These research areas broadly span language-conditional RL (i.e., instruction following, learning a reward function from instructions, language in the observation or action space) and language-assisted RL (language as domain knowledge, language to structure policies)~\citep{luketina2019survey}. Recent work has also explored audio as a modality in an agent's multisensory interaction with the world~\citep{dean2020see}. Modern robot systems are also equipped with multiple sensors to aid in their decision-making and there has been considerable research in learning multimodal representations from multiple sensors for robot manipulation~\citep{lee2020detect,lee2020multimodal,lee2019making}.

These multimodal problems are fundamentally different from those that are concerned with prediction tasks. Alongside the core challenges in learning complementary information and aligning entities in language instructions to those in the visual environment, there also lies the core challenge of learning \textit{actionable} representations that link to the set of actions that can be taken and their associated long-term rewards~\cite{luketina2019survey}. We plan to include these datasets in a future version of \names. We also encourage research in extending these multimodal tasks beyond language and vision to truly incorporate the diverse set of modalities humans use in everyday interactive tasks.

\vspace{-1mm}
\subsection{Models}
\vspace{-1mm}

By partitioning the structure of multimodal code into the distinct areas in Appendix~\ref{appendix:algos} (data processing, unimodal and multimodal model design, optimization objectives, and training structures), \codes\ enables easy addition of new innovations from all areas. It is easy to add new unimodal encoders as they are developed in areas such as computer vision and natural language processing. Similarly, it is extremely simple to add multimodal methods while ensuring compatibility with existing unimodal encoders, fusion paradigms, optimization objectives, and training structures. Please refer to Appendix~\ref{appendix:code} for code snippets changing multimodal models, optimization objectives, and training structures.

The authors maintain a reading list for topics in multimodal ML~\citep{readinglist} that is regularly updated for the latest advances in the area. We plan to periodically add proposed methods to the \codes\ toolkit with help from the community as well.

\vspace{-1mm}
\subsection{Evaluation}
\vspace{-1mm}

\names\ is designed with holistic evaluation in mind. Currently, \names\ supports evaluation for prediction performance, time and space complexity, and robustness to noisy and missing modalities. There are several other crucial evaluation dimensions that we plan to include in the following versions of the benchmark:

\vspace{-1mm}
\subsubsection{Uncertainty Estimates}
\vspace{-1mm}

There has been important work in building ML models that return uncertainty estimates along with their prediction targets~\citep{gal2016dropout,gneiting2007probabilistic} along with recent interest in building multimodal models with similar capabilities~\citep{brown2020uncertainty,xia2020uncertainty}. As ML models are increasingly deployed in real-world sensitive scenarios~\cite{ws-2016-nlp-social,journals/nle/Dale19,VELUPILLAI201811}, there is an increasing need to quantify when ML models do not know the right answer and potentially abstain~\citep{ziyin2019deep} or defer the prediction to a human expert~\cite{kompa2021second}. As future steps, we plan to also include evaluations of uncertainty predictions into \names, such as using the recently proposed Uncertainty Toolkit~\citep{uncertainty,chung2020beyond,tran2020methods}. This will enable the inclusion and evaluation of uncertainty-predicting multimodal models such as the ones proposed in~\citep{brown2020uncertainty,xia2020uncertainty}.

\vspace{-1mm}
\subsubsection{Robustness to Distribution Shifts}
\label{appendix:shifts}
\vspace{-1mm}

Distribution shifts, spanning shifts in dataset distributions and label distributions, are among core challenges currently preventing machine learning systems from being safely deployed in real-world settings~\cite{rabanser2018failing}. Subtle changes in the data distribution can significantly impact performance, a phenomenon exemplified by adversarial examples~\citep{szegedy2013intriguing}, and shifts in the label distribution can significantly compromise accuracy as well~\citep{zhang2013domain}.

Distribution shifts in multimodal settings have not been explored by the research community. Multimodal data can exhibit shifts in the marginal data distribution of each modality as well as in the joint distribution across modalities, which makes the problem inherently more complex. To enable research in benchmarking and analyzing distribution shift in multimodal settings, we plan to include:
\begin{enumerate}
    \item \textit{Data:} Data partitions (or new datasets) to \names\ that test for generalization across domains and subpopulations, in a manner similar to~\citep{koh2020wilds}. Building on the current datasets available in \names, some examples include affect recognition across different users, robotic manipulation across different physical robots, and medical diagnosis across different age groups.
    \item \textit{Algorithms:} On the algorithmic side, we plan to include currently established methods for distribution shift in a single modality (which has been the bulk of existing work) into \codes, which will enable both theoreticians and practitioners to analyze the new challenges that multimodal data brings to the study of distribution shift.
    \item \textit{Evaluation:} Finally, to evaluate robustness to distribution shift, we plan to build a standardized evaluation pipeline into \names\ (in a similar way for robustness tests currently implemented). We will also tap into insights from the experimental protocol in~\cite{rabanser2018failing} which includes evaluation metrics to detect dataset shift before attempting to correct it.  
\end{enumerate}

\vspace{-1mm}
\subsubsection{Fairness}
\label{appendix:fairness}
\vspace{-1mm}

To safely deploy human-centric multimodal models in real-world scenarios such as healthcare, HCI, legal systems, and social science, it is necessary to recognize the role they play in shaping social biases and stereotypes. Recent work has shown that word-level embeddings reflect and propagate \textit{social biases} present in training corpora~\cite{bolukbasi2016man,caliskan2017semantics}. Machine learning systems that incorporate these word embeddings can further amplify biases~\cite{barocas2016big} and unfairly discriminate against users, particularly those from disadvantaged social groups. Similar observations have been observed for datasets and models in the visual domain such as facial recognition~\citep{anastasi2005own} and image captioning~\citep{hendricks2018women} tasks, which has called for immediate efforts towards better documentation and risk analysis of both ML datasets~\citep{gebru2018datasheets} and models~\citep{mitchell2019model}.

We believe that the ability to make fair judgments is even more important in a multimodal setting for the following reasons:
\begin{enumerate}
    \item Human behavior is inherently multimodal. As a result, many research problems in multimodal learning involve human-centric data and tasks such as healthcare, affective computing, HCI, multimedia, human-robot interaction. As multimodal systems (such as emotion recognition systems) are deployed in the real world, it is crucial to characterize possible social biases they encode and design algorithms to mitigate these biases. Otherwise, real harm can be brought to under-represented populations which unfair machine learning models disproportionately harm~\cite{bolukbasi2016man}.
    \item While there has been a large body of work investigating the fairness of representations learned from language and images, there is little work currently investigating this for other modalities, as well as for the wide spectrum of multimodal models integrating multiple modalities which can potentially \textit{compound} biases stemming from each one~\citep{srinivasan2021worst}.
\end{enumerate}

There are many definitions of fairness and bias in ML and it is unclear which are important in which multimodal settings. While we do not have the best answer to conclusively evaluate for fairness in multimodal systems, we are making it a priority to include this feature in future versions of \names. In reference to~\citep{mehrabi2019survey}, certain dimensions of fairness we are currently exploring and plan to add to \names\ include:
\begin{enumerate}
    \item \textit{Data:} A better fine-grained understanding of bias in data, which we plan to achieve via human annotations for several multimodal datasets in \names\ (especially those that involve human-centric tasks such as affect recognition).
    \item \textit{Algorithms:} Algorithmic fairness, including training models that satisfy individual and group fairness, analyzing trained models from a geometric perspective (i.e., studying whether biases are encoded in representations learned by a model~\citep{bolukbasi2016man,liang2020fair}), and methods for pre-processing and post-processing data and models to satisfy fairness metrics.
    \item \textit{Evaluation:} Bias evaluation of trained multimodal models as well as those trained within a single modality, to determine the relationship between biases in a single modality versus those that manifest in multimodal problems, and comparing them to current progress in this direction on the language and vision modalities~\citep{ross2020measuring,srinivasan2021worst}.
\end{enumerate}
These tasks tackle benchmarking and analysis of biases in multimodal methods from different perspectives spanning data, algorithms, and evaluation, which make them compatible with our proposed modular framework in \names\ and \codes. We plan to include additional data annotations in the \names\ data loader, a suite of algorithms designed to mitigate bias for unimodal and multimodal models in \codes, and evaluation metrics for fairness in the \names\ evaluation pipeline.

\vspace{-1mm}
\subsection{Broader Outreach}
\vspace{-1mm}

\textbf{In workshops and competitions:} The authors have extensive experience in organizing challenges, workshops, and tutorials at leading ML, NLP, and computer vision conferences. Among these include large-scale challenges in multimodal language analysis at NAACL 2021 (\url{http://multicomp.cs.cmu.edu/naacl2021multimodalworkshop/}), ACL 2020 (\url{http://multicomp.cs.cmu.edu/acl2020multimodalworkshop/}), and ACL 2019 (\url{http://multicomp.cs.cmu.edu/acl2018multimodalchallenge/}). We plan to use \names\ as the subject of future workshops to accelerate reproducible research in multimodal learning. These workshops will focus on both new algorithms as well as careful analysis of existing algorithms in the field. Both directions will be accelerated via our resources: we plan to provide \names\ as a starting point for loading datasets and \codes\ as starter code for multimodal modeling, evaluation, and analysis.

\textbf{In academic courses:} We plan to use the \names\ benchmark as well as the standardized \codes\ codebase as an educational tool to support the Multimodal ML course taught annually at CMU (\url{https://cmu-multicomp-lab.github.io/mmml-course/fall2020/}). Students can choose to use one of the datasets provided in \names\ or add a new one to the current suite of multimodal datasets. When designing new algorithmic contributions, students can implement their approaches in the \codes\ toolkit which enables easy testing on multiple datasets, quick logging and analysis of results, and reproducible testing. This method of community-based expansion is also likely to see great leaps in the variety of datasets and models supported by this toolkit.

\textbf{Community-based expansion:} Finally, we plan to present a system for expanding the datasets and models in \names\ via input from the research community. Since \names\ is publicly released and will be regularly maintained, the existing starting benchmark, code, evaluation, and experimental protocols can greatly accelerate the addition of new datasets and models in the future. In the public GitHub (\dataurl), we have included a section on contributing to \names\ through either task proposals or additions of datasets and algorithms. The readme includes detailed instructions for adding new datasets and dataloaders, as well as new algorithms by modifying according to the code structure we have developed and standardized. The readme also contains details for writing a main function to test new data loaders and multimodal algorithms, and a test script to ensure compatibility with existing experiments. The authors will regularly monitor new proposals through this channel. Periodically, the authors will select popular task proposals (datasets and models) and add it into new versions of \names. The ease of loading datasets and evaluating models will naturally encourage interest in building new datasets and models on top of the toolkit. We further plan to encourage participants/students in our organized workshops and courses to use \names\ and contribute task proposals as well.

\end{document}